\documentclass{article}
\PassOptionsToPackage{numbers,sort&compress}{natbib}
\usepackage[main,final]{neurips_2026}

\makeatletter
\renewcommand{\@notice}{\enlargethispage{2\baselineskip}}
\makeatother

\usepackage[utf8]{inputenc} %
\usepackage[T1]{fontenc}    %
\usepackage{hyperref}       %
\usepackage{url}            %
\usepackage{booktabs}       %
\usepackage{amsfonts}       %
\usepackage{nicefrac}       %
\usepackage{microtype}      %
\usepackage{xcolor}         %
\usepackage{amsthm}

\usepackage{tocloft}
\usepackage{titletoc}

\usepackage{graphicx}

\usepackage[utf8]{inputenc}
\usepackage{amsthm}
\newtheorem{theorem}{Theorem}
\newtheorem{proposition}{Proposition}
\newtheorem{remark}{Remark}

\usepackage{wrapfig}
\usepackage[numbers,sort&compress]{natbib}

\usepackage{tcolorbox}
\usepackage{amsmath}
\usepackage{mathtools}
\usepackage{amssymb}

\usepackage{makecell}
\usepackage{multirow}
\usepackage{algorithm}
\usepackage{algorithmicx}
\usepackage{algcompatible}
\usepackage{algpseudocode}
\usepackage{caption}
\usepackage{xcolor}

\newtcolorbox{goalbox}{
    colback=gray!8,
    colframe=gray!45,
    boxrule=0.5pt,
    arc=2pt,
    left=6pt,
    right=6pt,
    top=6pt,
    bottom=6pt
}

\definecolor{darkgreen}{RGB}{0,100,0} 
\definecolor{linkgrayblue}{RGB}{70,90,110}
\definecolor{dustyrose}{RGB}{150,110,120}

\newcommand{\methodname}{ParetoSlider}

\hypersetup{hidelinks,pdftitle={ParetoSlider: Diffusion Models Post-Training for Continuous Reward Control}, pdfauthor={Shelly Golan, Michael Finkelson, Ariel Bereslavsky, Yotam Nitzan, Or Patashnik}}

\title{\methodname: Diffusion Models Post-Training for Continuous Reward Control}

\author{
  {\normalsize
  Shelly Golan$^{1}$ \enspace
  Michael Finkelson$^{1,2}$ \enspace
  Ariel Bereslavsky$^{1}$ \enspace
  Yotam Nitzan$^{3}$ \enspace
  Or Patashnik$^{1}$
  }\\[0.5em]
  {\normalfont\small
  $^{1}$Tel Aviv University \quad
  $^{2}$Lightricks \quad
  $^{3}$Adobe Research
  }
}

\begin{document}

\maketitle
\vspace{-10pt}
\begin{figure}[h!]
\vspace{-10pt}
\centering
\includegraphics[width=\textwidth]{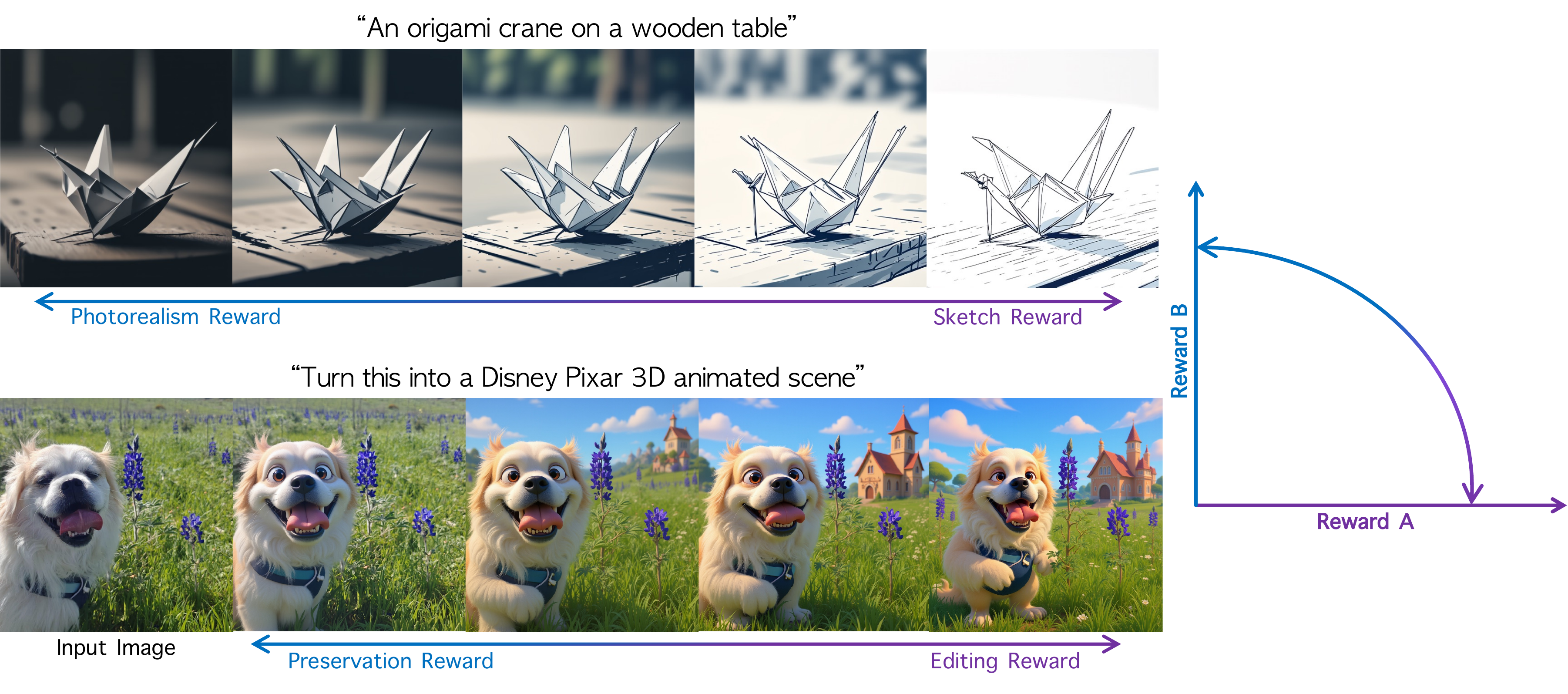}
\caption{\small ParetoSlider enables smooth inference-time
control over competing rewards trade-off via a single preference-conditioned model.
\textit{Top:} Text-to-image generation sliding between photorealism and sketch.
\textit{Bottom:} Image editing sliding between source preservation and prompt adherence.}
\end{figure}

\begin{abstract}
Reinforcement Learning (RL) post-training has become the standard for aligning generative models with human preferences, yet most methods rely on a single scalar reward. When multiple criteria matter, the prevailing practice of ``early scalarization'' collapses rewards into a fixed weighted sum. This commits the model to a single trade-off point at training time, providing no inference-time control over inherently conflicting goals -- such as prompt adherence versus source fidelity in image editing. We introduce \textbf{\methodname{}}, a multi-objective RL (MORL) framework for Diffusion Models post-training that allows continuous reward control during inference. By training the model with continuously varying preference weights as a conditioning signal the model learns how its denoising trajectory should change as the requested reward priority changes. Consequently, we enable users to navigate optimal trade-offs at inference time without retraining or maintaining multiple checkpoints. We evaluate \methodname{} across three state-of-the-art flow-matching backbones: SD3.5, FluxKontext, and LTX-2. Our single preference-conditioned model matches or exceeds the performance of baselines trained separately for fixed reward trade-offs, while uniquely providing fine-grained control over competing generative goals.
        
\end{abstract}
\vspace{-12pt}
\section{Introduction}
\vspace{-8pt}
\label{sec:intro}

Reinforcement Learning (RL) has emerged as the cornerstone for aligning Large Language Models (LLMs) with nuanced human intent, transforming them into highly capable systems across a vast array of real-world applications
\cite{ouyang2022traininglanguagemodelsfollow, shao2024deepseekmathpushinglimitsmathematical, brown2020languagemodelsfewshotlearners}.
This success is now driving a parallel paradigm shift in visual generative modeling \cite{ fan2023dpokreinforcementlearningfinetuning, black2024trainingdiffusionmodelsreinforcement, clark2024directlyfinetuningdiffusionmodels, wallace2023diffusionmodelalignmentusing}.
Recent advances \cite{liu2025flowgrpotrainingflowmatching, xue2025dancegrpounleashinggrpovisual, zheng2026diffusionnftonlinediffusionreinforcement} have successfully applied RL methods to diffusion and flow matching models to optimize intricate and occasionally subjective objectives such as aesthetic appeal, style fidelity, and adherence to text prompts \cite{kirstain2023pickapicopendatasetuser, wu2023humanpreferencescorev2, ma2025hpsv3widespectrumhumanpreference, xu2023imagerewardlearningevaluatinghuman}. 

In practice, these models are not judged based on a single metric. 
Rather, their utility depends on the simultaneous satisfaction of multiple rewards that jointly define the quality and utility of the generated content.
In instruction-based editing, for example, a model aims to maximize a reward for instruction adherence while satisfying a second reward for faithful preservation of the source image.
These objectives are inherently in tension: pushing for more aggressive editing often degrades structural preservation.

This conflict is a hallmark of multi-objective optimization (MOO), where there is typically no single ``perfect'' solution \cite{miettinen1999nonlinear, deb2011multi}.
In practice, MOO problems are often addressed through \textit{early scalarization}, which collapses multiple objectives into a fixed weighted sum \cite{miettinen1999nonlinear, deb2011multi,Roijers_2013,Hayes_2022}. 
This approach requires a costly search for weighting coefficients and compromises on a static operating point, thereby forcing a permanent compromise that precludes the flexibility required to adapt to varying user-defined preferences at inference time.
A more principled approach to MOO is identifying the \textit{Pareto front}: the set of optimal trade-offs where no single criterion can be improved without degrading another.

While recent efforts have begun addressing multi-objective alignment in visual generative models through RL (MORL), they typically face a trilemma of flexibility, efficiency, and scalability.
Some methods utilize Pareto-based selection during training, but remain limited to a fixed, static trade-off at inference-time \cite{lee2025capo, lee2024parrotparetooptimalmultirewardreinforcement}. 
Others achieve inference-time control through model interpolation, yet this requires training and storing multiple checkpoints -- a cost that scales linearly with the number of objectives \cite{cheng2025diffusionblendinferencetimemultipreference, rame2023rewardedsoupsparetooptimalalignment}. 
Finally, training-free steering approaches offer flexibility but suffer from heavy per-step sampling overhead \cite{jin2025inferencetimealignmentcontroldiffusion, xie2025dymotrainingfreediffusionmodel}.

To bridge this gap, we introduce \methodname{}, a controllable multi-objective reinforcement learning (MORL) framework for diffusion alignment.
Rather than restricting the model to a fixed balance of rewards during training, we introduce a preference vector $\omega$ that 
specifies the desired relative priority among reward objectives and is provided to the model as a conditioning signal.
Since the model is conditioned on the preference vector $\omega$, it has the foresight to produce results that abide by the current relative reward preference.
Consequently, the model learns how its denoising trajectory should change as the requested reward trade-off changes within a single set of parameters, rather than converging to a single static solution.

To efficiently optimize this continuous formulation, we build upon DiffusionNFT~\cite{zheng2026diffusionnftonlinediffusionreinforcement}, an RL fine-tuning framework for flow-matching models. We compute NFT losses for each reward independently and aggregate them based on preferences $\omega$ only at the loss level.
At inference time,varying $\omega$ exposes a continuous control knob for navigating the learned reward trade-off surface without the overhead of training multiple checkpoints or generating dedicated datasets.

We evaluate \methodname{} on three state-of-the-art flow-matching backbones: Stable Diffusion~3.5 \cite{sd35} for text-to-image synthesis, FluxKontext \cite{labs2025flux1kontextflowmatching} for instruction-based image editing, and LTX-2 \cite{hacohen2026ltx2efficientjointaudiovisual} for text-to-video generation.
Across all three domains, our single preference-conditioned model matches or exceeds the performance of multiple separately tuned early-scalarization baselines at their respective operating points, while enabling continuous inference-time control over reward weights.
Through extensive ablations, we compare conditioning mechanisms for injecting the preference vector into diffusion transformers, identifying design choices that enable a smooth and continuous transition between reward trade-offs.
Additionally, we compare scalarization strategies and loss aggregation formulations. We show that our aggregation method with per-reward losses yields more fine-grained controllability over the trade-off than baseline alternatives and we extend this analysis to non-convex reward trade-offs.

\vspace{-5pt}
\section{Related Work}
\vspace{-5pt}
\label{sec:related}

\paragraph{\textbf{RL Fine-Tuning of Diffusion Models.}}
\label{subsec:diffusion_FT}
RL-based post-training for generative models is commonly divided into \emph{offline} methods, which use static preference datasets such as diffusion DPO~\cite{wallace2023diffusionmodelalignmentusing}, and \emph{online} methods, which sample from the current policy and query reward functions during training. Online methods can explore beyond the support of a fixed dataset, making them attractive for reward-driven visual alignment.
Early online methods~\cite{black2024trainingdiffusionmodelsreinforcement, fan2023dpokreinforcementlearningfinetuning} applied policy gradients over full denoising trajectories, while differentiable-reward methods~\cite{clark2024directlyfinetuningdiffusionmodels} improved efficiency by back-propagating through reward models directly. More recently, GRPO~\cite{shao2024deepseekmathpushinglimitsmathematical} has been adapted to flow-matching models~\cite{liu2025flowgrpotrainingflowmatching,xue2025dancegrpounleashinggrpovisual,li2025mixgrpo}, replacing learned value networks with group-based normalization. DiffusionNFT~\cite{zheng2026diffusionnftonlinediffusionreinforcement} further shifts optimization to the forward process, using implicit velocity steering and a flow-matching loss to avoid storing or differentiating through sampled trajectories.
Despite these advances, existing online diffusion RL methods tend to optimize a single scalar reward or a fixed weighted combination of rewards, yielding one operating point with no inference-time control. \methodname{} extends DiffusionNFT to a controllable multi-reward setting, enabling continuous preference control along the supported Pareto front.

\vspace{-8pt}
\paragraph{\textbf{Multi-Objective RL.}}
\label{subsec:MORL}
Multi-objective RL (MORL) seeks policies that navigate trade-offs among conflicting objectives~\cite{10.1007/s10458-022-09552-y}. In LLM alignment, recent methods condition a single model on continuous preference weights, either by appending them to prompts~\cite{yang2024rewardsincontextmultiobjectivealignmentfoundation}, injecting them through adapters~\cite{zhong2024panaceaparetoalignmentpreference}, or incorporating them into preference-learning objectives~\cite{ren2025cosdpoconditionedoneshotmultiobjective}. This allows one model to cover many trade-offs and expose preferences as inference-time controls.
Multi-reward diffusion alignment remains comparatively less explored. Parrot~\cite{lee2024parrotparetooptimalmultirewardreinforcement} and Flow-Multi~\cite{lee2026flowmulti} use Pareto-based sample selection to balance rewards, but without explicit preference conditioning they converge to a fixed policy. Rewarded Soups~\cite{rame2023rewardedsoupsparetooptimalalignment} and Diffusion Blend~\cite{cheng2025diffusionblendinferencetimemultipreference} enable preference control by interpolating independently trained single-reward models or their denoising processes, so storage or inference complexity grows with the number of basis objectives. PROUD~\cite{yao2024proud} obtains Pareto-optimal samples through per-step gradient optimization, incurring substantial inference-time overhead.

\vspace{-8pt}
\paragraph{\textbf{Continuous Control in Generative Models.}}Traditional approaches to controllable generation focus on identifying or learning semantic directions within a model's latent space that can be modulated at inference time~\cite{gandikota2024concept, sridhar2024prompt, chiu2026text, parihar2025kontinuous}. While these methods provide smooth adjustment over specific visual concepts or attributes, they are fundamentally limited to the pre-existing capabilities of the frozen backbone and do not account for external, non-differentiable objectives. 
In contrast, \methodname{} shifts the paradigm from semantic steering to online reward optimization. Instead of modulating fixed concepts, our framework actively explores the trade-off space between competing reward functions during reinforcement learning. By conditioning the diffusion policy on the preference vector $\omega$, we internalize the entire Pareto front into the model's parameters. This allows for continuous navigation between optimized objectives---such as aesthetic quality versus prompt adherence---without the need for specialized data generation or the storage of multiple task-specific checkpoints.

\vspace{-5pt}
\section{Preliminaries}
\label{sec:prelim}
\vspace{-5pt}
\subsection{RL Formulation of Diffusion and Flow Models}
\vspace{-5pt}

Following prior work~\cite{black2024trainingdiffusionmodelsreinforcement, fan2023dpokreinforcementlearningfinetuning}, we formulate the iterative denoising process of a diffusion or flow-matching model as a sequential Markov decision process (MDP), defined by a tuple ($\mathcal{S}$, $\mathcal{A}$, $\mathcal{P}$, $\mathcal{R}$).
The state space $\mathcal{S}$ consists of all pairs $s_t = (x_t, c)$ of noisy latents $x_t$ at diffusion timestep $t$ and the conditioning signal $c$ (e.g., a text prompt or input image). 
The action space $\mathcal{A}$ consists of the model's per-step predictions $a_t$, which for flow-matching corresponds to the velocity vector $a_t=v_\theta(x_t,t,c)$. The transition dynamics $\mathcal{P}$ maps the pair($s_t$, $a_t$) to the next state $s_{t-1}$, corresponding to the sampling scheduler.
The diffusion model thus serves as the policy $\pi_\theta(x_{t-1} \mid x_t, c)$, mapping the current noisy state to a distribution over the next, less noisy state. 
The reward function $\mathcal{R}$ is nonzero only at the terminal step, where the final sample $x_0$ is evaluated by $M$ objectives, yielding $\mathbf{r}(x_0,c)=[r_1(x_0,c),\ldots,r_M(x_0,c)]^\top$.
Standard diffusion RL methods assume a scalar reward and optimize a single expected return. With multiple objectives, the conventional approach first sums the components of $\mathbf{r}(x_0,c)$ before applying a single-objective update, we refer to this as ``early scalarization''. This commits the policy to one fixed trade-off.
In the multi-objective setting, we instead define one expected return per objective,
$J_m(\pi)=\mathbb{E}_{c,\,x_0\sim\pi(\cdot\mid c)}[r_m(x_0,c)]$, and seek policies that approximate the pareto front across all $M$ objectives.

\vspace{-5pt}
\subsection{Pareto Dominance}
\vspace{-5pt}
\label{subsec:pareto_def}
When rewards compete, gains in one component generally come at the expense of another, and no single policy maximizes all components simultaneously. The relevant notion of optimality is therefore set-valued: instead of a single perfect policy, we seek an entire set of policies that approximate the Pareto-front.
We say that a policy $\pi$ \emph{dominates} another policy $\pi'$ (denoted $\pi \succ \pi'$) if it achieves an expected return that is no smaller on every objective and strictly larger on at least one:
\begin{equation}
\begin{aligned}
    \pi \succ \pi'
    \iff\;&
    \forall\, m \in \{1, \dots, M\},\; J_m(\pi) \geq J_m(\pi') \\
    &\land\; \exists\, l \in \{1,\dots,M\}
    \text{ s.t. } J_l(\pi) > J_l(\pi').
\end{aligned}
\end{equation}

\begin{figure}[t!]
    \centering
    \includegraphics[width=1\linewidth]{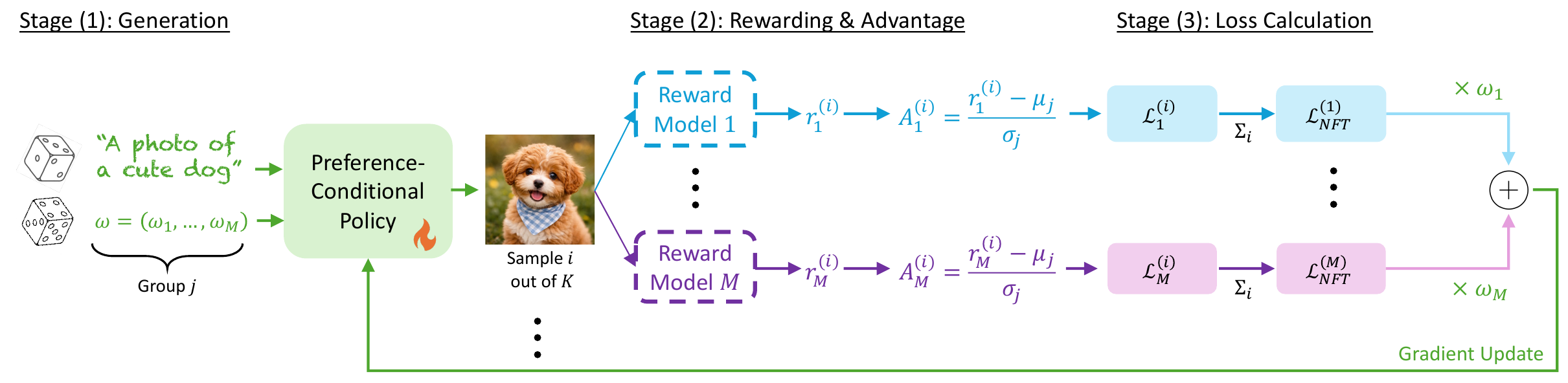}
    \vspace{-7pt}
    \caption{\small ParetoSlider training pipeline. (1) For each prompt and sampled $\omega$, the policy generates $K$ images, (2) which are scored by $M$ reward models and normalized into per-reward advantages. (3) A DiffusionNFT loss is computed per reward and aggregated with $\omega$ before the gradient update.}
    \vspace{-15pt}
\label{fig:method}
\end{figure}

\vspace{-8pt}
\subsection{DiffusionNFT}
\label{subsec:diffusion_nft}
\vspace{-5pt}
Prior RL-based fine-tuning of diffusion models, such as FlowGRPO, formulate reinforcement learning on the reverse sampling process \cite{liu2025flowgrpotrainingflowmatching, xue2025dancegrpounleashinggrpovisual}.
In particular, FlowGRPO optimization is carried out over a multi-step reverse-time trajectory. Each update depends on likelihood ratio terms accumulated across many denoising steps. This tends to be computationally expensive, since rewards must be propagated through long sampled trajectories.
DiffusionNFT~\cite{zheng2026diffusionnftonlinediffusionreinforcement} addresses these limitations by reformulating the policy optimization on the forward process rather than the reverse denoising process. Instead of optimizing a policy through reverse-time likelihood ratios over sampled denoising trajectories, they directly update the policy model via a standard flow-matching loss.

\vspace{-5pt}
\section{Method}
\label{sec:method}
\vspace{-5pt}
We introduce \methodname{}, a multi-objective reinforcement learning framework for diffusion models. 
Our goal is to enable users to continuously navigate the trade-off between a set of objectives, quantified by reward functions $r_1, \dots, r_M$, using a single controllable model.
To navigate the trade-offs between these objectives, we introduce an $M$-dimensional preference vector $\omega = [\omega_1, \dots, \omega_M]^\top$, where $\omega$ lies on the probability simplex, defined as follows:
\begin{equation}
    \omega = (\omega_1,\ldots,\omega_M)^\top \in \Delta^M,  \quad
    \Delta^M = \left\{\omega \in \mathbb{R}^M_{\geq 0}
        :
        \sum_{m=1}^M \omega_m = 1 \right\}.
\end{equation}
Optimizing this objective for different values of $\omega$ motivates a family of preference-induced policies corresponding to different reward trade-offs.
In \methodname{}, we amortize this family into a single preference-conditioned diffusion policy.
Under this formulation, different preference vectors $\omega$ specify different desired trade-offs among the reward functions. When the rewards are placed on a comparable scale, the desired policy is therefore a function of the specific weighting. 
Consequently, a static policy with no access to $\omega$ cannot represent the full family of preference-dependent outputs.
We therefore move beyond static policies and introduce a preference-conditioned diffusion policy $\pi_\theta(x_{t-1} \mid x_t, c, \omega)$.
By exposing the model to continuously varying preference $\omega$ during training, we enable the iterative denoising process to learn the mapping of any user-specified preference to its approximated point in the reward trade-off space.

In practice, raw reward functions often exhibit disparate numerical scales and variances.
To prevent ``loud'' rewards from hijacking the gradient and overshadowing the preference conditioning $\omega$, we move away from standard early scalarization in favor of a late-scalarization strategy, detailed as part of our training paradigm in \S\ref{subsec:training}. Then, in \S\ref{subsec:architecture} we describe the architectural mechanisms for conditioning the diffusion backbone on $\omega$.

\vspace{-5pt}
\subsection{Preference Guided Policy Training}
\label{subsec:training}
\vspace{-5pt}
We adopt DiffusionNFT (\S\ref{subsec:diffusion_nft}) as our base RL algorithm due to its strong reward-alignment performance and substantially improved training efficiency, reported to be 3$\times$ to 25$\times$ faster than FlowGRPO. 
Following this framework, our method relies on an iterative, three-stage process: (1) generating a batch of grouped visual samples using the current policy and scoring them across all reward functions, (2) transforming these raw rewards into advantages to stabilize the optimization, and (3) updating the policy using a combination of the DiffusionNFT loss and a KL-divergence loss.
To enable fine-grained control during inference, we modify each of these three stages to explicitly incorporate the preference vector $\omega$, as illustrated in Figure~\ref{fig:method}. During sample generation, the preference vector $\omega$ is provided as an explicit conditioning signal to the diffusion policy, so that each generated group is associated with a desired reward trade-off. During training, the preference $\omega$ is used only at the late aggregation stage to combine the resulting reward-induced DiffusionNFT losses. 
\begin{figure}[t]
    \centering
    \scriptsize
    \setlength{\tabcolsep}{0.002\textwidth}

    \begin{minipage}[t]{0.32\textwidth}
        \centering
        \begin{tabular}{c c c}
            \multicolumn{3}{c}{\textit{\makecell{``A hummingbird hovering near\\bright tropical flowers''}}} \\[2pt]
            \includegraphics[width=0.32\linewidth]{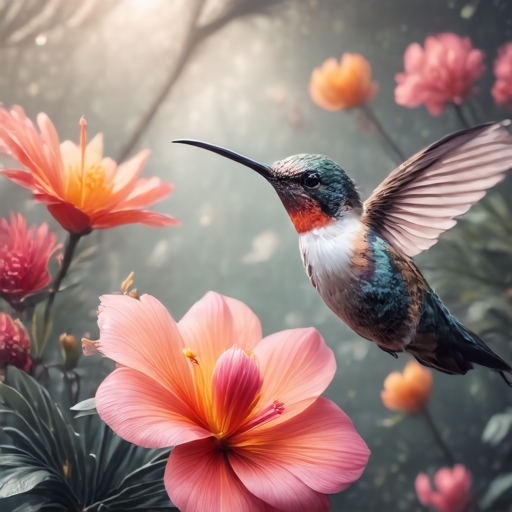} &
            \includegraphics[width=0.32\linewidth]{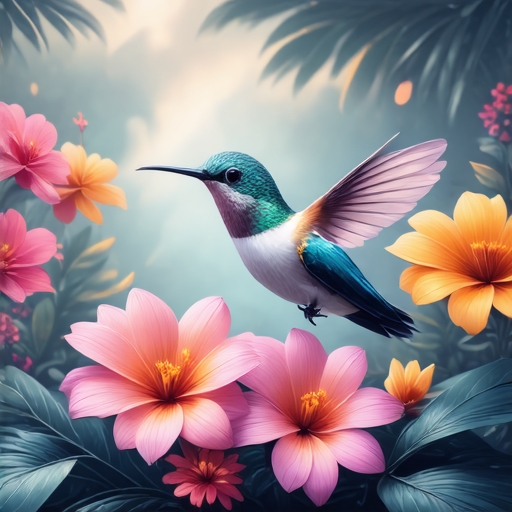} &
            \includegraphics[width=0.32\linewidth]{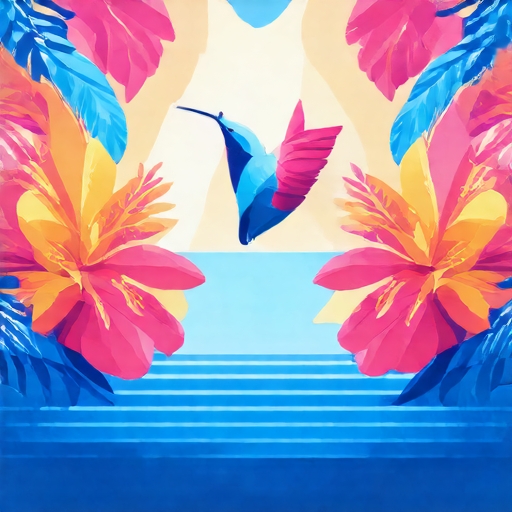} \\
            Photorealistic & $\xleftrightarrow{\hspace{1.1cm}}$ & Flat Vector Art
        \end{tabular}
    \end{minipage}\hfill
    \begin{minipage}[t]{0.32\textwidth}
        \centering
        \begin{tabular}{c c c}
            \multicolumn{3}{c}{\textit{\makecell{``A macro photo of a honeybee\\on a sunflower''}}} \\[2pt]
            \includegraphics[width=0.32\linewidth]{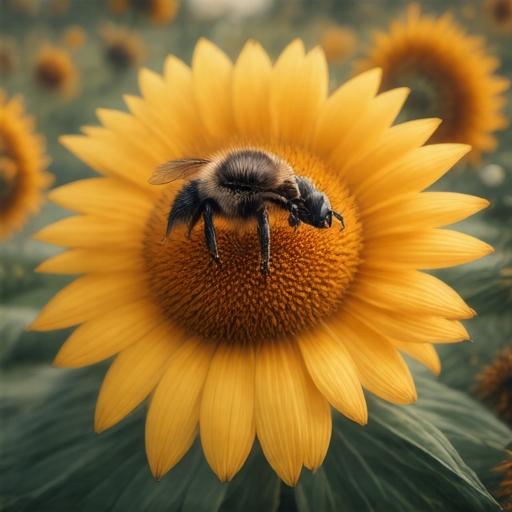} &
            \includegraphics[width=0.32\linewidth]{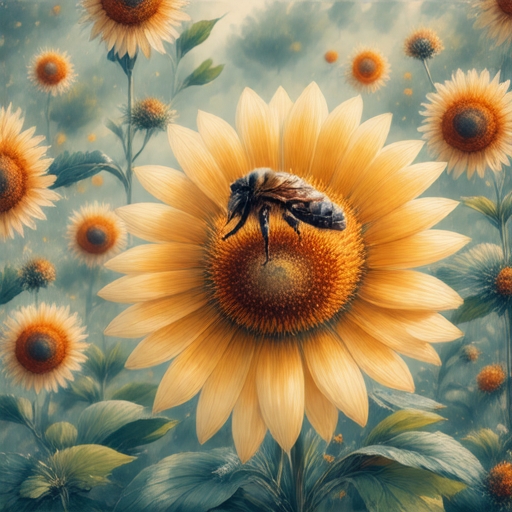} &
            \includegraphics[width=0.32\linewidth]{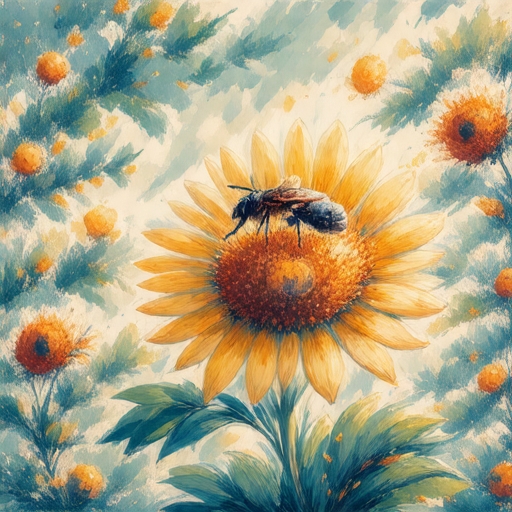} \\
            Photorealistic & $\xleftrightarrow{\hspace{1.1cm}}$ & Watercolor
        \end{tabular}
    \end{minipage}\hfill
    \begin{minipage}[t]{0.32\textwidth}
        \centering
        \begin{tabular}{c c c}
            \multicolumn{3}{c}{\textit{\makecell{``A girl riding a bicycle\\through a field of sunflowers''}}} \\[2pt]
            \includegraphics[width=0.32\linewidth]{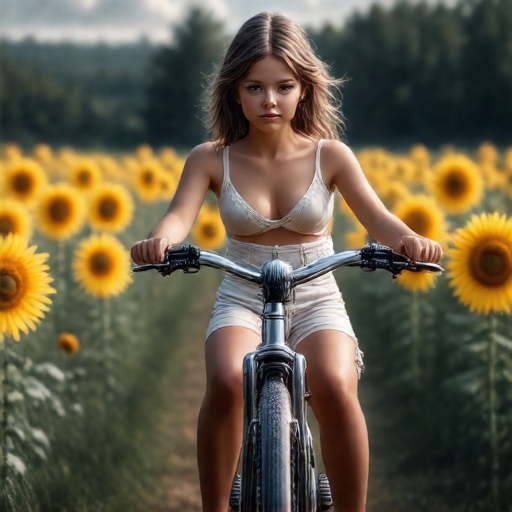} &
            \includegraphics[width=0.32\linewidth]{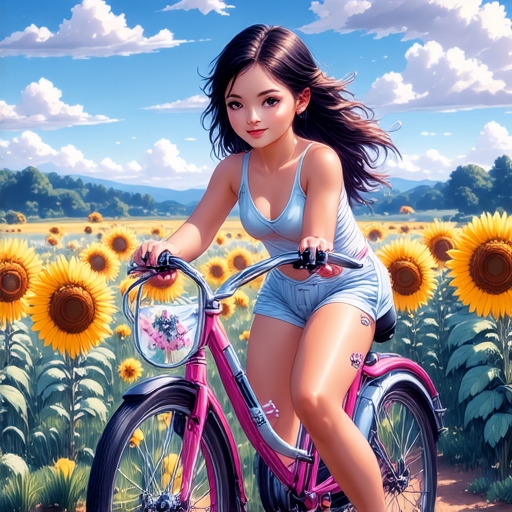} &
            \includegraphics[width=0.32\linewidth]{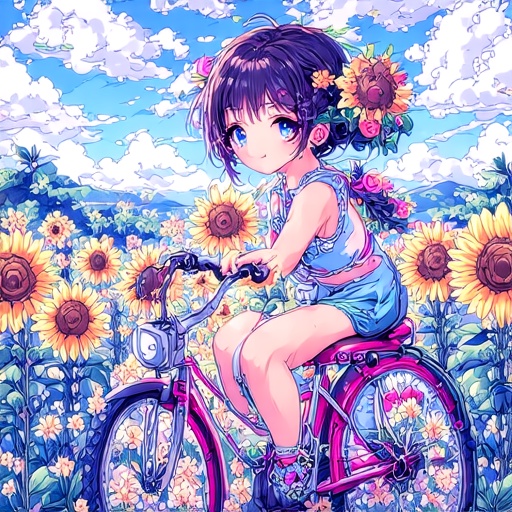} \\
            Photorealistic & $\xleftrightarrow{\hspace{1.1cm}}$ & Anime
        \end{tabular}
    \end{minipage}

    \vspace{3pt}

    \begin{minipage}[t]{0.32\textwidth}
        \centering
        \begin{tabular}{c c c}
            \multicolumn{3}{c}{\textit{\makecell{``A lighthouse on a green cliff\\overlooking a turquoise sea''}}} \\[2pt]
            \includegraphics[width=0.32\linewidth]{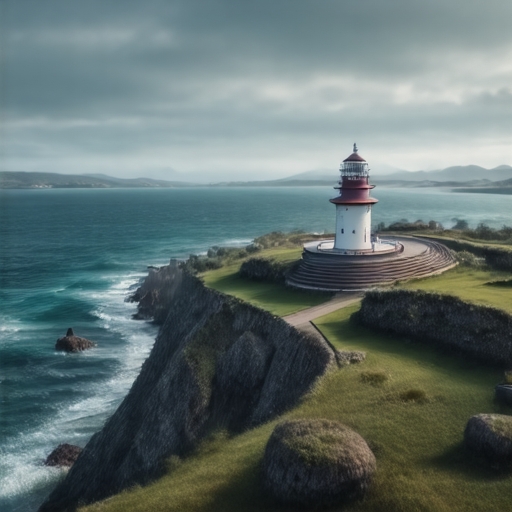} &
            \includegraphics[width=0.32\linewidth]{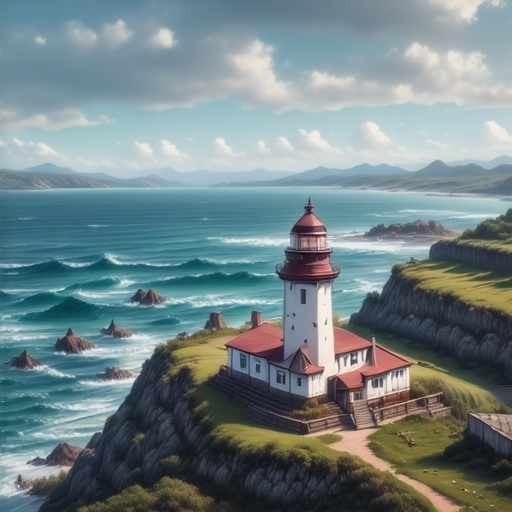} &
            \includegraphics[width=0.32\linewidth]{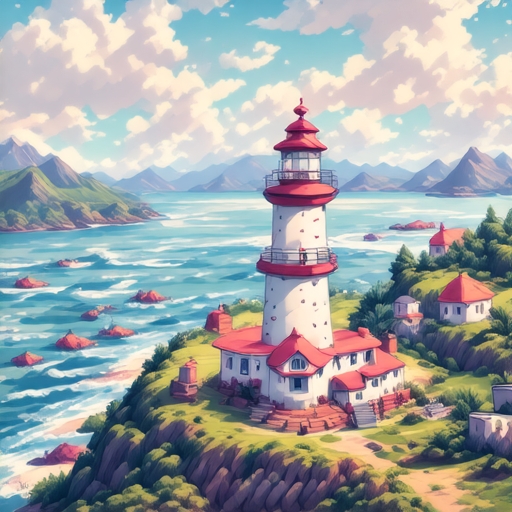} \\
            Photorealistic & $\xleftrightarrow{\hspace{1.1cm}}$ & Animated Scene
        \end{tabular}
    \end{minipage}\hfill
    \begin{minipage}[t]{0.32\textwidth}
        \centering
        \begin{tabular}{c c c}
            \multicolumn{3}{c}{\textit{\makecell{``An easter bunny on a spring day\\in a field holding a basket of easter eggs''}}} \\[2pt]
            \includegraphics[width=0.32\linewidth]{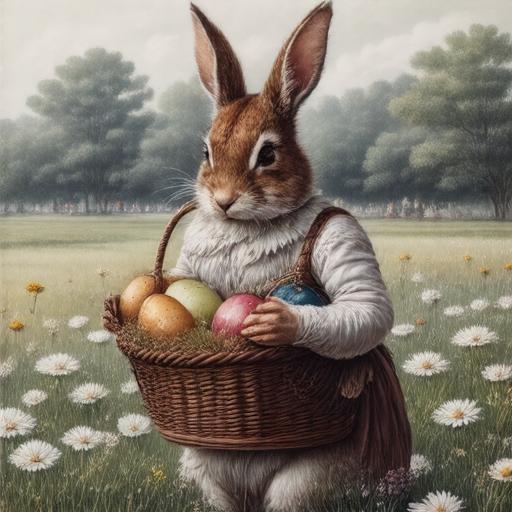} &
            \includegraphics[width=0.32\linewidth]{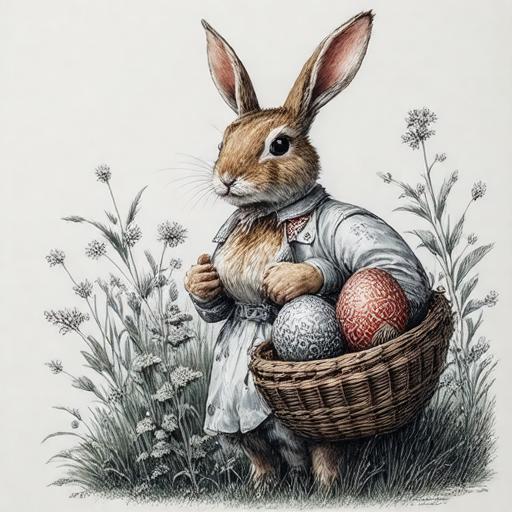} &
            \includegraphics[width=0.32\linewidth]{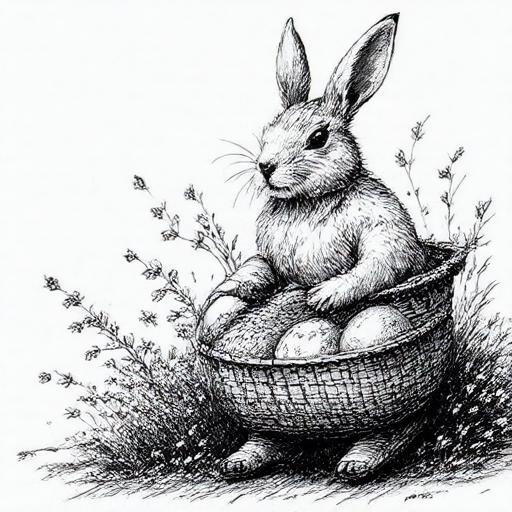} \\
            Photorealistic & $\xleftrightarrow{\hspace{1.1cm}}$ & Sketch
        \end{tabular}
    \end{minipage}\hfill
    \begin{minipage}[t]{0.32\textwidth}
        \centering
        \begin{tabular}{c c c}
            \multicolumn{3}{c}{\textit{\makecell{``A hot air balloon flying over\\a lavender field''}}} \\[2pt]
            \includegraphics[width=0.32\linewidth]{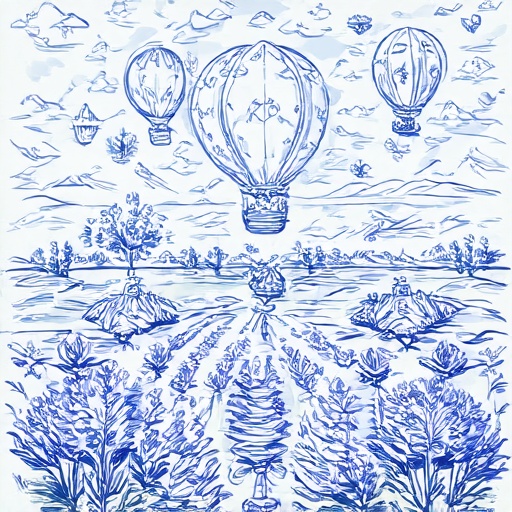} &
            \includegraphics[width=0.32\linewidth]{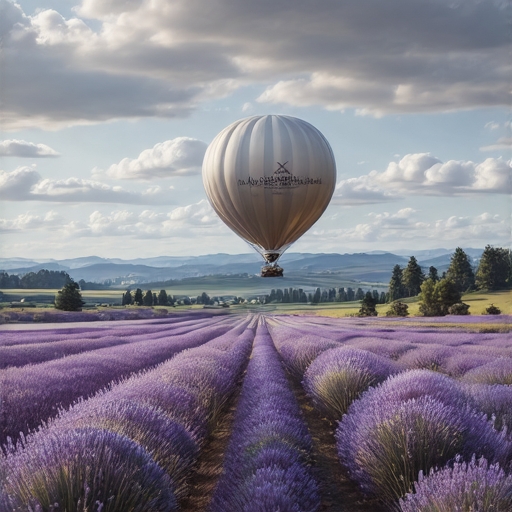} &
            \includegraphics[width=0.32\linewidth]{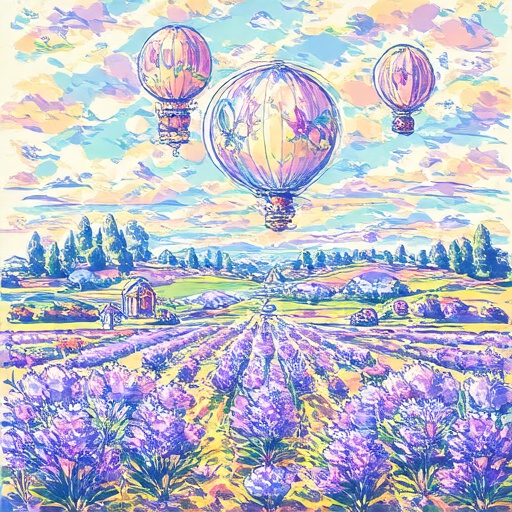} \\
             Sketch & Photorealistic & Watercolor
        \end{tabular}
    \end{minipage}
    \caption{\small Multi-objective style interpolation results. Each triplet shows images generated at three preference configurations spanning the Pareto front.}
    \label{fig:t2i_grid}
\end{figure}

\vspace{-5pt}
\paragraph{\textbf{Preference-Conditioned Group Generation.}}
Following the group-relative policy optimization strategy used in FlowGRPO, we adopt a group-based generation strategy to construct the batch. 
While standard frameworks like DiffusionNFT condition this generation solely on the input $c$, our policy must be explicitly conditioned on both $c$ and the preference vector $\omega$. To achieve this, for each input $c$, we draw a preference $\omega$ from a distribution over the simplex $\Delta^M$ (defined in \S\ref{sec:method}). The conditioned policy $\pi_\theta(x_{t-1} \mid x_t, c, \omega)$ then generates a group of $K$ samples for the pair $(c, \omega)$. Samples within the same group share both the input condition and the requested reward trade-off, so their rewards can be compared using group-relative statistics. We then proceed to the advantage calculation. For a group $j$ containing $K$ samples, generated by the same conditioning pair $(c,\omega)$, the advantage of a sample $i$ and the reward objective $m$ is defined as: 
\vspace{-4pt}
\begin{equation} 
    A_m^{(i)} =  (r_m^{(i)} - \mu_{j,m})/{(\sigma_{j,m} + \epsilon)},
    \label{eq:grpo_advantage}
\end{equation}
where $r_m^{(i)} = r_m(x_0^{(i)},c)$ is the value of reward channel $m$ assigned to the $i$-th sample in the group $j$, and $\mu_{j,m}$ and $\sigma_{j,m}$ are the mean and standard deviation of reward channel $m$ across the $K$ samples in that group. 
$\epsilon$ is a small constant added to prevent division by zero.

\vspace{-5pt}
\paragraph{\textbf{Preference-Weighted Policy Optimization.}}
\label{par:optimization}
For each generated sample $i$ and reward channel $m$, we first compute a per-objective standardized advantage $A_m^{(i)}$ across the $K$ samples in the same prompt-preference group (Eq.~\ref{eq:grpo_advantage}). We then map each advantage to a DiffusionNFT interpolation coefficient $\rho_m^{(i)}=\frac{1}{2}+\frac{1}{2} \mathrm{clip}\!\left(\frac{A_m^{(i)}}{\epsilon_{\mathrm{clip}}},-1, 1\right)$, where $\rho_m^{(i)} \in [0,1]$. Which yields a separate DiffusionNFT loss for each reward:
\begin{equation}
    \mathcal{L}_{m}^{(i)}
    =
    \rho_m^{(i)} \|v_+^{(i)}-v^{(i)}\|_2^2
    +
    \bigl(1-\rho_m^{(i)}\bigr)
    \|v_-^{(i)}-v^{(i)}\|_2^2 .
\end{equation}
The preference vector enters only at the final loss aggregation:
\begin{equation}
    \mathcal{L}_{\mathrm{NFT}}^{(i)} = \sum_{m=1}^{M}\omega_m\mathcal{L}_{m}^{(i)},
    \qquad
    \mathcal{L}_{\mathrm{NFT}} = \frac{1}{K}\sum_{i=1}^{K} \mathcal{L}_{\mathrm{NFT}}^{(i)} .
    \label{eq:w_nft}
\end{equation}
The full objective adds a KL penalty to the pretrained reference model,
$\mathcal{L} = \mathcal{L}_{\mathrm{NFT}} + \lambda_{\mathrm{KL}}\mathcal{L}_{\mathrm{KL}}$.

\vspace{-5pt}
\paragraph{\textbf{Preference-Weighted Aggregation.}}
Given a desired preference vector $\omega\in\Delta^{M}$, we aggregate the NFT loss vector from Equation \ref{eq:w_nft} into a single scalar signal. For sample $i$, we denote the vector of reward-induced DiffusionNFT losses $ \boldsymbol{\mathcal{L}}^{(i)} = \left(\mathcal{L}_1^{(i)},\ldots,\mathcal{L}_M^{(i)}\right)$. Our main method uses linear preference-weighted aggregation: $g_{\omega}^{\mathrm{LS}}\!\left(\boldsymbol{\mathcal{L}}^{(i)}\right) = \sum_{m=1}^{M}\omega_m\mathcal{L}_{m}^{(i)}$.
This scalarization is performed in loss space, not directly in reward space. Thus, \methodname{} suggests a practical diffusion post-training setting by learning a single policy whose denoising behavior changes with the preference vector $\omega$. To further explore how linear and nonlinear loss aggregation variants affect reward trade-off controllability we present empirical ablations Section \S\ref{subsec:ablations}.

\vspace{-5pt}
\subsection{Preference-Conditioning Architectures}
\label{subsec:architecture}
\vspace{-5pt}
\begin{wrapfigure}{r}{0.5\textwidth}
    \vspace{-\baselineskip}
    \centering
    \scriptsize
    \setlength{\tabcolsep}{0.5pt}
    \renewcommand{\arraystretch}{0.99}
    \newcommand{\editw}{0.19\linewidth}

    \begin{tabular}{@{}c c c c@{}}
        \multicolumn{4}{c}{\textit{``Change this portrait into a pixel-art style''}} \\[2pt]
        \includegraphics[width=\editw]{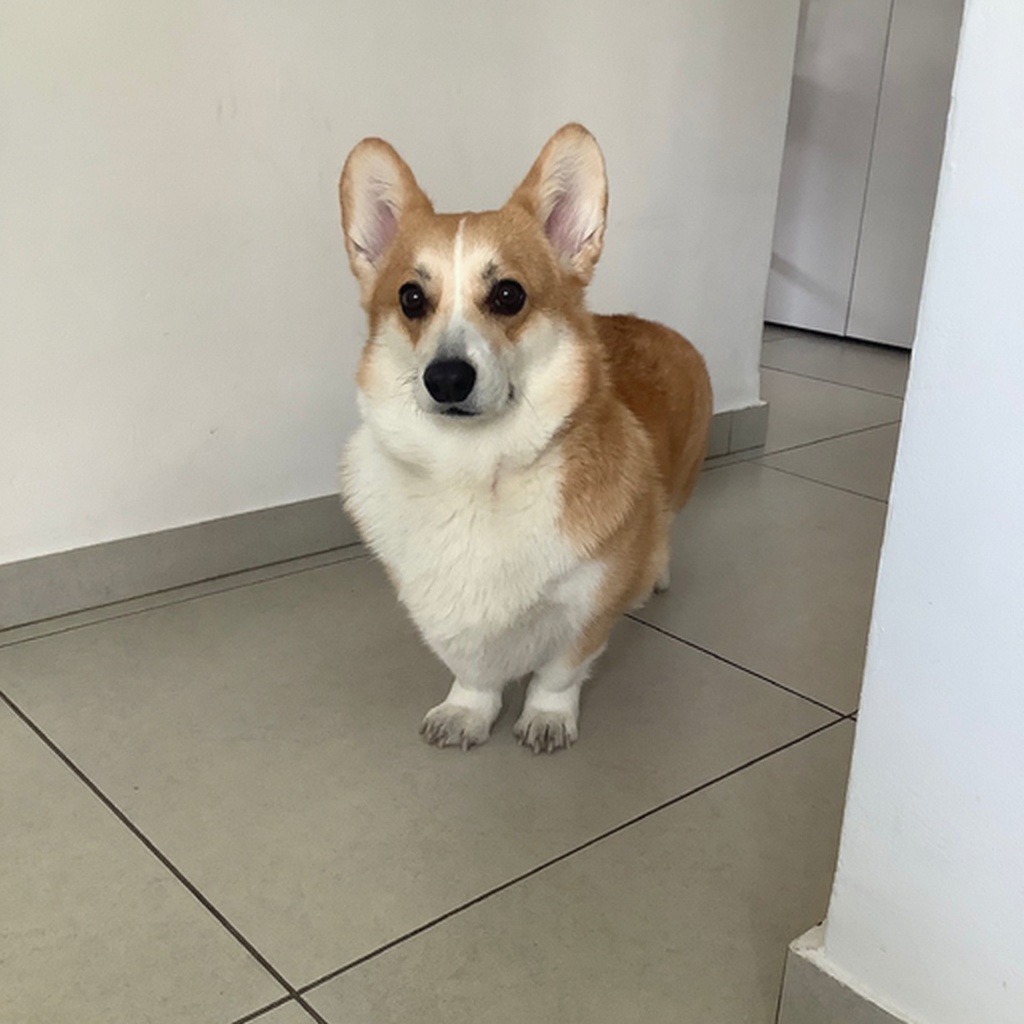} &
        \includegraphics[width=\editw]{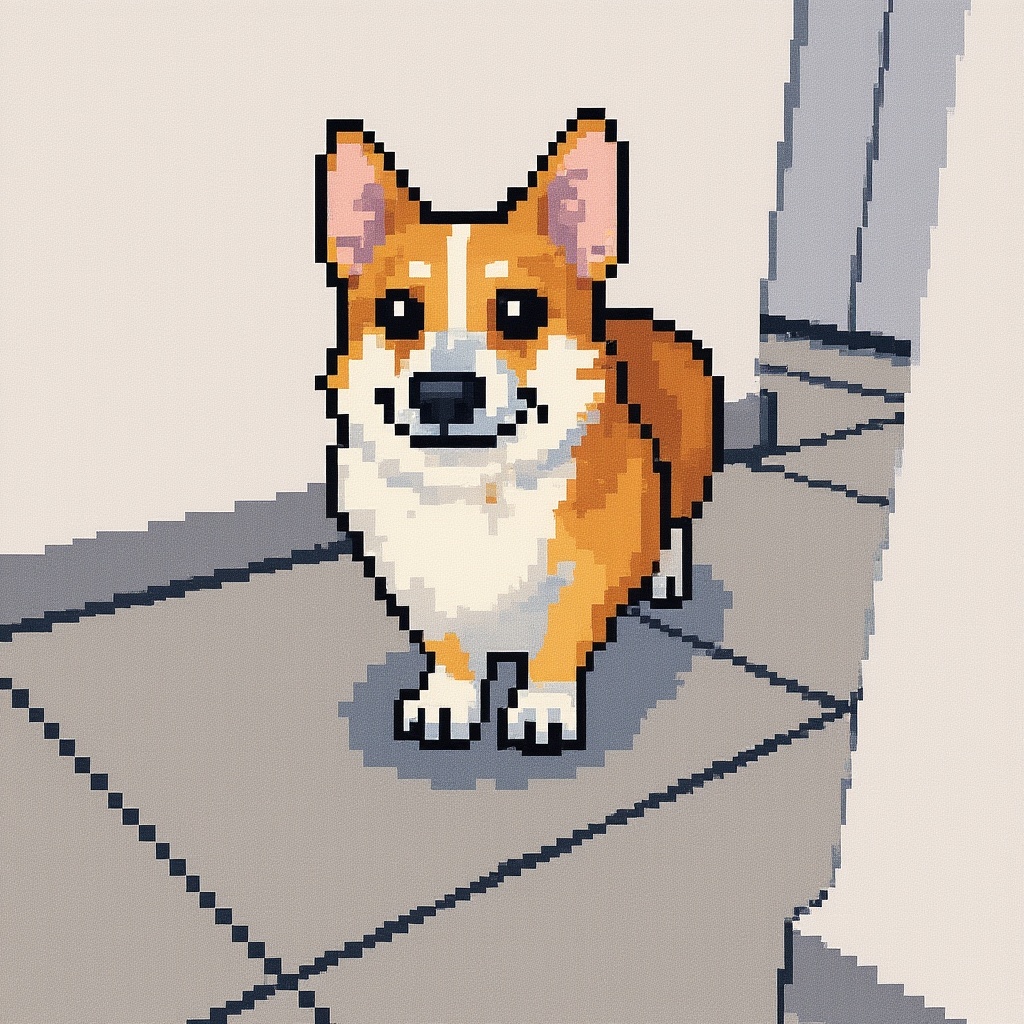} &
        \includegraphics[width=\editw]{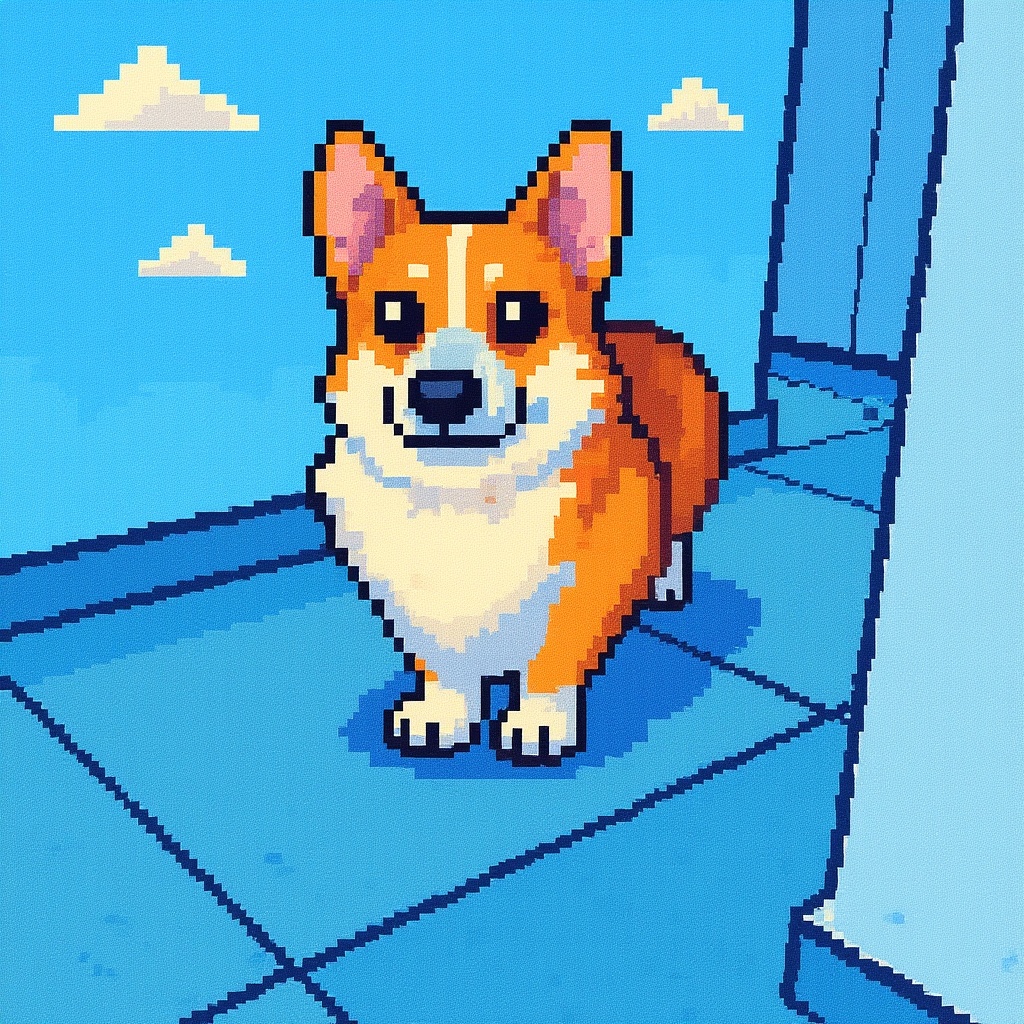} &
        \includegraphics[width=\editw]{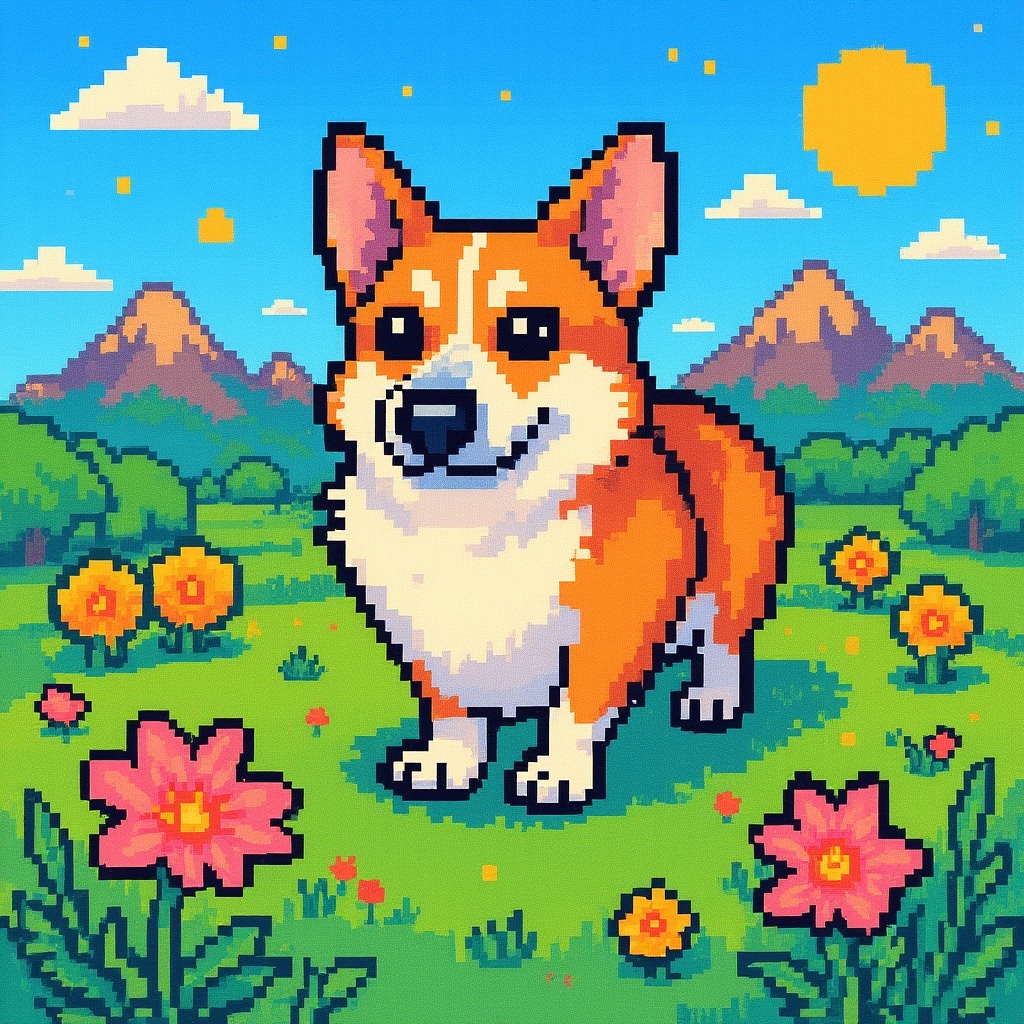} \\
        Input & Preserve & Balanced & Edit
    \end{tabular}

    \begin{tabular}{@{}c c c c@{}}
        \multicolumn{4}{c}{\textit{``Turn this into a 3D-rendered Disney Pixar scene''}} \\[2pt]
        \includegraphics[width=\editw]{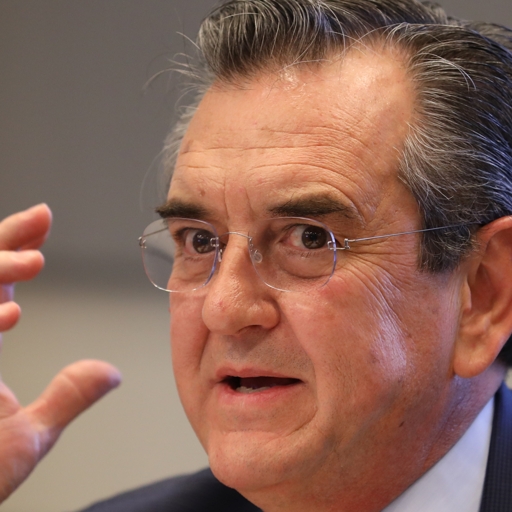} &
        \includegraphics[width=\editw]{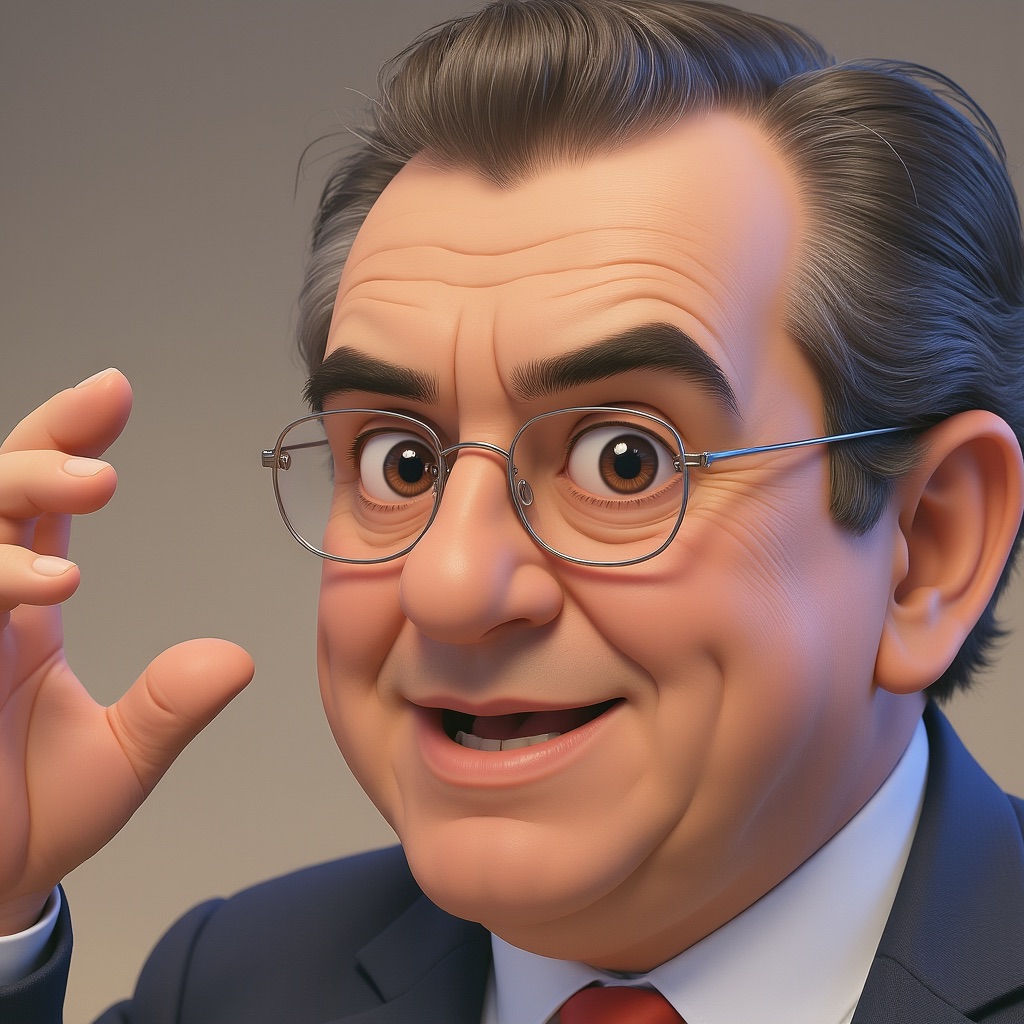} &
        \includegraphics[width=\editw]{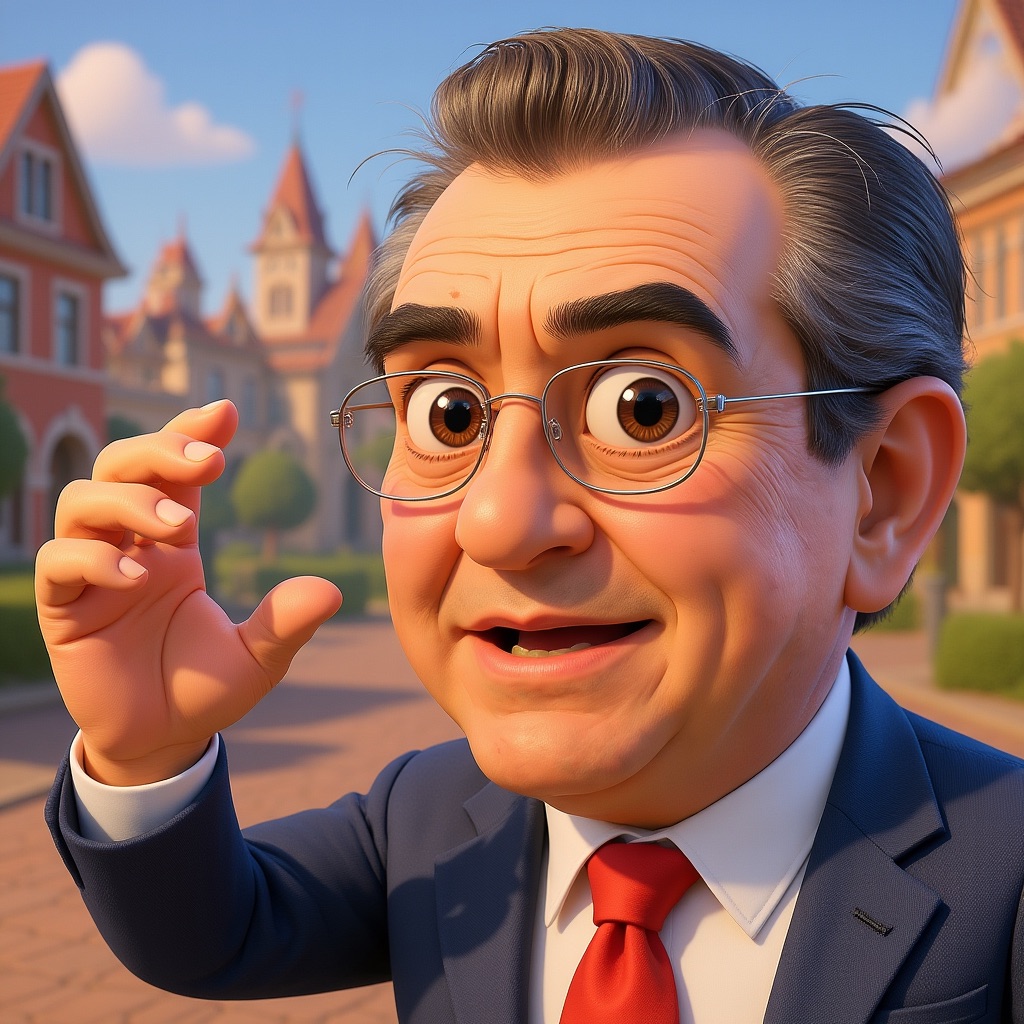} &
        \includegraphics[width=\editw]{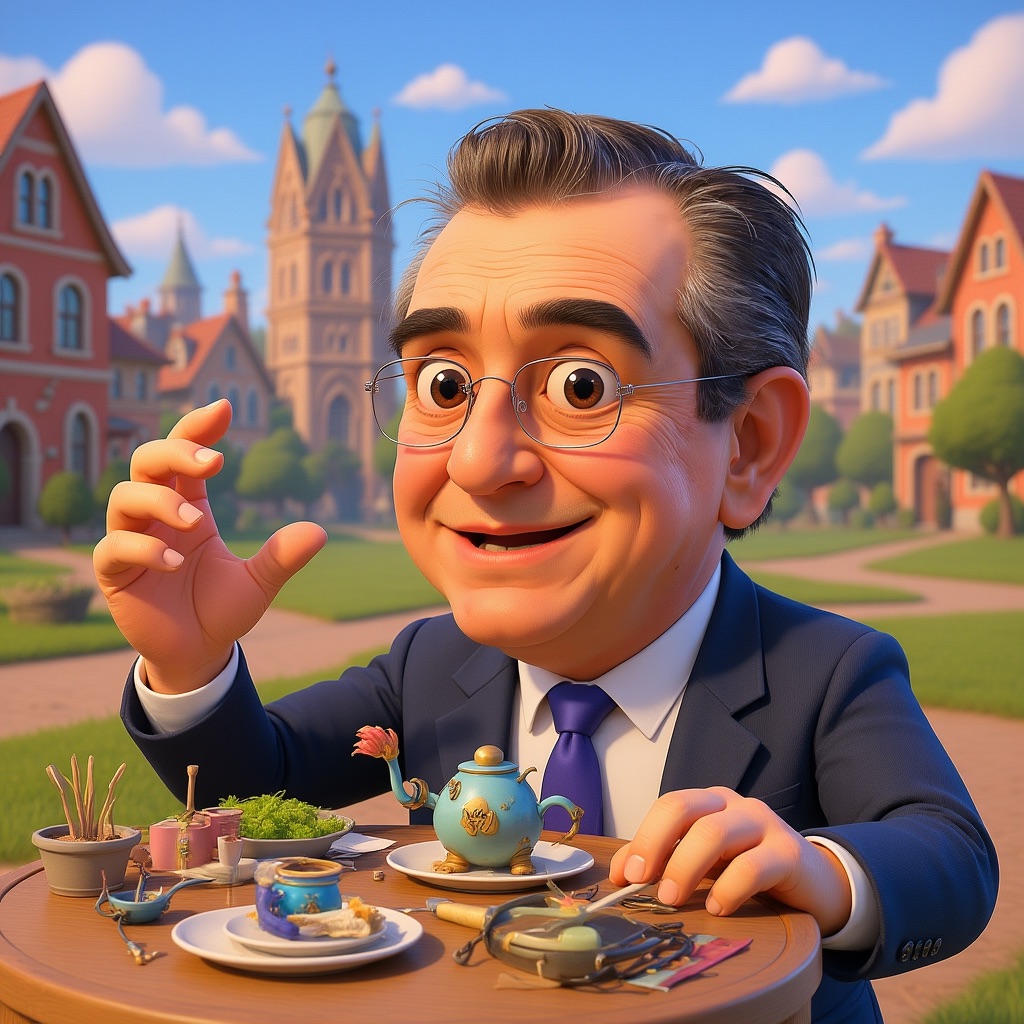} \\
        Input & Preserve & Balanced & Edit
    \end{tabular}

    \begin{tabular}{@{}c c c c@{}}
        \multicolumn{4}{c}{\textit{``Turn this woman into a warrior''}} \\[2pt]
        \includegraphics[width=\editw]{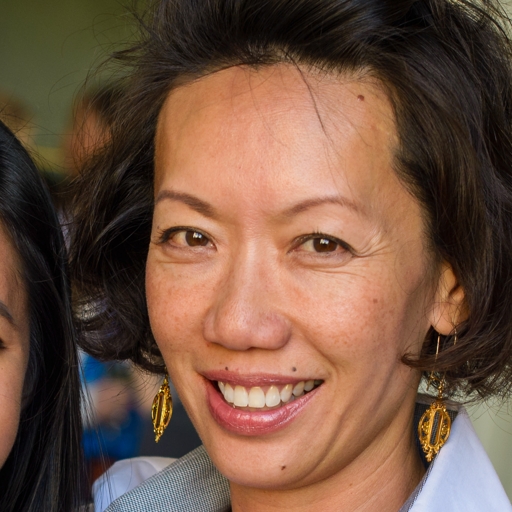} &
        \includegraphics[width=\editw]{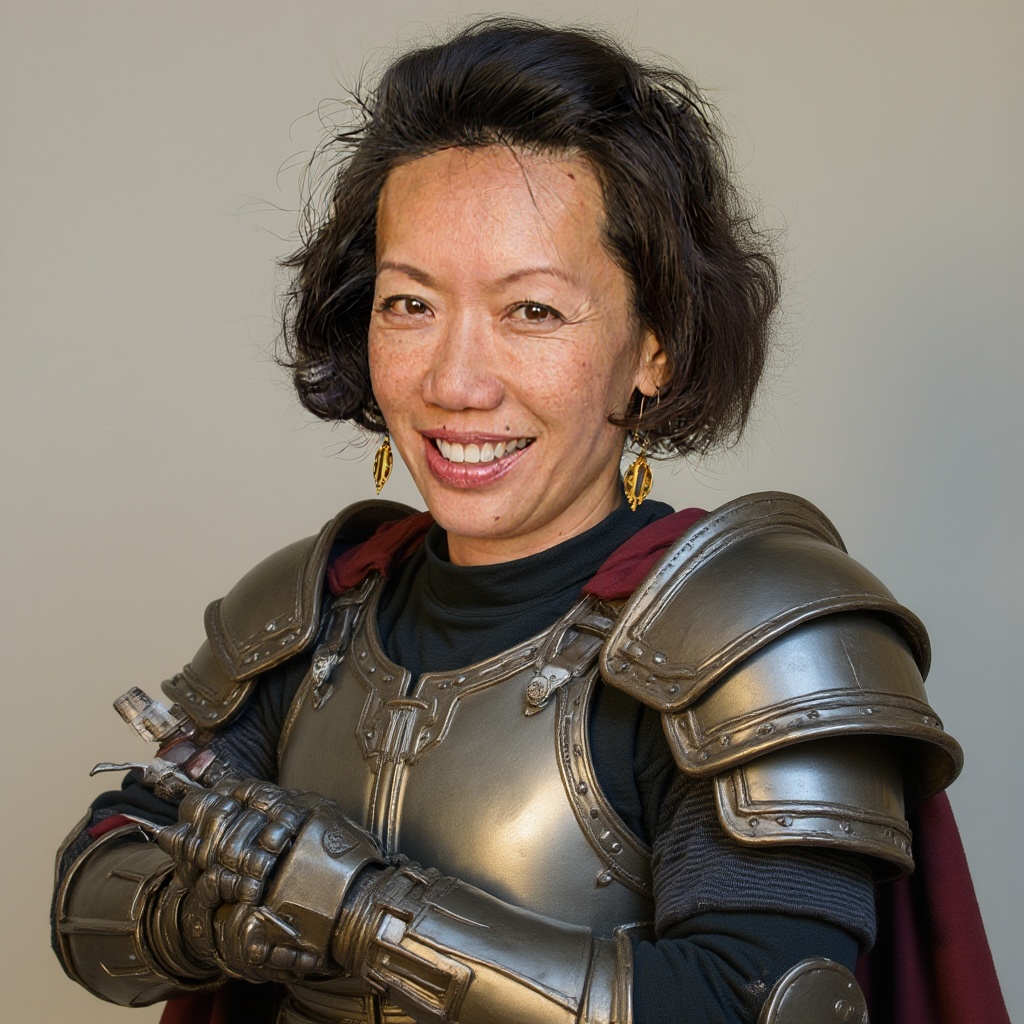} &
        \includegraphics[width=\editw]{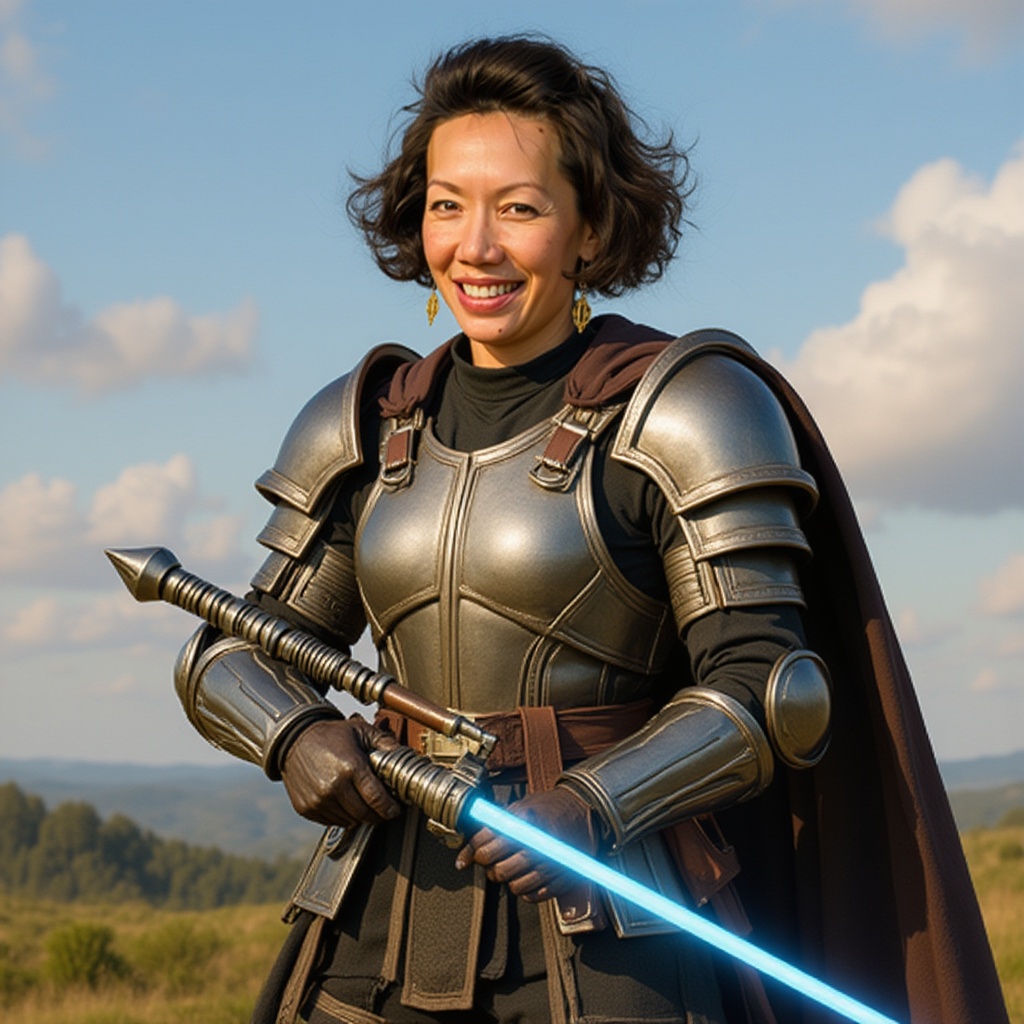} &
        \includegraphics[width=\editw]{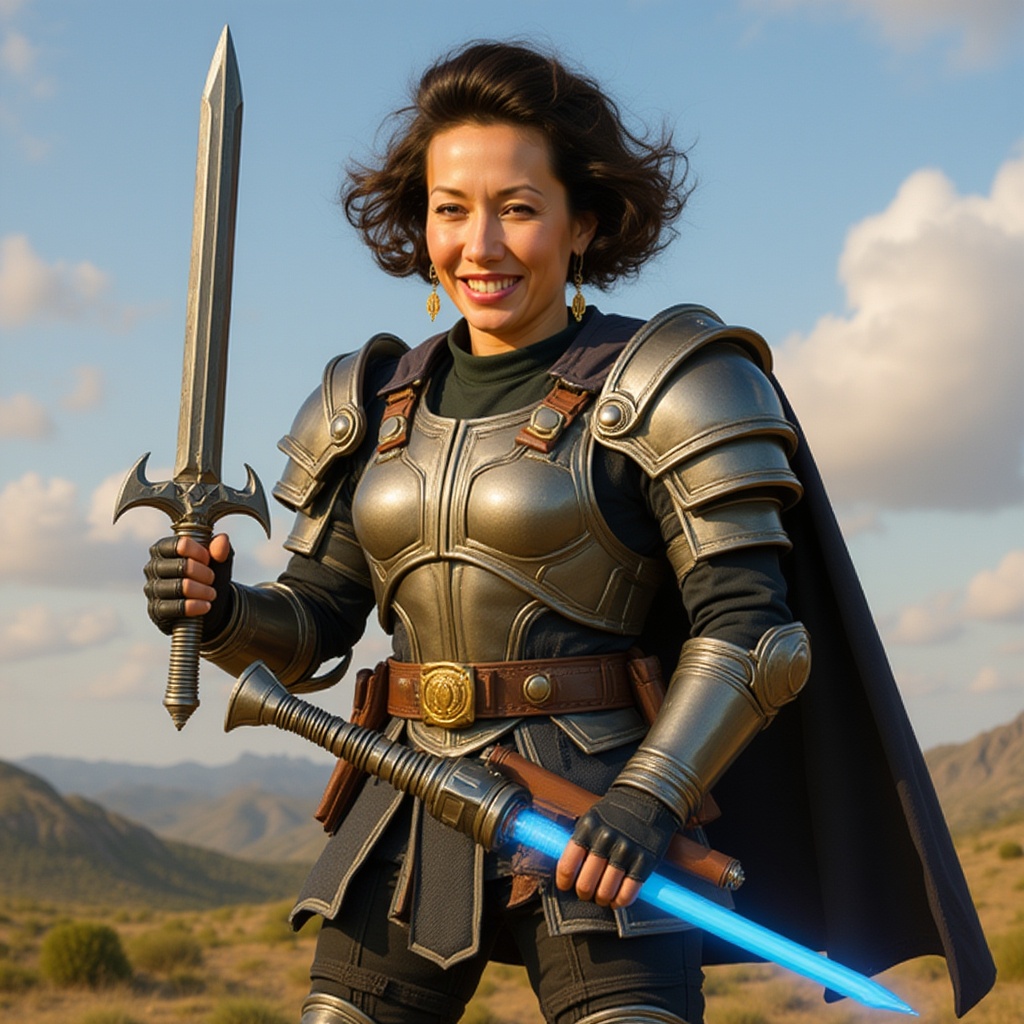} \\
        Input & Preserve & Balanced & Edit
    \end{tabular}

    \begin{tabular}{@{}c c c c@{}}
        \multicolumn{4}{c}{\textit{``Change the style of this image to a Ghibli scene''}} \\[2pt]
        \includegraphics[width=\editw]{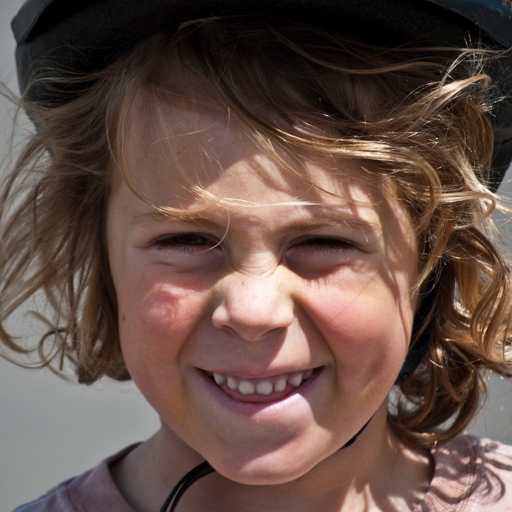} &
        \includegraphics[width=\editw]{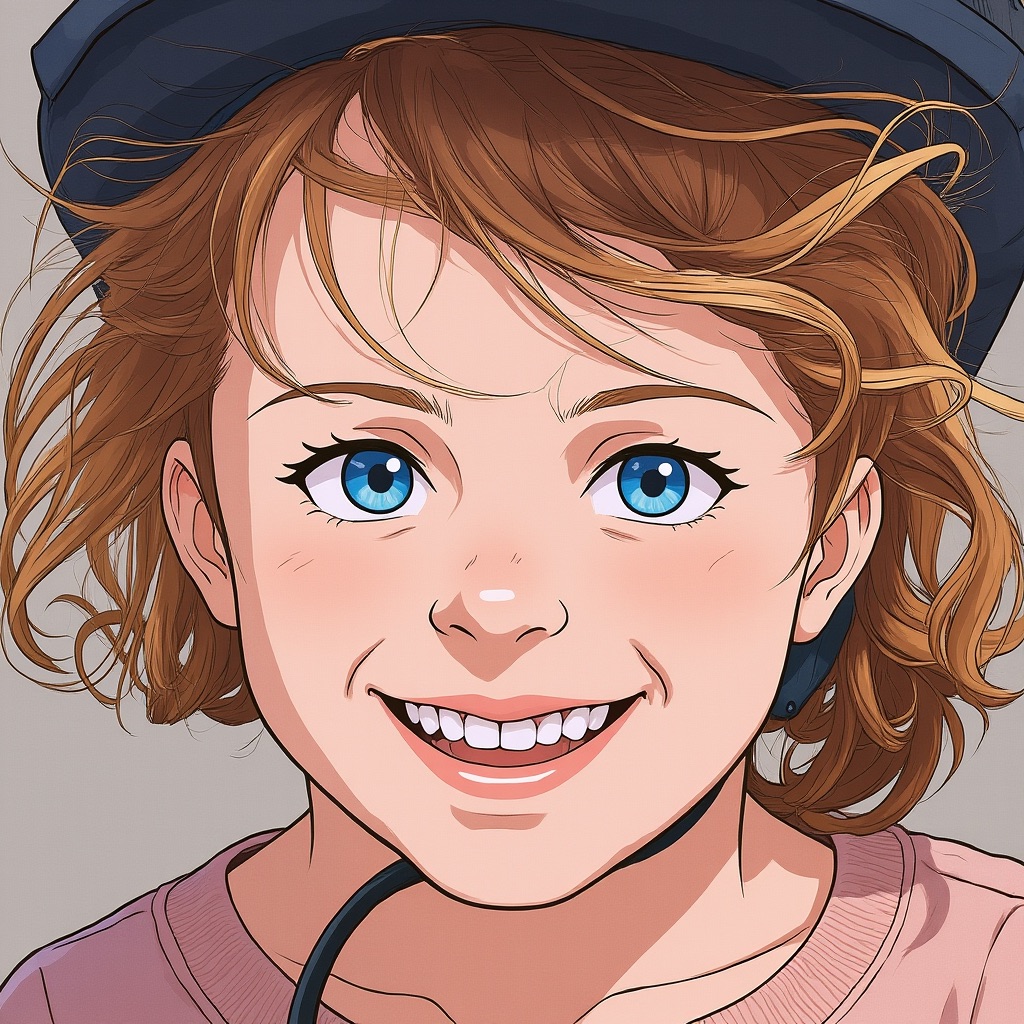} &
        \includegraphics[width=\editw]{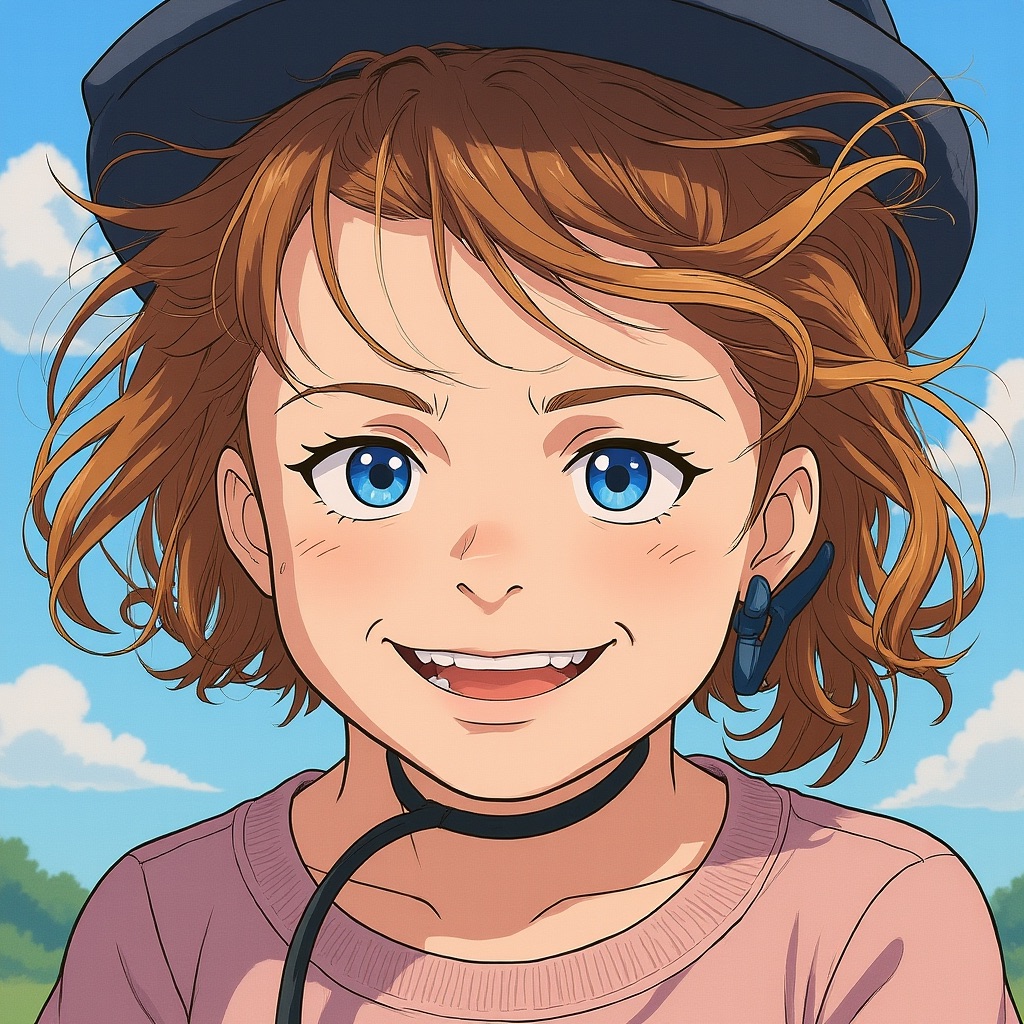} &
        \includegraphics[width=\editw]{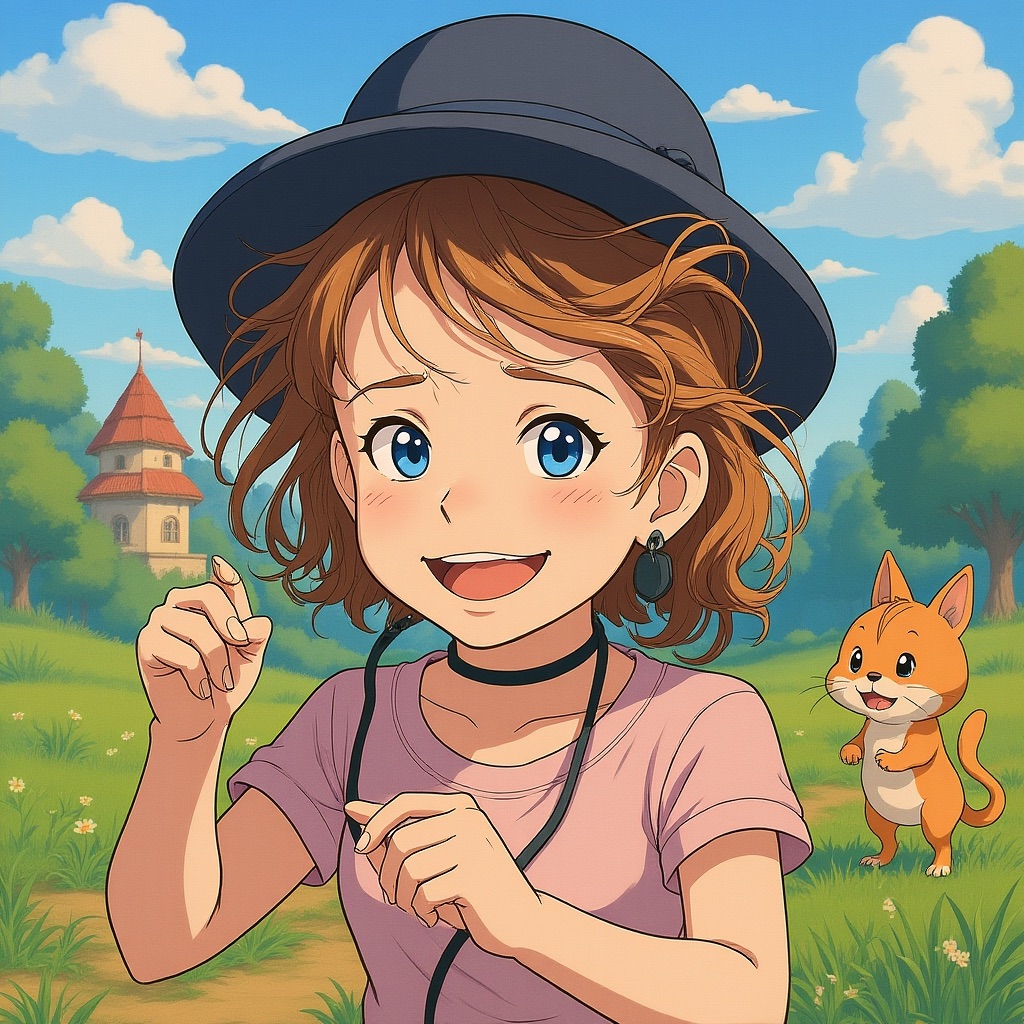} \\
        Input & Preserve & Balanced & Edit
    \end{tabular}

    \vspace{-4pt}
    \caption{\small Input preservation vs.\ instruction adherence, moving from full source preservation to full instruction adherence with a balanced midpoint.}
    \label{fig:editing_results}
\end{wrapfigure}

We inject the preference vector $\omega$ into the model through lightweight conditioning modules trained jointly with LoRA adapters. All conditioning modules are initialized so that their output is near-zero at the start of training, ensuring that preference conditioning is introduced gradually without disturbing the pretrained base model.
To demonstrate the versatility of our approach, we apply \methodname{} to three different base models (SD3.5, FluxKontext, LTX2), tailoring the integration to each architecture. 

\vspace{-10pt}
\paragraph{\textbf{Text-to-Image.}} 
We build upon the SD3.5~\cite{esser2024scaling} architecture, which processes image and text tokens through $L=18$ joint transformer blocks, each modulated by AdaLN parameters derived from a shared timestep embedding \texttt{temb}. In our text-to-image configuration, we inject $\omega$ through two complementary pathways: an implicit global signal via the timestep embedding, and a shared residual correction applied to the image stream of all transformer blocks. For stable early RL fine-tuning, the final linear layer of $f_{\mathrm{blk}}$ is initialized with weights drawn from $\mathcal{N}(0,\,10^{-3})$ and zero bias. We refer the reader to Appendix \ref{app:impl_sd3} for a thorough explanation regarding this implementation.
\begin{wrapfigure}{r}{0.5\textwidth}
    \vspace{-\baselineskip}
    \centering
    \setlength{\tabcolsep}{1pt}

    \begin{tabular}{c c c c @{\hskip 4pt} c c c}
        \raisebox{7pt}{\rotatebox{90}{\scriptsize Anime}} &
        \includegraphics[width=0.14\linewidth]{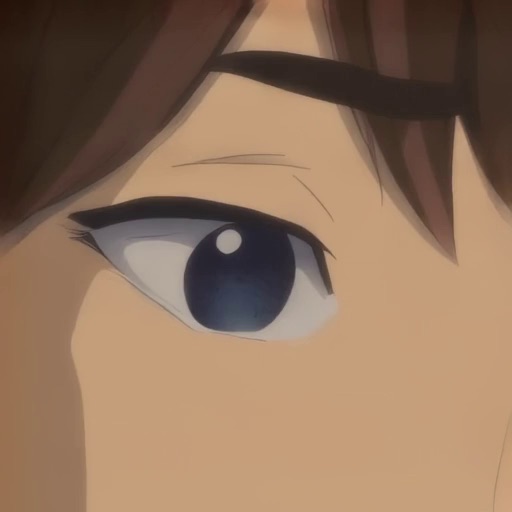} &
        \includegraphics[width=0.14\linewidth]{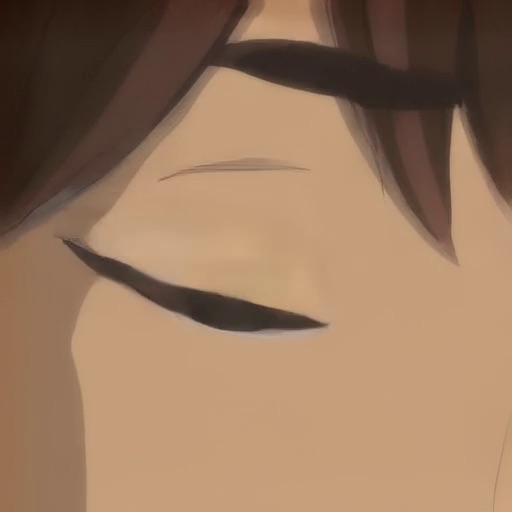} &
        \includegraphics[width=0.14\linewidth]{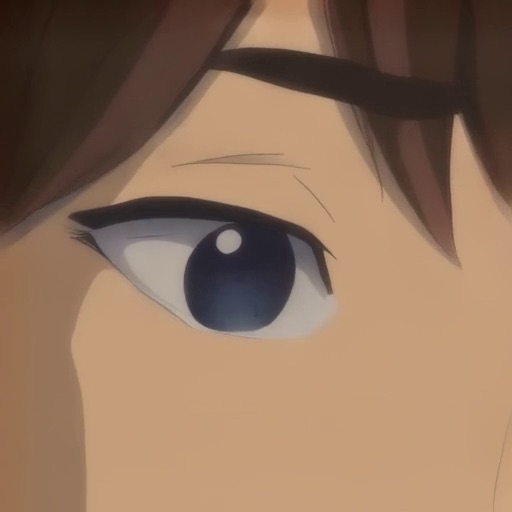} &
        \includegraphics[width=0.14\linewidth]{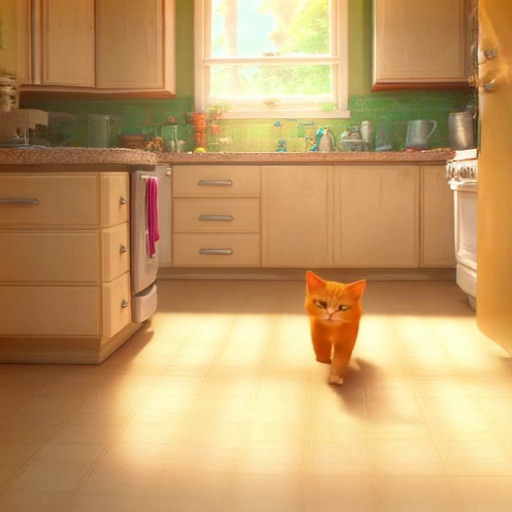} &
        \includegraphics[width=0.14\linewidth]{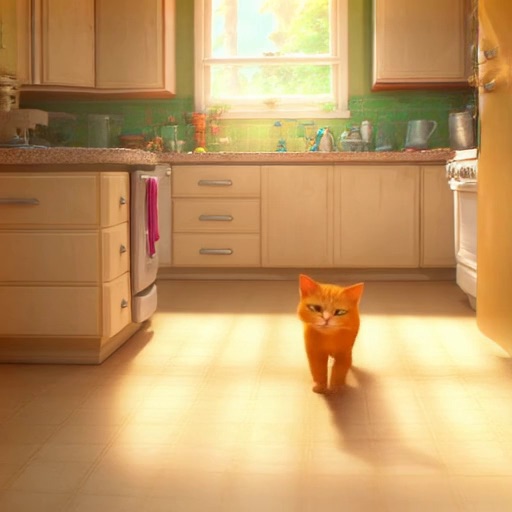} &
        \includegraphics[width=0.14\linewidth]{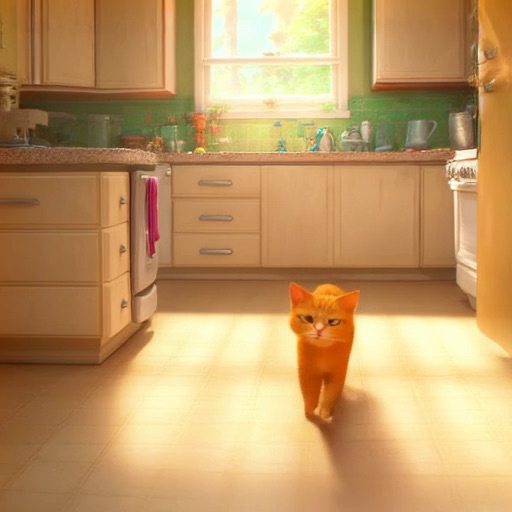} \\[-1pt]
        \raisebox{7pt}{\rotatebox{90}{\scriptsize Balanced}} &
        \includegraphics[width=0.14\linewidth]{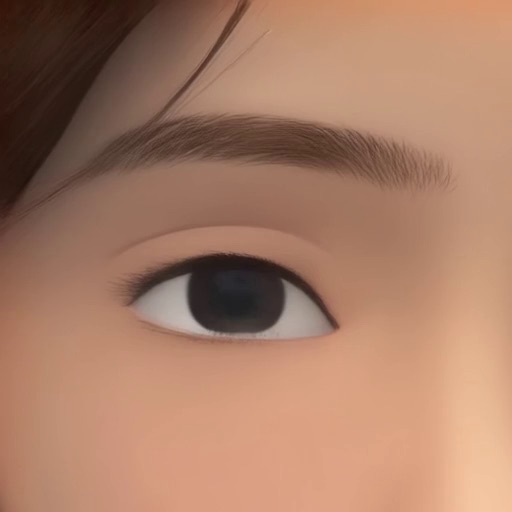} &
        \includegraphics[width=0.14\linewidth]{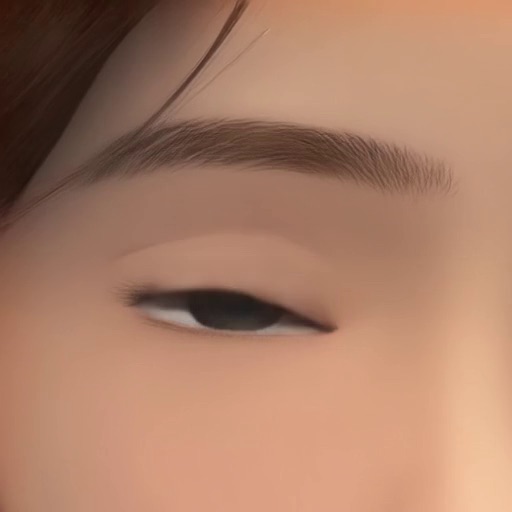} &
        \includegraphics[width=0.14\linewidth]{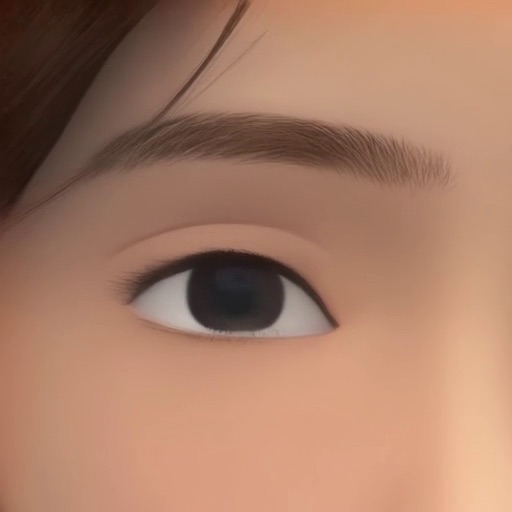} &
        \includegraphics[width=0.14\linewidth]{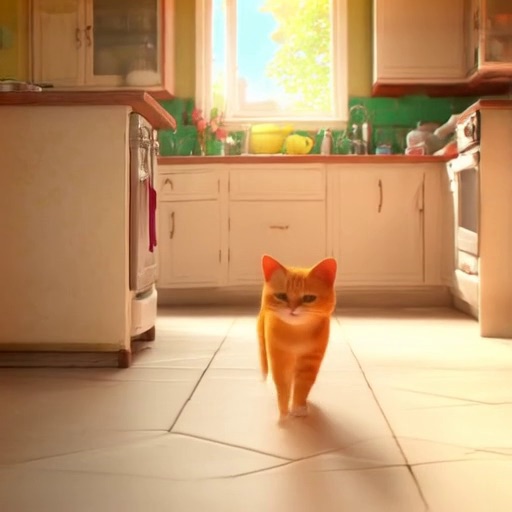} &
        \includegraphics[width=0.14\linewidth]{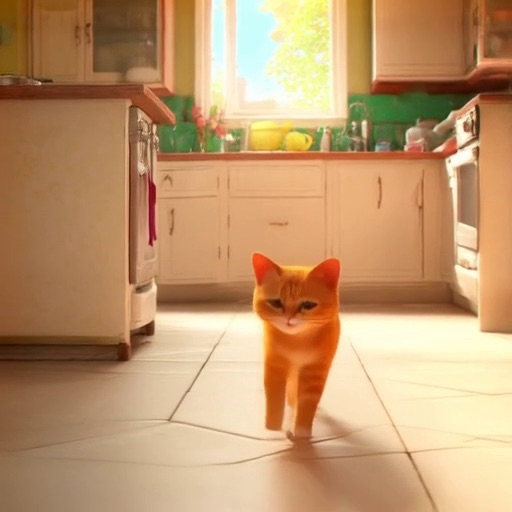} &
        \includegraphics[width=0.14\linewidth]{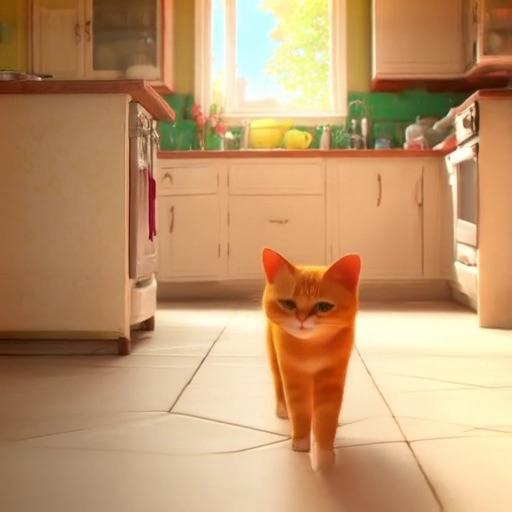} \\[-1pt]
        \raisebox{3pt}{\rotatebox{90}{\scriptsize Realistic}} &
        \includegraphics[width=0.14\linewidth]{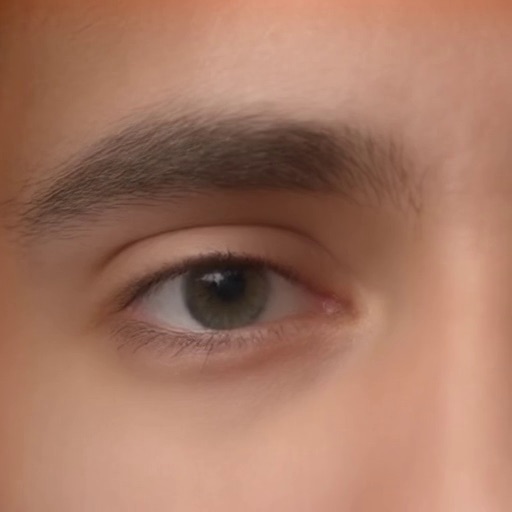} &
        \includegraphics[width=0.14\linewidth]{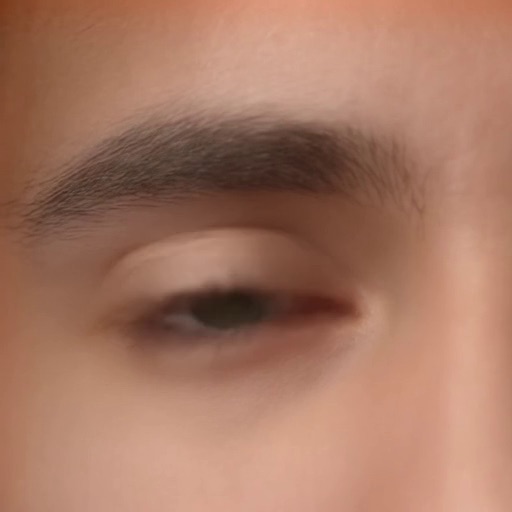} &
        \includegraphics[width=0.14\linewidth]{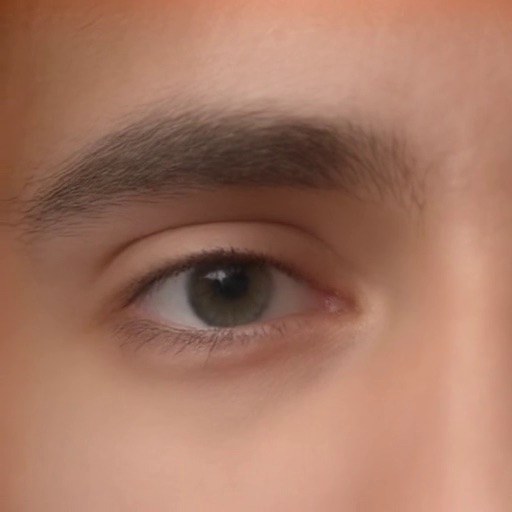} &
        \includegraphics[width=0.14\linewidth]{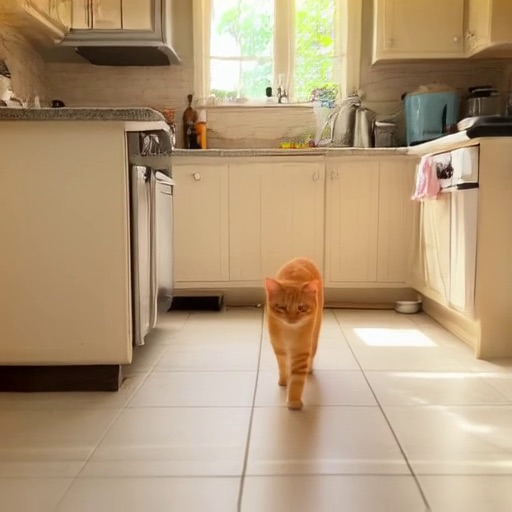} &
        \includegraphics[width=0.14\linewidth]{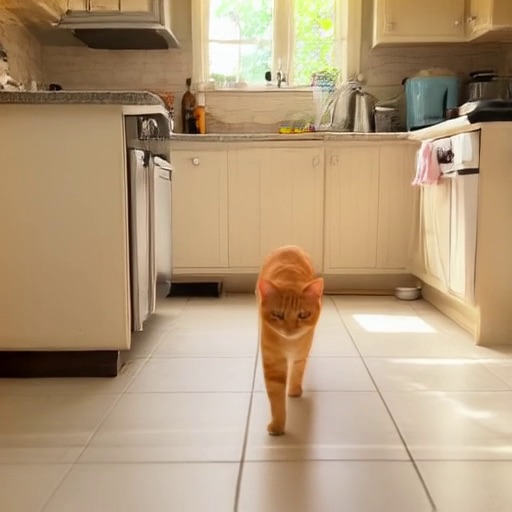} &
        \includegraphics[width=0.14\linewidth]{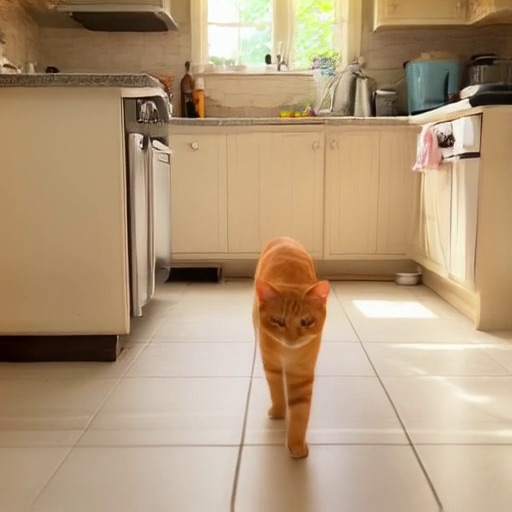} \\[2pt]
        & \multicolumn{3}{c}{\parbox{0.33\linewidth}{\centering\scriptsize \textit{``Extreme closeup of a human}\\\textit{eye blinking slowly.''}}} &
        \multicolumn{3}{c}{\parbox{0.33\linewidth}{\centering\scriptsize \textit{``A orange cat walking towards}\\\textit{the camera across a sunny kitchen floor.''}}} 
    \end{tabular}
    \vspace{-4pt}
    \caption{\small Text-to-video results with LTX2 on the animation-photorealism trade-off. Each block shows three frames for animation, balanced, and photorealistic preferences. A single model controls style at inference time through $\omega$.}
    \label{fig:ltx2_qualitative}
\end{wrapfigure}

\vspace{-10pt}
\paragraph{\textbf{Image-to-Image.}}
For image editing and personalization, we build upon FluxKontext~\cite{labs2025flux1kontextflowmatching}. Following the ablation studies by Prihar et al. \cite{parihar2025kontinuous}, we directly modulate the AdaLN scale and shift of the context (text) stream within each of the 19 dual-stream transformer blocks.  We first encode $\omega$ via sinusoidal embeddings and project it alongside the pooled text embedding $\bar{\mathbf{e}}_{\mathrm{text}}$:
$\mathrm{enc}(\omega) = \left[\sin(\pi\omega);\, \cos(\pi\omega)\right] \in \mathbb{R}^{2M},$ and then:
$(\Delta\boldsymbol{\gamma}_\omega,\, \Delta\boldsymbol{\beta}_\omega)
    = f_{\mathrm{ctx}}\!\left(
        [\mathbf{W}_{\mathrm{enc}}\,\mathrm{enc}(\omega);\, 
        \bar{\mathbf{e}}_{\mathrm{text}}]
    \right),$
where $f_{\mathrm{ctx}}$ is a three-layer MLP with hidden dimension $2048$. The resulting corrections are added residually to the existing AdaLN parameters of the context stream:
$\boldsymbol{\gamma} \leftarrow \boldsymbol{\gamma} + \Delta\boldsymbol{\gamma}_\omega, \quad \boldsymbol{\beta} \leftarrow \boldsymbol{\beta} + \Delta\boldsymbol{\beta}_\omega.$
Unlike SD3.5, where our final text-to-image configuration modulates the image stream via a shared residual signal reused across blocks, here we target the text stream, as FluxKontext's dual-stream architecture routes conditioning primarily through the context pathway.

\vspace{-5pt}
\paragraph{Text-to-Video}
For this task, we adopt the LTX-2 \cite{hacohen2026ltx2efficientjointaudiovisual} model and condition the policy through the same shared block-residual mechanism used for SD3.5 (Equation~\ref{eq:blk_residual}), targeting the video stream and using the LTX-2 inner dimension $d=3840$. The projector $f_{\mathrm{blk}}$ (sinusoidal PE followed by a four-layer MLP) is unchanged apart from its output width. 

\vspace{-5pt}
\subsection{Inference-Time Control}
\label{subsec:inference}
\vspace{-5pt}
At inference time, the user specifies the trade-off weights $\omega\in\Delta^M$ and the model generates samples through its standard denoising process, w.r.t. the user-preferences. No retraining, interpolation, or per-step gradient guidance is required. Varying $\omega$ continuously navigates the learned reward trade-off surface, providing a slider interface for multi-reward control.

\vspace{-5pt}
\subsection{Runtime and Memory Analysis}
\vspace{-5pt}
The main cost remains the shared transformer forward/backward passes, while the per-objective loss adds only an $O(R)$ scalar mixing steps. The preference-vector conditioning adds $\approx 7$M parameters, $<0.5\%$ of SD3.5-medium. Empirically, on $6{\times}$A100-80GB, the fixed-weight baseline takes $17.3$ minute/epoch and the conditional run takes $+0.7$ minute/epoch when $M=2$ and $+1.3$ minute/epoch $M=3$, highly negligible compared to multiple training procedures of the fixed-weight baseline.
\vspace{-5pt}
\section{Experiments}
\label{sec:experiments}
\vspace{-5pt}
\begin{wrapfigure}[32]{r}{0.5\textwidth}
\vspace{-25pt}
\vspace{-\baselineskip}
\centering
\setlength{\tabcolsep}{0.5pt}
\renewcommand{\arraystretch}{0.9}

\newcommand{\figW}{0.8\linewidth}
\newcommand{\imgw}{0.16\linewidth}

\includegraphics[width=\figW]{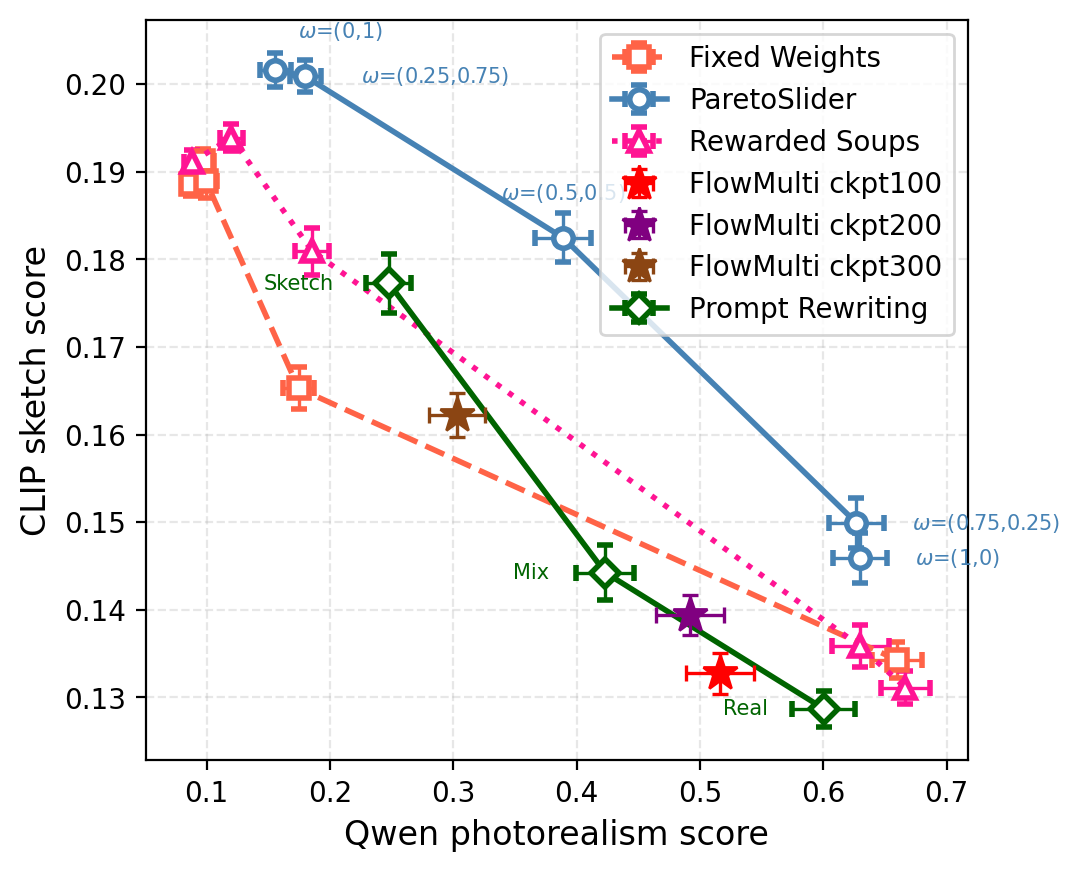}

\begin{tabular}{@{}c c c c c c@{}}
    \raisebox{10pt}{\rotatebox{90}{\scriptsize \color{blue}Ours}} &
    \includegraphics[width=\imgw]{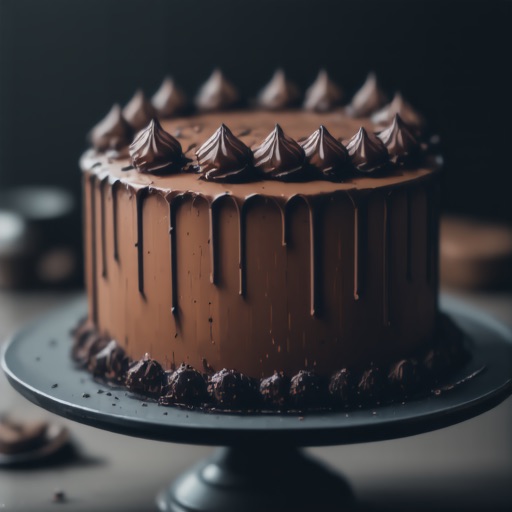} &
    \includegraphics[width=\imgw]{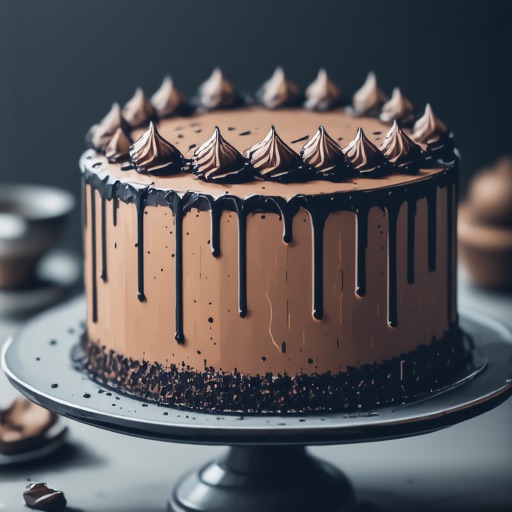} &
    \includegraphics[width=\imgw]{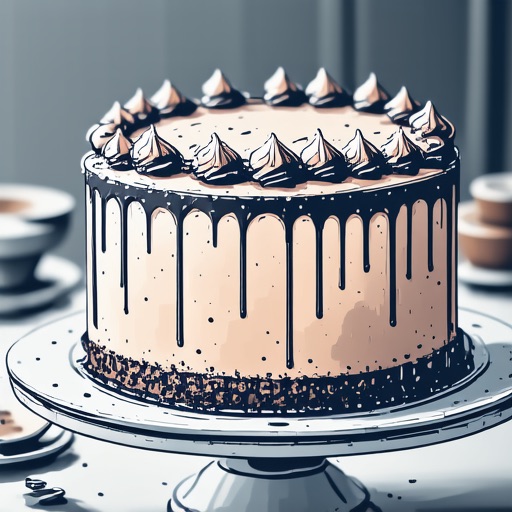} &
    \includegraphics[width=\imgw]{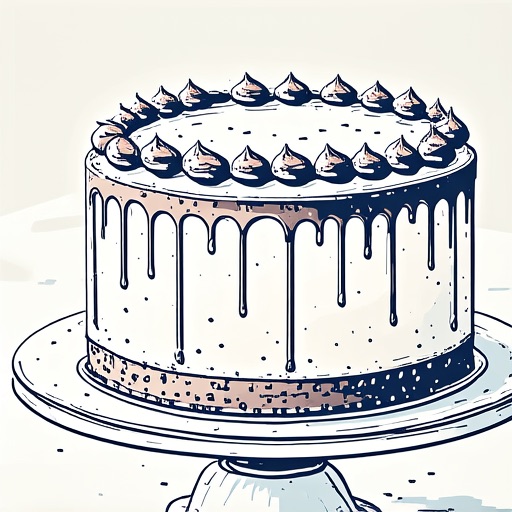} &
    \includegraphics[width=\imgw]{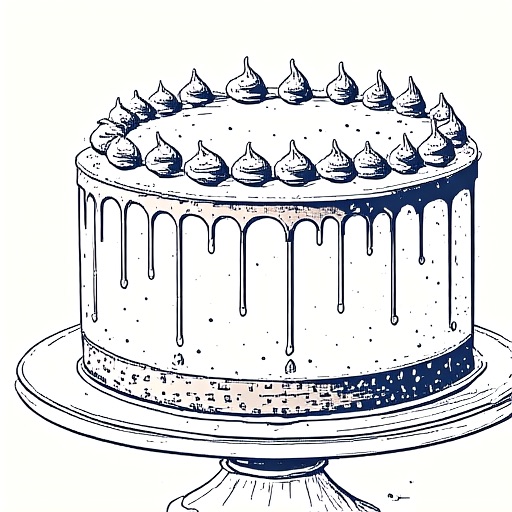} \\[-1pt]

    \raisebox{6pt}{\rotatebox{90}{\scriptsize \color{orange}\makecell{Fixed\\Weight}}} &
    \includegraphics[width=\imgw]{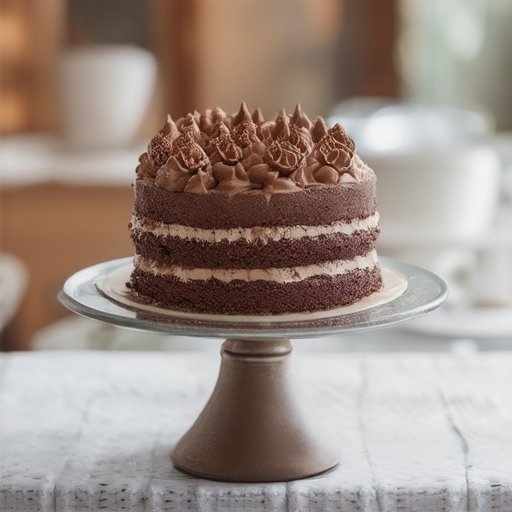} &
    \includegraphics[width=\imgw]{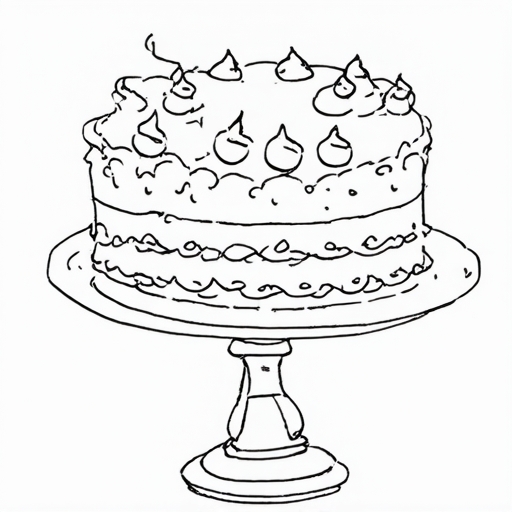} &
    \includegraphics[width=\imgw]{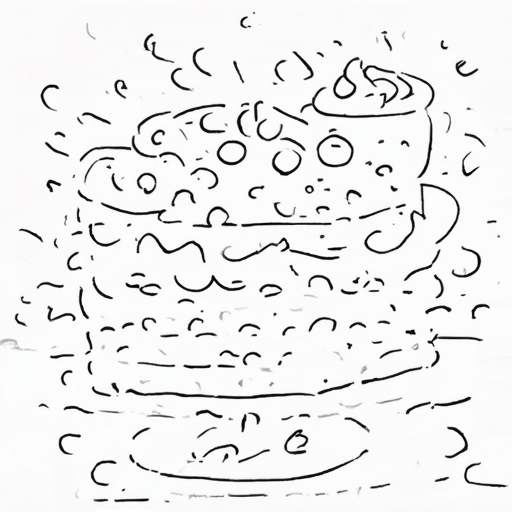} &
    \includegraphics[width=\imgw]{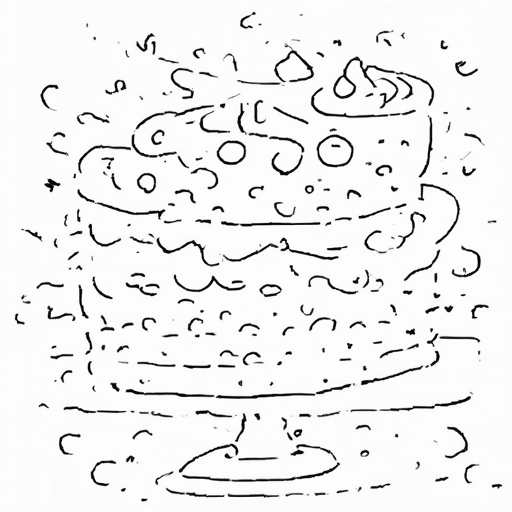} &
    \includegraphics[width=\imgw]{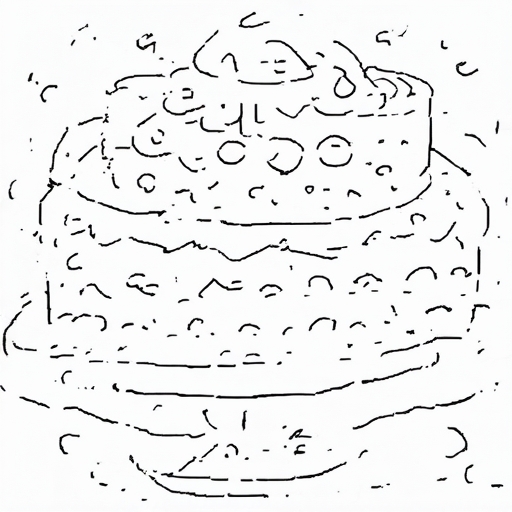} \\[-1pt]

    \raisebox{1pt}{\rotatebox{90}{\scriptsize \color{magenta}\makecell{Rewarded\\Soup}}} &
    \includegraphics[width=\imgw]{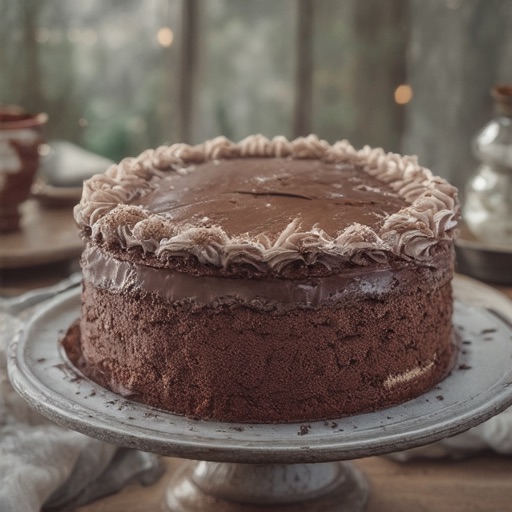} &
    \includegraphics[width=\imgw]{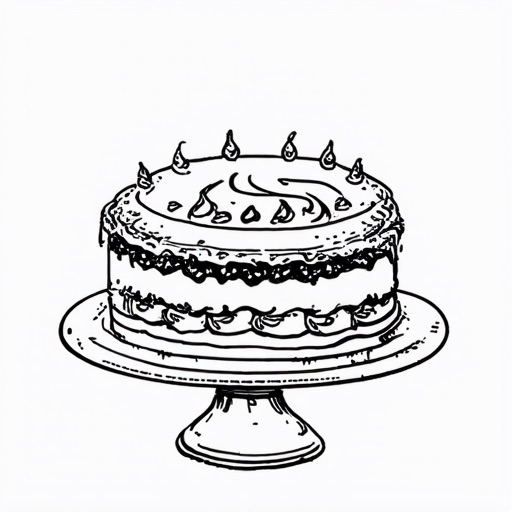} &
    \includegraphics[width=\imgw]{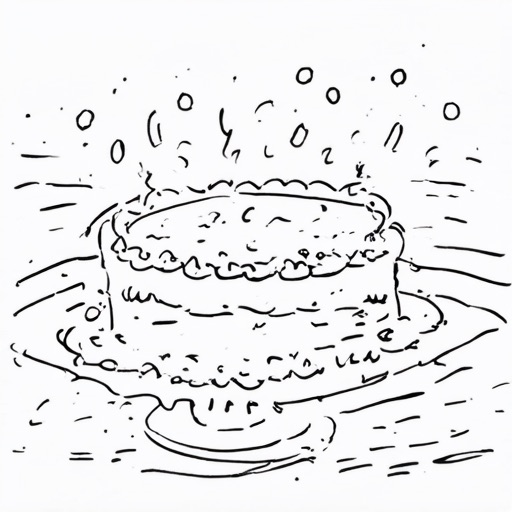} &
    \includegraphics[width=\imgw]{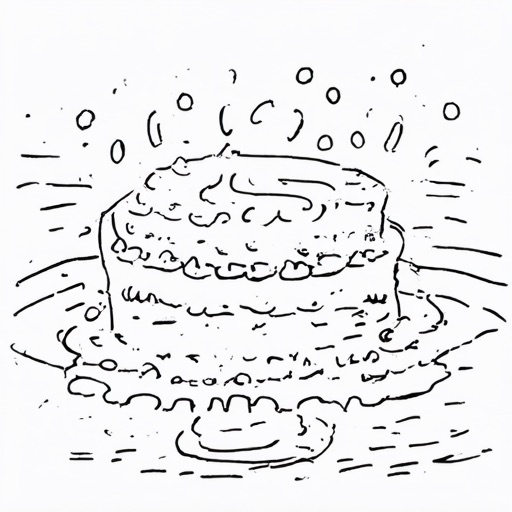} &
    \includegraphics[width=\imgw]{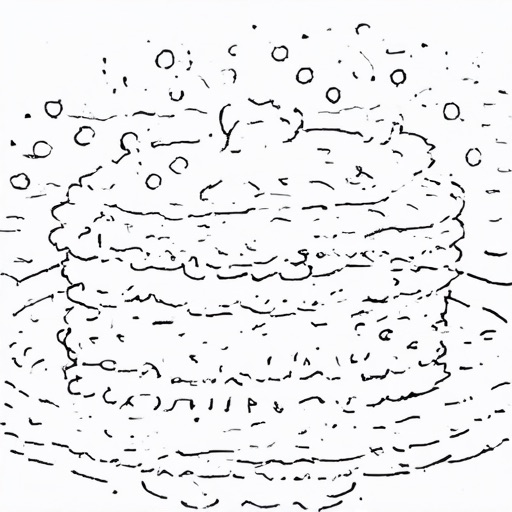} \\[-1pt]

    & \scriptsize Realistic
    & \multicolumn{3}{c}{\scriptsize $\xleftarrow{\hspace{0.05cm}}\;
    \omega=(\omega_{\mathrm{real}},\omega_{\mathrm{sketch}})
    \;\xrightarrow{\hspace{0.05cm}}$}
    & \scriptsize Sketch
\end{tabular}

\setlength{\tabcolsep}{0.15pt}
\newcommand{\smallimgw}{0.16\linewidth}

\begin{tabular}{@{}c c c c @{\hspace{2pt}} c c c c@{}}
    \raisebox{0pt}{\rotatebox{90}{\scriptsize FlowMulti}} &
    \includegraphics[width=\smallimgw]{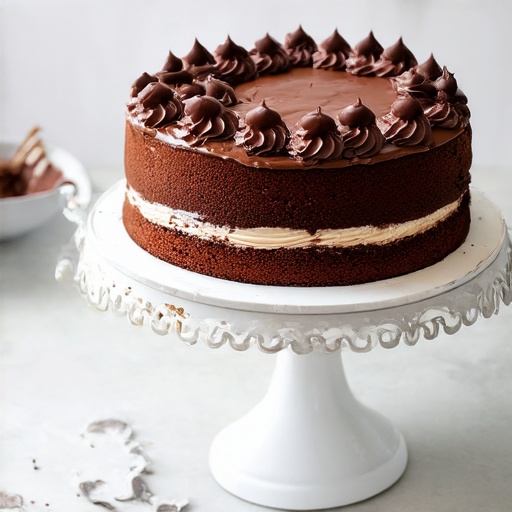} &
    \includegraphics[width=\smallimgw]{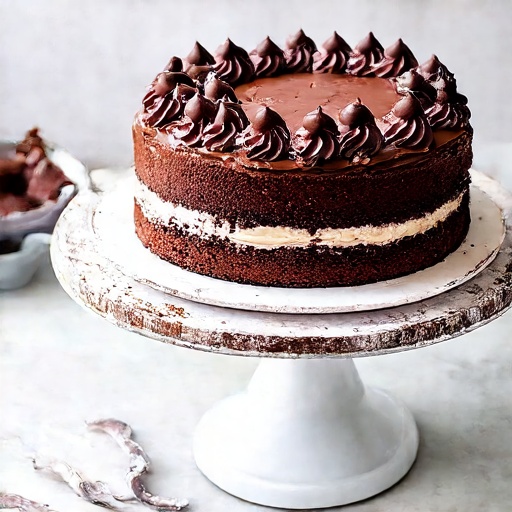} &
    \includegraphics[width=\smallimgw]{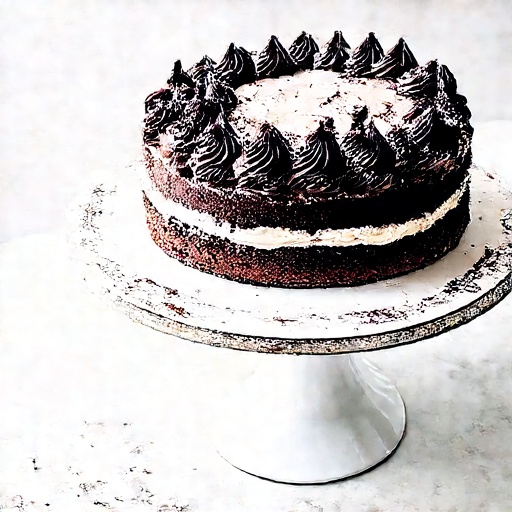} &
    \raisebox{0pt}{\rotatebox{90}{\scriptsize \color{darkgreen}Prompting}} &
    \includegraphics[width=\smallimgw]{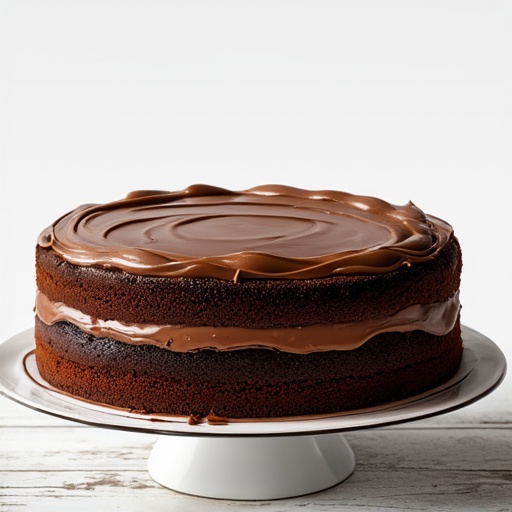} &
    \includegraphics[width=\smallimgw]{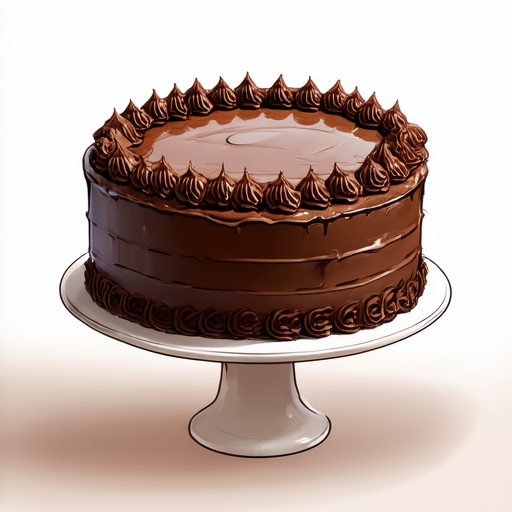} &
    \includegraphics[width=\smallimgw]{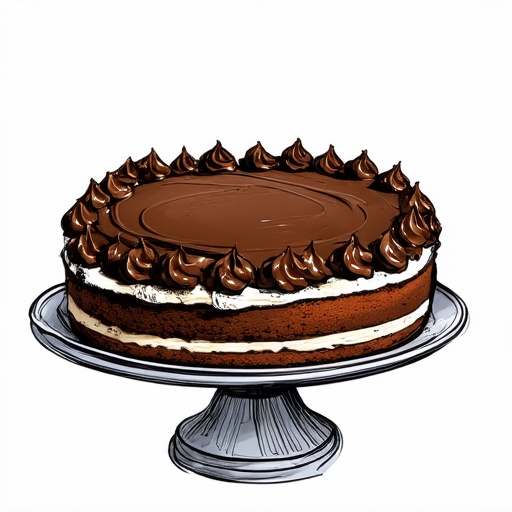} \\[-1pt]
    &
    \scriptsize \color{red}Epoch 100 &
    \scriptsize \color{purple}Epoch 200 &
    \scriptsize \color{brown}Epoch 300 &
    &
    \scriptsize Realistic &
    \scriptsize Mix &
    \scriptsize Sketch
\end{tabular}

\vspace{-4pt}
\caption{\small Top: Controllability comparison. Bottom: qualitative transitions. \methodname{} produces smooth, well-ordered transitions from photorealism to sketch as $\omega$ varies. Fixed Weights requires a separate model per point, RewardedSoups \cite{rame2023rewardedsoupsparetooptimalalignment} interpolates between two separately trained LoRAs, FlowMulti is a MORL framework that produces a static point on the Pareto front, and Prompting offers only coarse operating points. All methods are dominated by \methodname{}.}
\label{fig:pareto_and_baselines}
\end{wrapfigure}  

Our experiments are designed to answer three complementary questions. 
(1) The necessity of explicit preference conditioning: we investigate whether existing control mechanisms -- such as classifier-free guidance scales or prompt engineering -- can produce controllable trade-offs by comparing \methodname{} against various training and inference-time baselines (\S\ref{subsec:comp}).
(2) Controllability: we demonstrate that \methodname{} allows a smooth inference time control via qualitative and quantitative comparisons (\S\ref{subsec:quant_results}). In addition, we empirically evaluate coverage and convergence towards the supported Pareto front using the hypervolume (HV) indicator, where our method consistently dominates.
(3) The impact of core design choices: through targeted ablations, we isolate the specific contributions of our preference-conditioning architecture and different loss aggregation methods. (\S\ref{subsec:ablations}).

\vspace{-10pt}
\subsection{Qualitative Results}
\label{subsec:qual_results}
\vspace{-5pt}
We begin with qualitative results across tasks, illustrating how varying the preference vector $\omega$ produces coherent and continuous transitions along the learned reward trade-off surface. 
Figure~\ref{fig:t2i_grid} demonstrates our method in the text-to-image domain, showing smooth preference-controlled transitions between photorealism and several target styles, including flat vector art, watercolor, anime, animated scene, and sketch. The figure also presents how our method extends beyond two rewards and interpolates seamlessly between three  or more distinct styles (bottom right triplet and additional results in Figure \ref{fig:t2i_results_supp}).
Figure~\ref{fig:editing_results} applies our approach in the I2I domain to balance instruction adherence with input image preservation. For instance, the warrior transformation in the third row presents high preservation levels, fine-grained identity details like tiny freckles are faithfully maintained. However, as adherence to the prompt increases, the identity slightly shifts and these subtle details are eventually lost to the stronger stylistic edit.
Finally, Figure~\ref{fig:ltx2_qualitative} shows our method in the T2V domain on the LTX2 model, smoothly navigating the trade-off between photorealism and animation. Additional Results are provided in Appendix \S\ref{app:experiments}.
\begin{figure}[t!]
\vspace{-5pt}
    \centering
    \scriptsize
    \setlength{\tabcolsep}{0.5pt}
    \renewcommand{\arraystretch}{0.99}

    \begin{minipage}[t]{0.3\textwidth}
        \vspace{0pt}
        \centering
        \includegraphics[width=\linewidth]{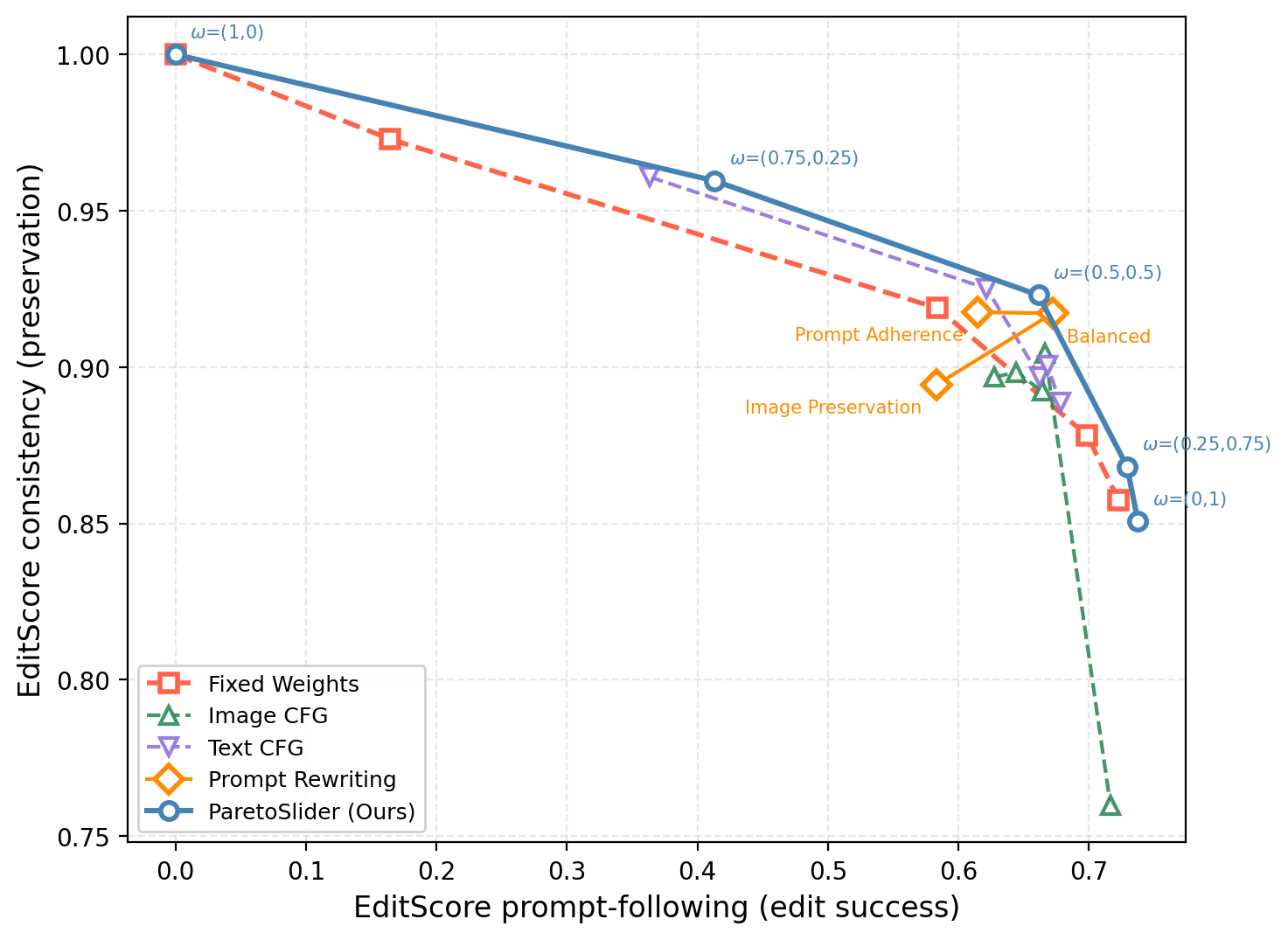}
    \end{minipage}
    \begin{minipage}[t]{0.65\textwidth}
        \vspace{0pt}
        \centering
        \newcommand{\editimgw}{0.135\linewidth}

        \begin{tabular}{@{}c c c c c c c c c@{}}
            &
            \multicolumn{8}{c}{%
                \scriptsize\textit{``Change this portrait to a 3D rendered Disney Pixar scene''}
            } \\

            \multirow{2}{*}{%
                \makecell[c]{%
                    \includegraphics[width=0.13\linewidth]{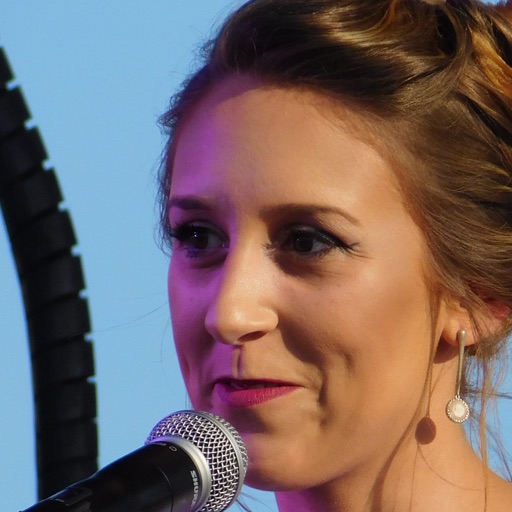}\\[-1pt]
                    \scriptsize Input
                }
            }
            &
            \raisebox{11pt}{\rotatebox{90}{\scriptsize \color{blue}Ours}} &
            \includegraphics[width=\editimgw]{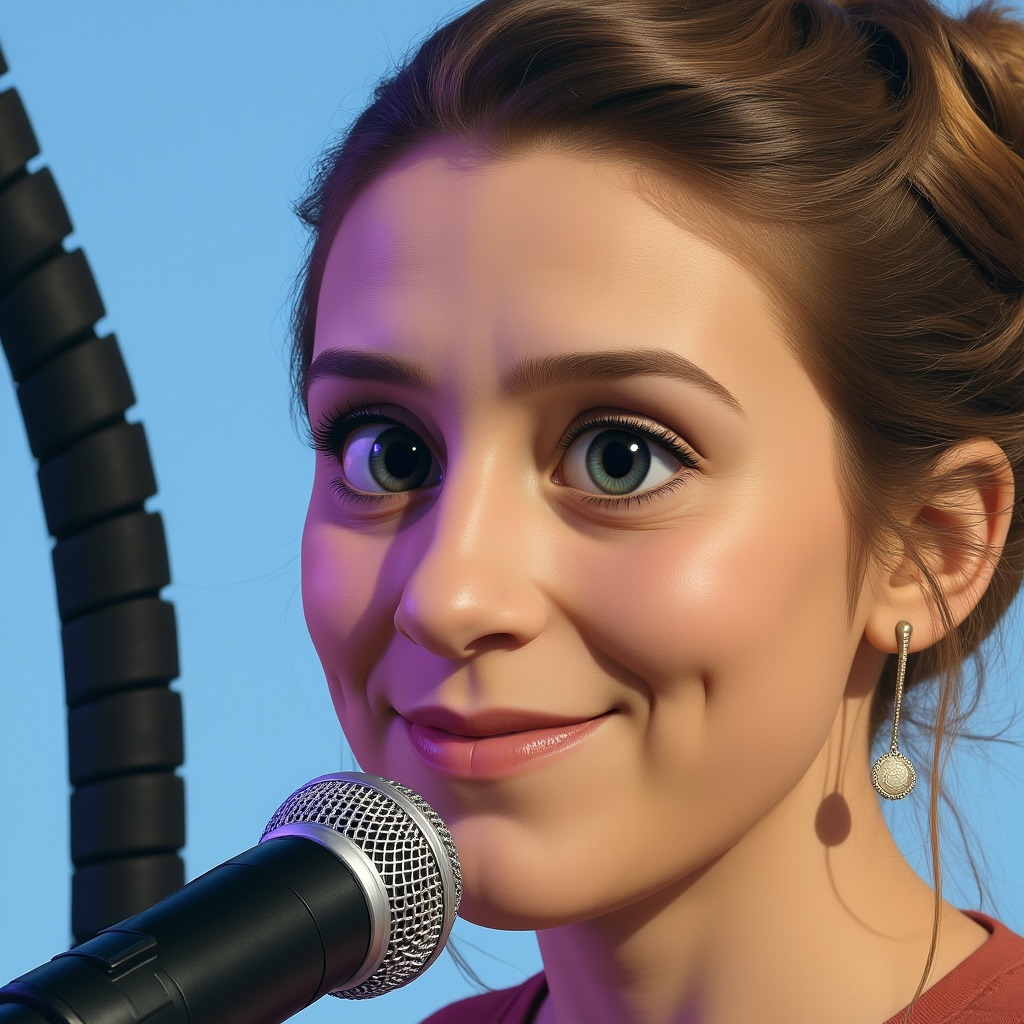} &
            \includegraphics[width=\editimgw]{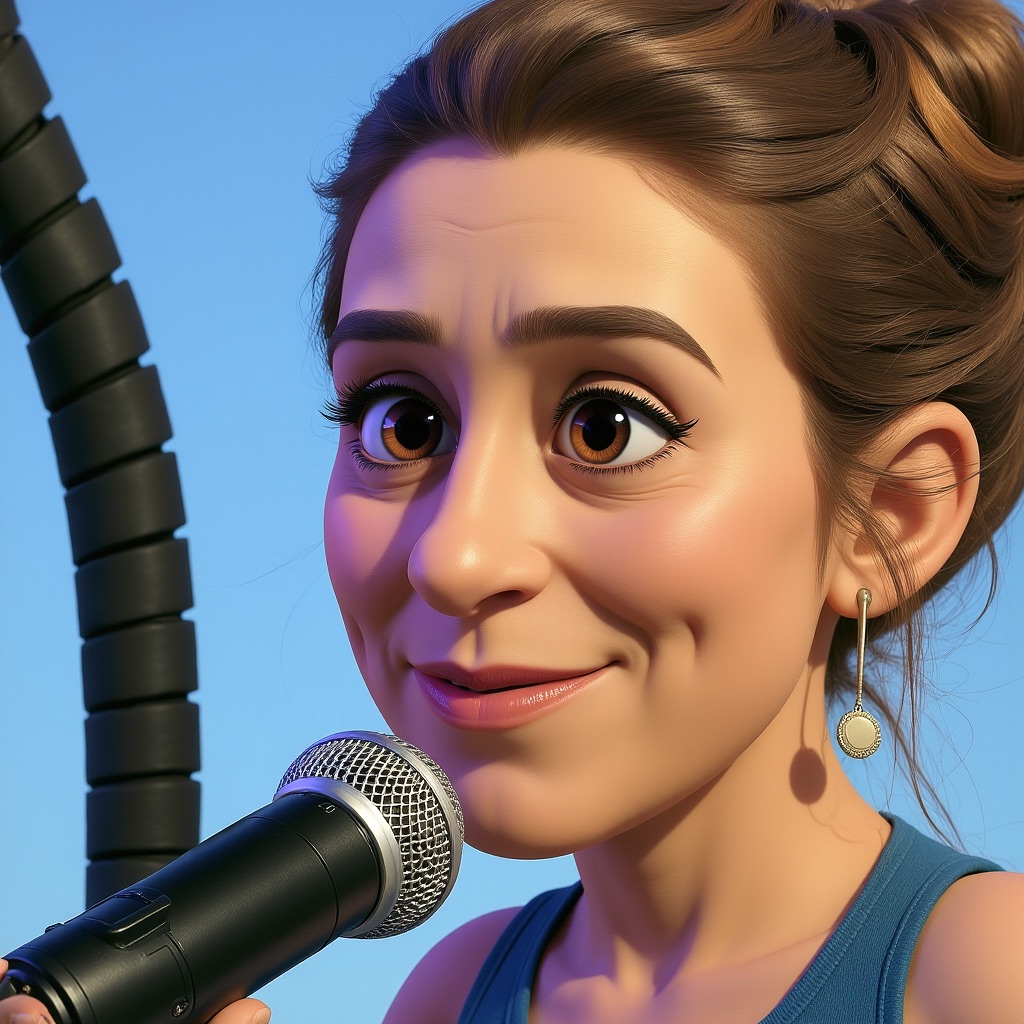} &
            \includegraphics[width=\editimgw]{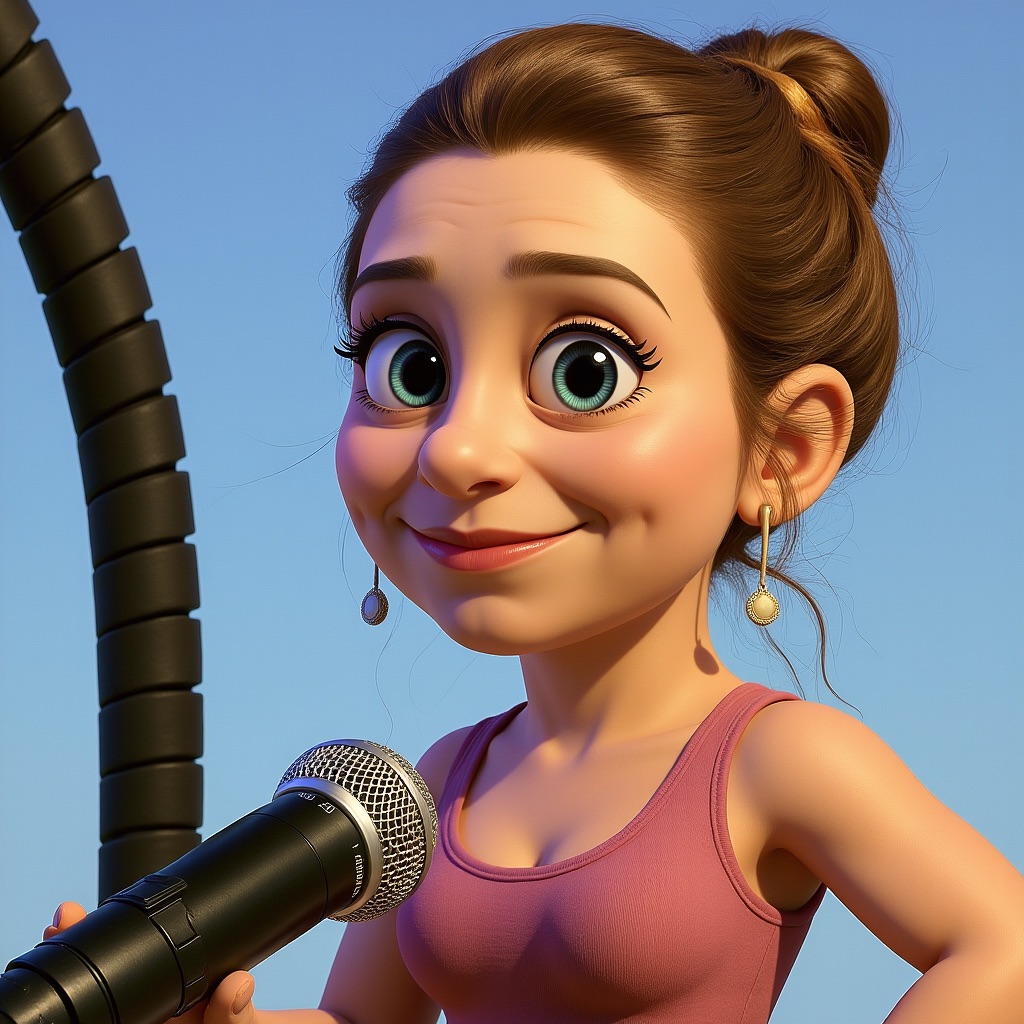} &
            \raisebox{2pt}{\rotatebox{90}{\scriptsize \color{darkgreen}ImageCFG}} &
            \includegraphics[width=\editimgw]{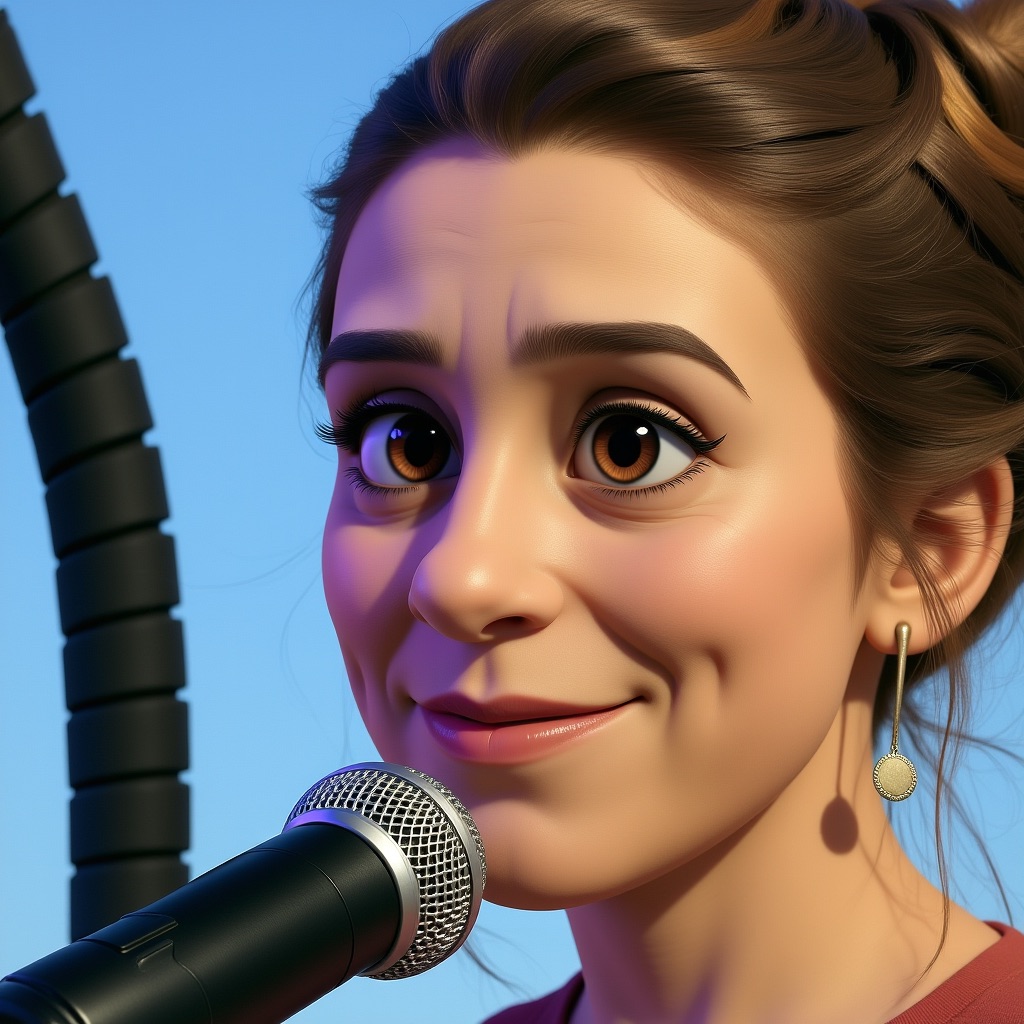} &
            \includegraphics[width=\editimgw]{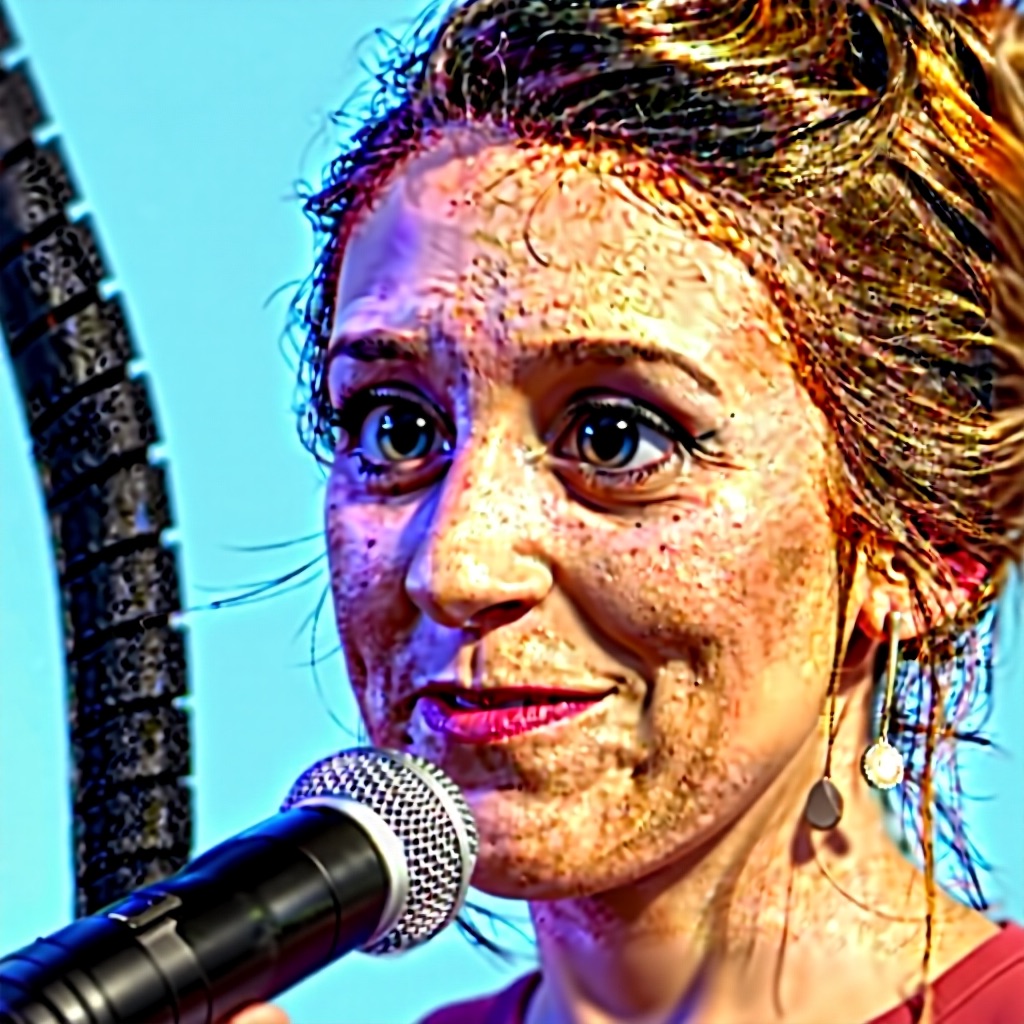} &
            \includegraphics[width=\editimgw]{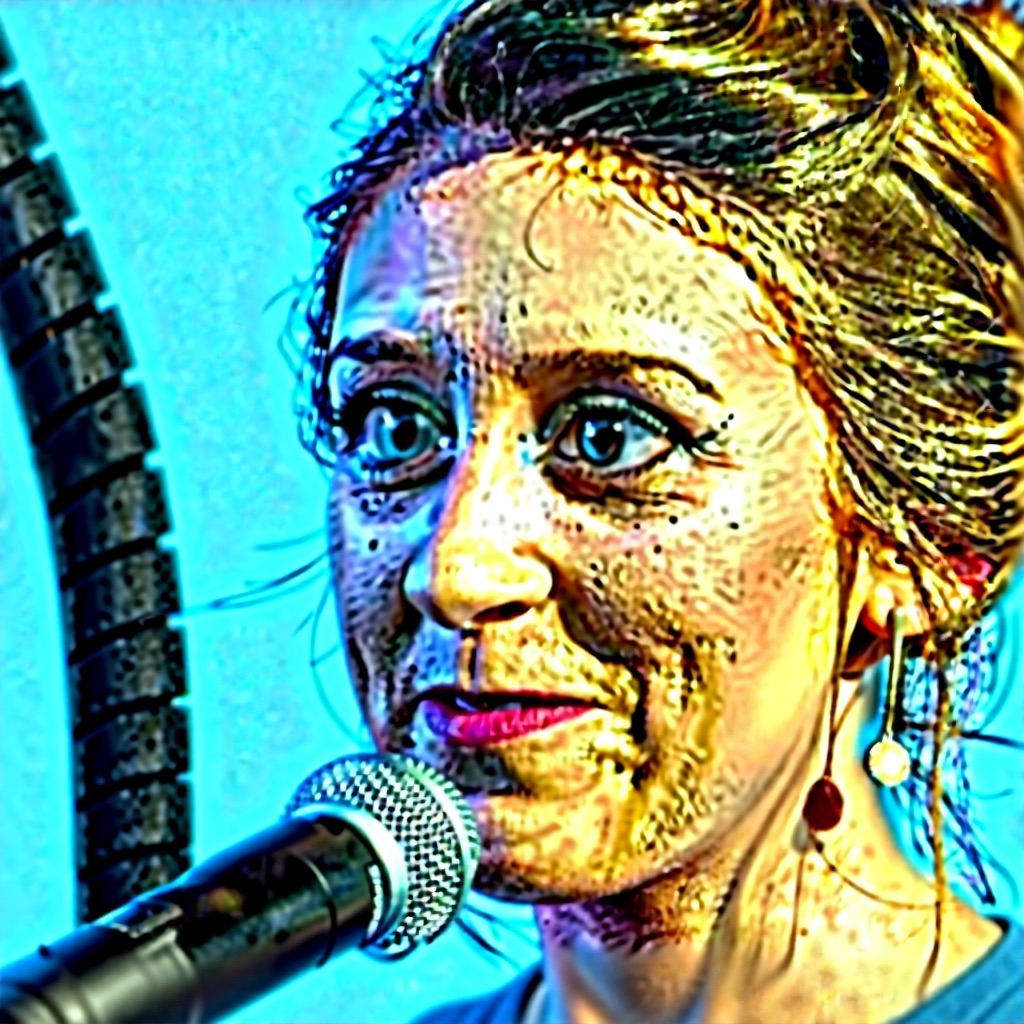} \\[-1pt]

            &
            \raisebox{0pt}{\rotatebox{90}{\scriptsize \color{orange}FixedWeight}} &
            \includegraphics[width=\editimgw]{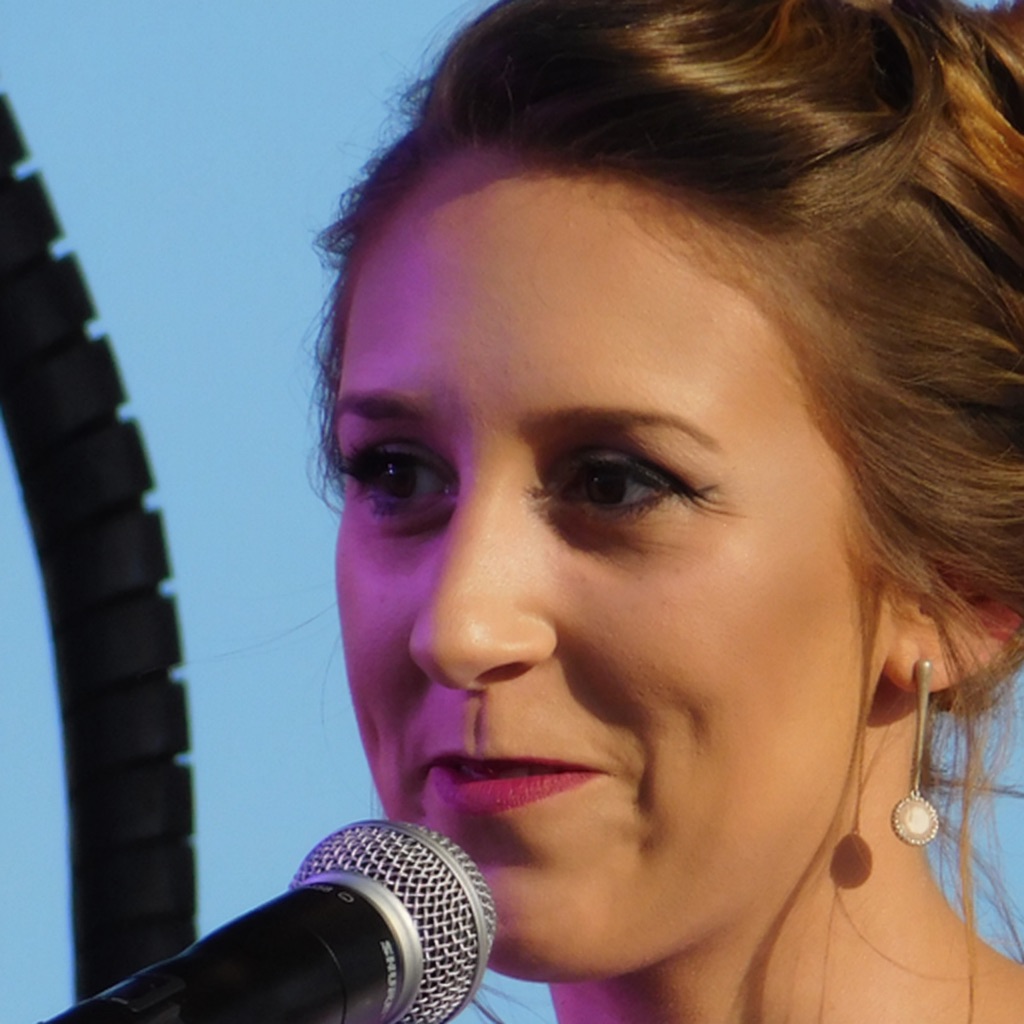} &
            \includegraphics[width=\editimgw]{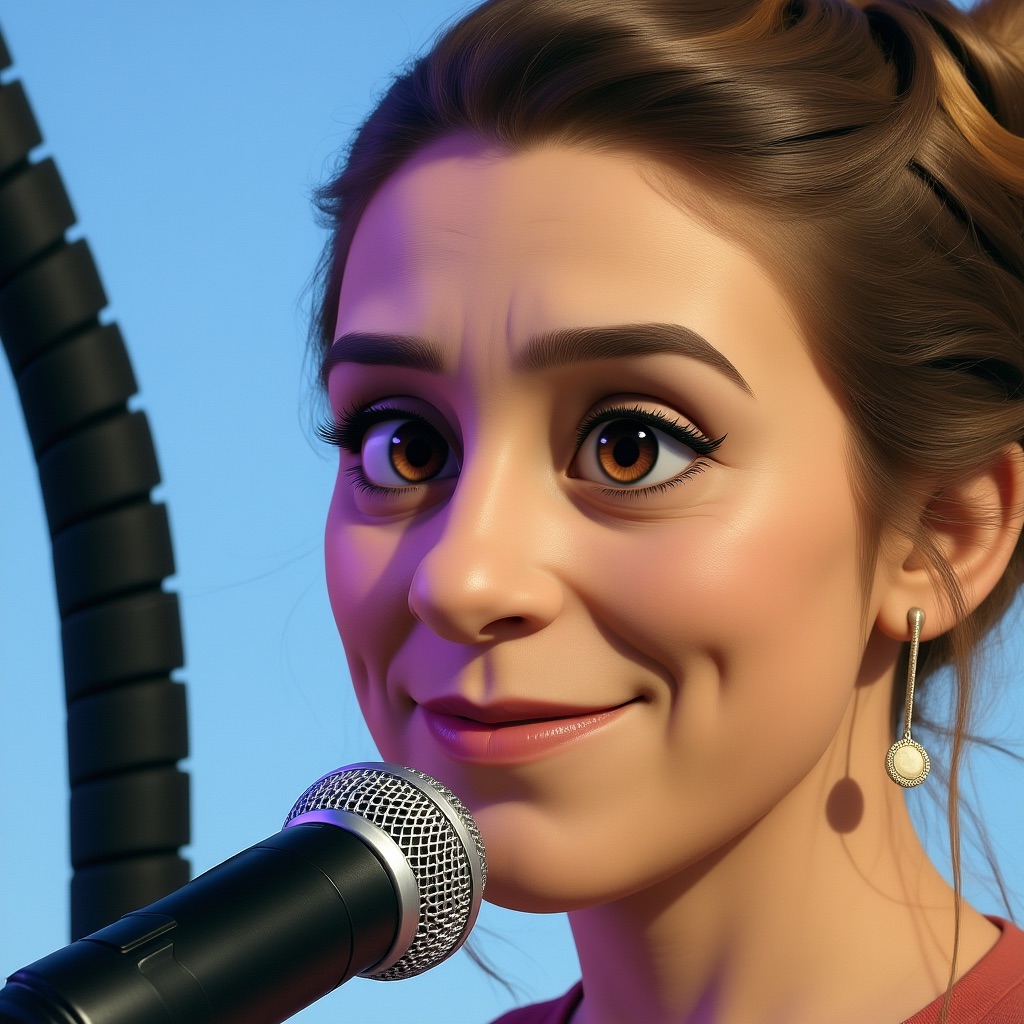} &
            \includegraphics[width=\editimgw]{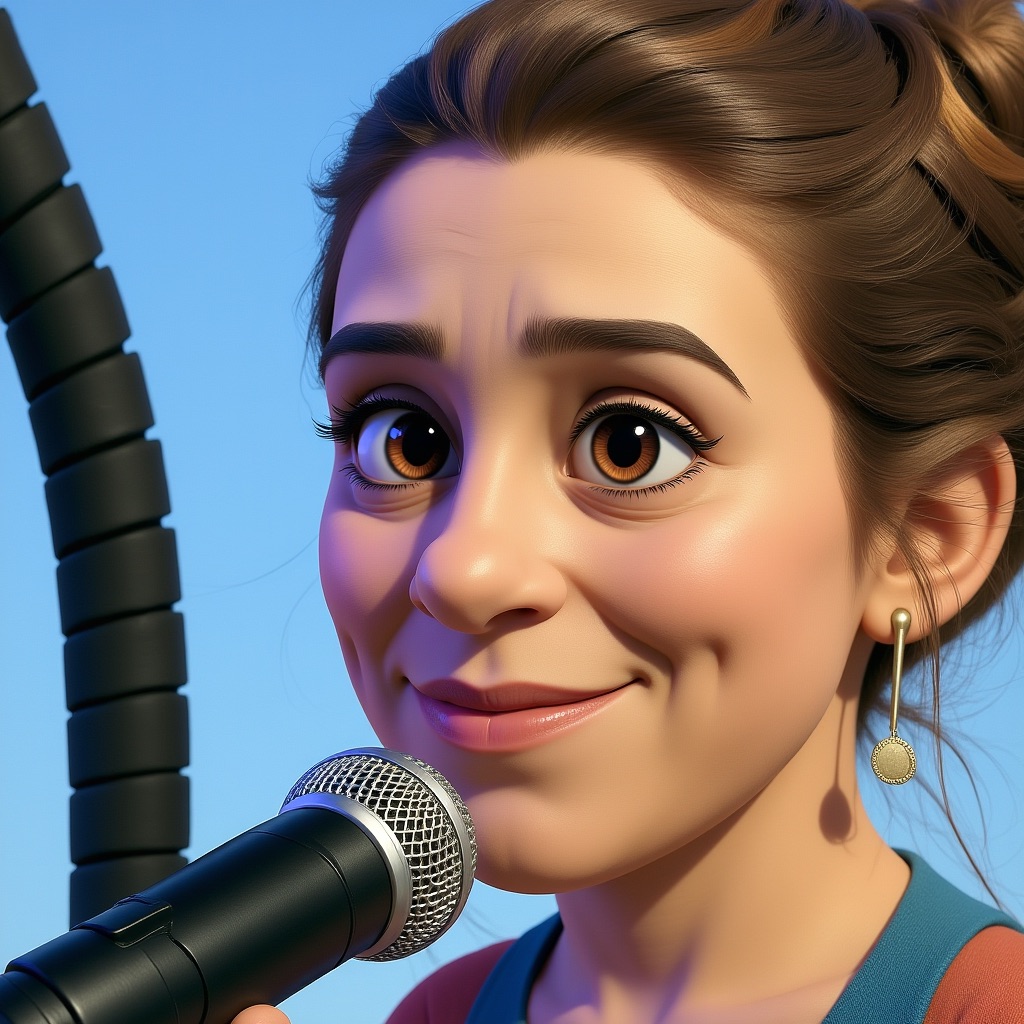} &
            \raisebox{5pt}{\rotatebox{90}{\scriptsize \color{purple}TextCFG}} &
            \includegraphics[width=\editimgw]{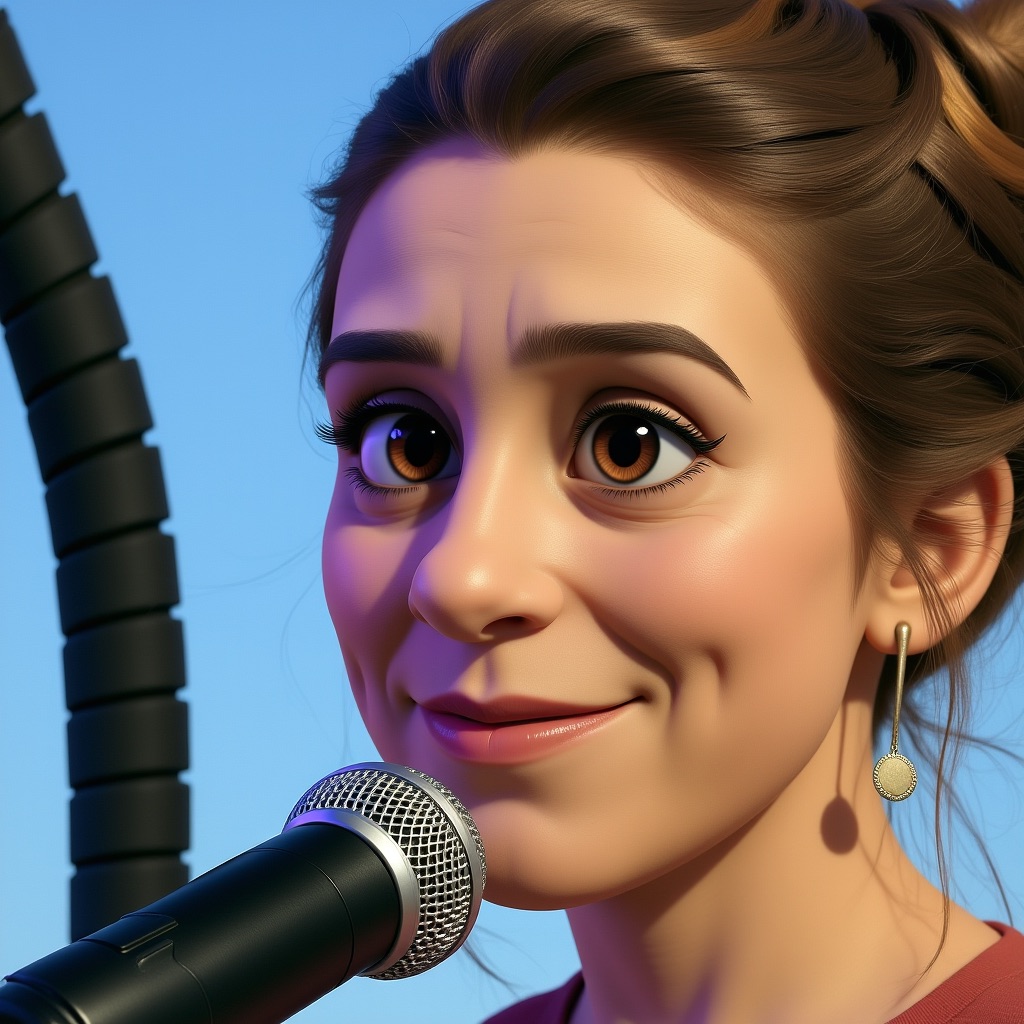} &
            \includegraphics[width=\editimgw]{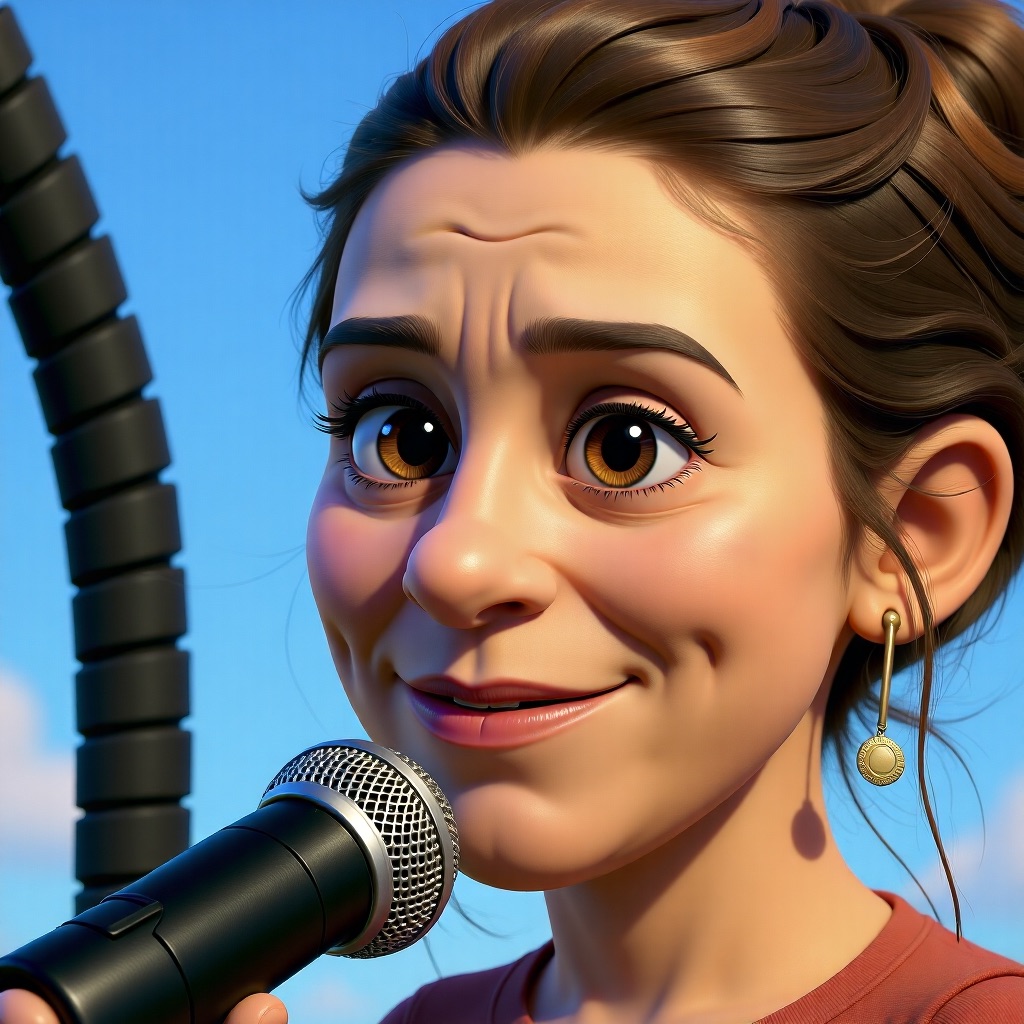} &
            \includegraphics[width=\editimgw]{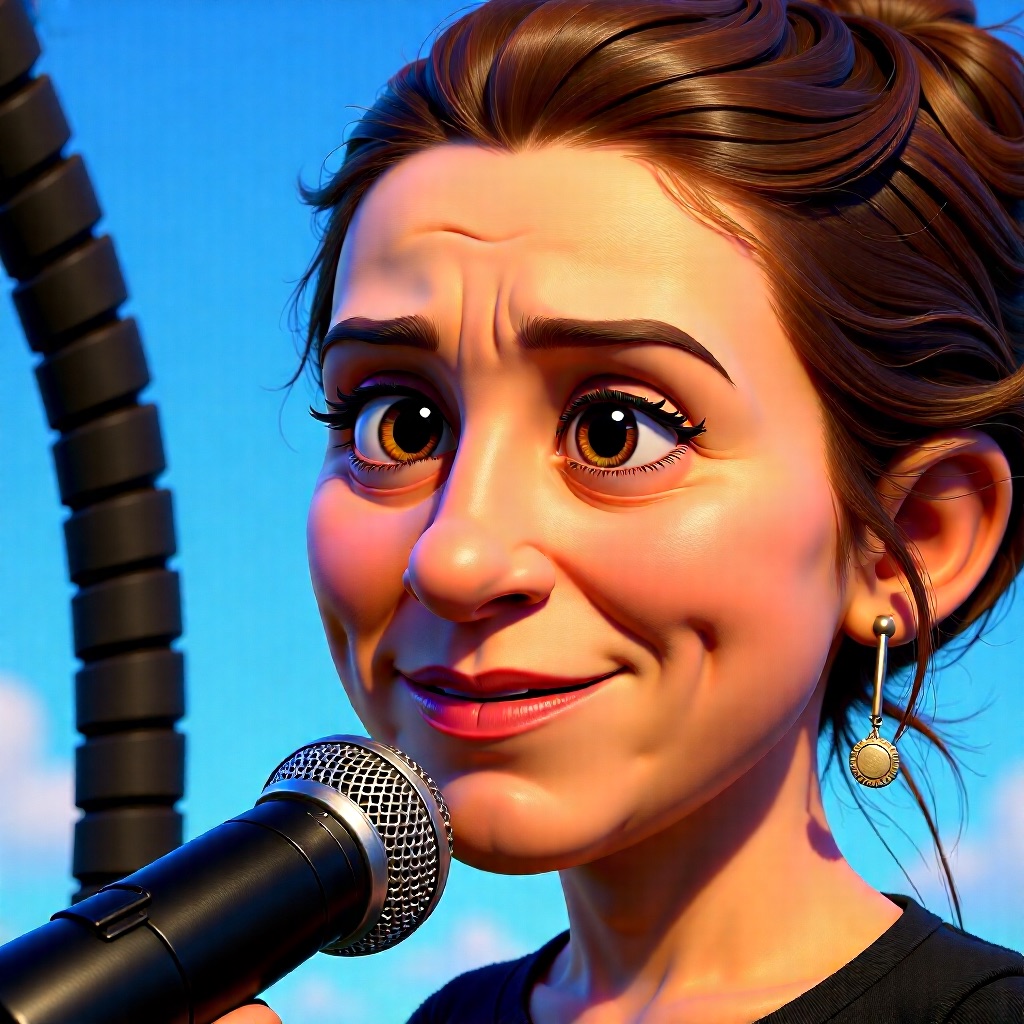} \\[-1pt]

            &
            &
            \scriptsize Preserve & \scriptsize Balanced & \scriptsize Edit &
            &
            \scriptsize Preserve & \scriptsize Balanced & \scriptsize Edit
        \end{tabular}
    \end{minipage}

    \vspace{-3pt}
    \caption{\small \textit{Left:} Controllability comparison. \textit{Right:} qualitative outputs as preferences shift from preservation to editing. ParetoSlider provides smooth control, while FixedWeights requires separate models and CFG baselines show artifact-prone trade-offs.}
    \label{fig:editing_comparison}
    \vspace{-5pt}
\end{figure}
\begin{figure}[t!]
    \centering
    \small
    \setlength{\tabcolsep}{0.002\textwidth}

    \begin{minipage}[t]{0.35\textwidth}
        \vspace{0pt}
        \centering
        \includegraphics[width=\linewidth]{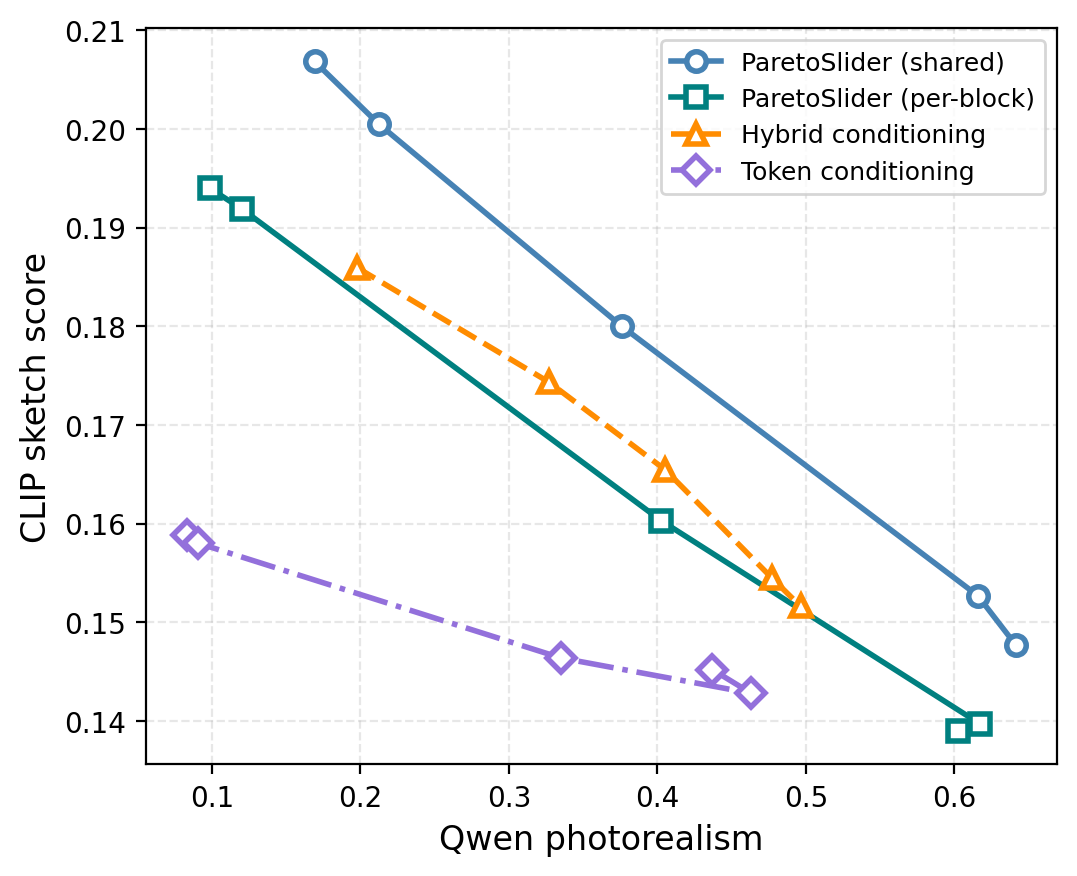}
    \end{minipage}\hfill
    \begin{minipage}[t]{0.65\textwidth}
        \vspace{4pt}
        \centering
        \begin{tabular}{c c c c c c c}
            \includegraphics[width=0.15\linewidth]{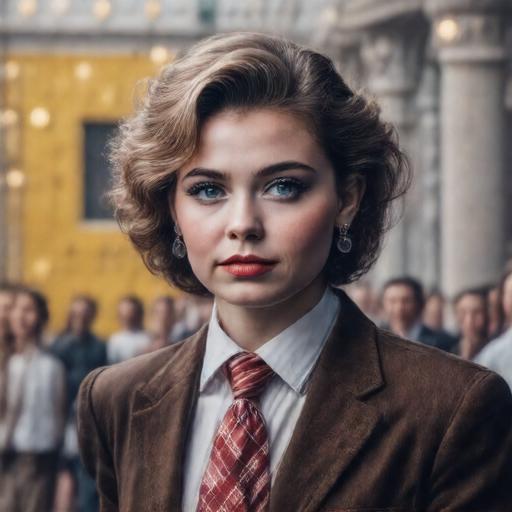} &
            \includegraphics[width=0.15\linewidth]{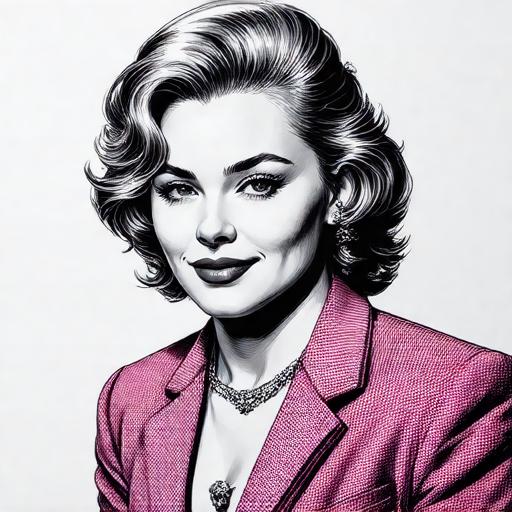} &
            \includegraphics[width=0.15\linewidth]{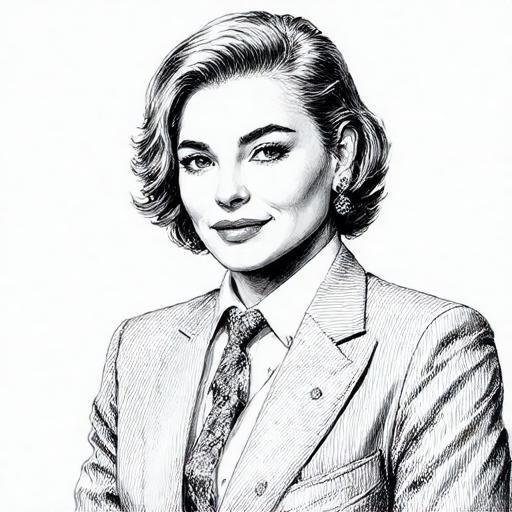} &
            \includegraphics[width=0.15\linewidth]{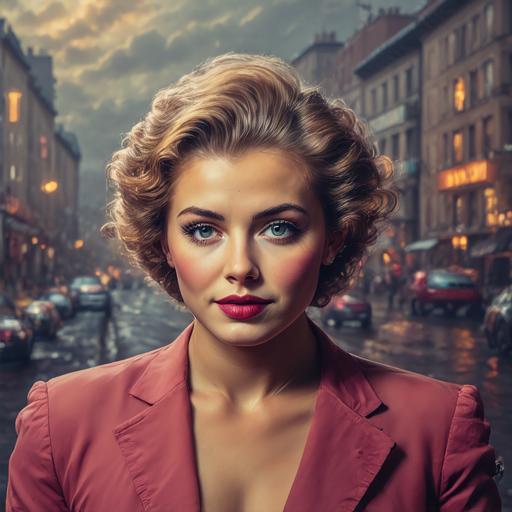} &
            \includegraphics[width=0.15\linewidth]{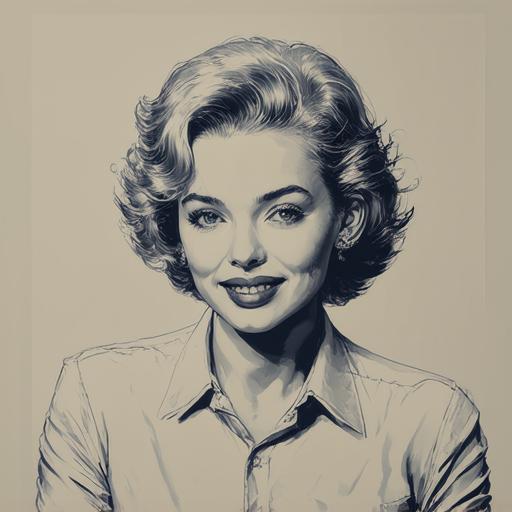} &
            \includegraphics[width=0.15\linewidth]{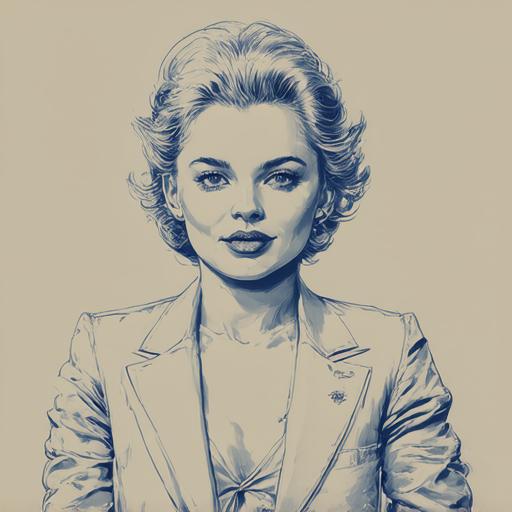} \\
            & Shared & & & Per & \\[0.5em]
            \includegraphics[width=0.15\linewidth]{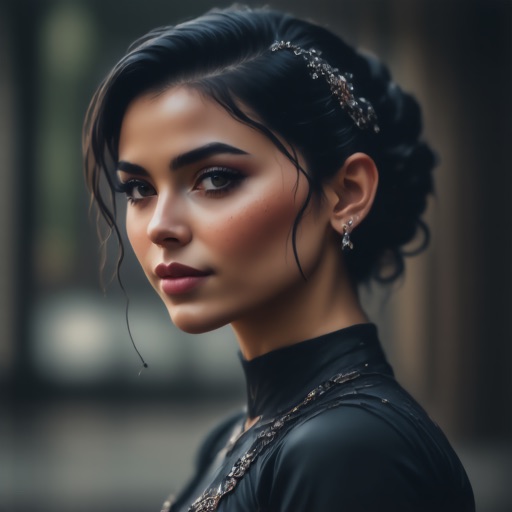} &
            \includegraphics[width=0.15\linewidth]{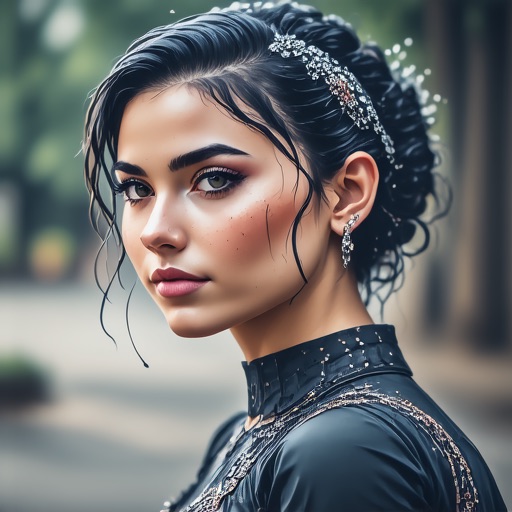} &
            \includegraphics[width=0.15\linewidth]{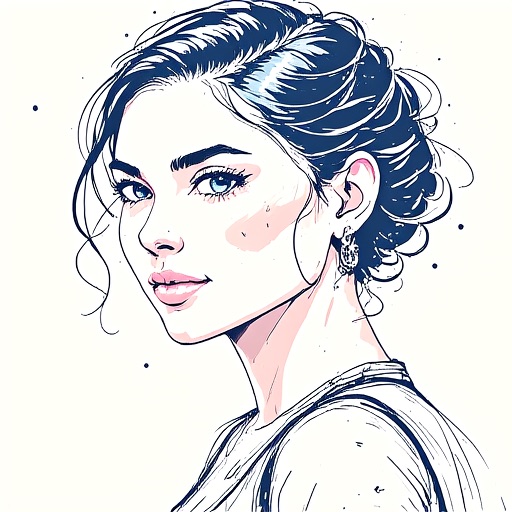} &
            \includegraphics[width=0.15\linewidth]{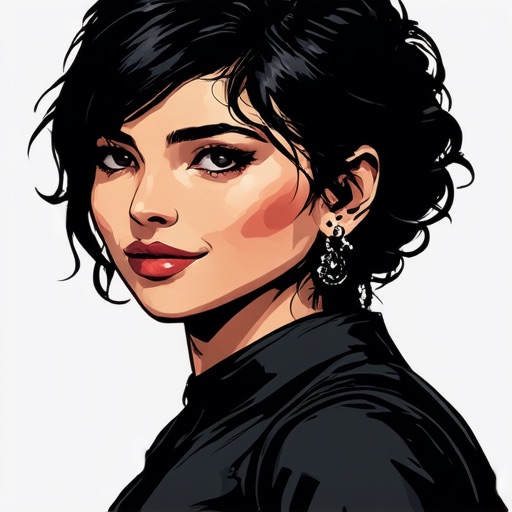} &
            \includegraphics[width=0.15\linewidth]{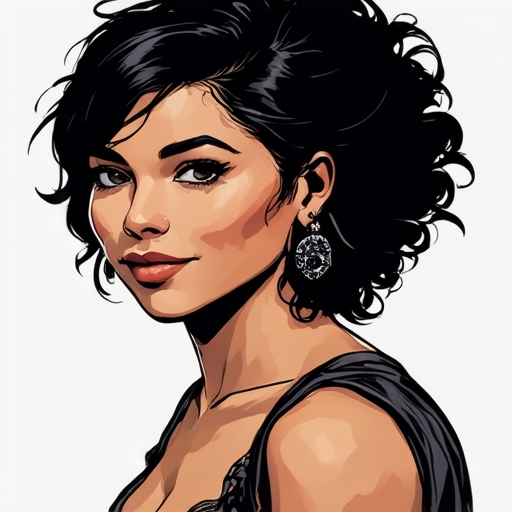} &
            \includegraphics[width=0.15\linewidth]{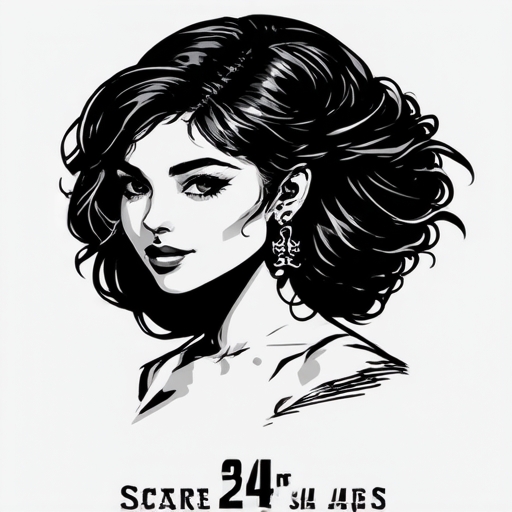} \\
            & Hybrid & & & Token & \\[0.4em]
           
        \end{tabular}
    \end{minipage}
    \vspace{-3pt}
    \caption{\small Ablation of SD3.5 preference-conditioning on the photorealism-sketch trade-off. \textit{Left:} Shared conditioning gives the strongest conditioning. \textit{Right:} Shared and per-block conditioning yield smoother transitions, while hybrid and token conditioning show weaker control and collapse toward the dominant metric.}
    \label{fig:ablation_condition}
    \vspace{-5pt}
\end{figure}

\subsection{Comparisons}
\label{subsec:comp}
\vspace{-5pt}
\paragraph{\textbf{Baselines T2I.}}
We compare against five baselines that represent natural alternatives to our preference-conditioned training. \textit{Fixed-Weights} uses the same DiffusionNFT training pipeline with a fixed weighted reward sum, requiring a separate training run per operating point. We train DiffusionNFT on an equal setting of hyperparameters and number of epochs as our approach. \textit{Flow-Multi} \cite{lee2026flowmulti} is a GRPO-based baseline with batch-wise Pareto non-dominated selection. This baseline was fine-tuned according to the setting detailed in the paper. Unlike our method, it still learns a single static policy and offers no inference-time control.  \textit{RewardedSoups}~\cite{rame2023rewardedsoupsparetooptimalalignment} trains separate models specialized for individual rewards and obtains intermediate trade-offs by linearly interpolating their weights. \textit{Prompt Rewriting} uses an LLM to rewrite prompts emphasizing each objective (e.g., appending photorealism or sketch descriptors), providing control through text alone.\begin{wrapfigure}[29]{r}{0.5\textwidth}
\vspace{-5pt}
    \centering
    \small
    \setlength{\tabcolsep}{0.5pt}
    \renewcommand{\arraystretch}{0.9}
    \newcommand{\figW}{0.8\linewidth}
    \newcommand{\imgw}{0.15\linewidth}
    
    \includegraphics[width=\figW]{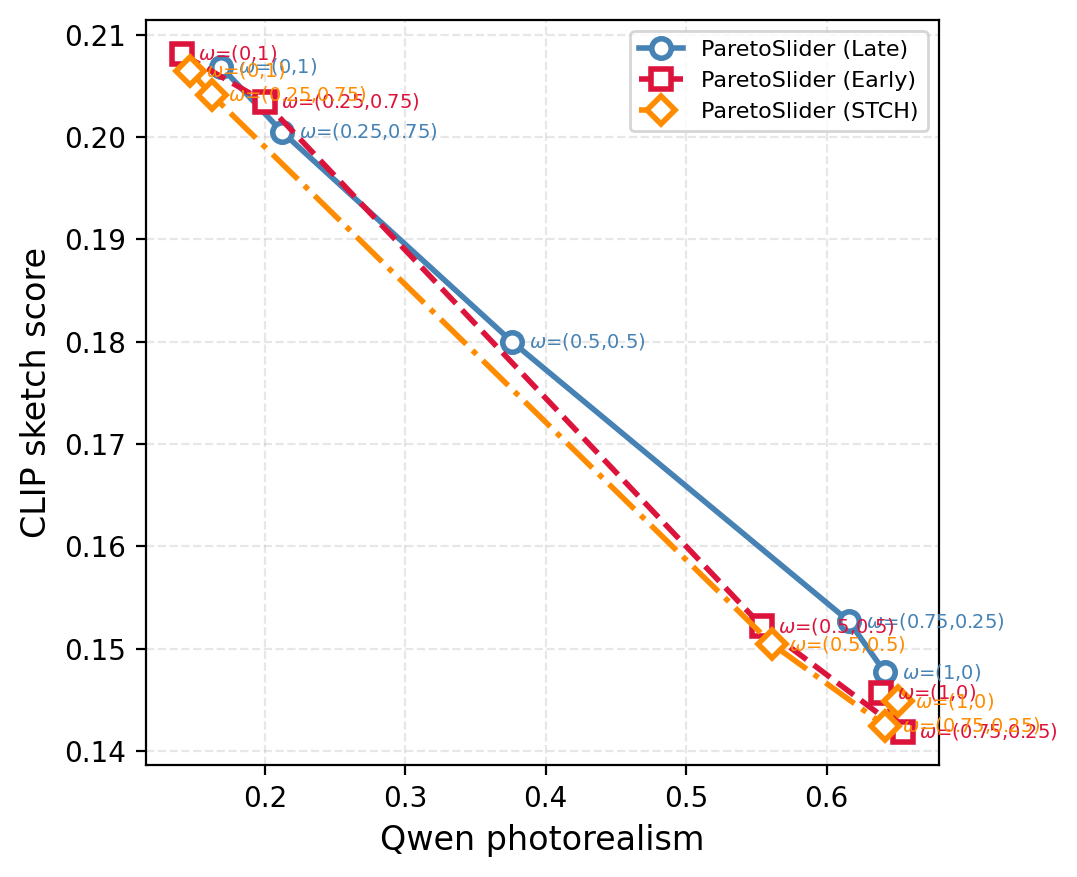}
    
    \begin{tabular}{@{}c c c c c c@{}}
        \raisebox{10pt}{\rotatebox{90}{\scriptsize \color{blue}Ours}} &
        \includegraphics[width=\imgw]{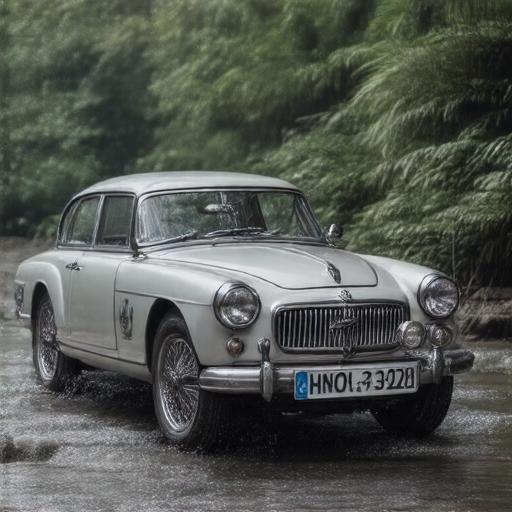} &
        \includegraphics[width=\imgw]{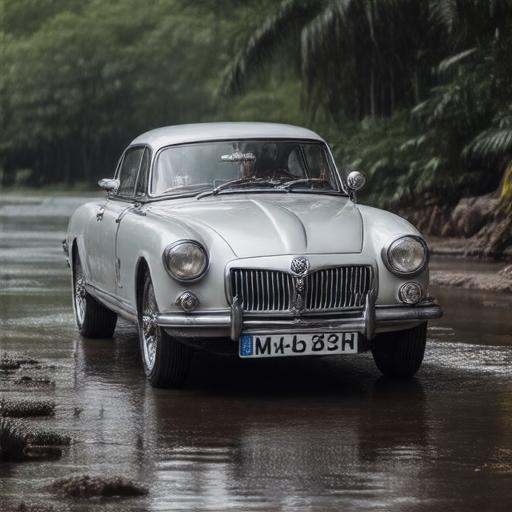} &
        \includegraphics[width=\imgw]{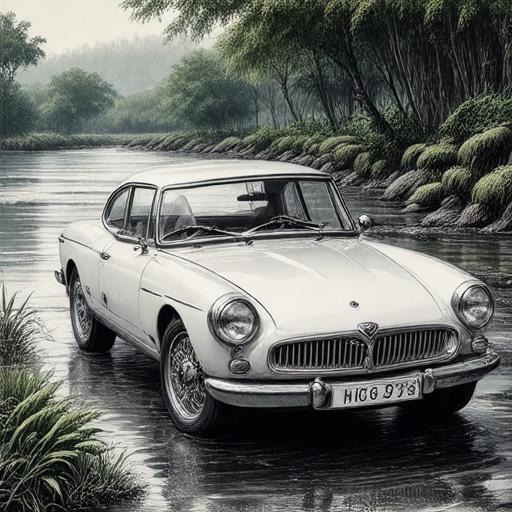} &
        \includegraphics[width=\imgw]{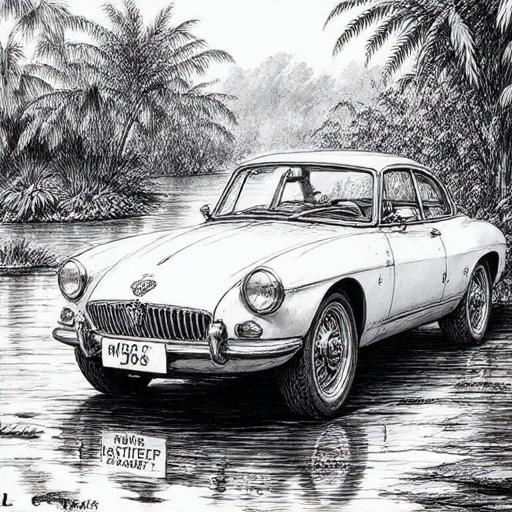} &
        \includegraphics[width=\imgw]{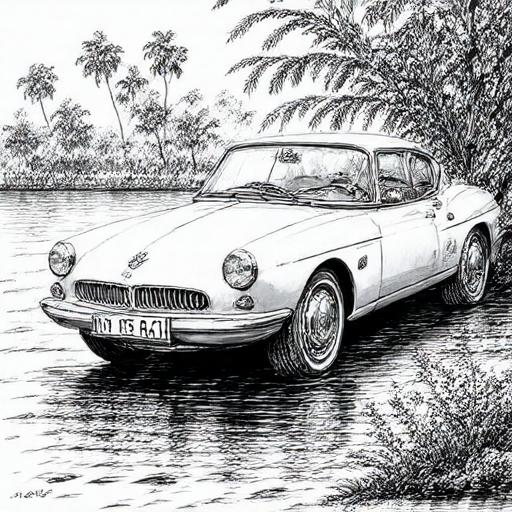} \\[-1pt]
        
        \raisebox{9pt}{\rotatebox{90}{\scriptsize \color{orange}Early}} &
        \includegraphics[width=\imgw]{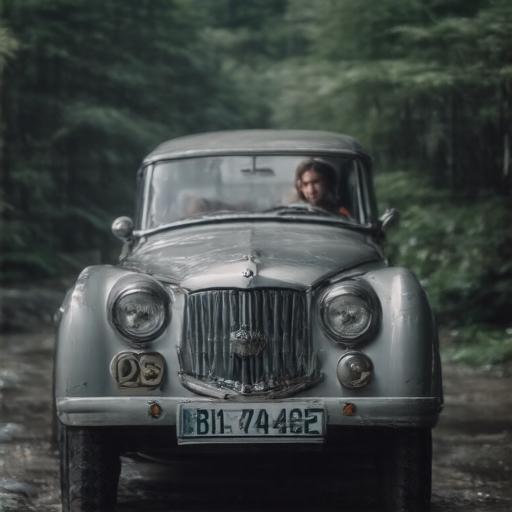} &
        \includegraphics[width=\imgw]{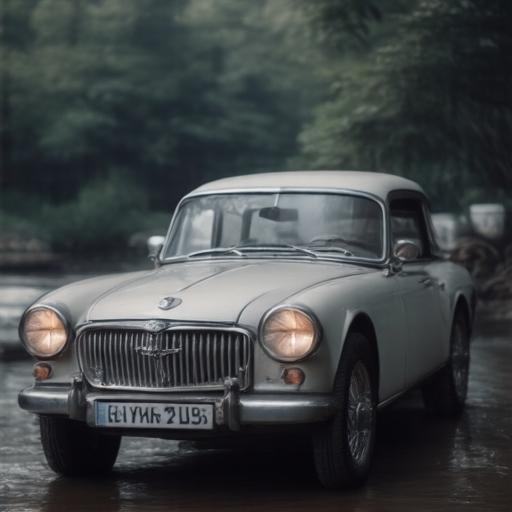} &
        \includegraphics[width=\imgw]{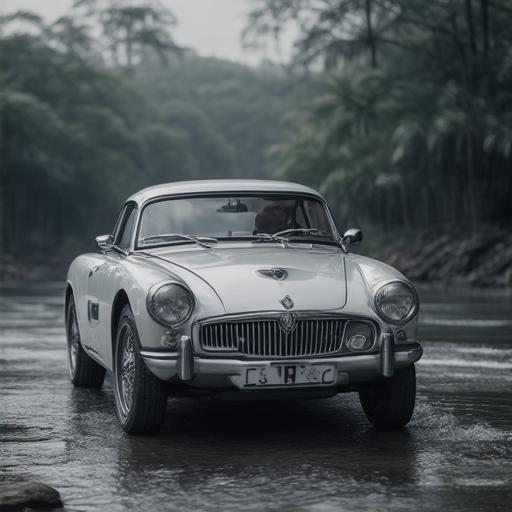} &
        \includegraphics[width=\imgw]{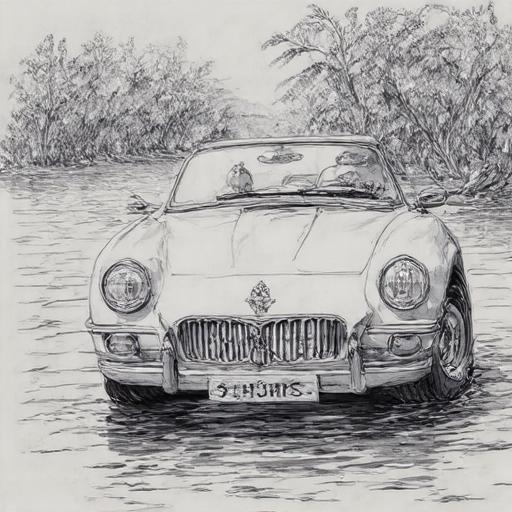} &
        \includegraphics[width=\imgw]{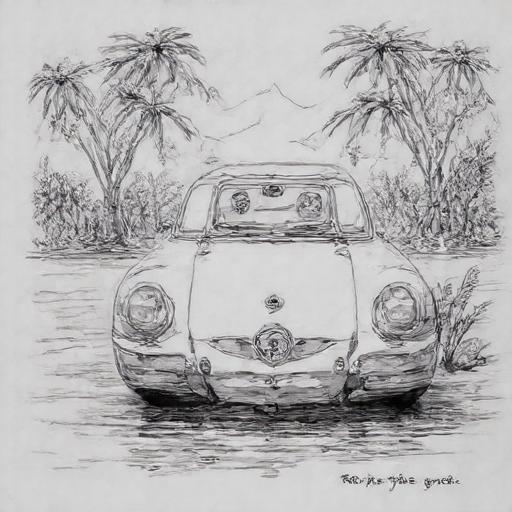} \\[-1pt]
        
        \raisebox{8pt}{\rotatebox{90}{\scriptsize \color{orange}STCH}} &
        \includegraphics[width=\imgw]{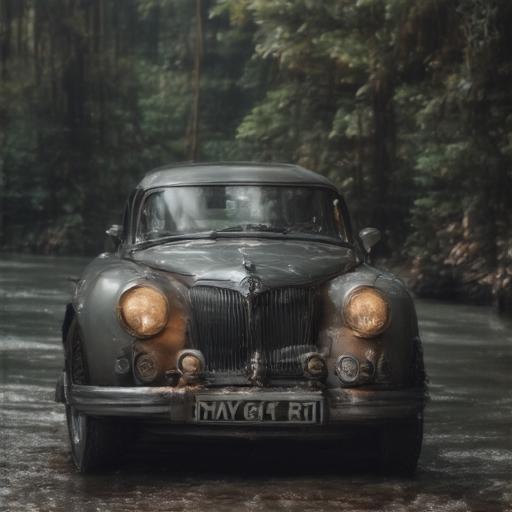} &
        \includegraphics[width=\imgw]{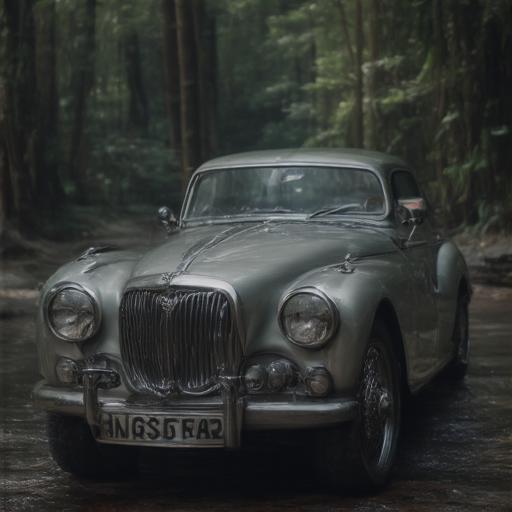} &
        \includegraphics[width=\imgw]{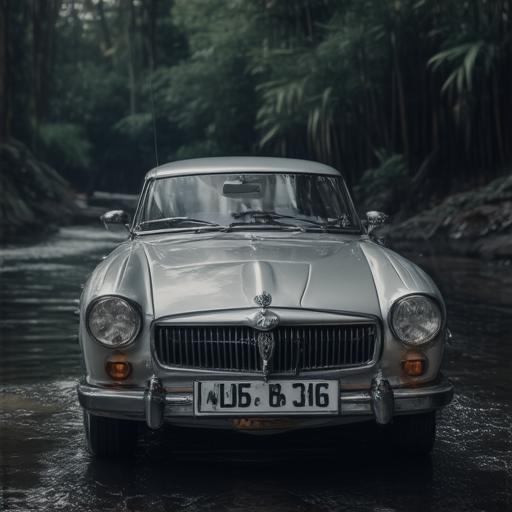} &
        \includegraphics[width=\imgw]{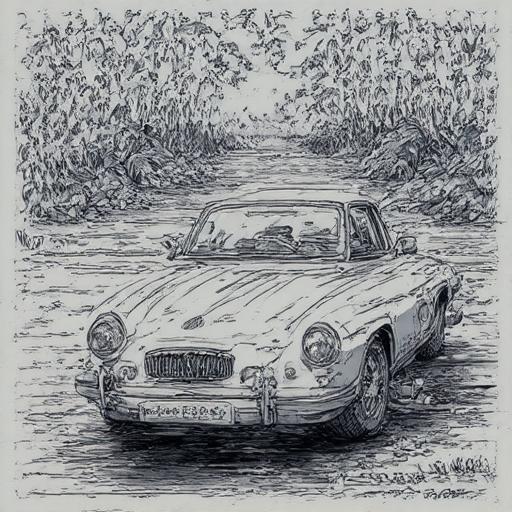} &
        \includegraphics[width=\imgw]{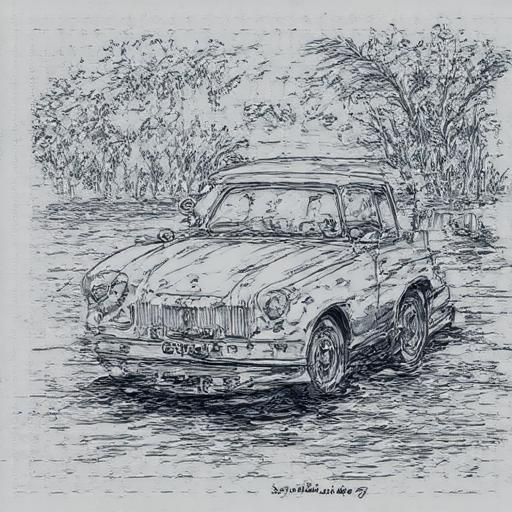} \\[-1pt]
        
        & \scriptsize Realistic
        & \multicolumn{3}{c}{\scriptsize $\xleftarrow{\hspace{0.05cm}}\;
        \omega=(\omega_{\mathrm{real}},\omega_{\mathrm{sketch}})
        \;\xrightarrow{\hspace{0.05cm}}$}
        & \scriptsize Sketch
    \end{tabular}
    
    \vspace{-4pt}
    \caption{\small \textit{Top:} Controllability comparison for late scalarization (ours), early scalarization, and STCH~\cite{lin2024smooth}.
    Only late scalarization produces a genuine middle-ground operating point at $\omega{=}(0.5, 0.5)$, while early and STCH collapse toward the realism corner.
    \textit{Bottom:} qualitative transitions from realistic to sketch as $\omega$ varies. The middle column shows that our method yields a true blend, while early and STCH ``jump'' between the two endpoints with no smooth interior.}
    \label{fig:ablation_losses}
    \vspace{-10pt}
\end{wrapfigure}
\textit{Diffusion Blend (DB-MPA)} constructs intermediate trade-offs by composing denoisers specialized for individual objectives. DB-MPA and RewardedSoups reuse the same two reward-endpoint specialists, trained with matched budgets and seeds. We compare both methods with ParetoSlider on a dense grid of 21 preference vectors, as detailed in Appendix~\ref{app:dense_sweep}.
As shown in Figure~\ref{fig:pareto_and_baselines} (right), ParetoSlider produces smooth, coherent transitions across the full preference spectrum, while Fixed-Weights collapses toward the dominant reward and both Flow-Multi and Prompt Rewriting offer only coarse, isolated operating points. Quantitatively, Figure~\ref{fig:pareto_and_baselines} shows that our single preference-conditioned model traces a well-ordered trade-off that consistently dominates all baselines.

\vspace{-5pt}
\paragraph{\textbf{Baselines I2I.}}
For image-to-image editing, we compare against inference-time baselines that already expose practical control knobs over the edit-preservation trade-off. 
Following the dual-guidance controls demonstrated in InstructPix2Pix~\cite{brooks2023instructpix2pix}, \textit{Text-CFG} sweeps the text classifier-free guidance scale, increasing adherence to the editing instruction at the cost of stronger deviations from the source image, while \textit{Image-CFG} sweeps the image guidance scale to strengthen source preservation and suppress larger edits. 
We also compare against \textit{Prompt Rewriting}, which uses an LLM to reformulate the editing instruction so as to emphasize either edit strength or source fidelity. These baselines test whether standard guidance controls and instruction engineering are sufficient to recover the desired trade-off without explicit preference-conditioned training. 
Additionally, we fine-tune five DiffusionNFT checkpoints separately trained on uniformly spaced trade-off points along the Pareto front. As shown in Figure~\ref{fig:editing_comparison} (right), while this produces a reasonable trade-off, the editing quality is weaker than our approach. Additionally, increasing the Image-CFG and Text-CFG scales progressively strengthens source preservation at the cost of visual artifacts. 
The controllability comparison in Figure~\ref{fig:editing_comparison} (left) shows that our preference-conditioned model produces a smooth and consistent trade-off curve. Our method covers a broader range of the trade-off space than Text CFG and consistently dominates FixedWeights, which requires training a separate model for each operating point, resulting in superior coverage overall.

\vspace{-5pt}
\subsection{Quantitative Results}
\label{subsec:quant_results}
\vspace{-5pt}
In all of the experiments we report held-out metrics to prevent a circular evaluation and demonstrate the robustness of our method. The main paper presents a single set of metrics and Appendix \S\ref{app:experiments} presents three aditional sets of evaluation metrics. We report three complementary metrics: Hypervolume (HV), monotonicity, and perceptual smoothness. HV~\cite{zitzler2002multiobjective} measures the dominated volume of the non-dominated solution set after global min-max normalization of reward scores, using the origin $(0,0)$ as the reference point. The ``ND'' column reports the number of non-dominated operating points retained by each method, following \S\ref{subsec:pareto_def}. Monotonicity is measured by the Spearman rank correlation between the requested trade-off vector $\omega$ and the measured rewards, averaged over the two objectives. Smoothness is measured as the maximum LPIPS distance between adjacent operating points along the preference vector $\omega$ sweep, capturing the largest perceptual jump, where lower values indicate smoother visual transitions. Additional quantative results for T2I and I2I tasks are presented in Appendix \S\ref{app:quant_extended}.
\paragraph{Dense preference control.}
We additionally evaluate 21 evenly spaced preference vectors on the photorealism--sketch task, comparing ParetoSlider with RewardedSoups and Diffusion Blend (DB-MPA). Under held-out evaluators, ParetoSlider achieves signed Spearman correlations of $-1.00$ for photorealism and $+1.00$ for sketch as the sketch preference increases. It also retains stronger interior trade-offs where $\omega=(0.5,0.5)$. Appendix~\ref{app:dense_sweep} reports the full sweep, local monotonicity, and adjacent perceptual distances.

\begin{table}
\centering
\small
\caption{\small Quantitative T2I comparison on the photorealism vs. sketch trade-off using held-out metrics. ParetoSlider achieves the highest hypervolume and lowest worst-case perceptual jump (smoothness) while maintaining near-perfect monotonic control.}
\label{tab:hypervolume_t2i}
\setlength{\tabcolsep}{3pt}
\begin{tabular}{lcccc} 
\toprule
\textbf{Method} & \textbf{HV} $\uparrow$ & \textbf{Monotonicity} $\uparrow$ & \textbf{Smoothness} $\downarrow$ & ND  \\
\midrule
ParetoSlider (ours)  & $\mathbf{0.024 \pm .002}$ & $\underline{0.962 \pm .035}$ & $\mathbf{0.529 \pm .007}$ & 5 \\
Prompt Rewriting & $\underline{0.011 \pm .001}$  & $\mathbf{1.000 \pm .000}$ & $\underline{0.615 \pm .009}$ & 3 \\
RewardedSoups \cite{rame2023rewardedsoupsparetooptimalalignment} & $0.010 \pm .002$  & $0.952 \pm .014$  & $0.625 \pm .007$ & 4 \\
FixedWeights  & $0.007 \pm .002$   & $0.870 \pm .088$  & $0.666 \pm .005$ & 2 \\
\bottomrule
\end{tabular}
\vspace{-10pt}
\end{table}

\vspace{-5pt}
\subsection{Ablation Studies}
\label{subsec:ablations}
\vspace{-5pt}
We ablate two independent design choices on SD3.5 (T2I generation): the preference conditioning method and the multi-objective loss formulation. Each experiment varies one axis while fixing the other in our default setting. 
Qualitative transitions and Pareto front comparisons for both ablation choice are presented in Figures~\ref{fig:ablation_condition} and \ref{fig:ablation_losses}.
As can be seen both qualitatively and quantitatively, our method outperforms both variants, producing images that faithfully adhere to the preference vector $\omega$.

\vspace{-5pt}
\paragraph{\textbf{Conditioning Method.}}
\vspace{-5pt}
We compare four architectures for injecting the preference vector $\omega$ into the transformer, differing in both the location and mechanism of conditioning. The \textit{Shared} and \textit{Per-block} variants apply modulation-based conditioning, as described in \S\ref{subsec:architecture} and \S\ref{app:impl_sd3}. The \textit{Token} variant projects $\omega$ into learnable tokens that are prepended to the text sequence. The \textit{Hybrid} variant combines timestep conditioning with AdaLN modulation. Full details are provided in Appendix \S\ref{app:impl_details}. 

\vspace{-10pt}
\paragraph{\textbf{Loss Formulation.}}
We evaluate our linear late-scalarization formulation against two alternative strategies: early scalarization, which remains the na\"ive default in multi-objective alignment, and Tchebycheff (TCH) late-scalarization~\cite{lin2024smooth}, which utilizes a non-linear max-based aggregation.
As seen in Fig.~\ref{fig:ablation_losses}, both baselines fail to find a stable middle ground; instead, the models collapse to one of the rewards at any given point, effectively destroying the fine-grained controllability required for a smooth trade-off.
In Appx.~\S\ref{app:theory}, we provide a theoretical proof of the non-equivalence between early and late scalarization and demonstrate that early scalarization fails to preserve the preference signal during training.
We speculate that TCH scalarization fails due to its max-based nature, which induces discontinuous optimization dynamics that degrade smooth inference-time transitions.
Additional analysis is presented in Appendix \S\ref{app:experiments}.

\vspace{-5pt}
\section{Conclusions}
\vspace{-5pt}
We presented ParetoSlider, a controllable multi-objective RL post-training framework that enables continuous inference-time control over trade-offs between competing rewards in diffusion and flow-matching models.
Rather than committing to a fixed operating point at training time, ParetoSlider conditions a single model on a preference vector $\omega$, amortizing an approximation of the reward trade-off space into a single set of parameters. We evaluated ParetoSlider across three state-of-the-art backbones spanning T2I synthesis, image editing, and T2V generation. In all settings, a single preference-conditioned model matches or exceeds multiple separately trained fixed-weight baselines, while providing smooth inference-time control. We believe ParetoSlider establishes a scalable paradigm for a controllable alignment of visual generative models, and opens the door to richer user-facing control interfaces where non-expert users can intuitively navigate complex reward trade-offs at inference time.

\vspace{-5pt}

\begin{ack} 
\vspace{-5pt}
We thank Ofir Schlisselberg and Itay Nakash for their early feedback and helpful suggestions.
We also thank NVIDIA for their generous support through the NVIDIA Academic Grant program, which provided GPU hours via Brev for this research.
\end{ack}
\clearpage

{
    \small
    \bibliographystyle{ieeenat_fullname}
    \bibliography{refs}
}

\clearpage

\appendix
\clearpage
\thispagestyle{empty}

\vspace*{1.5cm}

\begin{center}
    {\Huge \bfseries Appendix}\\[0.4cm]
    {\Large Supplementary Material}\\[1.2cm]
    \rule{0.75\linewidth}{0.6pt}
\end{center}

\vspace{0.8cm}

\begingroup
\hypersetup{linkcolor=black}
\startcontents[appendix]
\printcontents[appendix]{}{1}{}
\endgroup

\vfill

\begin{center}
    \rule{0.75\linewidth}{0.6pt}
\end{center}

\clearpage
\setcounter{section}{0}

\section{Implementation Details}
\label{app:impl_details}

\subsection{Implementation of Text-to-Image Generation}
\label{app:impl_sd3}

We build our text-to-image policy on Stable Diffusion 3.5 Medium \cite{esser2024scaling}, freezing the VAE and text encoders to adapt only the transformer backbone via LoRA and our preference-conditioning modules \cite{hu2022lora, hanparameter}. The model is conditioned on an explicit preference vector $\omega \in \mathbb{R}^{M}$, where $M$ is the number of reward objectives.

\vspace{-5pt}
\paragraph{\textbf{Reference, current, and old policies.}}
Training maintains three policies: a trainable current policy $\theta$, an exponential moving average (EMA) old policy $\theta_{\mathrm{old}}$ for implicit velocity construction, and a frozen reference policy $\theta_{\mathrm{ref}}$ for KL regularization. The EMA update is defined as:
$$\theta_{\mathrm{old}} \leftarrow \lambda \theta_{\mathrm{old}} + (1-\lambda)\theta$$

\vspace{-5pt}
\paragraph{\textbf{Backbone and trainable parameters.}}
We instantiate a preference-conditioned variant of the SD3.5 transformer by copying the pretrained transformer weights into a modified architecture. 
A two-layered MLP $f_{\mathrm{time}}\colon \mathbb{R}^M \to \mathbb{R}^{d}$ maps the preference vector into the timestep-embedding space and adds it directly: $\tilde{\mathbf{t}}_{\mathrm{emb}} = \mathbf{t}_{\mathrm{emb}} + f_{\mathrm{time}}(\omega)$.
Because every transformer block derives its AdaLN scale, shift, and gating parameters from $\tilde{\mathbf{t}}_{\mathrm{emb}}$, this injection broadcasts preference information to all $L$ blocks simultaneously.

A projector network $f_{\mathrm{blk}}$ first encodes $\omega$ via sinusoidal positional embeddings and then maps the result through a four-layer MLP to produce a shared modulation vector $\boldsymbol{\delta}_\omega = f_{\mathrm{blk}}(\mathrm{enc}(\omega)) \in \mathbb{R}^{d}$. In each transformer block $\ell$, $\boldsymbol{\delta}_\omega$ is injected into the image-stream hidden states after the feed-forward layer, gated by the block's native AdaLN gating parameter $\mathbf{g}^{(\ell)}$:
\begin{equation} 
\mathbf{h}^{(\ell)} \leftarrow \mathbf{h}^{(\ell)} + \mathbf{g}^{(\ell)} \odot \boldsymbol{\delta}_\omega, \label{eq:blk_residual}
\end{equation}
where $\mathbf{h}^{(\ell)}$ denotes the image-stream hidden states at block $\ell$. By reusing the block's own gating mechanism, the modulation participates in the same per-block scaling learned during pretraining, which we find stabilizes training. The same $\boldsymbol{\delta}_\omega$ is shared across all $L$ blocks, keeping the parameter overhead minimal.

\paragraph{\textbf{Preference-conditioning variants.}}
We compare three additional preference-conditioning designs for the SD3.5 transformer which are presented in \S\ref{subsec:ablations} and Figure \ref{fig:ablation_condition}. 

A per-block residual variant that increases the output dimension of $f_{\mathrm{blk}}$ from $d$ to $L\cdot d$. The resulting vector is sliced into block-specific residual directions $\{\boldsymbol{\delta}_{\omega}^{(\ell)}\}_{\ell=1}^{L}$, where $\boldsymbol{\delta}_{\omega}^{(\ell)}\in\mathbb{R}^{d}$. Each block-specific vector is injected at the same post-feed-forward residual location and gated by the corresponding native
$\mathbf{g}_{\mathrm{mlp}}^{(\ell)}$. This variant preserves the same injection mechanism as the shared residual model, but allows the preference modulation direction to vary with depth. It increases the projector parameter count by a factor of $L$, while leaving the backbone and LoRA configuration unchanged.

As a token-conditioning ablation, we replace residual image-stream modulation with preference tokens inserted into the text-conditioning stream. The preference vector is projected into a small set of learned tokens that are prepended to the text encoder hidden states before the SD3.5 context projection.
Concretely, a learnable base token matrix $\mathbf{E}_{\mathrm{base}}\in\mathbb{R}^{N_t\times d_c}$ is combined with an MLP projection of the preference vector,
\begin{equation}
\mathbf{P}_{\omega} = \mathbf{E}_{\mathrm{base}} + f_{\mathrm{token}}(\omega),
\end{equation}
where $\mathbf{P}_{\omega}\in\mathbb{R}^{N_t\times d_c}$ and $d_c$ is the joint-attention context dimension. The preference tokens are concatenated with the original text context and participate in cross-attention throughout all transformer blocks. We initialize $\mathbf{E}_{\mathrm{base}}$ from a zero-mean Gaussian with standard deviation $0.01$, and initialize the final layer of $f_{\mathrm{token}}$ near zero with Gaussian standard deviation $10^{-3}$ so that the conditioning strength increases smoothly during training. The number of preference tokens $N_t$ is configurable. We used $N_t=4$ in our ablations.

Finally, we evaluate an ``Hybrid'' variant that removes block-level residual modulation altogether. In this setting, $f_{\mathrm{blk}}$ is dropped. A separate two-layer MLP $f_{\mathrm{out}}\colon\mathbb{R}^{M}\!\to\!\mathbb{R}^{2d}$ maps the preference vector to output modulation parameters $(\boldsymbol{\alpha}_{\omega},\boldsymbol{\beta}_{\omega})$, which are applied after the final transformer block and before the unpatchify projection:
\begin{equation}
\mathbf{h} \leftarrow \mathbf{h}\odot(1+\boldsymbol{\alpha}_{\omega}) + \boldsymbol{\beta}_{\omega}.
\end{equation}
Although this design is parameter-efficient and leaves the internal transformer blocks unchanged, we found it to be highly prone to reward hacking in practice. Because the preference-dependent transformation is applied only at the output of the transformer, the model can exploit the reward signal through late-stage
feature shifts rather than learning a robust preference-conditioned denoising process throughout the network.

\vspace{-5pt}
\begin{table*}[t]
\centering
\caption{\small Hyperparameters for the main SD3.5 text-to-image experiments.}
\label{tab:sd3_hparams}
\begin{tabular}{ll}
\toprule
\textbf{Category} & \textbf{Value} \\
\midrule
Backbone & Stable Diffusion 3.5 Medium \\
Trainable parameters & LoRA adapters + preference-conditioning modules \\
LoRA rank & 32 \\
Warm start & CLIPScore + PickScore training 6 epochs \\
Training epochs & 12 \\
\midrule
Resolution & $512 \times 512$ \\
Training denoising steps & 25 \\
Evaluation denoising steps & 40 \\
Sampler & DPM2 \\
Guidance scale & 1.0 \\
Noise level & 0.7 \\
\midrule
Repeated samples per prompt & 24 \\
Preference subgroups per prompt & 3 \\
\midrule
Optimizer & AdamW \\
Learning rate & $3 \times 10^{-4}$ \\
Adam $\beta_1, \beta_2$ & $(0.9,\,0.999)$ \\
Weight decay & $10^{-4}$ \\
Adam $\epsilon$ & $10^{-8}$ \\
Gradient clipping & 1.0 \\
Advantage clipping $\epsilon_{\mathrm{clip}}$ & 5 \\
KL coefficient $\lambda_{\mathrm{KL}}$ & 0.0001 \\
Implicit velocity coefficient $\beta$ & 0.1 \\
Mixed precision & fp16 \\
\bottomrule
\end{tabular}
\vspace{-10pt}
\end{table*}

\vspace{-5pt}
\paragraph{\textbf{Sampling and optimization setup.}}
We summarize the main SD3.5 text-to-image hyperparameters in Table~\ref{tab:sd3_hparams}. Unless noted otherwise, all text-to-image experiments use this configuration, including the sampling setup, optimizer settings, repeated samples per prompt, and optimization coefficients. For the main conditioned photorealism-versus-sketch setup, we initialize from a warm-start checkpoint trained on PickScore \cite{kirstain2023pickapicopendatasetuser} and CLIPScore \cite{hessel2021clipscore} with the original DiffusionNFT \cite{zheng2026diffusionnftonlinediffusionreinforcement} setup, which enables stable high-quality generation already at guidance scale $1.0$.

\begin{table*}[t]
\centering
\caption{\small Hyperparameters for the main FLUXKontext image-editing experiments.}
\label{tab:kontext_hparams}
\begin{tabular}{ll}
\toprule
\textbf{Category} & \textbf{Value} \\
\midrule
Backbone & FLUX.1-Kontext-dev \\
Trainable parameters & LoRA adapters + preference projector \\
Conditioning method & AdaLN context \\
LoRA rank & 32 \\
LoRA alpha & 64 \\
Training epochs & 12 \\
\midrule
Resolution & $512 \times 512$ \\
Training denoising steps & 10 \\
Evaluation denoising steps & 15 \\
Sampler & DPM2 \\
Guidance scale & 2.5 \\
Noise level & 0.7 \\
\midrule
Repeated samples per prompt & 24 \\
Preference subgroups per prompt & 3 \\
\midrule
Optimizer & AdamW \\
Learning rate & $3 \times 10^{-4}$ \\
Adam $\beta_1, \beta_2$ & $(0.9,\,0.999)$ \\
Weight decay & $10^{-4}$ \\
Adam $\epsilon$ & $10^{-8}$ \\
Gradient clipping & 1.0 \\
Advantage clipping $\epsilon_{\mathrm{clip}}$ & 5 \\
KL coefficient $\lambda_{\mathrm{KL}}$ & 0.0001 \\
Implicit velocity coefficient $\beta$ & 0.1 \\
Mixed precision & bf16 \\
\bottomrule
\end{tabular}
\end{table*}

\vspace{-5pt}
\subsection{Implementation of Image-to-Image Tasks}
\label{app:impl_flux}

\paragraph{\textbf{Backbone and trainable parameters.}}
For instruction-based image editing, like in SD3, the VAE, text encoders, and base transformer weights remain entirely frozen throughout training. The model is adapted exclusively through LoRA injected into the transformer's attention layers, alongside a lightweight preference projector used for context modulation.
\vspace{-10pt}
\paragraph{\textbf{Preference conditioning.}}
We condition FLUXKontext \cite{labs2025flux1kontextflowmatching} in the transformer modulation space. Concretely, the preference vector is mapped by a lightweight projector to a modulation vector of dimension $2d$, which is split into scale and shift terms and added to the AdaLN parameters of the context stream inside the dual-stream transformer blocks. The image stream is not directly modulated, and the single-stream blocks are left unchanged. This design follows the same general principle as KontinuousKontext, which projects an external control signal into the model's modulation space, but here the scalar edit-strength control is replaced by a multi-dimensional preference vector.

\vspace{-5pt}
\subsection{Implementation of Text-to-Video Tasks}
\label{app:impl_video}
For this task, we adopt the LTX-2 \cite{hacohen2026ltx2efficientjointaudiovisual} model and condition the policy through the same shared block-residual mechanism used for SD3.5 (Equation~\ref{eq:blk_residual}), targeting the video stream and using the LTX-2 inner dimension $d=3840$. The projector $f_{\mathrm{blk}}$ (sinusoidal PE followed by a four-layer MLP) is unchanged apart from its output width. For stable early RL fine-tuning, the final linear layer of $f_{\mathrm{blk}}$ is initialized with weights drawn from $\mathcal{N}(0,\,10^{-3})$ and zero bias.

\vspace{-5pt}
\begin{table*}[t]
\centering
\caption{\small Hyperparameters for the main LTX-2 text-to-video experiments.}
\label{tab:ltx2_hparams}
\begin{tabular}{ll}
\toprule
\textbf{Category} & \textbf{Value} \\
\midrule
Backbone & LTX-2 (19B parameters) \\
Text encoder & Gemma 3 12B IT \cite{team2024gemma} \\
Trainable parameters & LoRA adapters + preference modules \\
Conditioning method & Shared block-residual (gated by native FF gate) \\
LoRA rank / alpha & 32 / 32 \\
Preference projector & Sinusoidal PE $\to$ MLP ($2M{\to}768{\to}3840$, 4 layers, ReLU) \\
Training prompts & 1{,}000 \\
\midrule
Resolution & $512 \times 512$ \\
Number of frames & 41 \\
Frame rate & 25\,fps \\
Training denoising steps & 20 \\
Evaluation denoising steps & 50 \\
Guidance scale (evaluation) & 4.0 \\
Timestep sampling & Shifted logit-normal \\
\midrule
Repeated samples per prompt & 24 \\
Preference subgroups per prompt & 2 \\
Timesteps per sample & 5 \\
\midrule
Optimizer & AdamW \\
Learning rate & $3 \times 10^{-4}$ \\
Adam $\beta_1, \beta_2$ & $(0.9,\,0.999)$ \\
Weight decay & $10^{-4}$ \\
Adam $\epsilon$ & $10^{-8}$ \\
Gradient clipping & 1.0 \\
Advantage clipping $\epsilon_{\mathrm{clip}}$ & 5 \\
KL coefficient $\lambda_{\mathrm{KL}}$ & 0.1 \\
Implicit velocity coefficient $\beta$ & 0.1 \\
EMA decay & 0.9 \\
Mixed precision & bf16 \\
\bottomrule
\end{tabular}
\vspace{-10pt}
\end{table*}

\vspace{-5pt}
\paragraph{\textbf{Algorithm.}}
Algorithm~\ref{alg:method_nft} details the full training procedure of our \methodname{} framework. To clearly delineate our contributions, the base DiffusionNFT optimization steps are written in black, while our multi-objective ParetoSlider additions, including preference sampling, per-channel group normalization, and scalarized losses, are highlighted in {\color{blue} blue}.

\begin{algorithm*}[t]
\caption{\methodname{} Fine-tuning (Extensions made to DiffusionNFT are in {\color{blue} blue})}
\label{alg:method_nft}
\begin{algorithmic}[1]
\Require Policy $v_\theta$ {\color{blue} (preference-conditioned)}, reference $v_{\mathrm{ref}}$, prompt dataset $\mathcal{D}$.
\Require {\color{blue} Reward functions $r_1, \dots, r_M$}, Group size $K$, clip $\epsilon_{\mathrm{clip}}$, step size $\beta$, inner epochs $N$.

\While{not converged}
    \State \textcolor{gray}{\textit{// 1. Sampling \& {\color{blue} Multi-Objective} Scoring}}
    \State Sample batch of prompts $\mathcal{C} \sim \mathcal{D}$
    \For{each prompt $c \in \mathcal{C}$}
        \State {\color{blue} Sample preference: $\omega^{(c)} \sim p(\omega)$ \quad \textcolor{gray}{\textit{// Structured vertex/edge/interior}}}
        \State Generate $K$ samples: $\{x_{\color{blue} 0,\omega}^{(c,i)}\}_{i=1}^K \sim \pi_\theta(\cdot \mid c {\color{blue}, \omega^{(c)}})$
        \State Evaluate rewards: {\color{blue} $r_{m,\omega}^{(c,i)} = r_m(x_{0,\omega}^{(c,i)}, c)$ for each $m = 1, \dots, M$}
    \EndFor

    \State \textcolor{gray}{\textit{// 2. {\color{blue} Late-Scalarization: Per-Channel} Group Normalization}}
    \For{each prompt $c \in \mathcal{C}$}
        \For{{\color{blue} objective $m = 1, \dots, M$}}
            \State $\mu_{\color{blue} m,\omega} \leftarrow \mathrm{mean}(\{r_{\color{blue} m,\omega}^{(c,\cdot)}\}), \quad \sigma_{\color{blue} m,\omega} \leftarrow \mathrm{std}(\{r_{\color{blue} m,\omega}^{(c,\cdot)}\})$
            \State $A_{\color{blue} m,\omega}^{(c,i)} \leftarrow (r_{\color{blue} m,\omega}^{(c,i)} - \mu_{\color{blue} m,\omega})\, /\, (\sigma_{\color{blue} m,\omega} + \epsilon)$ \quad for each $i$
        \EndFor
    \EndFor

    \State \textcolor{gray}{\textit{// 3. Policy Update}}
    \State $\theta_{\mathrm{old}} \leftarrow \theta$ \quad \textcolor{gray}{\textit{// Snapshot for implicit velocities}}
    \For{epoch $= 1, \dots, N$}
        \For{each sample $(c, i)$}
            \State \textcolor{gray}{\textit{// a. {\color{blue} Per-objective} interpolation weights}}
            \For{{\color{blue} $m = 1, \dots, M$}}
                \State $\rho_{\color{blue} m,\omega}^{(c,i)} \leftarrow 0.5 + 0.5 \cdot \mathrm{clip}(A_{\color{blue} m,\omega}^{(c,i)} / \epsilon_{\mathrm{clip}},\; -1,\; 1)$
            \EndFor

            \State \textcolor{gray}{\textit{// b. Flow matching with implicit velocity steering}}
            \State Sample $t \sim \mathcal{U}(0,1)$, \; $\xi \sim \mathcal{N}(\mathbf{0}, \mathbf{I})$
            \State $x_t \leftarrow (1 - t)\, x_{\color{blue} 0,\omega}^{(c,i)} + t\, \xi$, \quad $v \leftarrow \xi - x_{\color{blue} 0,\omega}^{(c,i)}$
            \State $v_+ \leftarrow (1 - \beta)\, v_{\theta_{\mathrm{old}}}(x_t, c {\color{blue}, \omega^{(c)}}, t) + \beta\, v_\theta(x_t, c {\color{blue}, \omega^{(c)}}, t)$
            \State $v_- \leftarrow (1 + \beta)\, v_{\theta_{\mathrm{old}}}(x_t, c {\color{blue}, \omega^{(c)}}, t) - \beta\, v_\theta(x_t, c {\color{blue}, \omega^{(c)}}, t)$

            \State \textcolor{gray}{\textit{// c. {\color{blue} Per-objective losses, scalarized by preference}}}
            \For{{\color{blue} $m = 1, \dots, M$}}
                \State $\mathcal{L}_{\color{blue} m} \leftarrow \rho_{\color{blue} m,\omega}^{(c,i)} \| v_+ - v \|_2^2 + (1 - \rho_{\color{blue} m,\omega}^{(c,i)}) \| v_- - v \|_2^2$
            \EndFor
            \State $\mathcal{L}_{\mathrm{policy}} \leftarrow {\color{blue} \sum_{m=1}^{M} \omega_m^{(c)} \cdot} \mathcal{L}_{\color{blue} m}$
            \State $\mathcal{L}_{\mathrm{total}} \leftarrow \mathcal{L}_{\mathrm{policy}} + \lambda_{\mathrm{KL}} \| v_\theta(x_t, c {\color{blue}, \omega^{(c)}}, t) - v_{\mathrm{ref}}(x_t, c, t) \|_2^2$
            \State $\theta \leftarrow \theta - \eta\, \nabla_\theta \mathcal{L}_{\mathrm{total}}$
        \EndFor
    \EndFor
    \State $\theta_{\mathrm{old}} \leftarrow \mathrm{EMA}(\theta_{\mathrm{old}}, \theta)$
\EndWhile
\end{algorithmic}
\end{algorithm*}

\vspace{-5pt}
\subsection{Preference Sampling}
\label{app:preference_sampling}
During training, we sample $K$ images per conditioning signal and preference vector $\omega$. For text-to-image and text-to-video tasks, this conditioning signal is a text prompt, while for image-to-image tasks, it is a fixed pair of a source image and an edit instruction.

To sample $\omega$, we draw from a symmetric Dirichlet distribution, $\mathrm{Dir}(1, \dots, 1)$. However, because a continuous distribution has zero probability of sampling the exact boundaries of the simplex, we explicitly force the selection of these critical regions. With a fixed probability, we override the interior sample with either a vertex (a one-hot vector) or an edge (a $\mathrm{Dir}(1, 1)$ mixture over exactly two randomly chosen objectives when $M>2$). This structured sampling guarantees comprehensive coverage of the entire multi-objective trade-off space. Finally, to maintain synchronization across distributed workers, the preference sampling is strictly deterministic for a given prompt and training step.

\begin{table*}[t]
\centering
\caption{\small Example samples from our FFHQ-based image-editing dataset. Each sample contains an edit instruction, a task type, and the source caption from which the instruction was derived.}
\label{tab:editing_dataset_examples}
\resizebox{\linewidth}{!}{%
\begin{tabular}{p{5.6cm} p{2.4cm} p{5.4cm}}
\toprule
\textbf{Instruction} & \quad \textbf{Task type} & \textbf{Source caption} \\
\midrule
Convert this photograph into a street graffiti mural
&\quad
style transfer
&
a photography of a man and woman taking a selfie
\\
\midrule
Change the hair color to silver
&\quad
hair changes
&
a photography of a man talking to another man in a room
\\
\midrule
Place this person in a vintage Parisian cafe
&\quad
background
&
a photography of a woman with a blue umbrella smiling
\\
\bottomrule
\end{tabular}%
}
\end{table*}

\begin{table*}[t]
\centering
\caption{\small Example prompts from the text-to-video training corpus.}
\label{tab:t2v_prompt_examples}
\begin{tabular}{p{10.8cm}}
\toprule
\textbf{Prompt} \\
\midrule
A sleek spaceship drifting silently through an asteroid field with distant stars in the background. \\
\midrule
A street food vendor flipping crispy crepes on a hot griddle at a bustling night market. \\
\midrule
A traceur performing a wall flip off a brick building in slow motion. \\
\bottomrule
\end{tabular}
\vspace{-10pt}
\end{table*}

\vspace{-5pt}
\subsection{Reward Functions}
\label{app:rewards}
\paragraph{\textbf{Text-to-Image.}}
For text-to-image post-training, we derive our training reward signals from off-the-shelf reward models. We categorize our evaluators into three types: First, to measure general image-text alignment and quality, we utilize PickScore \cite{kirstain2023pickapicopendatasetuser}. Second, for abstract stylistic attributes, such as watercolor, animation etc., we prompt Vision-Language Models (VLMs) like Qwen2.5-VL \cite{bai2023qwen}. Finally, for highly structured styles like sketch rendering, we employ domain classification models via SigLIP \cite{zhai2023sigmoid}.

Our main ablations focus on a two-objective photorealism-versus-sketch setting, which provides a particularly clear trade-off because improving one objective often degrades the other. We train with the PickScore \cite{kirstain2023pickapicopendatasetuser} reward using the prompt: \textit{``A photorealistic, high quality, 4K, camera-captured snapshot of [prompt].''} and the structured sketch reward. 
To ensure the robustness of the quantitative evaluation in Fig.\ref{fig:ablation_condition} and avoid potential over-optimization artifacts, we validate our framework using a diverse set of evaluation metrics distinct from those guiding the training process.
Specifically, we use Qwen2.5-VL \cite{bai2023qwen} for the sketch style and CLIPScore \cite{hessel2021clipscore} for photorealism.
The use of other metrics than those used during training, tests whether the learned controllable frontier reflects genuine behavioral change rather than overfitting to the specific reward used during optimization.

For the CLIP-based photorealism evaluations, we append a prefix to the base prompt: \textit{``A photorealistic, high quality, 4K, camera-captured snapshot of [prompt].''} For the Qwen VLM sketch evaluations, we query the model with the following zero-shot evaluation template in \ref{box:qwensketch}.

\vspace{-5pt}
\paragraph{\textbf{Image-to-Image.}}
For instruction-based image editing, the fundamental multi-objective trade-off lies between executing the requested edit (instruction adherence) and maintaining the fidelity of the unedited regions of the source image (preservation). 

To quantify these attributes during training, we employ Qwen2.5-VL-based editing rewards \cite{bai2023qwen} that can assess both the success of the edit. The model is queried with both the source and edited images using the evaluation prompt presented in \ref{box:edit}.
To measure preservation we employ CLIP image-to-image cosine similarity reward between the source and edited images:
$$\text{Similarity}_{\text{CLIP}} = \frac{\phi(x_{\text{edit}}) \cdot \phi(x_{\text{src}})}{\|\phi(x_{\text{edit}})\| \|\phi(x_{\text{src}})\|}$$
where $\phi$ represents the CLIP visual encoder. This heavily penalizes unnecessary modifications to the background or unrelated elements. 
\\

Finally, to ensure the recovered controllable frontier reflects genuine editing capabilities rather than exploitation of the training rewards, our robustness evaluations rely on an orthogonal, held-out evaluator. Specifically, we utilize VIEScore \cite{ku2024viescore} (Visual Instruction-guided Explainable Score) powered by GPT-4o \cite{hurst2024gpt}. 

\vspace{-5pt}
\paragraph{\textbf{Text-to-Video.}}
For text-to-video post-training, we define two competing style objectives using UnifiedReward-2.0~\cite{wang2025unified}, a Qwen2.5-VL-based 7B \cite{bai2023qwen} vision-language model. Eight frames are uniformly sampled from each generated video and passed as images to the VLM together with a rubric-style prompt that evaluates both style conformity and content alignment on a 0--5 integer scale, which is then normalized to a continuous $[0,1]$ score. Both rewards share the same underlying model and only the evaluation prompt differs.
The \emph{photorealism} reward evaluates whether the sampled frames resemble real camera footage. The \emph{animation} reward evaluates whether the frames exhibit the stylistic hallmarks of 3D animated films. Together, these two objectives define a clear stylistic trade-off: improving photorealism typically degrades the animation score and vice versa, making them a natural pair for evaluating controllable multi-objective video generation. 

\vspace{-5pt}
\subsection{Datasets}
\label{app:datsets}

\paragraph{\textbf{Text-to-Image.}}
For text-to-image generation, we use prompt-only datasets. To ensure a direct and fair comparison with our primary baseline, we utilized the same Pickscore dataset as used in DiffusionNFT.

\vspace{-12pt}
\paragraph{\textbf{Image-to-Image.}}
For instruction-based image editing, we construct a custom instruction set derived from the FFHQ-512 captions \cite{karras2019style}. We utilize Claude 4.6 Opus to parse each source caption for semantic facial attributes and subsequently generate diverse, contextually appropriate edit instructions. Each generated sample records the instruction, the source image index, and the specific edit category, see Table~\ref{tab:editing_dataset_examples} for representative examples. 
Additionally, Figure~\ref{fig:i2i_results_supp} presents results of our model trained on the general editing dataset, EditScore~\cite{luo2025editscore}.

\vspace{-10pt}
\paragraph{\textbf{Text-to-Video.}}
For text-to-video post-training, we use a prompt-only corpus of 1{,}000 prompts generated with Claude 4.6 Opus. The prompts are diverse and medium-short in length, covering a broad range of scenes, entities, and motion patterns. We present few examples in Table \ref{tab:t2v_prompt_examples}.

\vspace{-5pt}
\section{Theoretical Analysis}
\label{app:theory}
\vspace{-5pt}

This section formalizes two design choices behind \methodname{}'s preference-weighted training objective. At a high level, the question is where the preference vector $\omega$ should enter the learning process. 
The distinction matters because the DiffusionNFT update is not a linear function of the reward advantage. The advantage is passed through a clipped steering map $\rho$ that determines how strongly the update moves toward the reward-favored velocity target versus the disfavored one. Since $\rho$ is piecewise affine rather than globally affine, averaging advantages before applying $\rho$ is generally not the same as applying $\rho$ to each objective and averaging the resulting losses. Proposition~\ref{prop:nft-noncommute} makes this non-equivalence explicit and gives a concrete numerical example.
A second concern is that early scalarization can distort the intended meaning of $\omega$ even before $\rho$ enters the picture. If raw reward channels have different variances, normalizing only after taking a weighted sum does not preserve the requested relative weights on the standardized reward signals. High-variance rewards receive larger effective coefficients, while low-variance rewards are attenuated. Proposition~\ref{prop:earlyvslate} makes this distortion quantitative. Standardizing each channel before aggregation, as in Eq.~\ref{eq:w_nft}, removes this variance-induced bias and lets $\omega$ act directly on comparable per-objective training signals.

Finally, we clarify what can and cannot be concluded from weighted-sum scalarization. The classical weighted-sum argument (Theorem~\ref{thm:scalarization_pareto}) implies Pareto optimality of any exact global optimum of the scalarized reward objective when the weights are
strictly positive. The same argument applies to global minimizers of the scalarized surrogate loss (Theorem~\ref{thm:loss_scalarization_pareto}), but
only in the surrogate-loss space. Theorem~\ref{thm:conditional_reward_pareto} states sufficient alignment and near-optimality assumptions
under which an approximate reward-space guarantee follows.

\begin{proposition}[Early scalarization and late loss aggregation are non-equivalent under DiffusionNFT]
\label{prop:nft-noncommute}
Let $P^{(i)}=\|v_+^{(i)}-v^{(i)}\|_2^2$ and $N^{(i)}=\|v_-^{(i)}-v^{(i)}\|_2^2$. Define the early-scalarized DiffusionNFT loss as
\begin{equation}
\mathcal{L}_{\mathrm{early}}^{(i)}(\omega)
= \rho(A_\omega^{(i)})\,P^{(i)}
+ \bigl(1-\rho(A_\omega^{(i)})\bigr)\,N^{(i)},
\qquad
A_\omega^{(i)} = \sum_m \omega_m A_m^{(i)}.
\end{equation}
Let $\mathcal{L}_{\mathrm{late}}^{(i)}(\omega)$ be the late-aggregation loss from Equation~\ref{eq:w_nft}. Then
\begin{equation}
\mathcal{L}_{\mathrm{late}}^{(i)}(\omega)
- \mathcal{L}_{\mathrm{early}}^{(i)}(\omega)
=
\Bigl[\,\sum_m \omega_m\, \rho(A_m^{(i)})
       - \rho\!\bigl(\textstyle\sum_m \omega_m A_m^{(i)}\bigr)\Bigr]
\bigl(P^{(i)}-N^{(i)}\bigr).
\label{eq:late_minus_early}
\end{equation}
The two losses coincide when $P^{(i)} = N^{(i)}$ or when $\sum_m \omega_m \rho(A_m^{(i)}) = \rho(\sum_m \omega_m A_m^{(i)})$. The second condition holds, for example, when all $\{A_m^{(i)}\}_m$ and their convex combination $A_\omega^{(i)}$ lie inside a common affine piece of $\rho$. Outside such cases, early scalarization and late loss aggregation generally produce different update directions.
\end{proposition}

\begin{proof}
The per-objective DiffusionNFT loss is
\[
\mathcal{L}_m^{(i)}
=
\rho(A_m^{(i)})\,P^{(i)}
+
\bigl(1 - \rho(A_m^{(i)})\bigr)\,N^{(i)} .
\]
Therefore,
\[
\mathcal{L}_{\mathrm{late}}^{(i)}(\omega)
=
\sum_m \omega_m \mathcal{L}_m^{(i)}
=
\sum_m \omega_m
\Bigl[
\rho(A_m^{(i)})\,P^{(i)}
+
\bigl(1-\rho(A_m^{(i)})\bigr)\,N^{(i)}
\Bigr].
\]
Since $P^{(i)}$ and $N^{(i)}$ do not depend on $m$, and $\sum_m \omega_m = 1$, we can collect terms:
\[
\mathcal{L}_{\mathrm{late}}^{(i)}(\omega)
=
\bar\rho_{\mathrm{late}}^{(i)}\, P^{(i)}
+
\bigl(1 - \bar\rho_{\mathrm{late}}^{(i)}\bigr)\, N^{(i)},
\qquad
\bar\rho_{\mathrm{late}}^{(i)}
:=
\sum_m \omega_m \rho(A_m^{(i)}).
\]
Subtracting $\mathcal{L}_{\mathrm{early}}^{(i)}(\omega)$ gives
\[
\begin{aligned}
\mathcal{L}_{\mathrm{late}}^{(i)}(\omega)
-
\mathcal{L}_{\mathrm{early}}^{(i)}(\omega)
&=
\bigl[
\bar\rho_{\mathrm{late}}^{(i)}
-
\rho(A_\omega^{(i)})
\bigr] P^{(i)}
-
\bigl[
\bar\rho_{\mathrm{late}}^{(i)}
-
\rho(A_\omega^{(i)})
\bigr] N^{(i)}
\\
&=
\bigl[
\bar\rho_{\mathrm{late}}^{(i)}
-
\rho(A_\omega^{(i)})
\bigr]
\bigl(P^{(i)} - N^{(i)}\bigr),
\end{aligned}
\]
which is Eq.~\ref{eq:late_minus_early}. The bracket vanishes when $\rho$ is affine on the convex hull of $\{A_1^{(i)}, \dots, A_M^{(i)}\}$. It is generically nonzero when the involved advantages span multiple affine pieces of $\rho$, such as when some objectives are clipped while their weighted average or other objectives remain unclipped.
\end{proof}

To make the gap concrete, take $M = 2$, $\omega = (\tfrac{1}{2}, \tfrac{1}{2})$, $\epsilon_{\mathrm{clip}} = 1$, $A_1^{(i)} = 2$, and $A_2^{(i)} = 0$.
Channel $1$ is saturated, so $\rho(A_1^{(i)}) = 1$, while $\rho(A_2^{(i)}) = \tfrac{1}{2}$. Hence $\sum_m \omega_m \rho(A_m^{(i)}) = \tfrac{3}{4}$. The mixed advantage is $A_\omega^{(i)} = 1$, which gives $\rho(A_\omega^{(i)}) = 1$. Substituting into Eq.~\ref{eq:late_minus_early} yields:
\[
\mathcal{L}_{\mathrm{late}}^{(i)}
-
\mathcal{L}_{\mathrm{early}}^{(i)}
=
\tfrac{1}{4}(N^{(i)} - P^{(i)}),
\]
which is nonzero whenever the two flow-matching residuals disagree. This regime, where one objective is saturated and the other is not, is exactly where the choice of aggregation matters most. The behavior also distinguishes our formulation from the LLM variants MO-GRPO~\cite{ichihara2025mogrpo},
GDPO~\cite{liu2026gdpo}, and Panacea~\cite{zhong2024panaceaparetoalignmentpreference}, which aggregate normalized advantages before the policy update. In contrast, \methodname{} keeps the channels separate through the clipped steering map and aggregates only the resulting per-objective losses.

\begin{proposition}[Early scalarization distorts the preference ratio when reward channels have unequal variances]
\label{prop:earlyvslate}
Let $\Sigma$ be the within-group reward covariance and assume $\sigma_m = \sqrt{\Sigma_{mm}} > 0$ for every channel $m$. The early-scalarized advantage:
\[
A^{(i)}_{\mathrm{early}}(\omega)
=
\frac{\omega^{\top}(\mathbf r^{(i)} - \mu)}
{\sqrt{\omega^{\top}\Sigma\omega} + \epsilon}
\]
admits the decomposition
\begin{equation}
A^{(i)}_{\mathrm{early}}(\omega)
=
\sum_{m=1}^{M}
\alpha_m(\omega, \Sigma)\,
\frac{r_m^{(i)} - \mu_m}{\sigma_m},
\qquad
\alpha_m(\omega, \Sigma)
=
\frac{\omega_m \sigma_m}
{\sqrt{\omega^{\top}\Sigma\omega} + \epsilon}.
\label{eq:effective_weight}
\end{equation}
Consequently, for any two channels $m,n$ with $\omega_m,\omega_n > 0$,
\[
\frac{\alpha_m(\omega, \Sigma)}
{\alpha_n(\omega, \Sigma)}
=
\frac{\omega_m \sigma_m}
{\omega_n \sigma_n}.
\]
Thus the requested ratio $\omega_m / \omega_n$ is preserved on the standardized signal only when $\sigma_m = \sigma_n$. Otherwise, the effective ratio is distorted by a factor of $\sigma_m / \sigma_n$, which amplifies high-variance channels and attenuates low-variance ones.
\end{proposition}

\begin{proof}
Let $\tilde r^{(i)} = \mathbf r^{(i)} - \mu$. Then
\[
\begin{aligned}
A^{(i)}_{\mathrm{early}}(\omega)
&=
\frac{\omega^\top \tilde r^{(i)}}
{\sqrt{\omega^\top \Sigma \omega} + \epsilon}
\\
&=
\sum_{m=1}^M
\frac{\omega_m \sigma_m}
{\sqrt{\omega^\top \Sigma \omega} + \epsilon}
\cdot
\frac{\tilde r_m^{(i)}}{\sigma_m}
\\
&=
\sum_{m=1}^M
\alpha_m(\omega, \Sigma)
\cdot
\frac{r_m^{(i)} - \mu_m}{\sigma_m},
\end{aligned}
\]
which proves Eq.~\ref{eq:effective_weight}. The ratio
\[
\frac{\alpha_m(\omega, \Sigma)}
{\alpha_n(\omega, \Sigma)}
=
\frac{\omega_m \sigma_m}
{\omega_n \sigma_n}
\]
follows by cancelling the common normalizer.
\end{proof}

This variance-induced reweighting provides a simple mechanism for the reward-imbalance or reward-hijacking behavior observed with early scalarization in multi-objective RL~\cite{ichihara2025mogrpo, liu2026gdpo,zhong2024panaceaparetoalignmentpreference}. By standardizing each channel before aggregation, the late loss aggregation update in Eq.~\ref{eq:w_nft} lets $\omega$ act directly on comparable per-objective training signals.

\begin{theorem}[Scalarized optima are Pareto optimal in reward space]
\label{thm:scalarization_pareto}
Let $\Pi$ be the feasible policy class and let $J(\pi) = \big(J_1(\pi),\ldots,J_M(\pi)\big)^\top$ denote the vector of expected returns for a policy $\pi \in \Pi$. For $\omega \in \Delta^{M-1}$, define
\[
J_\omega(\pi)
=
\sum_{m=1}^M \omega_m J_m(\pi).
\]
If $\omega_m > 0$ for all $m$, then any maximizer
\[
\pi^\star_\omega
\in
\arg\max_{\pi \in \Pi} J_\omega(\pi)
\]
is Pareto optimal. If some entries of $\omega$ are zero, then $\pi^\star_\omega$ is at least weakly Pareto optimal.
\end{theorem}

\begin{proof}
Suppose first that $\omega_m > 0$ for all $m$, and assume for contradiction that $\pi^\star_\omega$ is not Pareto optimal. Then there exists $\pi' \in \Pi$ with
\[
J_m(\pi') \ge J_m(\pi^\star_\omega)
\quad
\text{for every } m,
\]
and
\[
J_j(\pi') > J_j(\pi^\star_\omega)
\quad
\text{for at least one } j.
\]
Multiplying each inequality by the corresponding positive weight and summing over $m$ gives
\[
J_\omega(\pi')
>
J_\omega(\pi^\star_\omega),
\]
which contradicts the optimality of $\pi^\star_\omega$.

If some entries of $\omega$ are zero, suppose instead that there exists $\pi' \in \Pi$ with
\[
J_m(\pi') > J_m(\pi^\star_\omega)
\quad
\text{for every } m.
\]
Since at least one $\omega_m$ is strictly positive, multiplying by the weights and summing again yields
\[
J_\omega(\pi')
>
J_\omega(\pi^\star_\omega),
\]
again contradicting optimality. Hence $\pi^\star_\omega$ is weakly Pareto optimal.
\end{proof}

This is the classical weighted-sum scalarization result in multi-objective optimization~\cite{miettinen1999nonlinear}, used recently in preference-conditioned LLM alignment~\cite{zhong2024panaceaparetoalignmentpreference}. We restate it here as a preliminary result on which we base our claims below.

We also state the corresponding result for the loss-based surrogate that DiffusionNFT actually optimizes. The conclusion concerns Pareto optimality
with respect to the per-objective surrogate losses, not directly with respect to the original expected returns. A policy is Pareto optimal in loss
space if no other feasible policy can reduce one objective's loss without increasing at least one other objective's loss.

\begin{theorem}[Scalarized loss minimizers are Pareto optimal in loss space]
\label{thm:loss_scalarization_pareto}
Let $\Pi$ be the feasible policy class and let
\[
\mathcal L(\pi)
=
(\mathcal L_1(\pi), \ldots, \mathcal L_M(\pi))^\top
\]
denote the vector of per-objective surrogate losses. For $\omega \in \Delta^{M-1}$, define
\begin{equation}
\mathcal L_\omega(\pi)
=
\sum_{m=1}^M
\omega_m \mathcal L_m(\pi).
\end{equation}
If $\omega_m > 0$ for all $m$, then any global minimizer
\begin{equation}
\pi_\omega^\star
\in
\arg\min_{\pi \in \Pi}
\mathcal L_\omega(\pi)
\end{equation}
is Pareto optimal with respect to minimizing $(\mathcal L_1, \ldots, \mathcal L_M)$. If some entries of $\omega$ are zero, then $\pi_\omega^\star$ is weakly Pareto optimal in loss space.
\end{theorem}

\begin{proof}
Suppose first that $\omega_m > 0$ for all $m$, and assume for contradiction that $\pi^\star_\omega$ is not Pareto optimal in loss space. Then there exists $\pi' \in \Pi$ with
\[
\mathcal L_m(\pi')
\le
\mathcal L_m(\pi^\star_\omega)
\quad
\text{for every } m,
\]
and
\[
\mathcal L_j(\pi')
<
\mathcal L_j(\pi^\star_\omega)
\quad
\text{for at least one } j.
\]
Multiplying each inequality by the corresponding positive weight and summing over $m$ gives
\[
\mathcal L_\omega(\pi')
<
\mathcal L_\omega(\pi^\star_\omega),
\]
which contradicts the global optimality of $\pi^\star_\omega$.

If some entries of $\omega$ are zero, suppose instead that there exists
$\pi' \in \Pi$ with
\[
\mathcal L_m(\pi')
<
\mathcal L_m(\pi^\star_\omega)
\quad
\text{for every } m.
\]
Since at least one $\omega_m$ is strictly positive, multiplying by the weights and summing again yields
\[
\mathcal L_\omega(\pi')
<
\mathcal L_\omega(\pi^\star_\omega),
\]
again contradicting optimality. Hence $\pi^\star_\omega$ is weakly Pareto optimal in loss space.
\end{proof}

\begin{remark}
\label{rmk:loss-vs-reward}
Theorem~\ref{thm:loss_scalarization_pareto} concerns only the surrogate losses optimized by DiffusionNFT. It does not, by itself, imply Pareto optimality with respect to the original expected reward vector $J(\pi)$. That conclusion would require an additional alignment condition ensuring that reducing $\mathcal L_m$ corresponds to increasing $J_m$, for instance a monotone relationship between $\mathcal L_m$ and $J_m$. We do not establish this condition in generality. Instead, we rely on Theorem~\ref{thm:scalarization_pareto} together with the empirical Pareto sweeps in Section~\ref{sec:experiments} as the practical justification for \methodname{}'s reward-space behavior. We leave a formal alignment analysis to future work.
\end{remark}

\begin{theorem}[Conditional approximate Pareto optimality in reward space]
\label{thm:conditional_reward_pareto}
Let $\Pi_\kappa$ denote the feasible policy class under a fixed KL-divergence budget $\kappa$ from the reference model. For each objective $m$, let $\mathcal{L}_m(\pi)$ denote the population DiffusionNFT surrogate loss under the same prompt distribution used to define $J_m(\pi)$.

Assume that, for every $\pi\in\Pi_\kappa$,
\begin{equation}
    \mathcal{L}_m(\pi)=f_m(J_m(\pi)),
    \qquad m=1,\ldots,M,
\end{equation}
where each $f_m$ is strictly decreasing and its inverse $f_m^{-1}$ is $K$-Lipschitz on the attainable loss range.

Fix a preference vector $\omega$ satisfying $\sum_{m=1}^{M}\omega_m=1$ and $\omega_m\geq\gamma>0$ for every $m$, and define
\begin{equation}
    \mathcal{L}_\omega(\pi)
    =\sum_{m=1}^{M}\omega_m\mathcal{L}_m(\pi).
\end{equation}
Suppose the learned conditional policy $\hat{\pi}_\omega\in\Pi_\kappa$ satisfies
\begin{equation}
    \mathcal{L}_\omega(\hat{\pi}_\omega)
    \leq
    \inf_{\pi\in\Pi_\kappa}\mathcal{L}_\omega(\pi)
    +\varepsilon,
    \qquad \varepsilon\geq 0.
    \label{eq:surrogate_near_optimality}
\end{equation}

Then any $\pi\in\Pi_\kappa$ satisfying $J_m(\pi)\geq J_m(\hat{\pi}_\omega)$ for every $m$ also satisfies
\begin{equation}
    0
    \leq J_m(\pi)-J_m(\hat{\pi}_\omega)
    \leq \frac{K\varepsilon}{\omega_m}
    \leq \frac{K\varepsilon}{\gamma},
    \qquad m=1,\ldots,M.
    \label{eq:reward_pareto_bound}
\end{equation}
Consequently, no feasible policy can weakly improve all rewards while improving any objective by more than $\delta=K\varepsilon/\gamma$. When $\varepsilon=0$, $\hat{\pi}_\omega$ is Pareto optimal in reward space within $\Pi_\kappa$.
\end{theorem}

\begin{proof}
Consider any $\pi\in\Pi_\kappa$ such that $J_m(\pi)\geq J_m(\hat{\pi}_\omega)$ for every $m$. Since each $f_m$ is strictly decreasing,
\begin{equation}
    d_m
    :=\mathcal{L}_m(\hat{\pi}_\omega)
      -\mathcal{L}_m(\pi)
    \geq 0.
\end{equation}
By Eq.~\eqref{eq:surrogate_near_optimality},
\begin{align}
    \sum_{m=1}^{M}\omega_m d_m
    &=
    \mathcal{L}_\omega(\hat{\pi}_\omega)
    -\mathcal{L}_\omega(\pi)
    \nonumber\\
    &\leq \varepsilon.
\end{align}
Because all terms are nonnegative and
$\omega_m>0$, this implies
\begin{equation}
    d_m\leq\frac{\varepsilon}{\omega_m}
    \qquad\text{for every }m.
\end{equation}
Applying the $K$-Lipschitz property of $f_m^{-1}$ gives
\begin{align}
    0
    &\leq J_m(\pi)-J_m(\hat{\pi}_\omega)
    \nonumber\\
    &=
    \left|
    f_m^{-1}(\mathcal{L}_m(\pi))
    -
    f_m^{-1}(\mathcal{L}_m(\hat{\pi}_\omega))
    \right|
    \nonumber\\
    &\leq Kd_m
    \leq\frac{K\varepsilon}{\omega_m}
    \leq\frac{K\varepsilon}{\gamma}.
\end{align}
This establishes Eq.~\eqref{eq:reward_pareto_bound}. For $\varepsilon=0$, any policy that weakly improves every reward must leave every reward unchanged, which excludes strict Pareto dominance.
\end{proof}

\paragraph{Scope of the guarantee.}
Theorem~\ref{thm:conditional_reward_pareto} gives sufficient conditions for transferring surrogate-loss near-optimality to approximate reward-space non-dominance. The error $\varepsilon$ may account for optimization, estimation, and amortization errors, provided that the population near optimality condition in Eq.~\eqref{eq:surrogate_near_optimality} holds.

These assumptions are not established as general properties of DiffusionNFT. In particular, the clipped, stochastic, and on-policy surrogate is not
generally known to be a strictly decreasing function of expected reward. We therefore assess the learned reward-space trade-offs empirically in Section~5 and Appendix~D.

At simplex boundaries, where some $\omega_m=0$, the coordinate-wise bound applies only to objectives with positive weights. No such bound follows for zero-weight objectives. When $\varepsilon=0$, the policy is at least weakly Pareto optimal, since a strict improvement in every reward would
strictly decrease the weighted surrogate loss. For $\varepsilon>0$, exact weak Pareto optimality is not guaranteed.

Finally, this result establishes conditional non-dominance, not coverage of the entire Pareto frontier. Linear scalarization cannot generally recover unsupported regions of a non-convex surrogate-loss frontier.

\vspace{-5pt}
\section{Extended Related Work}
\vspace{-5pt}

\paragraph{\textbf{Offline Multi-Preference Fine-Tuning of Diffusion Models.}}
\label{app:offline_multireward_diffusion}

Offline alignment has emerged as a data-driven alternative to online RL. Relying on static pairwise comparisons rather than explicit reward models makes these methods practically appealing. Diffusion-DPO~\cite{wallace2024diffusiondpo}, for example, first adopted DPO for diffusion models by deriving likelihood surrogates from the denoising objective. However, while this reduces computational overhead, it comes at the cost of exploration: offline methods are fundamentally bounded by their training datasets and struggle to adapt to preferences unseen during training.

Of greater relevance to our work are offline methods targeting the multi-objective setting. CaPO~\cite{lee2025capo} constructs training pairs by selecting from Pareto frontiers across multiple reward models, jointly optimizing competing criteria within a single DPO objective. PPD~\cite{dang2025personalizedpreferencefinetuningdiffusion} trains a preference-conditioned diffusion model that interpolates between multiple objectives at inference time, conditioning on per-user embeddings extracted by a VLM from few-shot pairwise examples. Unlike scalar reward optimization, PPD genuinely supports multi-reward control within a single model. However, because preferences are represented as learned user identities rather than explicit reward weights, the reachable control space is limited to interpolations between seen user embeddings instead of arbitrary points on the continuous reward simplex. \methodname{}, in contrast, conditions directly on explicit preference vectors $\omega$ and trains online, actively exploring the objective space to cover a broader and more precise Pareto frontier.

\vspace{-5pt}
\section{Extended Experiments}
\label{app:experiments}
\vspace{-5pt}

\paragraph{\textbf{Evaluation protocol.}}
Unless otherwise stated, quantitative results are computed on 100 prompts randomly sampled from the test set. The standard two-objective comparisons evaluate five preference vectors. Appendix~\ref{app:dense_sweep} additionally evaluates a dense grid of 21 preference vectors. For the Fixed-Weights baseline, we train 5 separate DiffusionNFT models, each with a different fixed reward weighting, using the same hyperparameters, initialization checkpoint, and training duration of 12 epochs as \methodname{}. FlowMulti is trained for 300 steps following the official implementation details in the original paper. For the Prompt Rewriting baseline, we use Gemini 3 to generate 3 rewritten prompts per test prompt: one emphasizing photorealism, one emphasizing sketch, and one requesting a balanced blend of the two, using the templates shown in Table \ref{tab:prompt_rewriting_examples}. Similarly, for the image-to-image editing task, the rewrites emphasize prompt adherence, source preservation, or a balanced blend (see Table~\ref{tab:editing_prompt_rewriting_examples}).

\begin{figure}
    \centering
    \includegraphics[width=1\linewidth]{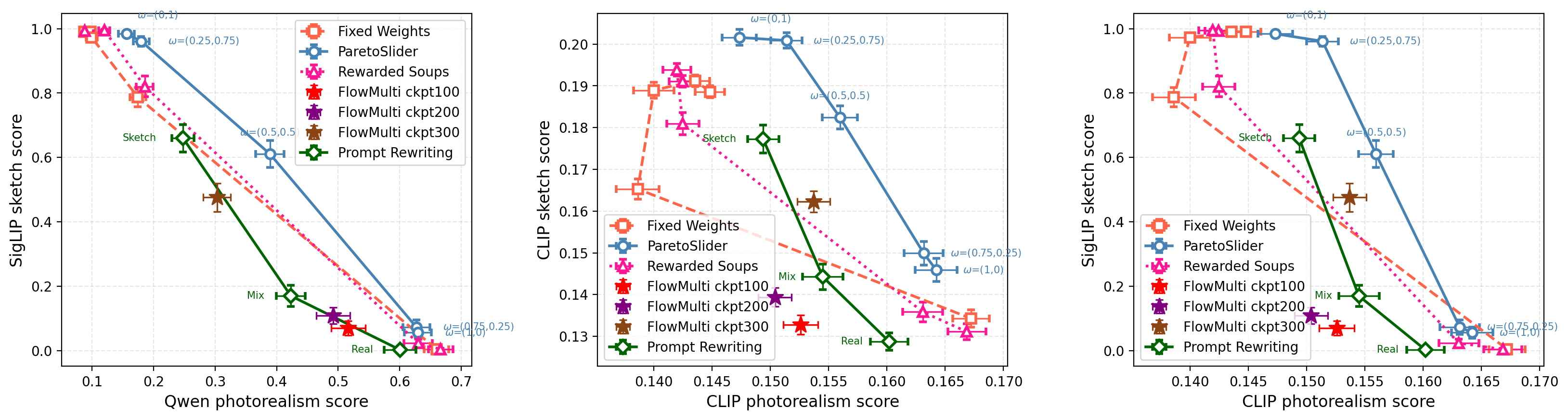}
    \caption{Robust comparison of baseline methods on text-to-image generation under alternative evaluation metrics.}
    \label{fig:baselines_other}
    \vspace{-5pt}
\end{figure}

\begin{figure}
    \centering
    \includegraphics[width=1\linewidth]{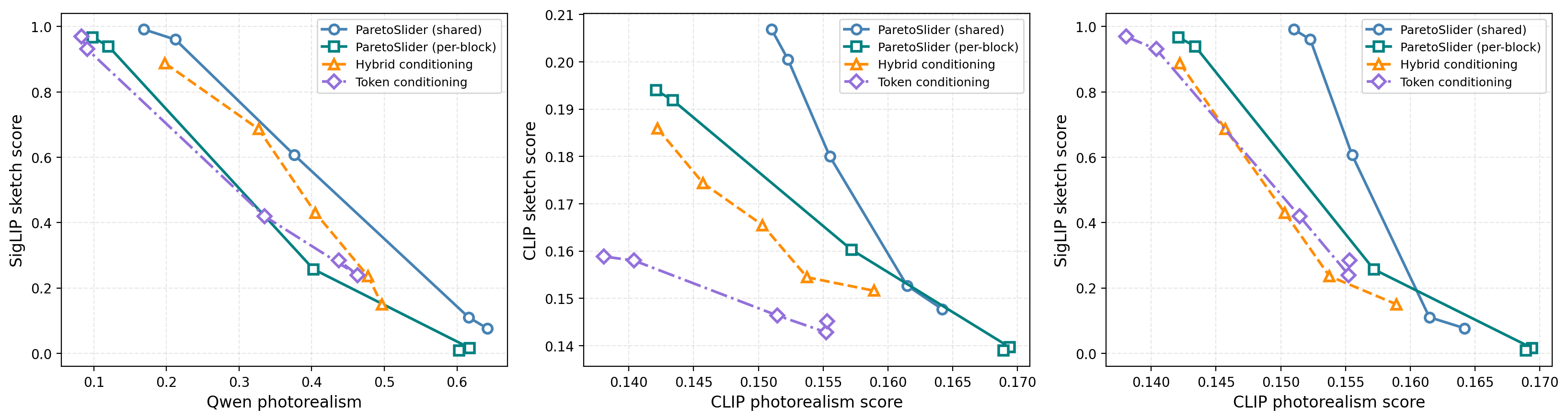}
    \caption{Robustness of conditioning architectures under alternative evaluation metrics.}
    \label{fig:cond_ablation_other}
\end{figure}

\begin{figure}
    \centering
    \includegraphics[width=1\linewidth]{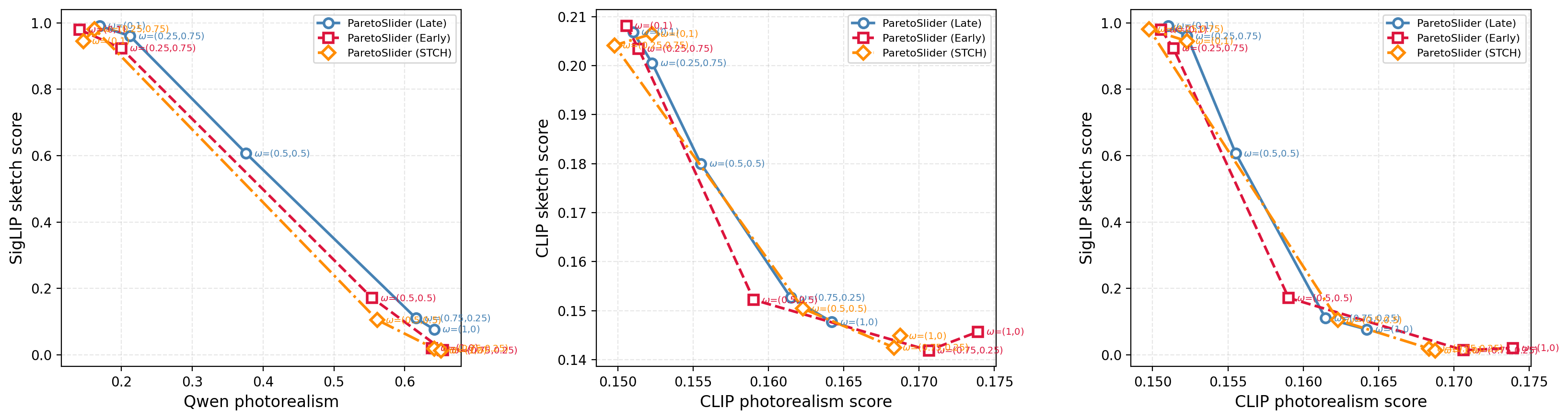}
    \caption{Robustness of scalarization strategies under alternative evaluation metrics.}
    \label{fig:loss_other}
    \vspace{-10pt}
\end{figure}

To emphasize the robustness of our method to different reward model scoring we present additional plots demonstrating that are method consistently outperforms all baselines as shown in Figures \ref{fig:baselines_other}, \ref{fig:cond_ablation_other}, \ref{fig:loss_other}. 

\begin{table*}[t]
\centering
\caption{\small Example prompt rewrites used for the Prompt Rewriting baseline in the photorealism-versus-sketch setting. For each original prompt, Gemini 3 produces a photorealistic rewrite, a sketch rewrite, and a balanced rewrite.}
\label{tab:prompt_rewriting_examples}
\resizebox{\linewidth}{!}{%
\begin{tabular}{p{3.5cm} p{4.2cm} p{4.2cm} p{4.2cm}}
\toprule
\textbf{Original prompt} & \textbf{Photorealistic rewrite} & \textbf{Sketch rewrite} & \textbf{Balanced rewrite} \\
\midrule
a young male cyborg with white hair sitting down on a throne in a dystopian world, epic
&
Young male cyborg, white hair, sitting on a throne, dystopian world, epic, photorealistic, cinematic lighting, high detail, 8k, sharp focus.
&
Sketch of a young male cyborg with white hair sitting on a throne in a dystopian world, highly detailed, epic composition.
&
A young male cyborg with white hair sitting on a throne in a dystopian world, epic, equally blended photorealistic and sketch styles.
\\
\midrule
a charcoal sketch of a city skyline
&
Photorealistic city skyline, high resolution, sharp focus, cinematic lighting, architectural detail.
&
City skyline, charcoal sketch.
&
City skyline, 50\% photorealistic and 50\% charcoal sketch.
\\
\midrule
enigmatic black square building on top of a purple hill, smoke stacks
&
Photorealistic, enigmatic black square building, purple hill summit, smoke stacks, cinematic lighting, high detail, 8k resolution.
&
Sketch of an enigmatic black square building with smoke stacks on top of a purple hill.
&
Enigmatic black square building on top of a purple hill with smoke stacks, equally blended photorealistic and sketch styles.
\\
\bottomrule
\end{tabular}%
}
\end{table*}

\begin{table*}[t]
\centering
\caption{\small Example instruction rewrites used for the Prompt Rewriting baseline in the Instruction Adherence vs.\ Image Preservation setting for image editing. For each original instruction, Gemini 3 produces a rewrite emphasizing prompt adherence, a rewrite emphasizing image preservation, and a balanced rewrite.}
\label{tab:editing_prompt_rewriting_examples}
\resizebox{\linewidth}{!}{%
\begin{tabular}{p{3.5cm} p{4.2cm} p{4.2cm} p{4.2cm}}
\toprule
\textbf{Original instruction} & \textbf{Adherence rewrite} & \textbf{Preservation rewrite} & \textbf{Balanced rewrite} \\
\midrule
Change the background to a tropical beach at sunset
&
Completely transform the background into a vibrant tropical beach at sunset, ensuring the change is bold and unmistakably clear.
&
Subtly update the background to a tropical beach at sunset while strictly preserving the original image's content and structure.
&
Change the background to a tropical beach at sunset while maintaining the original subject and composition.
\\
\midrule
Restyle this portrait with a gothic Victorian theme
&
Boldly and unmistakably restyle this portrait with an intense, fully-realized gothic Victorian theme.
&
Subtly restyle this portrait with a gothic Victorian theme while strictly preserving the original content, structure, and visual identity.
&
Restyle this portrait with a gothic Victorian theme while preserving the original person's likeness, pose, and the image's overall structure.
\\
\midrule
turn this portrait into pointillist style
&
Transform this portrait into a bold, unmistakable pointillist style with clearly visible dots throughout.
&
Gently apply a pointillist style to this portrait while strictly preserving its original content, structure, and visual identity.
&
Apply a pointillist style to this portrait while maintaining the original subject's features and overall composition.
\\
\bottomrule
\end{tabular}%
}
\end{table*}

\subsection{Qualitative Results}

Figures \ref{fig:t2i_results_supp} and \ref{fig:i2i_results_supp} show qualitative results of our continuous preference control framework. In both cases, the model smoothly adjusts the output according to the specified preference, demonstrating controllable transitions between competing objectives. In the last row of Figure \ref{fig:t2i_results_supp}, we observe that the colors gradually become more saturated, notably the color of the dress transforms from a muted gray to a saturated pink. Similarly, in the first row of Figure \ref{fig:i2i_results_supp}, we observe that higher preservation reward allows for very light edits, while higher editing reward alters the entire scene.

To strongly demonstrate the robustness of our learned Pareto frontiers, we compute the hypervolume across two distinct sets of evaluation metrics for each task. Text-to-Image is evaluated with a SigLIP classifier \cite{zhai2023sigmoid} and a combination of Qwen2.5-VL \cite{bai2023qwen} and CLIPScore \cite{hessel2021clipscore}, while Image-to-Image is evaluated using EditScore \cite{luo2025editscore}.

Consistent with the visual findings in Fig.~7 of the main paper, our \methodname{} achieves the highest HV, outperforming all baselines across all metric sets in both the text-to-image and image-to-image settings.

\subsection{Quantitative Results}
\label{app:quant_extended}

The main paper reports a single held-out metric pair (Qwen Photorealism$\times$CLIP sketch)  for clarity. Here we report the full quantitative comparison across four held-out metric pairs to verify that the rankings observed in the main paper are robust to evaluator choice. The four pairs are:
Pair A: Qwen2.5-VL photorealism $\times$ SigLIP sketch, Pair B: CLIP photo ensemble $\times$ CLIP sketch, Pair C: Qwen2.5-VL photorealism $\times$ CLIP sketch (in the main paper), Pair D: CLIP photo ensemble $\times$ SigLIP sketch.

All values are bootstrap mean $\pm$ standard error over $1000$ resamples of the $N\!=\!100$ test prompts. We omit FlowMulti from these tables: as a single-point method, it does not produce a sweep across $\omega$, so controllability, monotonicity, and calibration metrics are not defined.

\paragraph{\textbf{Hypervolume.}}
Table~\ref{tab:hv_robust_appendix} reports HV on each of the four pairs and the mean across pairs. \methodname{} achieves the highest HV on every metric pair. The ranking is preserved across pairs: Prompt Rewriting is consistently second, RewardedSoups is consistently third, and Fixed-Weights is consistently last.

\begin{table*}[t]
\centering
\small
\caption{\small Hypervolume (HV) on the SD3.5 photorealism vs.\ sketch trade-off (T2I), evaluated under four held-out evaluator pairs. All values are bootstrap mean $\pm$ standard error over $1000$ resamples of $N\!=\!100$ prompts. \methodname{} achieves the highest HV on every pair; the right column reports the mean across pairs. The Pair C column reproduces the headline result of Table~\ref{tab:hypervolume_t2i} in the main paper.}
\small
\label{tab:hv_robust_appendix}
\setlength{\tabcolsep}{1pt}
\begin{tabular}{lccccc}
\toprule
\textbf{Method} & \textbf{Pair A} $\uparrow$ & \textbf{Pair B} $\uparrow$ & \textbf{Pair C} $\uparrow$ & \textbf{Pair D} $\uparrow$ & \textbf{Mean across pairs} $\uparrow$ \\
& {\scriptsize Qwen$\times$SigLIP} & {\scriptsize CLIP$\times$CLIP} & {\scriptsize Qwen$\times$CLIP} & {\scriptsize CLIP$\times$SigLIP} & \\
\midrule
\methodname{} (ours)            & $\mathbf{0.247 \pm .015}$ & $\mathbf{0.0014 \pm .0001}$ & $\mathbf{0.024 \pm .002}$ & $\mathbf{0.017 \pm .001}$ & $\mathbf{0.072 \pm .004}$ \\
Prompt Rewriting               & $0.145 \pm .013$          & $0.0006 \pm .0001$          & $0.011 \pm .001$          & $0.008 \pm .002$          & $0.041 \pm .003$ \\
Rewarded Soups (LoRA)          & $0.109 \pm .010$          & $0.0005 \pm .0001$          & $0.010 \pm .002$          & $0.006 \pm .002$          & $0.031 \pm .003$ \\
Fixed-Weights                  & $0.085 \pm .008$          & $0.0006 \pm .0001$          & $0.007 \pm .002$          & $0.007 \pm .002$          & $0.025 \pm .002$ \\
\bottomrule
\end{tabular}
\end{table*}

\paragraph{\textbf{Monotonicity.}}
Table~\ref{tab:monotonicity_robust_appendix} reports per-pair Spearman $\rho$ averaged across the photo and sketch objectives for each metric pair. \methodname{} maintains a high monotonicity on every pair. Fixed-Weights and RewardedSoups exhibit substantial degradation on the CLIP-photo objective (Pairs B and D), where the photo-axis reward is non-monotone with respect to $\omega$.

\begin{table*}[t]
\centering
\small
\caption{\small Monotonicity (Spearman $\rho$ between requested $\omega$ and measured rewards, averaged over both objectives) on each of the four held-out evaluator pairs. Values are bootstrap mean $\pm$ standard error.}
\label{tab:monotonicity_robust_appendix}
\setlength{\tabcolsep}{6pt}
\begin{tabular}{lcccc}
\toprule
\textbf{Method} & \textbf{Pair A} $\uparrow$ & \textbf{Pair B} $\uparrow$ & \textbf{Pair C} $\uparrow$ & \textbf{Pair D} $\uparrow$ \\
\midrule
\methodname{} (ours)            & $0.971 \pm .031$ & $0.976 \pm .030$ & $0.962 \pm .035$ & $0.985 \pm .026$ \\
Prompt Rewriting               & $1.000 \pm .001$ & $1.000 \pm .000$ & $1.000 \pm .001$ & $1.000 \pm .001$ \\
Rewarded Soups (LoRA)          & $0.961 \pm .015$ & $0.855 \pm .073$ & $0.952 \pm .014$ & $0.864 \pm .074$ \\
Fixed-Weights                  & $0.955 \pm .047$ & $0.408 \pm .060$ & $0.870 \pm .088$ & $0.493 \pm .064$ \\
\bottomrule
\end{tabular}
\\[2pt]
\end{table*}

\paragraph{\textbf{Quality Preservation.}}
A common concern with preference-conditioned models is that varying the slider may degrade general image quality at intermediate operating points. Table~\ref{tab:stability_appendix} shows that the held-out quality metrics (LAION aesthetic predictor and CLIPScore, neither used during training) remain essentially flat across the $\omega$ sweep for every method. 

\begin{table*}[t]
\centering
\small
\caption{\small Quality preservation along the $\omega$ sweep. Reward range columns show min/max bootstrap-mean per-$\omega$ value (wider = the slider successfully moves the targeted reward). Quality columns show the mean across $\omega$ of held-out quality metrics not used during training, with the across-$\omega$ max prompt-level $\sigma$ in parentheses (smaller $\sigma$ indicates the slider does not destabilise quality).}
\label{tab:stability_appendix}
\setlength{\tabcolsep}{4pt}
\begin{tabular}{lcccc}
\toprule
& \multicolumn{2}{c}{\textbf{Reward range across $\omega$}} & \multicolumn{2}{c}{\textbf{Quality mean (max prompt $\sigma$)}} \\
\cmidrule(lr){2-3} \cmidrule(lr){4-5}
\textbf{Method} & Qwen-photo & Qwen-sketch & Aesthetic & CLIPScore \\
\midrule
\methodname{} (ours)            & $0.155 \to 0.630$  & $0.152 \to 0.783$  & $0.568 \pm .002$ ($\sigma \le .053$) & $0.254 \pm .002$ ($\sigma \le .050$) \\
Prompt Rewriting               & $0.247 \to 0.601$  & $0.117 \to 0.739$  & $0.594 \pm .003$ ($\sigma \le .049$) & $0.278 \pm .002$ ($\sigma \le .044$) \\
Rewarded Soups (LoRA)          & $0.088 \to 0.667$  & $0.059 \to 0.844$  & $0.580 \pm .002$ ($\sigma \le .050$) & $0.250 \pm .002$ ($\sigma \le .050$) \\
Fixed-Weights                  & $0.087 \to 0.660$  & $0.095 \to 0.859$  & $0.534 \pm .002$ ($\sigma \le .048$) & $0.244 \pm .002$ ($\sigma \le .048$) \\
\bottomrule
\end{tabular}
\end{table*}

\paragraph{\textbf{Image-to-Image Evaluation.}}
All preference-conditioned and Fixed-Weights models are trained on our FFHQ-derived editing instruction dataset (\S\ref{app:datsets}) using a Qwen2.5-VL-based editing reward for adherence and CLIP image-image cosine similarity for source preservation, and evaluated with the held-out EditScore~\cite{luo2025editscore} judge (\texttt{prompt\_following} for adherence, \texttt{consistency} for preservation). Values are bootstrap mean $\pm$ standard error over $1000$ resamples of $N\!=\!100$ prompts. 

\begin{table}[t]
\centering
\small
\caption{\small Quantitative comparison on the FluxKontext instruction adherence vs.\ source consistency trade-off (I2I), evaluated with EditScore. Values are bootstrap mean $\pm$ standard error over $1000$ resamples of $N\!=\!100$ prompts. }
\label{tab:i2i_main}
\setlength{\tabcolsep}{6pt}
\begin{tabular}{lccc}
\toprule
\textbf{Method} & \textbf{HV} $\uparrow$ & \textbf{Monotonicity} $\uparrow$ & \textbf{Smoothness} $\downarrow$ \\
\midrule
\methodname{} (ours)            &$\mathbf{0.169 \pm .023}$          & $\mathbf{0.991 \pm .034}$         & $\underline{0.219 \pm .018}$  \\
Fixed Weights                  & $\underline{0.167 \pm .024}$ & $\underline{0.988 \pm .034}$ &  $0.277 \pm .012$  \\
Text CFG                       & $0.162 \pm .023$          & $0.891 \pm .072$         &  $\mathbf{0.162 \pm .013}$\\
Prompt Rewriting               & $0.157 \pm .023$          & $0.756 \pm .304$ & $0.301 \pm .016$  \\
Image CFG                      & $0.138 \pm .022$          & $0.789 \pm .161$         &  $0.349 \pm .025$\\
\bottomrule
\end{tabular}
\\[2pt]
\end{table}

\subsection{Loss Ablations Extended}
For completeness, we recall the form of the smooth Tchebycheff (STCH) scalarization \cite{lin2024smooth} used in the loss-formulation ablation
of \S\ref{subsec:ablations} and Fig.~\ref{fig:ablation_losses}. Given per-objective NFT losses $\mathcal{L}_1, \ldots, \mathcal{L}_M$, a preference vector $\omega \in \Delta^{M}$, and a smoothing parameter $\mu > 0$, the STCH-aggregated loss is
\begin{equation}
    g^{\mu}_{\text{STCH}}\!\left(\mathcal{L}; \omega\right)
    \;=\; \mu \log\!\left( \sum_{m=1}^{M} \exp\!\left(
        \frac{\omega_m\,\mathcal{L}_m}{\mu}
    \right) \right),
    \label{eq:stch}
\end{equation}
which recovers the non-smooth Tchebycheff aggregation $\max_m \omega_m \mathcal{L}_m$ as $\mu \to 0^{+}$ and progressively softens the max as $\mu$ grows~\cite{lin2024smooth}. In the STCH variant of \methodname{}, we replace the linear late aggregation $\mathcal{L}_{\text{NFT}}^{(i)} = \sum_{m=1}^{M} \omega_m \mathcal{L}_m^{(i)}$ of Eq.~\ref{eq:w_nft} with Eq.~\ref{eq:stch} and leave every other component of the pipeline unchanged. We use $\mu = 5$ throughout the STCH ablation.

For fairness, every loss-formulation variant - early scalarization, STCH, and our late linear aggregation was trained for the 12 epochs and with identical hyperparameters and warm-start checkpoint. However, we note that STCH max-based concentration produces optimization dynamics that we assume empirically interact poorly with our preference-conditioned framework. 

\begin{figure}[t]
    \centering
    \scriptsize
    \setlength{\tabcolsep}{1pt}

    \begin{tabular}{cccccc}
        \multicolumn{5}{c}{\textit{``An easter bunny on a spring day in a field holding a basket of easter eggs''}} \\[-1pt]
        \raisebox{15pt}{\rotatebox{90}{\small Qwen2.5-VL}}
        \includegraphics[width=0.19\linewidth]{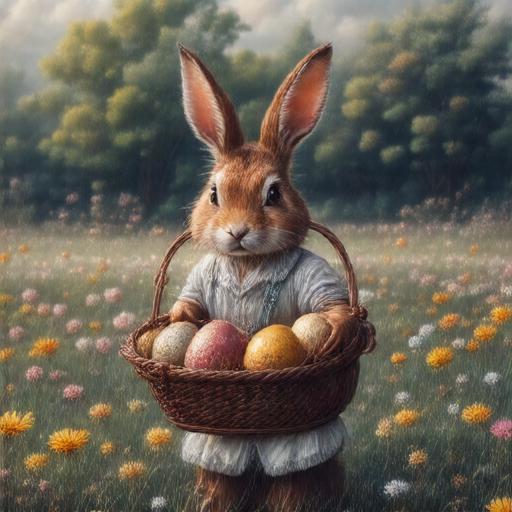} &
        \includegraphics[width=0.19\linewidth]{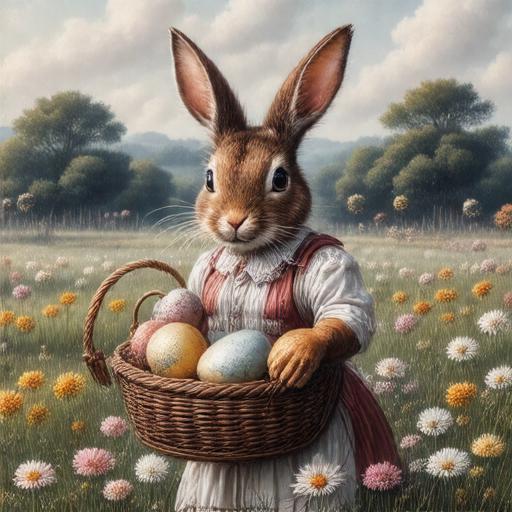} &
        \includegraphics[width=0.19\linewidth]{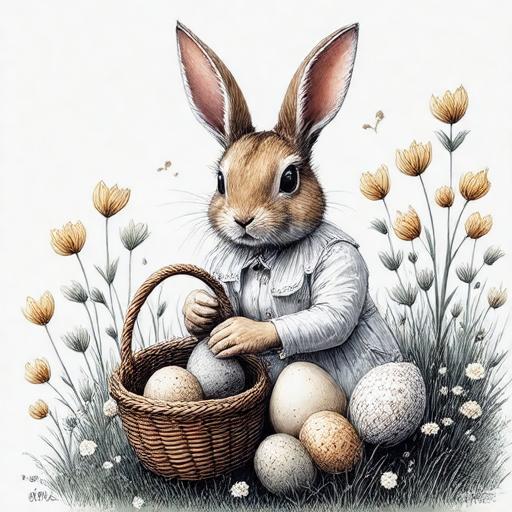} &
        \includegraphics[width=0.19\linewidth]{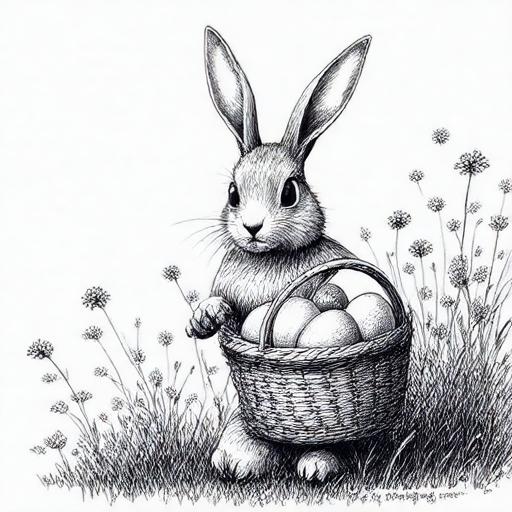} &
        \includegraphics[width=0.19\linewidth]{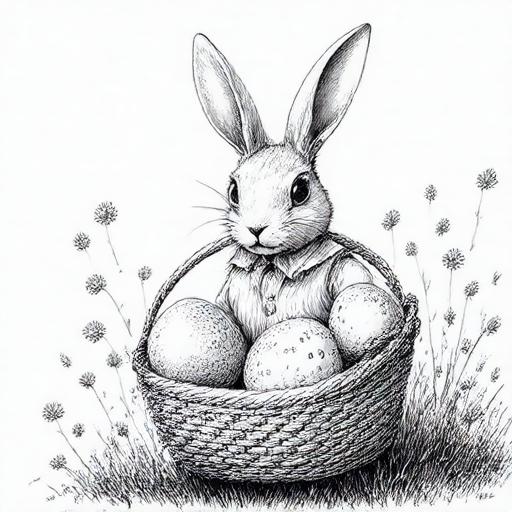} \\
        \raisebox{22pt}{\rotatebox{90}{\small SigLIP}}
        \includegraphics[width=0.19\linewidth]{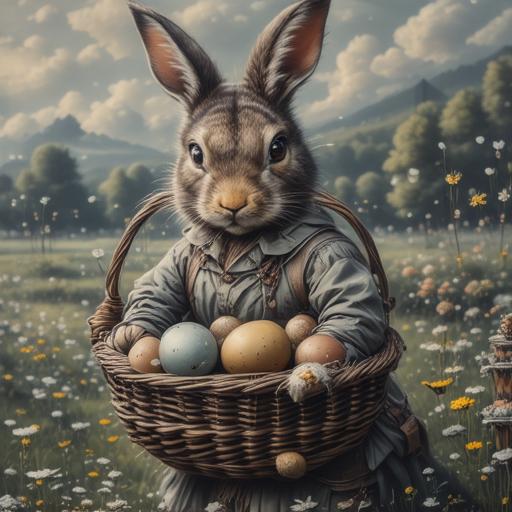} &
        \includegraphics[width=0.19\linewidth]{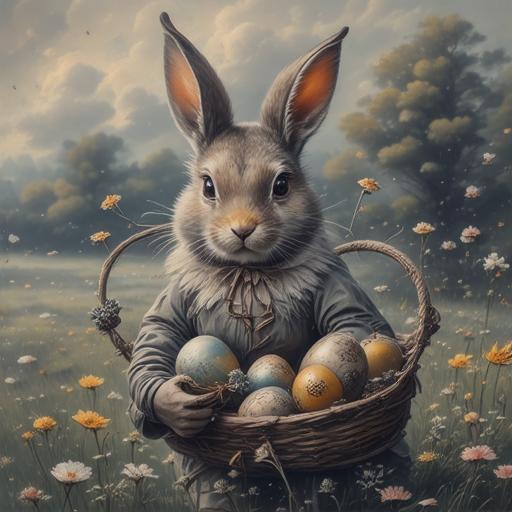} &
        \includegraphics[width=0.19\linewidth]{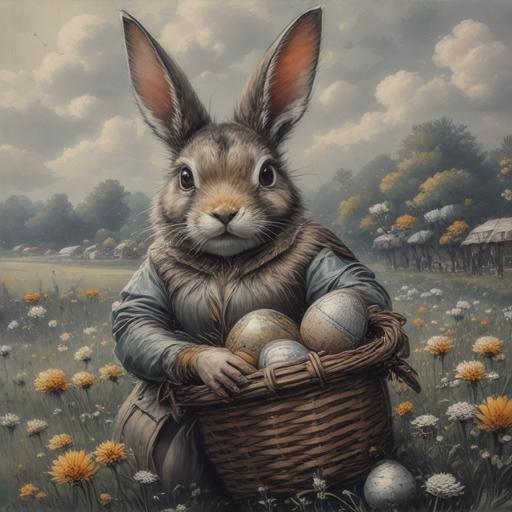} &
        \includegraphics[width=0.19\linewidth]{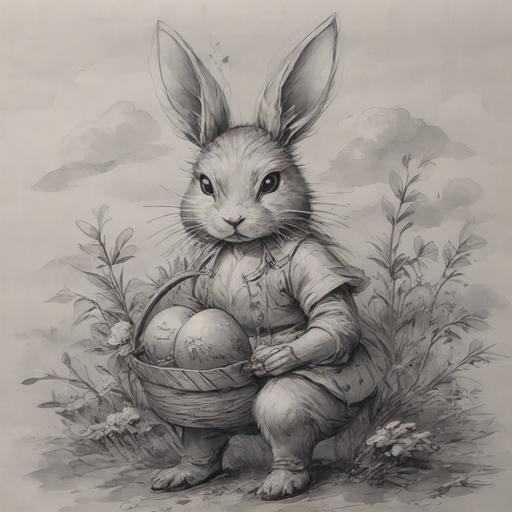} &
        \includegraphics[width=0.19\linewidth]{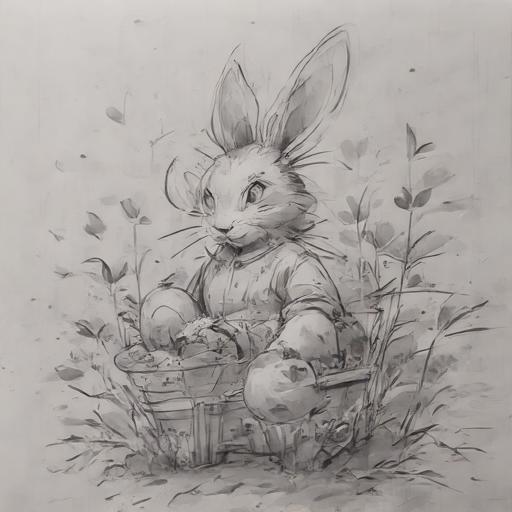} \\
        
    \end{tabular}

    \caption{\small Different sketch reward models results.}
    \label{fig:perobj_combined}
\end{figure}
\subsection{The Effect of Reward Models}

The reward model has a direct impact on the model learned during post-training. Since each reward captures a different notion of what constitutes a ``good'' image, optimizing against different rewards can lead to systematically different generations, even under the same prompt and model architecture. In practice, this means that reward choice affects not only the final score, but also the style, edit strength, realism, and semantic emphasis of the produced outputs.

Figure \ref{fig:perobj_combined} presents qualitative examples demonstrating this effect. Keeping the prompt and preference fixed, we compare outputs obtained using Qwen2.5-VL and SigLIP classifier for sketch score. 

\subsection{Dense Preference Sweeps and Diffusion Blend Comparison}
\label{app:dense_sweep}

\paragraph{Evaluation protocol.}
We evaluate the photorealism--sketch trade-off on 100 test prompts using 21 evenly spaced preference vectors, $\omega=(1-w,w)$ with $w\in\{0,0.05,\ldots,1\}$. We compare ParetoSlider with RewardedSoups and the DB-MPA variant of Diffusion Blend. Both baselines reuse the same two reward-endpoint specialists: RewardedSoups interpolates their parameters, whereas DB-MPA composes their denoisers. ParetoSlider instead learns the interior trade-offs through preference-conditioned training.

All three methods are evaluated at the same 21 preferences. Fixed-Weights specialists are evaluated only at the five preferences for which separate models were trained. We do not interpolate between these specialists. The models are trained using PickScore and Qwen sketch rewards. Here, evaluation uses Qwen photorealism and CLIP sketch scores, both held out from training.

\paragraph{Local control and perceptual smoothness.}
Table~\ref{tab:dense_control} reports global rank correlations, local monotonicity, adjacent reward changes, and adjacent LPIPS distances. Spearman correlations are signed with respect to $w$: the desired direction is negative for photorealism and positive for sketch. A monotone adjacent step follows the expected direction for the corresponding objective. Largest reversal measures the largest change against that direction. Adjacent reward changes and LPIPS characterize the magnitude of transitions between neighboring preferences.

ParetoSlider achieves the desired rank ordering to the reported precision, with 19 of 20 adjacent steps following the expected direction on each held-out objective. Its maximum adjacent LPIPS is lower than RewardedSoups ($0.18$ versus $0.25$). DB-MPA achieves a lower maximum adjacent LPIPS of $0.05$, showing that smooth transitions alone do not establish the quality of the resulting reward trade-offs.

\begin{table}[t]
\centering
\small
\setlength{\tabcolsep}{4pt}
\caption{
Control and smoothness on the 21-point preference sweep. Paired reward entries report photorealism / sketch. The final row instead reports mean / maximum adjacent LPIPS. Reward statistics use held-out Qwen photorealism and CLIP sketch scores.
}
\label{tab:dense_control}
\begin{tabular}{lccc}
\toprule
Metric & ParetoSlider & RewardedSoups & DB-MPA \\
\midrule
Signed Spearman
  & $-1.00 / +1.00$
  & $-1.00 / +0.95$
  & $-1.00 / +0.97$ \\
Monotone adjacent steps
  & $19/20 \,/\, 19/20$
  & $19/20 \,/\, 15/20$
  & $19/20 \,/\, 16/20$ \\
Largest reversal
  & $0.003 / 0.000$
  & $0.006 / 0.003$
  & $0.001 / 0.001$ \\
Maximum adjacent change
  & $0.12 / 0.015$
  & $0.16 / 0.017$
  & $0.09 / 0.007$ \\
Mean adjacent change
  & $0.024 / 0.003$
  & $0.030 / 0.004$
  & $0.029 / 0.003$ \\
Range across sweep
  & $0.48 / 0.064$
  & $0.58 / 0.068$
  & $0.58 / 0.063$ \\
Adjacent LPIPS: mean / max
  & $0.09 / 0.18$
  & $0.10 / 0.25$
  & $0.04 / 0.05$ \\
\bottomrule
\end{tabular}
\end{table}

\paragraph{Interior trade-offs.}
Table~\ref{tab:dense_scores} reports the full held-out reward sweep. At the balanced preference
$\omega=(0.5,0.5)$, ParetoSlider achieves a photorealism score of $0.50$, compared with $0.25$ for DB-MPA, while both achieve a sketch score of $0.16$ at the reported precision.
RewardedSoups and the balanced Fixed-Weights specialist achieve higher sketch scores of $0.19$, but substantially lower photorealism scores of $0.13$ and $0.10$, respectively. These results show how the methods differ in their interior trade-offs, beyond the smoothness of their preference sweeps.

\begin{table}[t]
\centering
\small
\setlength{\tabcolsep}{5pt}
\caption{
Full held-out evaluation on the 21-point preference sweep. The preference vector is $\omega=(1-w,w)$. Each cell reports Qwen photorealism / CLIP sketch. Dashes indicate preferences for which no Fixed-Weights specialist was trained. Scores are rounded to two decimal places. Local-control statistics in Table~\ref{tab:dense_control} retain finer precision.
}
\label{tab:dense_scores}
\begin{tabular}{ccccc}
\toprule
$w$ & ParetoSlider & RewardedSoups & DB-MPA & Fixed-Weights \\
\midrule
0.00 & 0.63 / 0.14 & 0.65 / 0.13 & 0.65 / 0.13 & 0.65 / 0.13 \\
0.05 & 0.63 / 0.14 & 0.65 / 0.13 & 0.63 / 0.13 & -- \\
0.10 & 0.63 / 0.14 & 0.65 / 0.13 & 0.63 / 0.13 & -- \\
0.15 & 0.63 / 0.14 & 0.63 / 0.13 & 0.61 / 0.14 & -- \\
0.20 & 0.62 / 0.14 & 0.62 / 0.13 & 0.61 / 0.14 & -- \\
0.25 & 0.62 / 0.14 & 0.59 / 0.14 & 0.56 / 0.15 & 0.12 / 0.18 \\
0.30 & 0.61 / 0.14 & 0.53 / 0.14 & 0.51 / 0.15 & -- \\
0.35 & 0.59 / 0.14 & 0.37 / 0.16 & 0.45 / 0.16 & -- \\
0.40 & 0.58 / 0.15 & 0.25 / 0.18 & 0.38 / 0.16 & -- \\
0.45 & 0.56 / 0.15 & 0.16 / 0.18 & 0.29 / 0.16 & -- \\
0.50 & 0.50 / 0.16 & 0.13 / 0.19 & 0.25 / 0.16 & 0.10 / 0.19 \\
0.55 & 0.38 / 0.18 & 0.12 / 0.19 & 0.20 / 0.16 & -- \\
0.60 & 0.30 / 0.19 & 0.11 / 0.19 & 0.17 / 0.16 & -- \\
0.65 & 0.25 / 0.19 & 0.11 / 0.19 & 0.14 / 0.16 & -- \\
0.70 & 0.22 / 0.19 & 0.10 / 0.20 & 0.12 / 0.17 & -- \\
0.75 & 0.20 / 0.19 & 0.09 / 0.20 & 0.10 / 0.17 & 0.07 / 0.20 \\
0.80 & 0.18 / 0.19 & 0.08 / 0.20 & 0.09 / 0.18 & -- \\
0.85 & 0.17 / 0.20 & 0.08 / 0.20 & 0.07 / 0.18 & -- \\
0.90 & 0.16 / 0.20 & 0.07 / 0.20 & 0.07 / 0.19 & -- \\
0.95 & 0.16 / 0.20 & 0.07 / 0.20 & 0.07 / 0.19 & -- \\
1.00 & 0.16 / 0.20 & 0.07 / 0.19 & 0.07 / 0.19 & 0.07 / 0.19 \\
\bottomrule
\end{tabular}
\end{table}

\section{Limitations}
\label{sec:limit}
ParetoSlider inherits a fundamental dependency on the quality of the reward models used during training. If a reward model fails to capture the true target objective, for instance, a sketch reward that responds to gray scale images rather than the genuine sketch style, we will not be able to recover the Pareto front and might optimize the wrong objective. In the multi-reward setting this risk is compounded: a single poorly specified reward can distort the entire trade-off surface. This highlights the importance of careful reward design and validation, particularly for abstract or subjective objectives where proxy rewards are hardest to specify.

\begin{figure*}
    \setlength{\tabcolsep}{0.002\textwidth}
    \scriptsize
    \centering

    \begin{tabular}{c c c c c c c}
         \multicolumn{5}{c}{\textit{``a firefighter holding a dalmatian puppy''}} \\
        \includegraphics[width=0.18\linewidth]{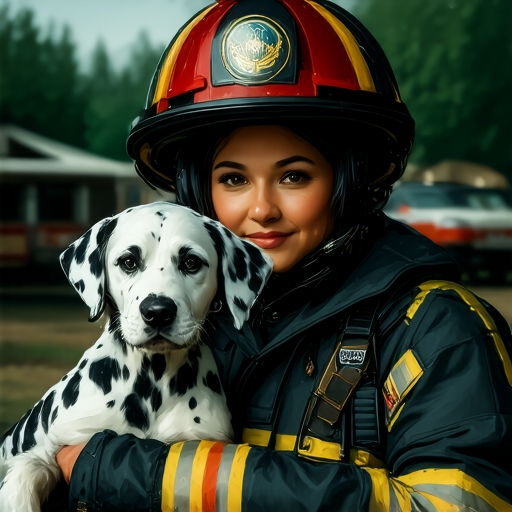} & 
        \includegraphics[width=0.18\linewidth]{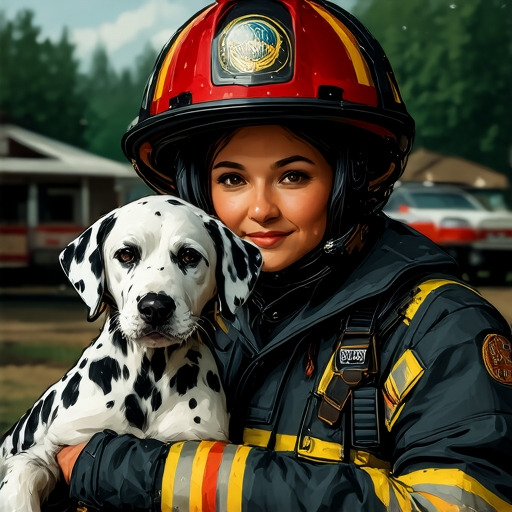} & 
        \includegraphics[width=0.18\linewidth]{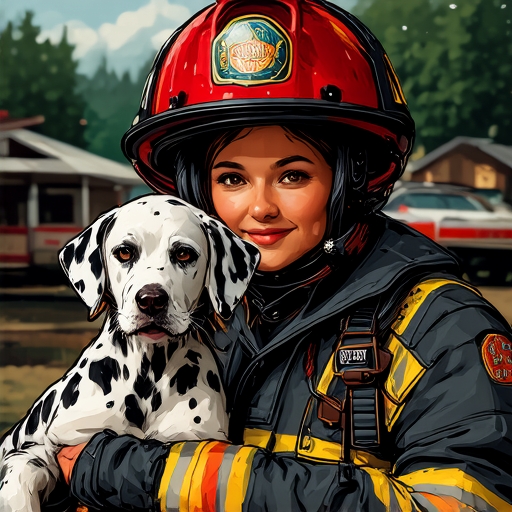} & 
        \includegraphics[width=0.18\linewidth]{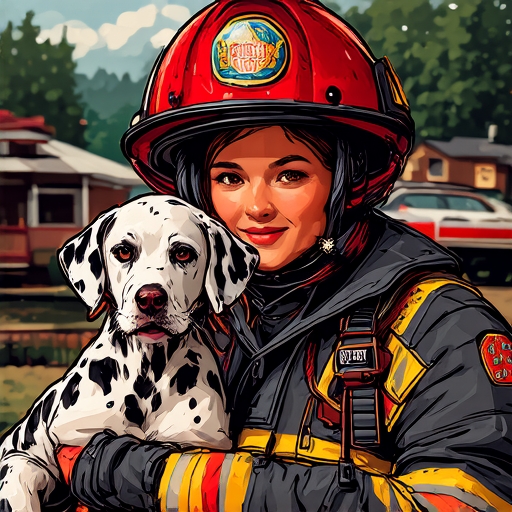}
        &
        \includegraphics[width=0.18\linewidth]{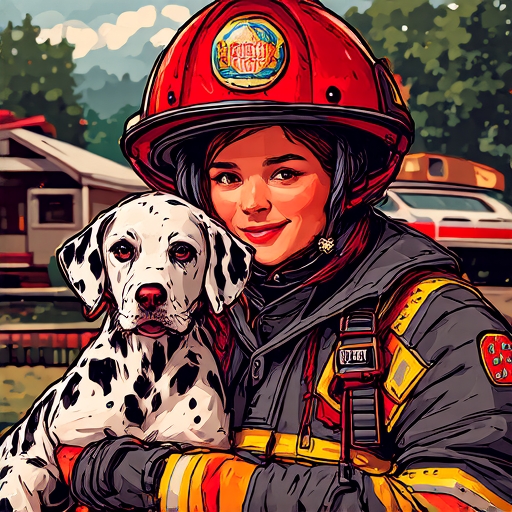}
        \\
        \multicolumn{5}{c}{\textit{``an astronaut floating above Earth''}} \\
        \includegraphics[width=0.18\linewidth]{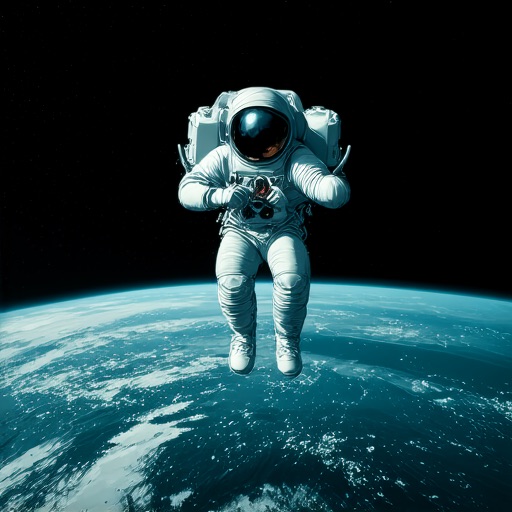} & 
        \includegraphics[width=0.18\linewidth]{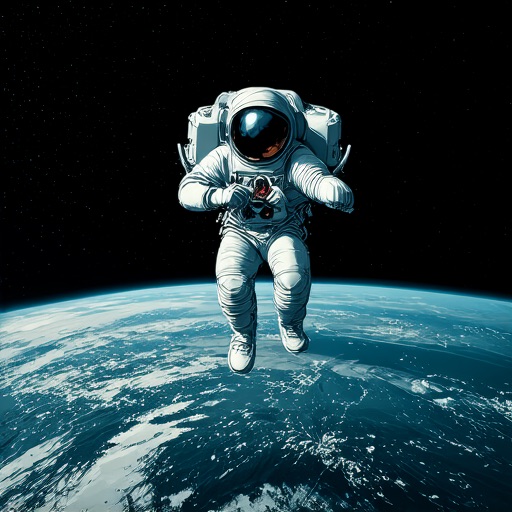} & 
        \includegraphics[width=0.18\linewidth]{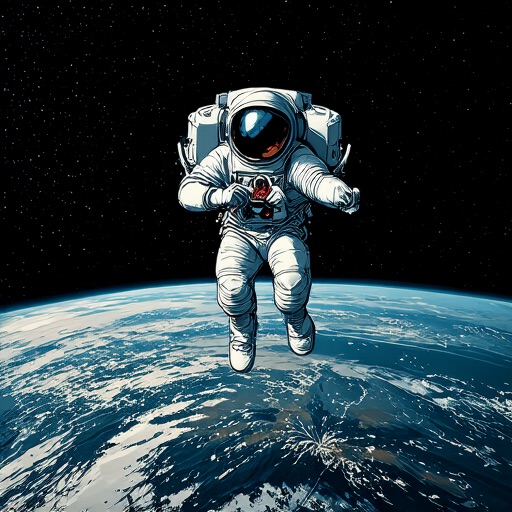} & 
        \includegraphics[width=0.18\linewidth]{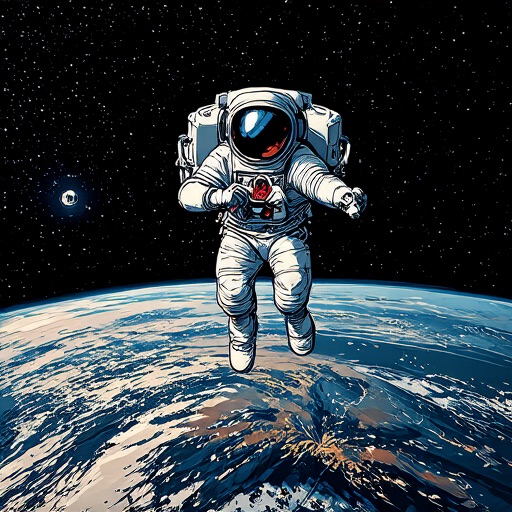}
        &
        \includegraphics[width=0.18\linewidth]{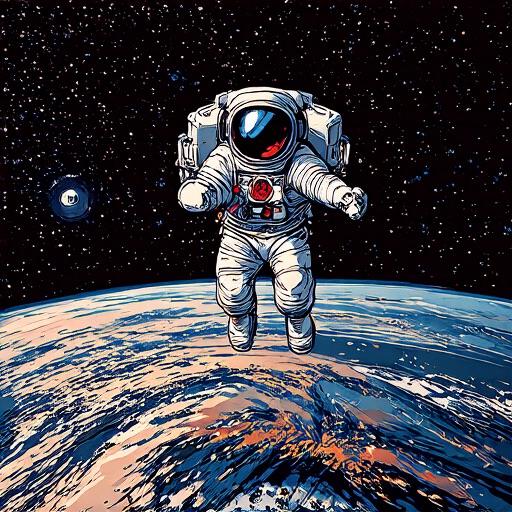}
        \\
        \multicolumn{5}{c}{$\text{Photorealistic} \xleftrightarrow{\hspace{6cm}} \text{Digital Art}$}
        \\
        \\
        \multicolumn{5}{c}{\textit{``a glass of orange juice next to a stack of pancakes''}} \\
        \includegraphics[width=0.18\linewidth]{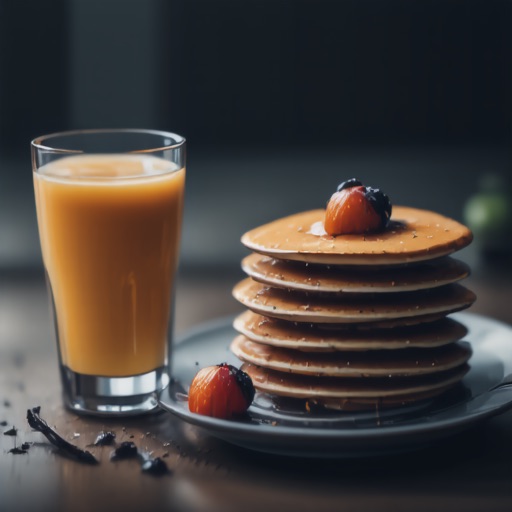} & 
        \includegraphics[width=0.18\linewidth]{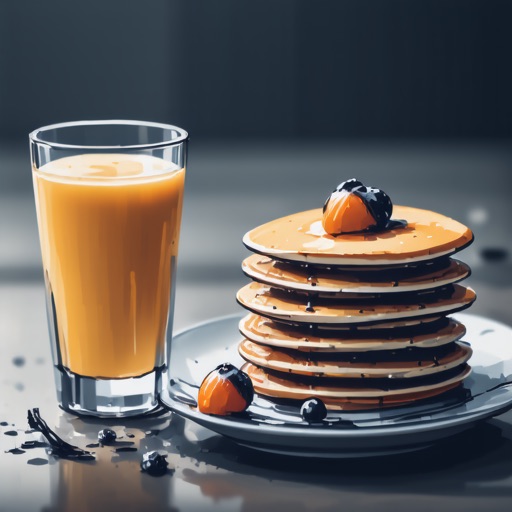} & 
        \includegraphics[width=0.18\linewidth]{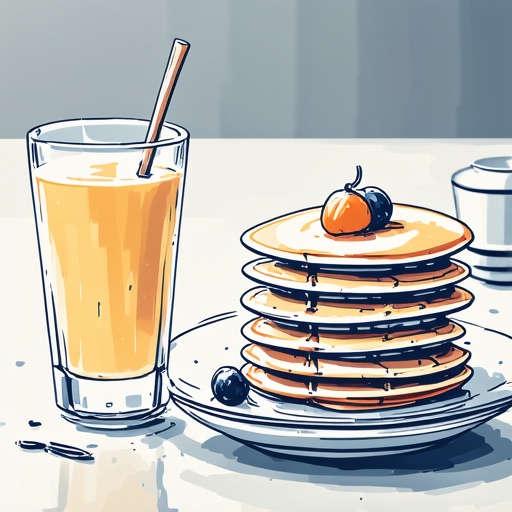} & 
        \includegraphics[width=0.18\linewidth]{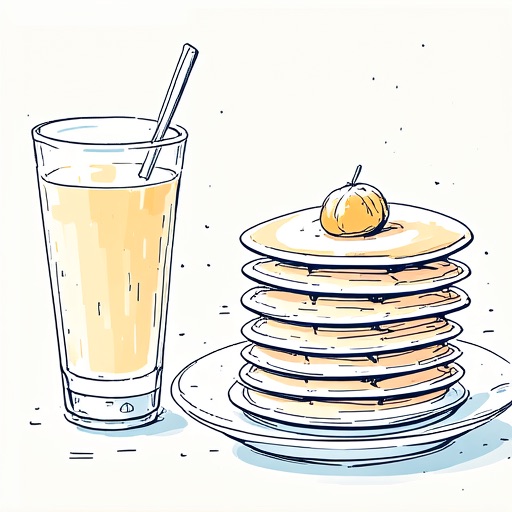}
        &
        \includegraphics[width=0.18\linewidth]{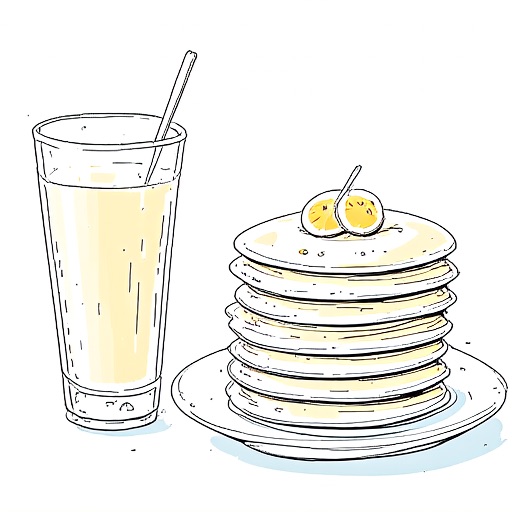}
        \\
        \multicolumn{5}{c}{\textit{``a grand piano in an empty concert hall''}} \\
        \includegraphics[width=0.18\linewidth]{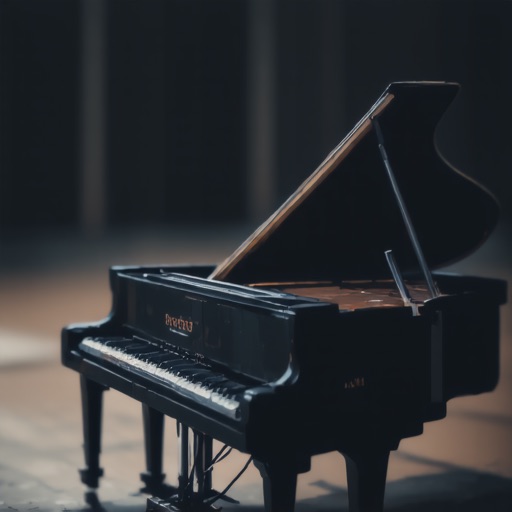} & 
        \includegraphics[width=0.18\linewidth]{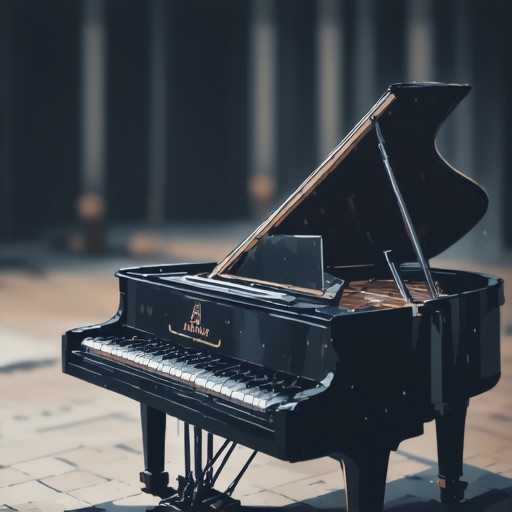} & 
        \includegraphics[width=0.18\linewidth]{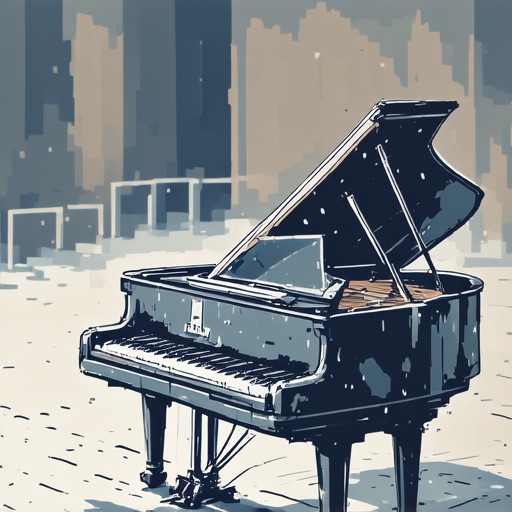} & 
        \includegraphics[width=0.18\linewidth]{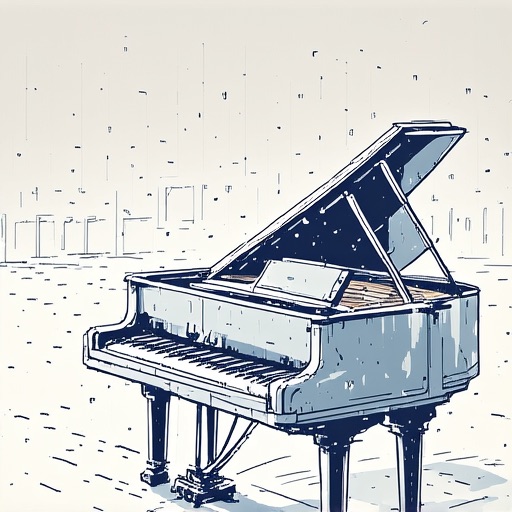}
        &
        \includegraphics[width=0.18\linewidth]{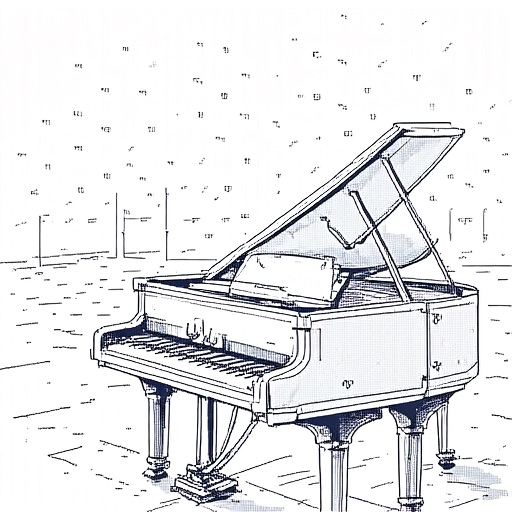}
         \\
        \multicolumn{5}{c}{$\text{Photorealistic} \xleftrightarrow{\hspace{6cm}} \text{Sketch}$}
        \\
        \\
        \multicolumn{5}{c}{\textit{``a living room with a sofa, a coffee table, a TV, and a window''}} \\
        \includegraphics[width=0.18\linewidth]{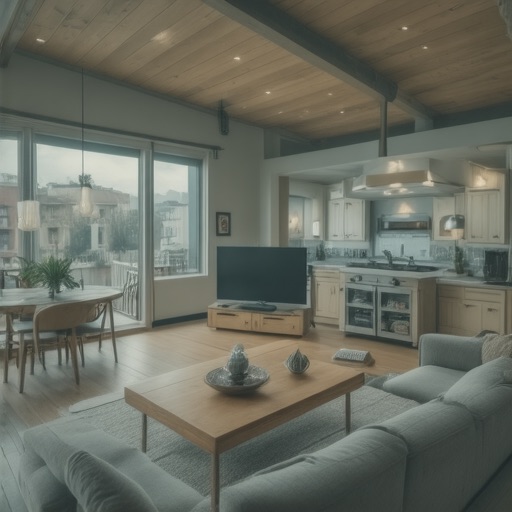} & 
        \includegraphics[width=0.18\linewidth]{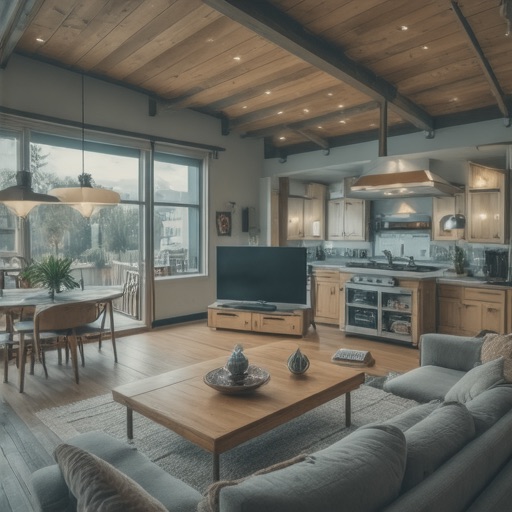} & 
        \includegraphics[width=0.18\linewidth]{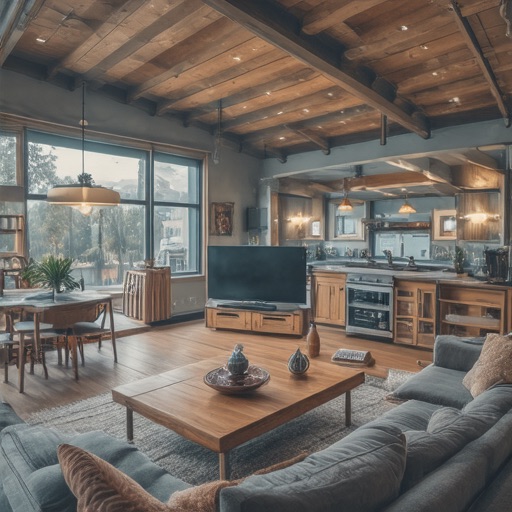} & 
        \includegraphics[width=0.18\linewidth]{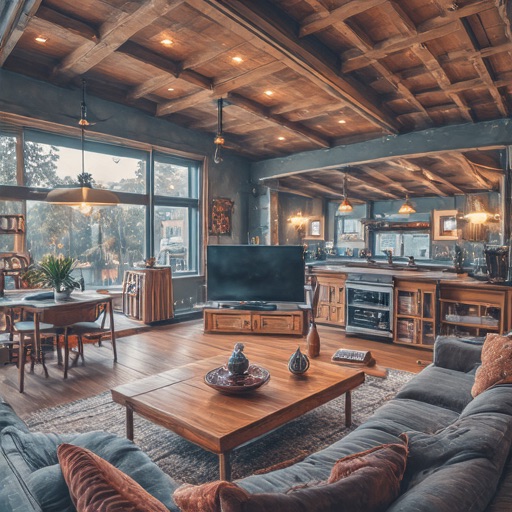}
        &
        \includegraphics[width=0.18\linewidth]{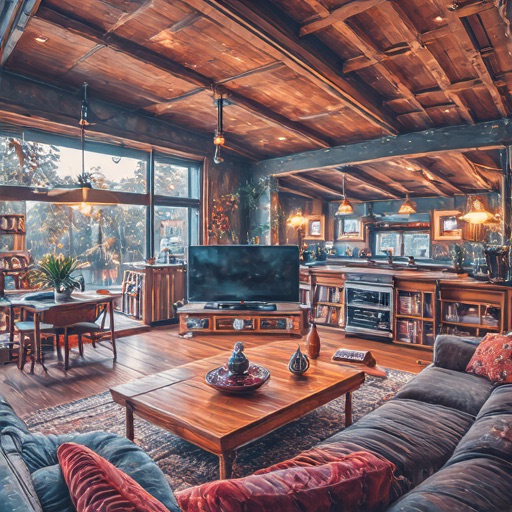}
        \\
        \multicolumn{5}{c}{\textit{``a studio headshot of a ballerina''}} \\
        \includegraphics[width=0.18\linewidth]{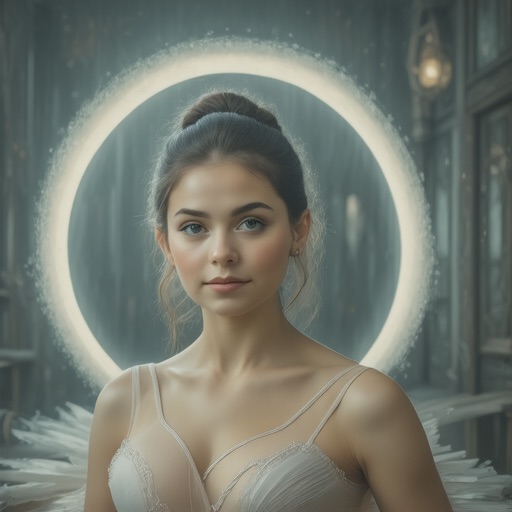} & 
        \includegraphics[width=0.18\linewidth]{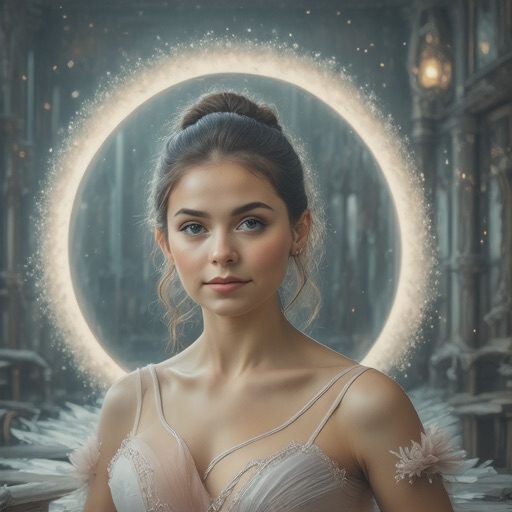} & 
        \includegraphics[width=0.18\linewidth]{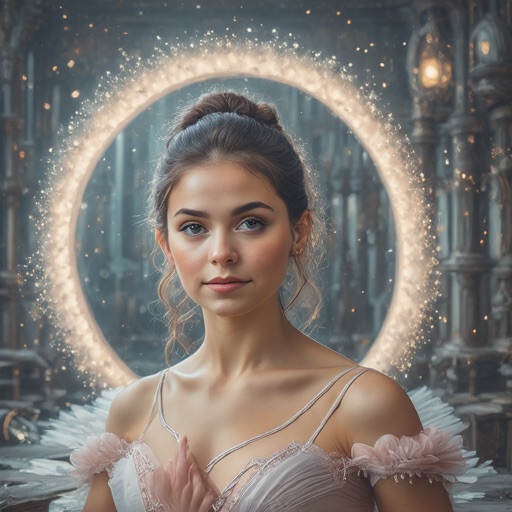} & 
        \includegraphics[width=0.18\linewidth]{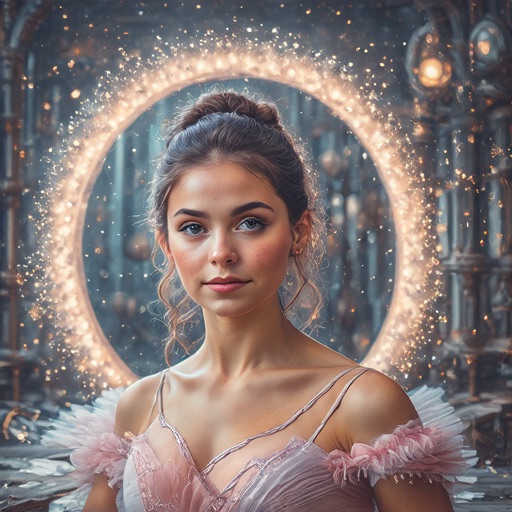}
        &
        \includegraphics[width=0.18\linewidth]{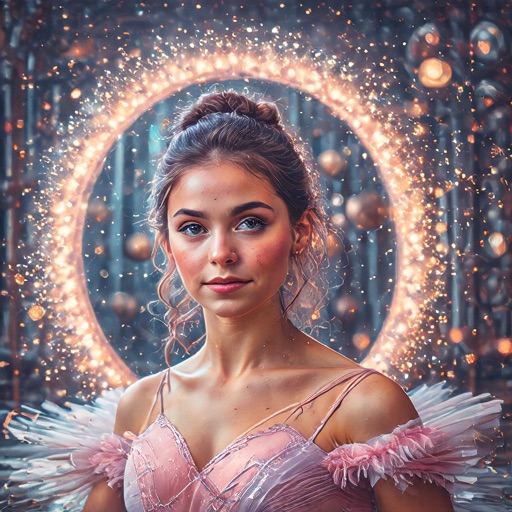}
        \\
        \multicolumn{5}{c}{$\text{Muted Pallette} \xleftrightarrow{\hspace{4cm}} \text{Saturated Pallette}$}
        \\

    \end{tabular}

    \caption{
    Our results for continuous preference control in text-to-image generation.
    }
    \label{fig:t2i_results_supp}
\end{figure*}

\begin{figure*}
    \setlength{\tabcolsep}{0.002\textwidth}
    \scriptsize
    \centering

    \begin{tabular}{c c c c c c c}
         \multicolumn{5}{c}{\textit{Disney Pixar Style}} \\
        \includegraphics[width=0.16\linewidth]{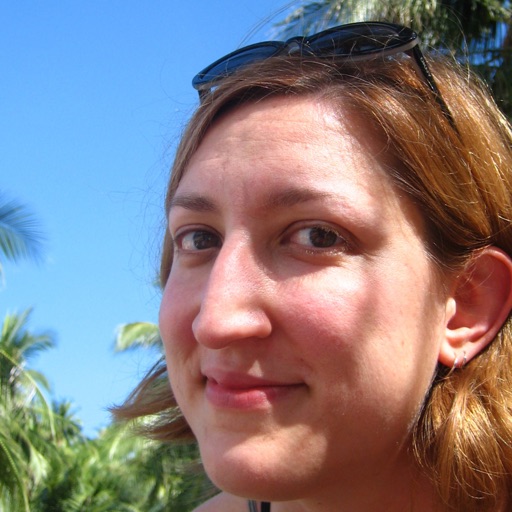} & 
        \includegraphics[width=0.16\linewidth]{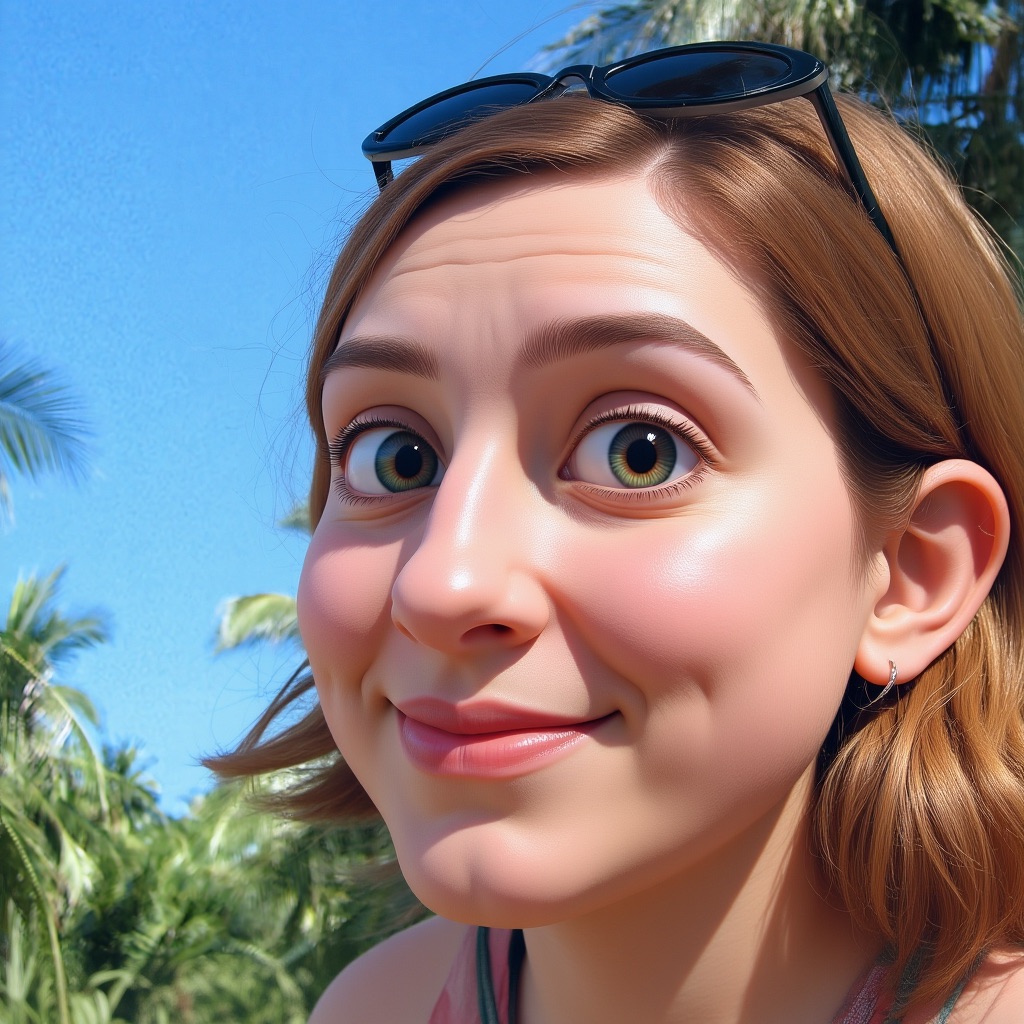} & 
        \includegraphics[width=0.16\linewidth]{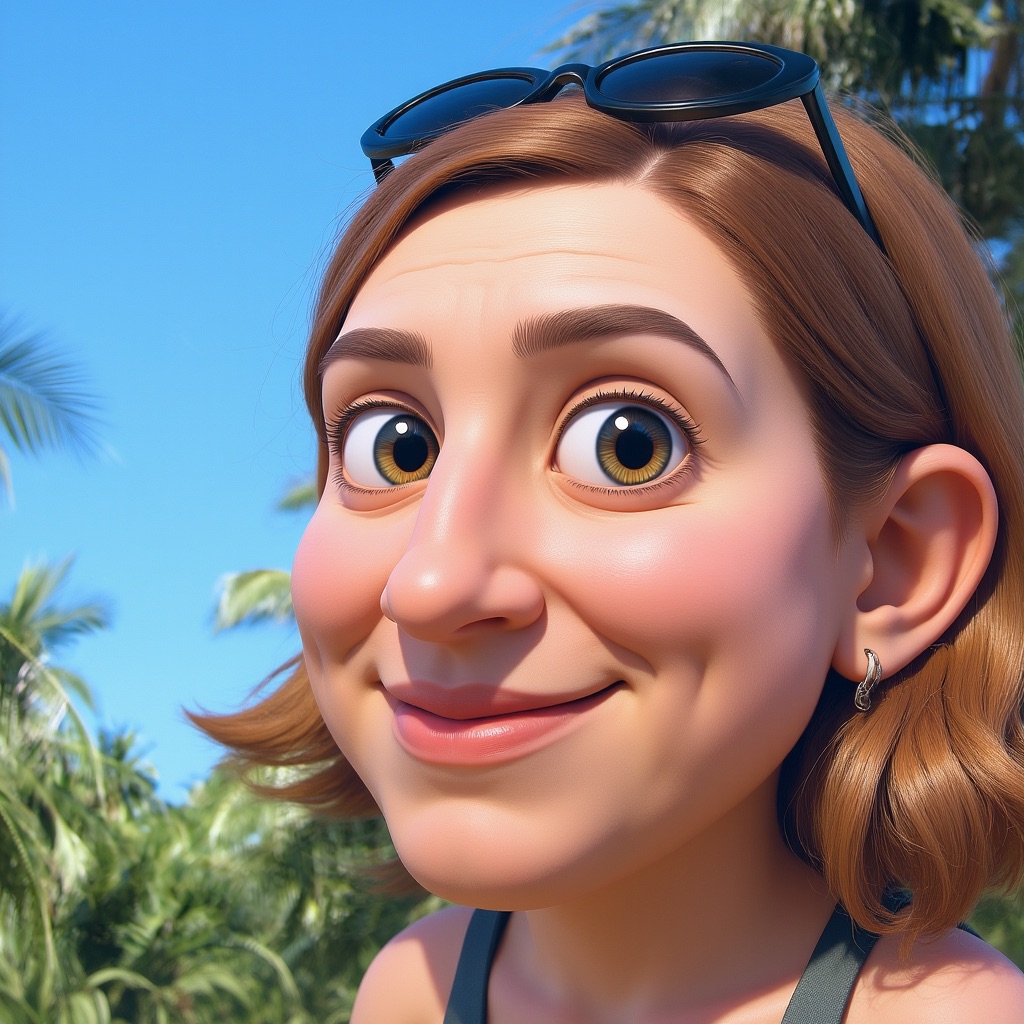} & 
        \includegraphics[width=0.16\linewidth]{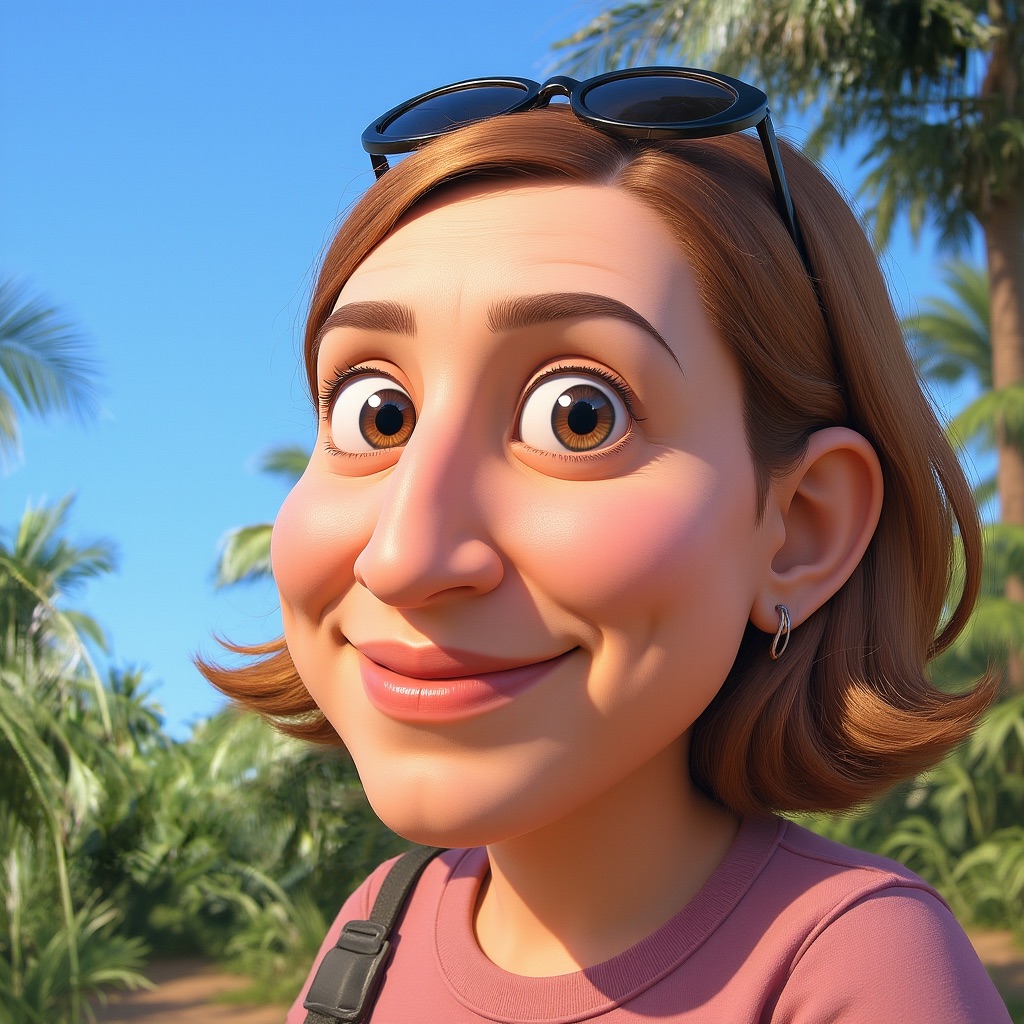}
        &
        \includegraphics[width=0.16\linewidth]{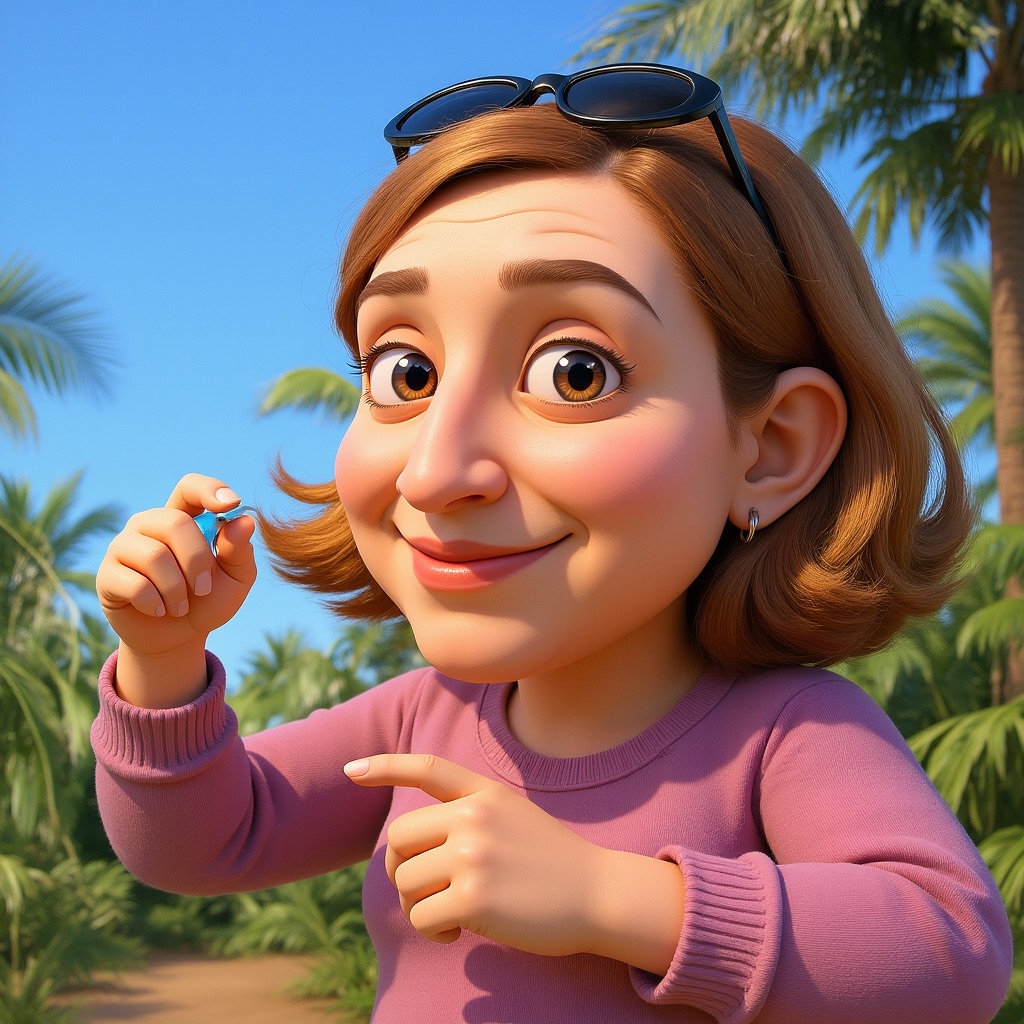}
        \\
        \multicolumn{5}{c}{\textit{Pixel Art}} \\
        \includegraphics[width=0.16\linewidth]{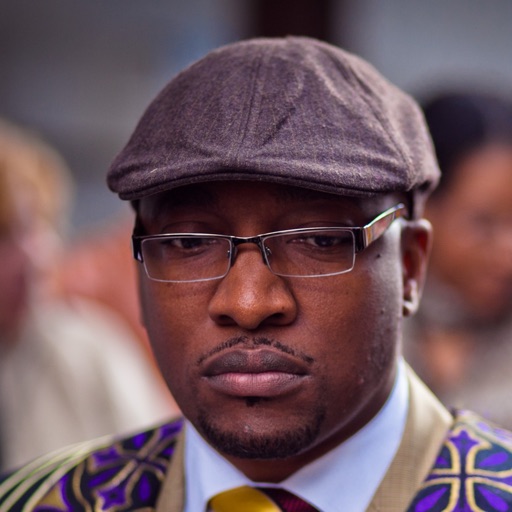} & 
        \includegraphics[width=0.16\linewidth]{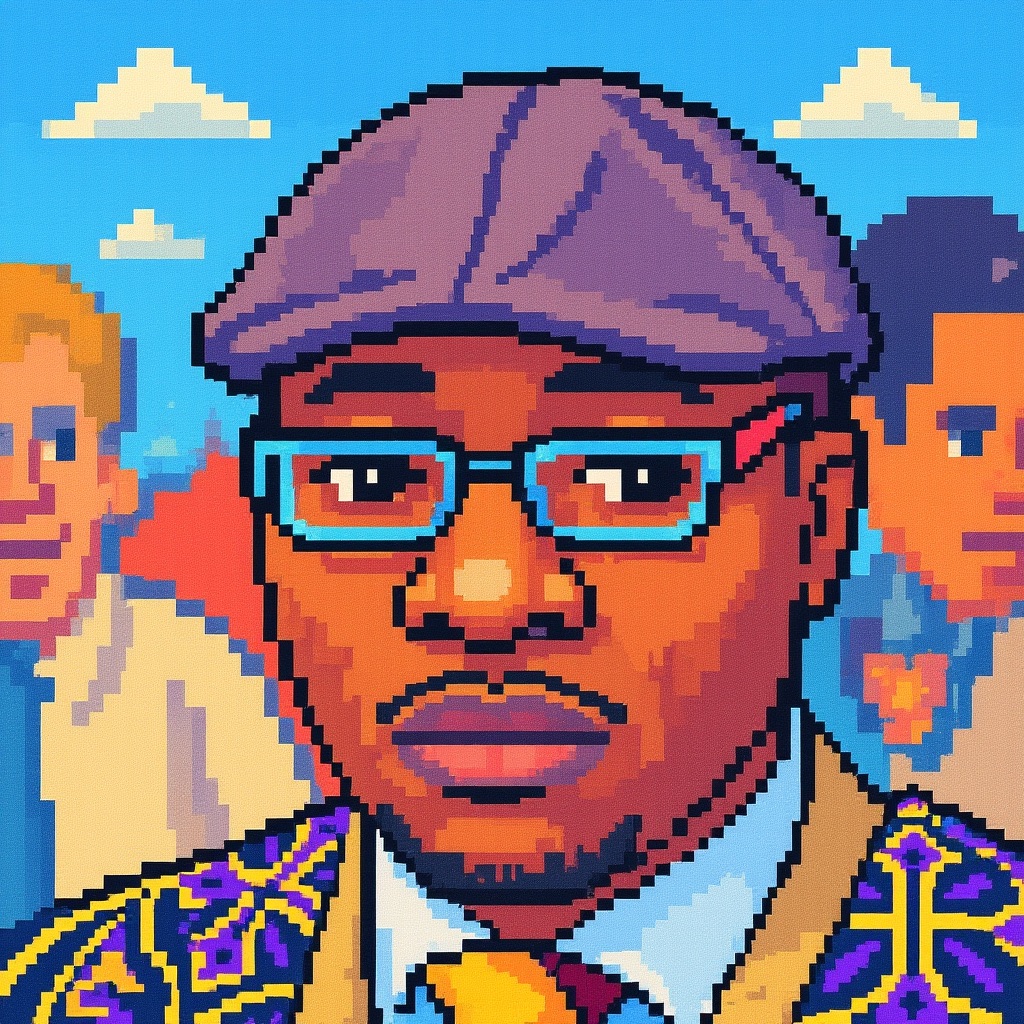} & 
        \includegraphics[width=0.16\linewidth]{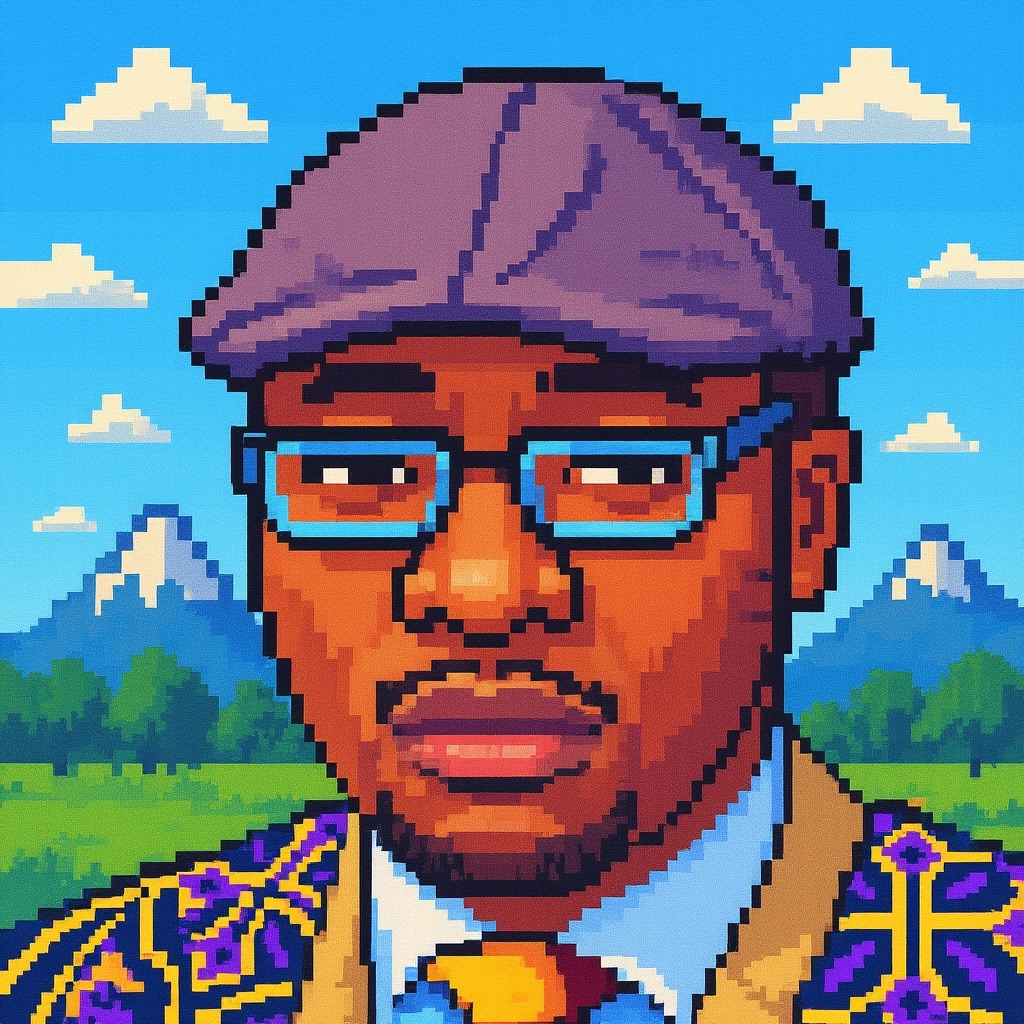} & 
        \includegraphics[width=0.16\linewidth]{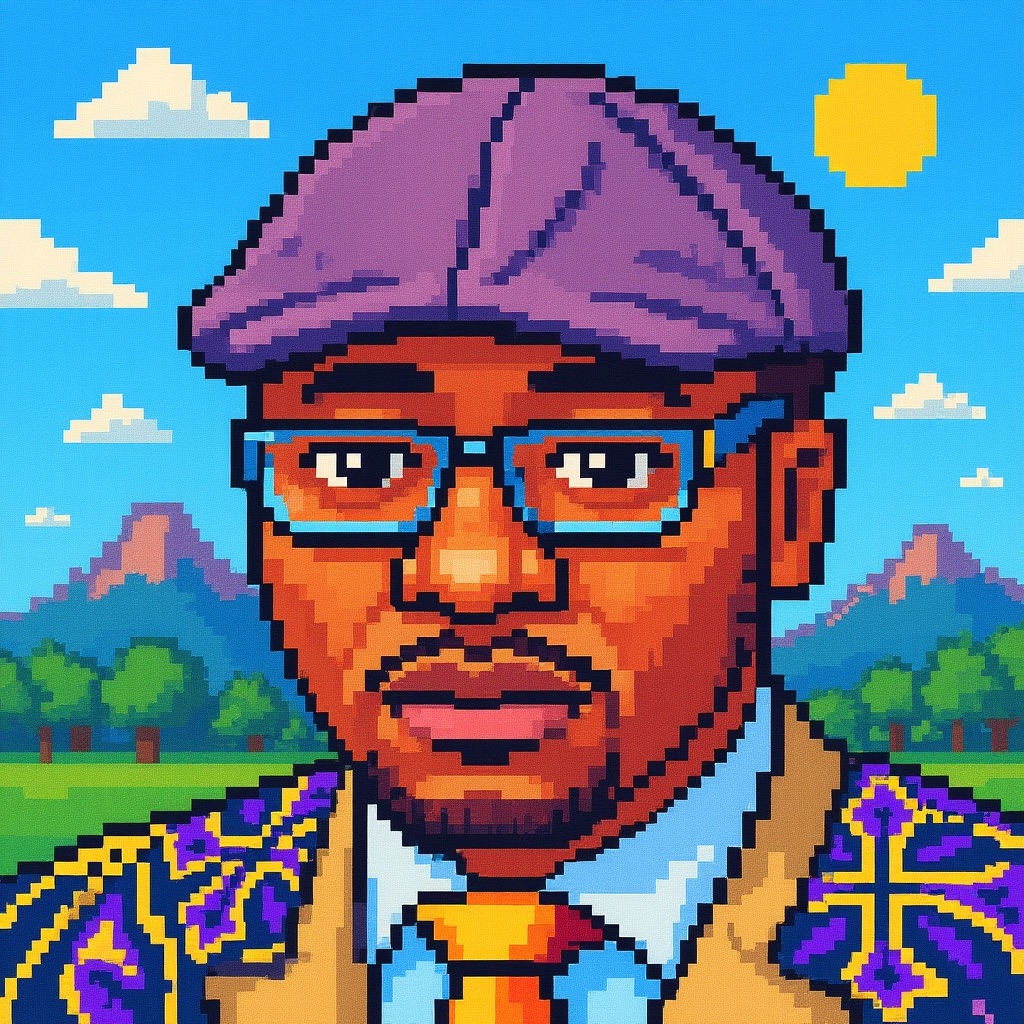}
        &
        \includegraphics[width=0.16\linewidth]{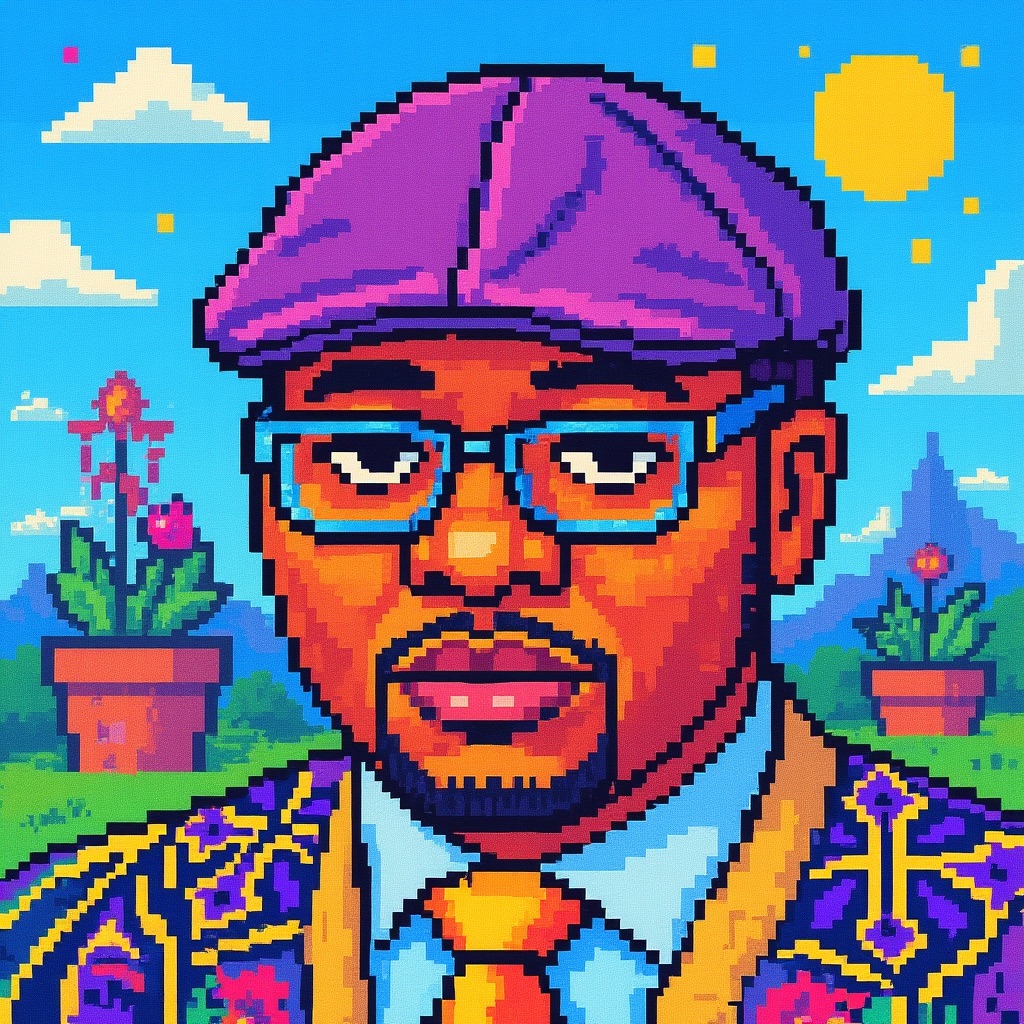}
        \\
        \multicolumn{5}{c}{\textit{Lego Mini-Figure}} \\
        \includegraphics[width=0.16\linewidth]{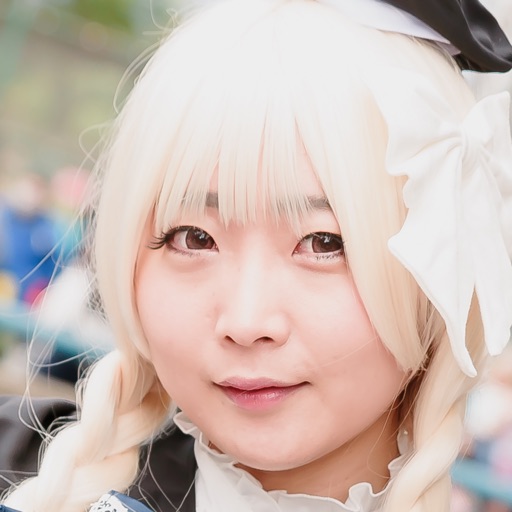} & 
        \includegraphics[width=0.16\linewidth]{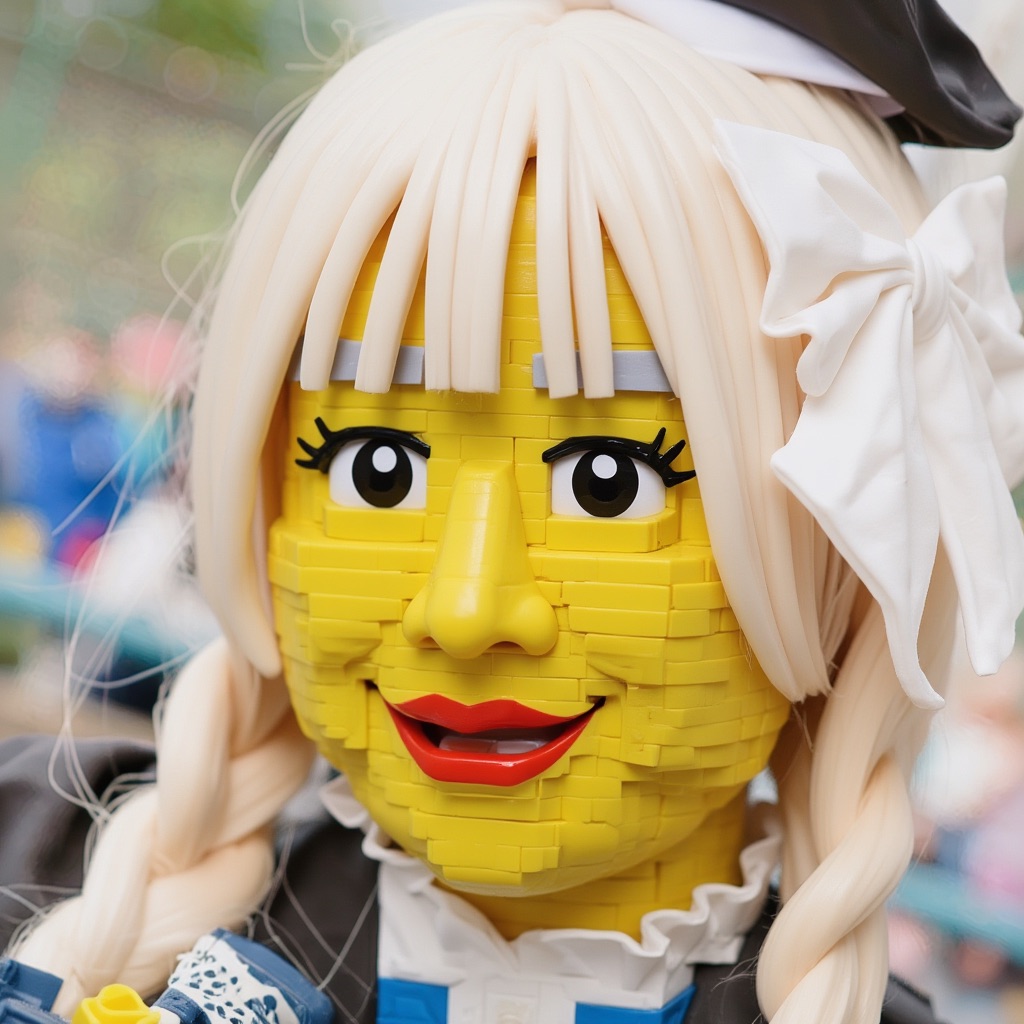} & 
        \includegraphics[width=0.16\linewidth]{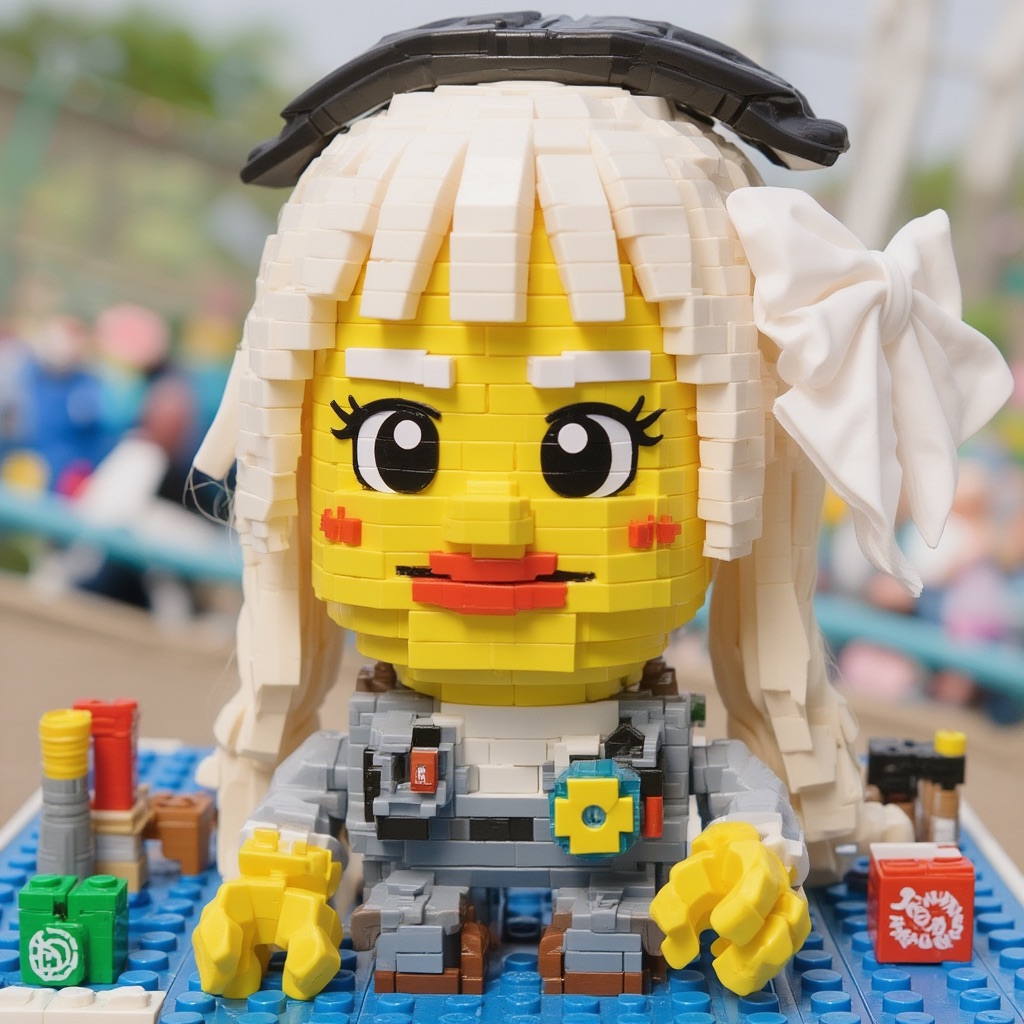} & 
        \includegraphics[width=0.16\linewidth]{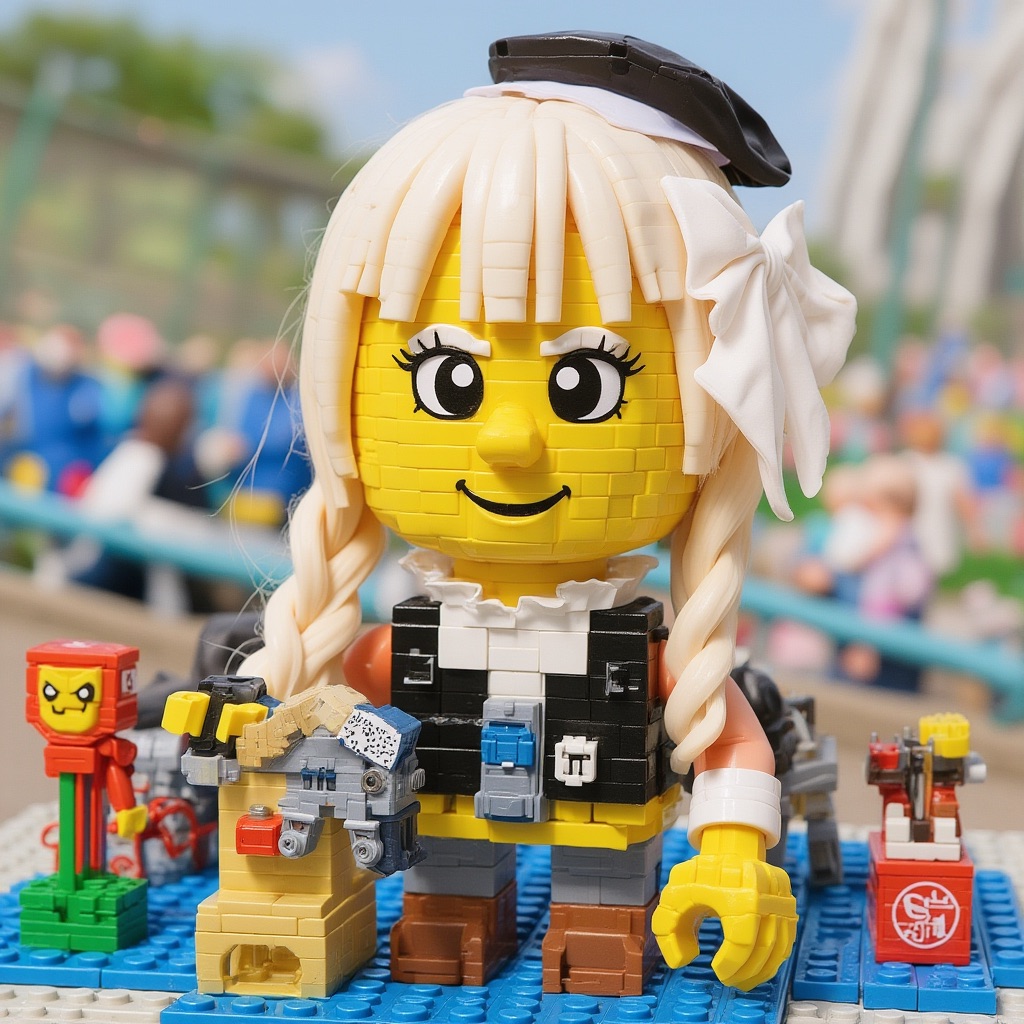}
        &
        \includegraphics[width=0.16\linewidth]{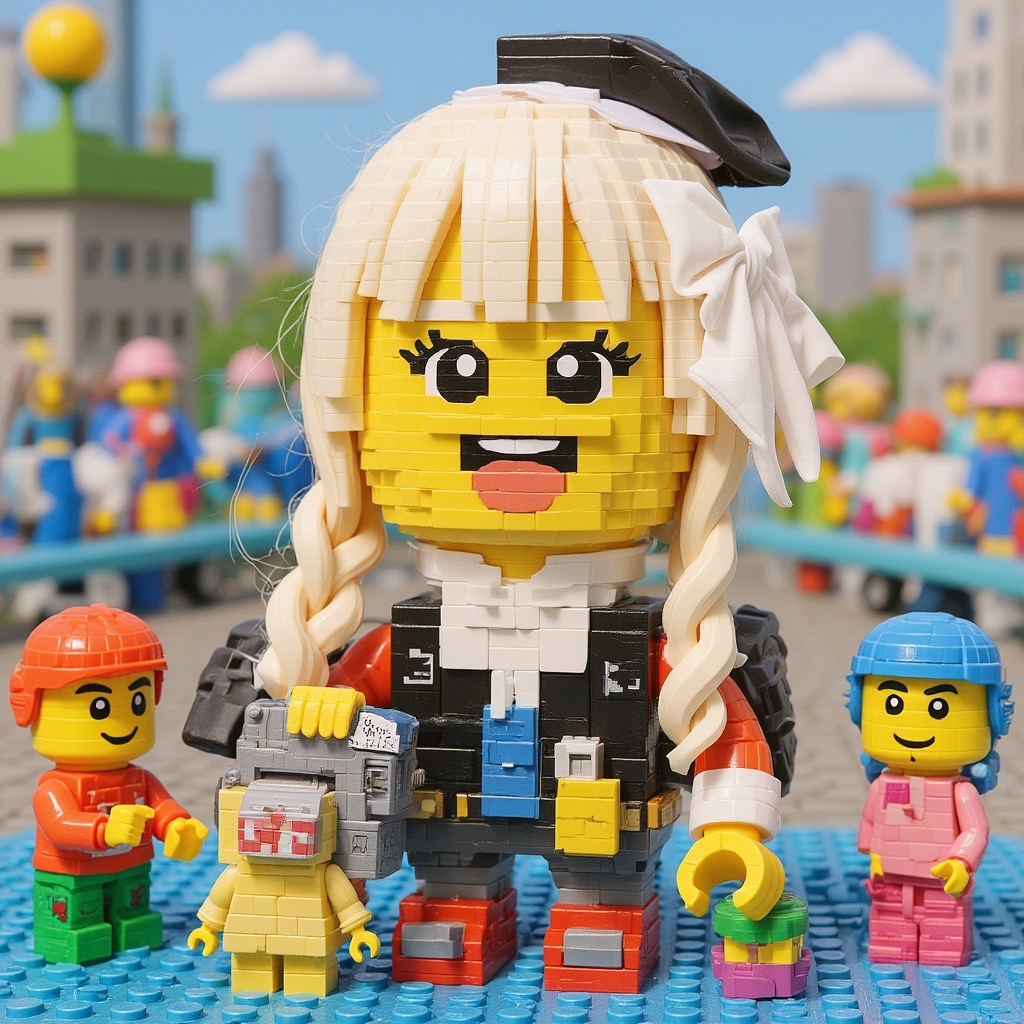}
        \\
        \multicolumn{5}{c}{\textit{Anime Character}} \\
        \includegraphics[width=0.16\linewidth]{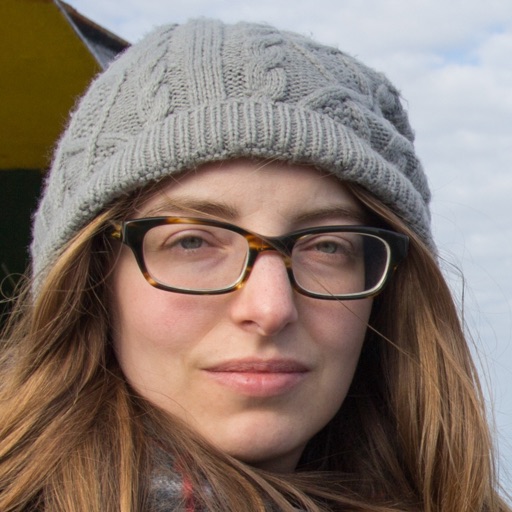} & 
        \includegraphics[width=0.16\linewidth]{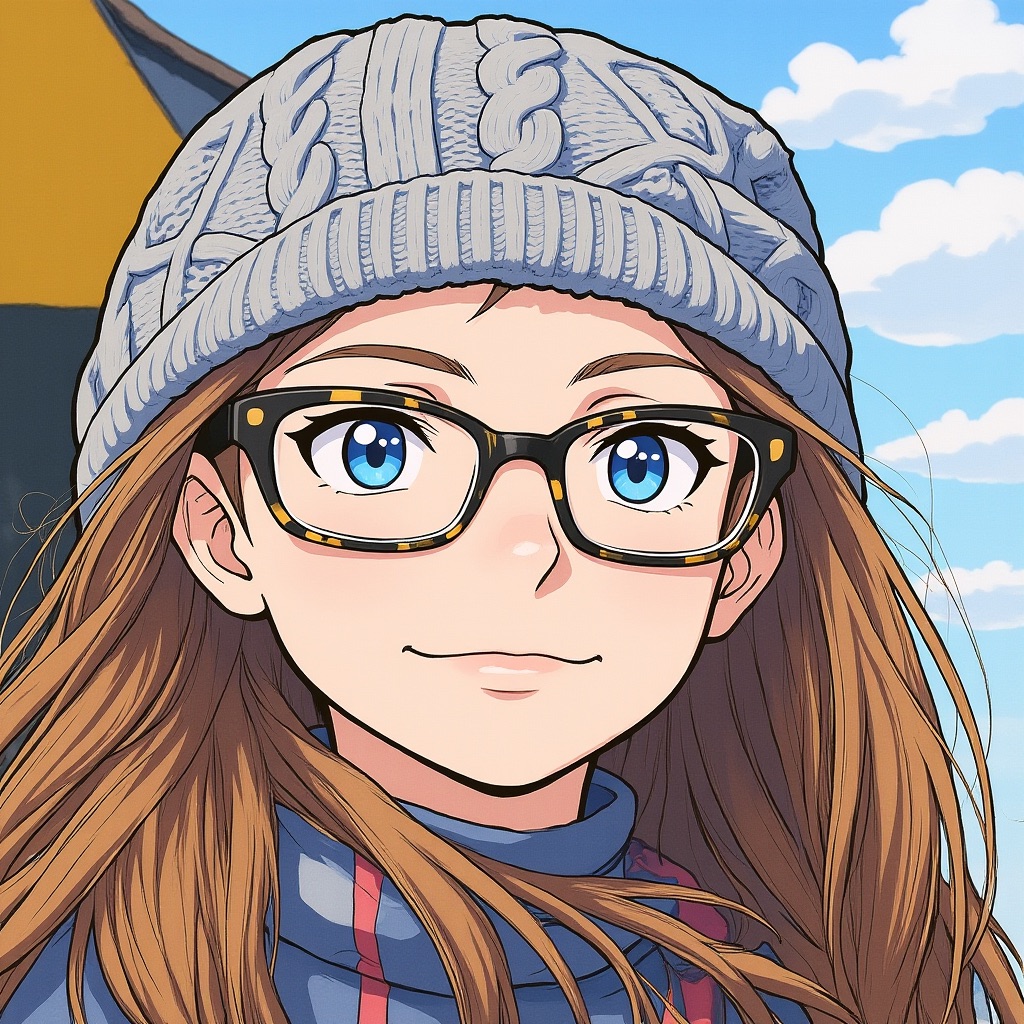} & 
        \includegraphics[width=0.16\linewidth]{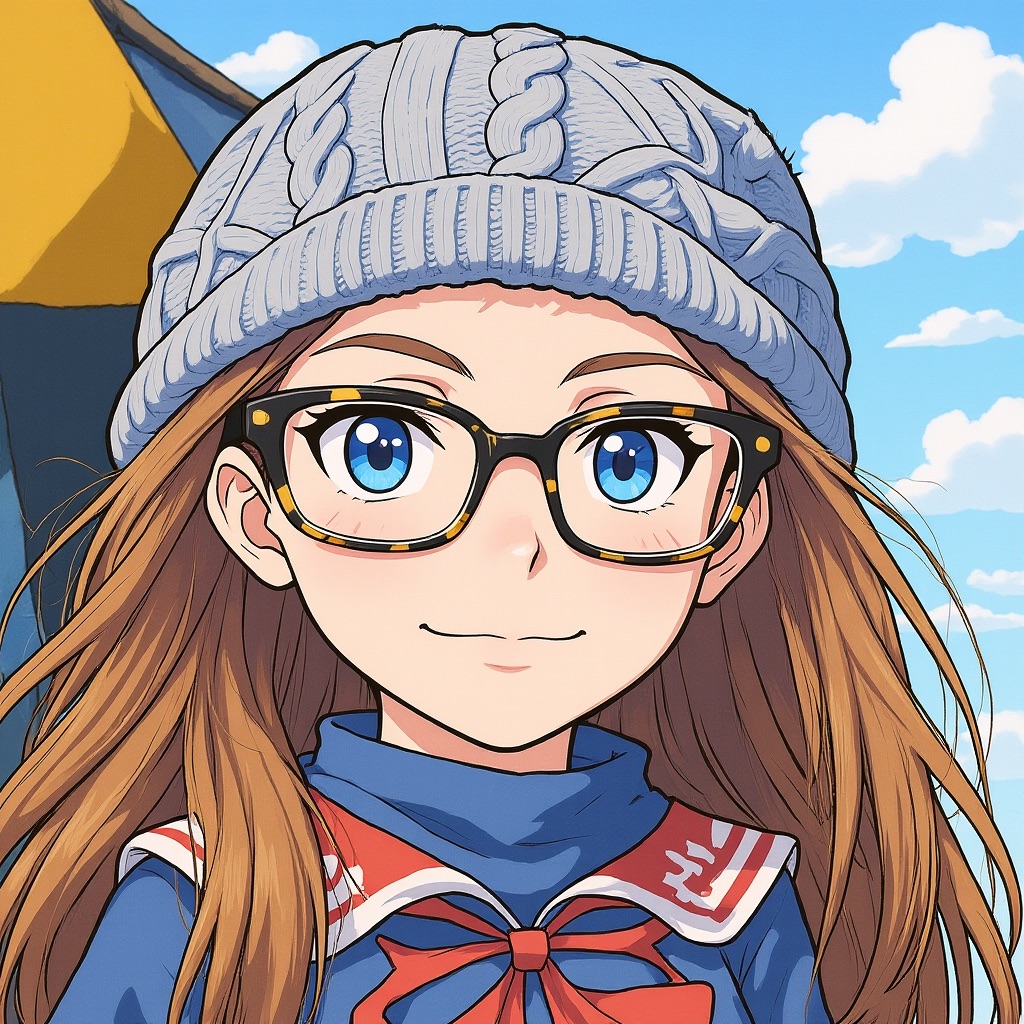} & 
        \includegraphics[width=0.16\linewidth]{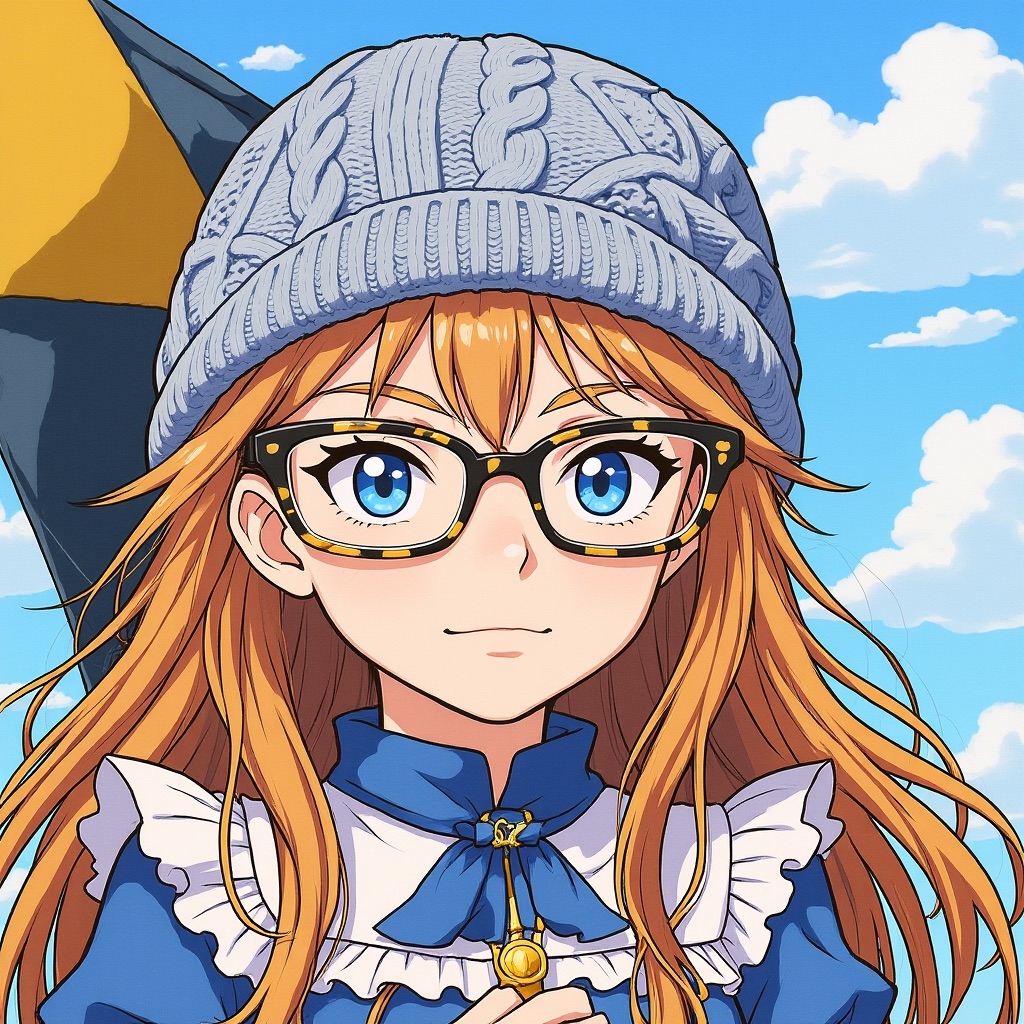}
        &
        \includegraphics[width=0.16\linewidth]{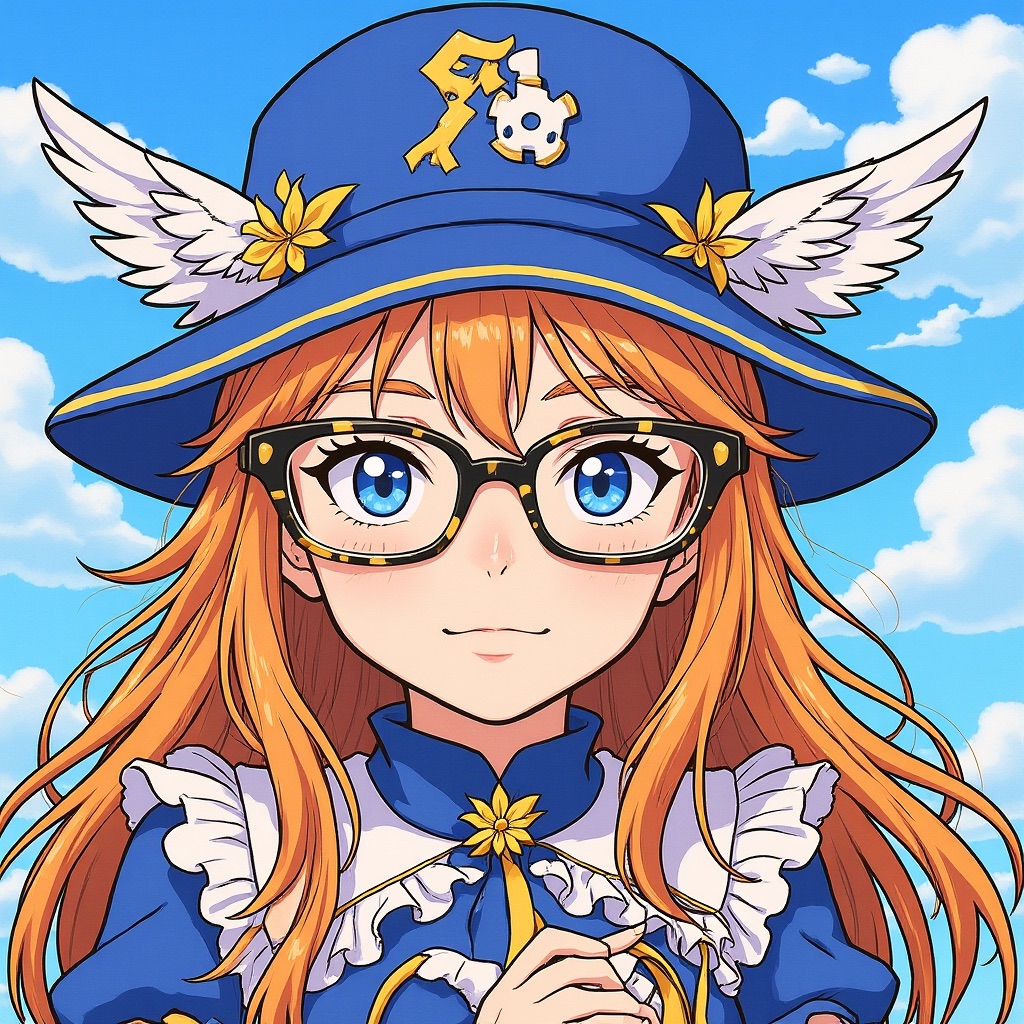}
        \\
        \multicolumn{5}{c}{\textit{Claymation Character}} \\
        \includegraphics[width=0.16\linewidth]{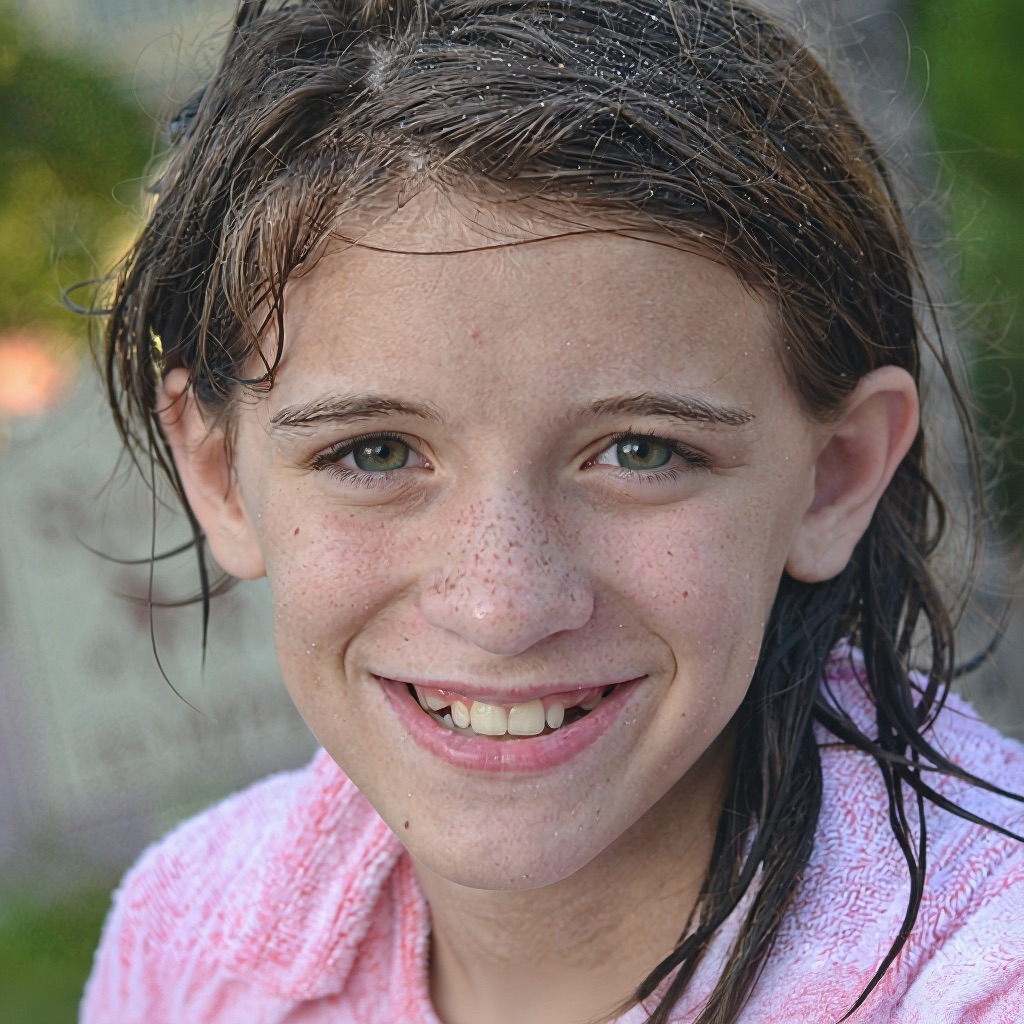} & 
        \includegraphics[width=0.16\linewidth]{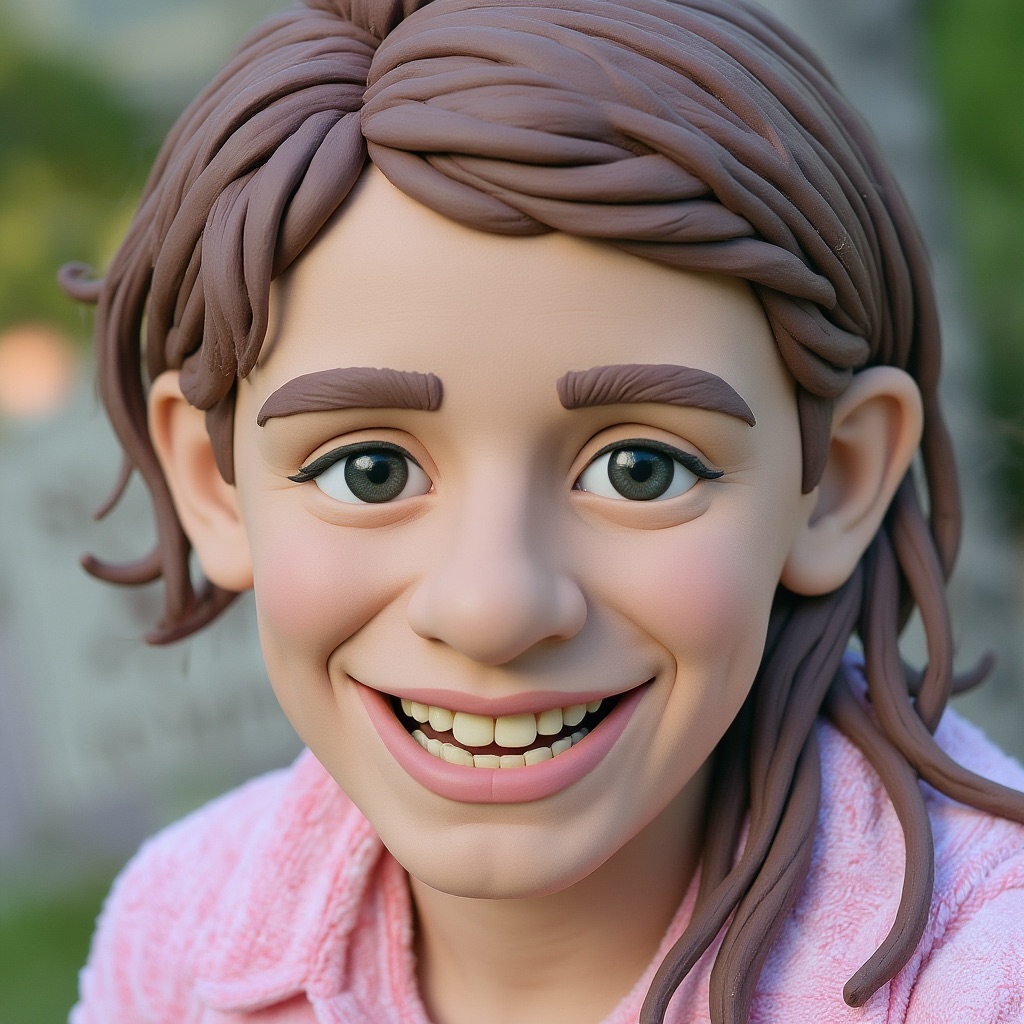} & 
        \includegraphics[width=0.16\linewidth]{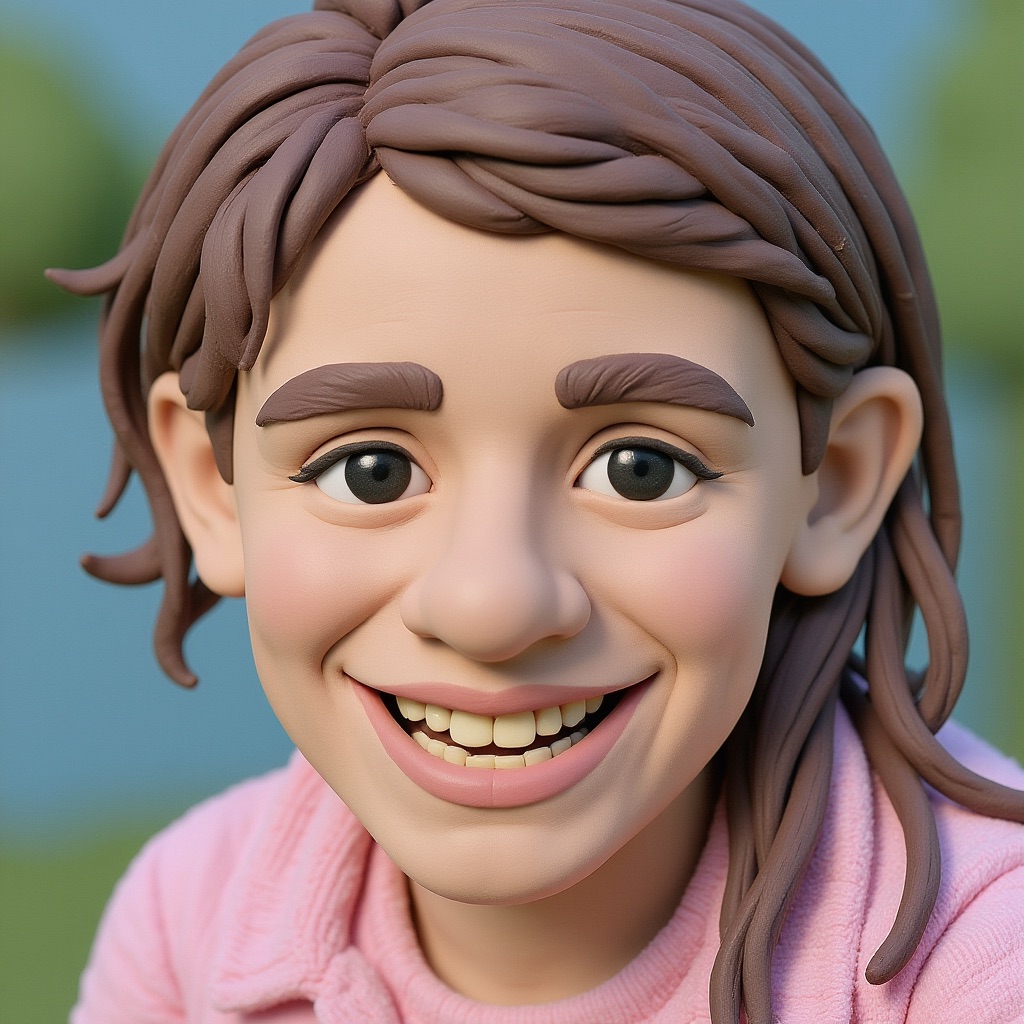} & 
        \includegraphics[width=0.16\linewidth]{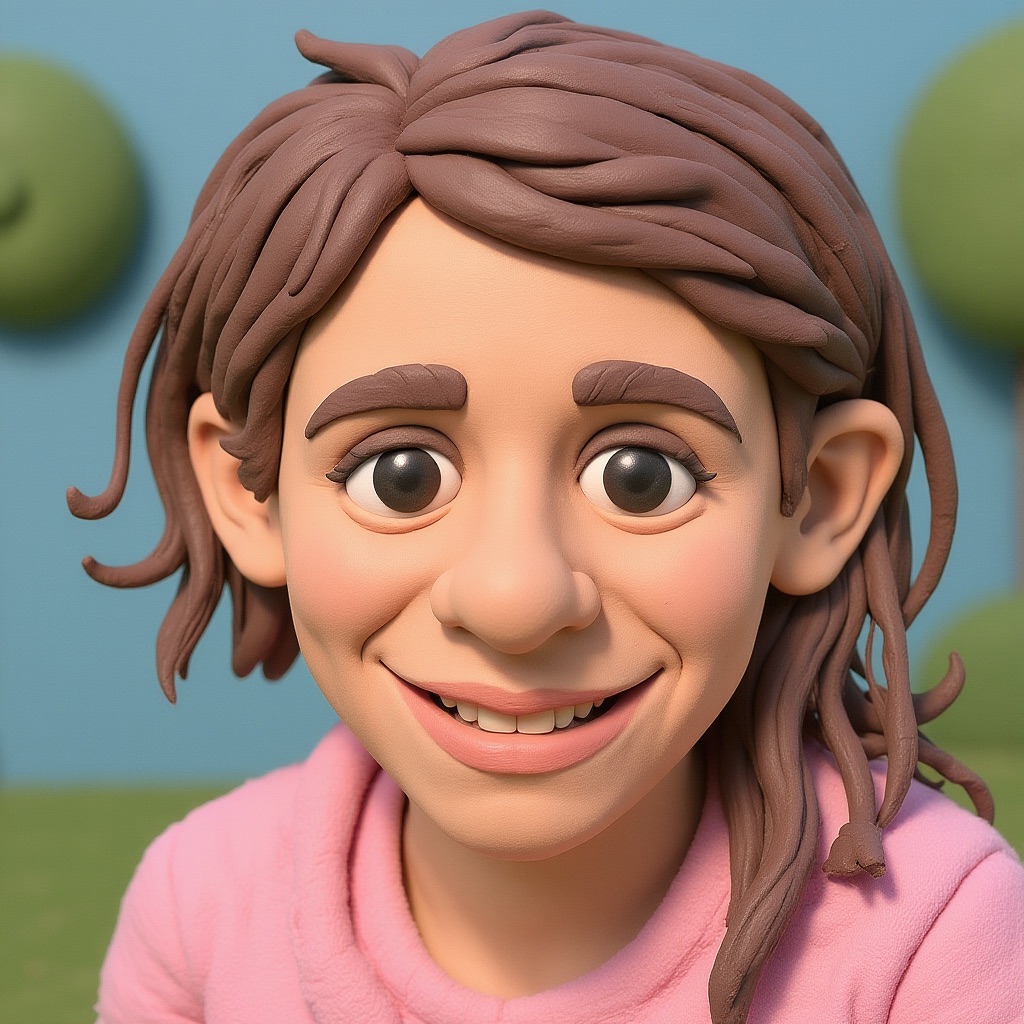}
        &
        \includegraphics[width=0.16\linewidth]{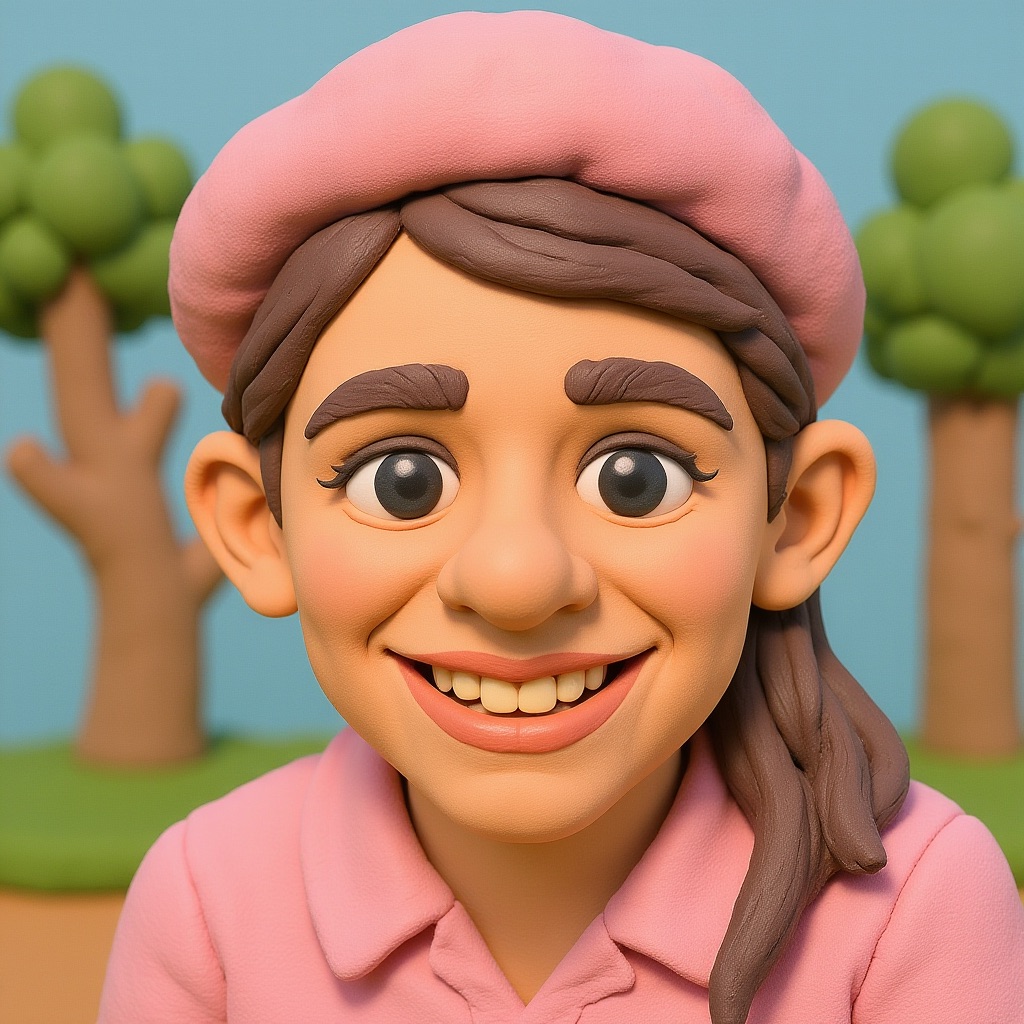}
        \\

        \multicolumn{5}{c}{\textit{Change the sofa to red}} \\
        \includegraphics[width=0.16\linewidth]{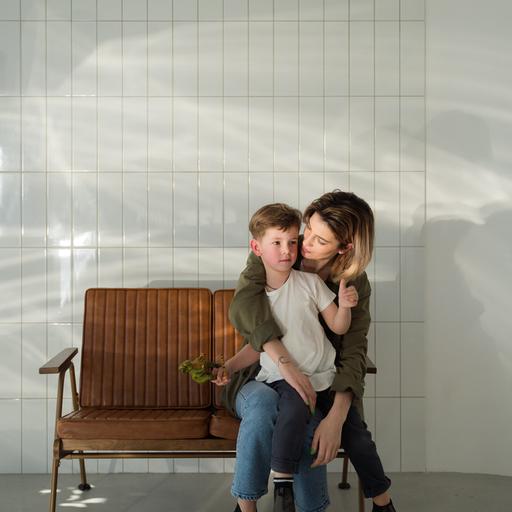} & 
        \includegraphics[width=0.16\linewidth]{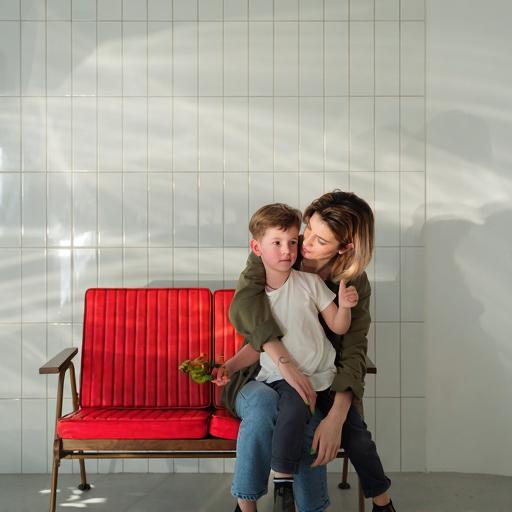} & 
        \includegraphics[width=0.16\linewidth]{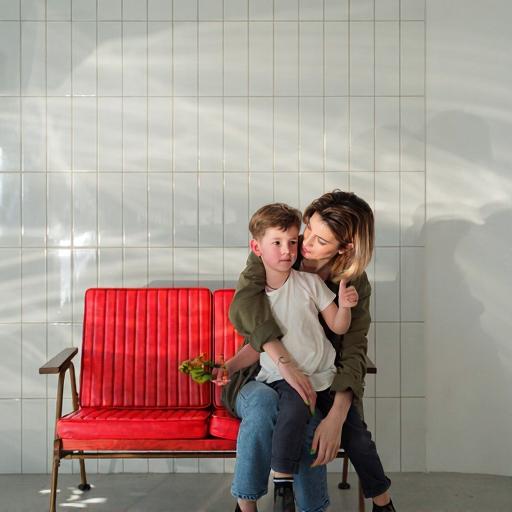} & 
        \includegraphics[width=0.16\linewidth]{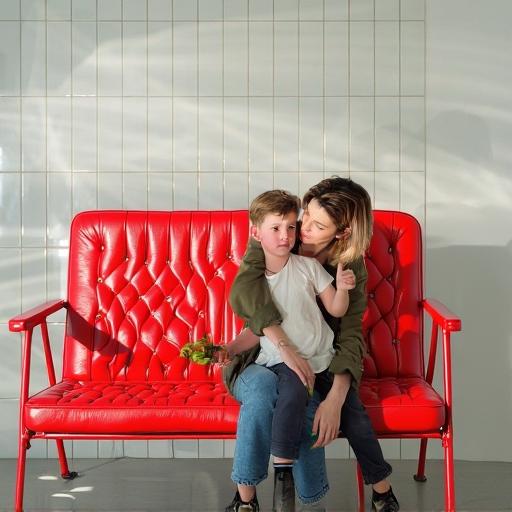}
        &
        \includegraphics[width=0.16\linewidth]{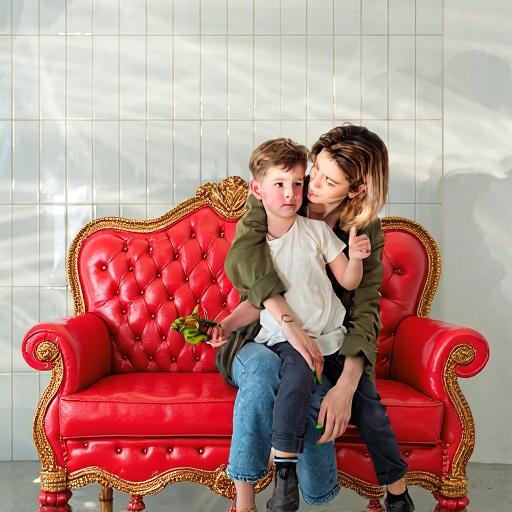}
        \\

        \multicolumn{5}{c}{\textit{Stylize as a geometric low-poly 3D render}} \\
        \includegraphics[width=0.16\linewidth]{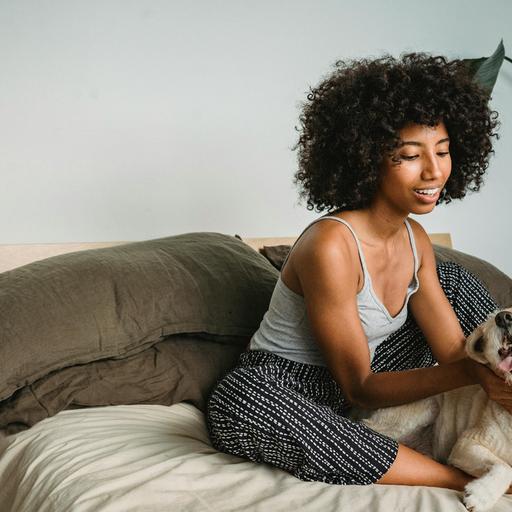} & 
        \includegraphics[width=0.16\linewidth]{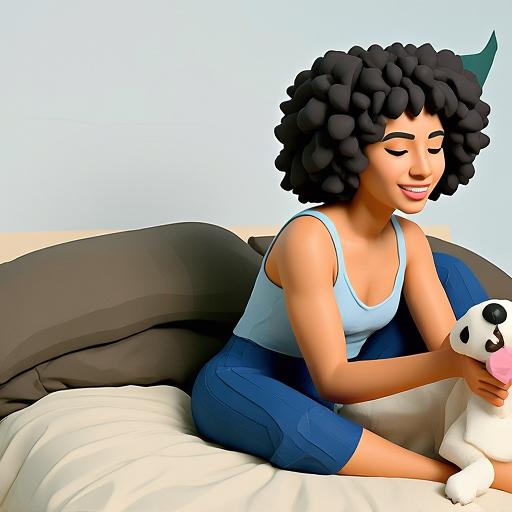} & 
        \includegraphics[width=0.16\linewidth]{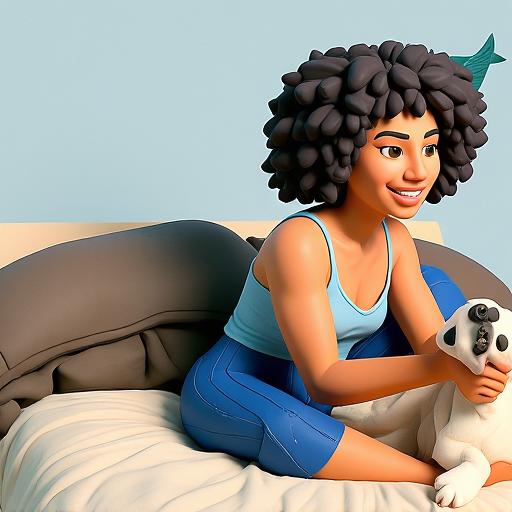} & 
        \includegraphics[width=0.16\linewidth]{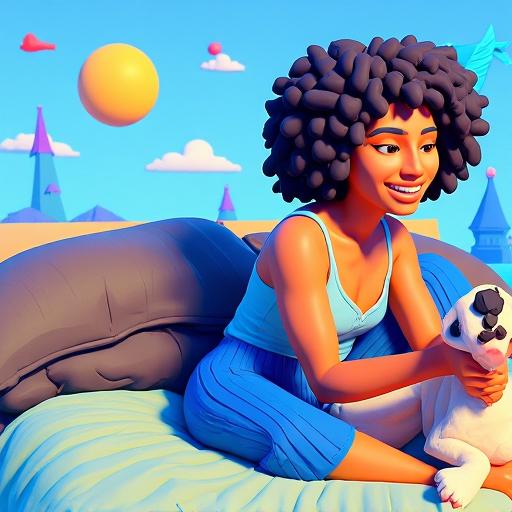}
        &
        \includegraphics[width=0.16\linewidth]{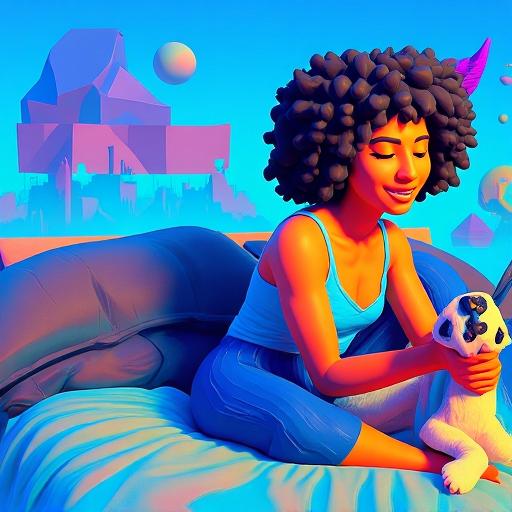}
        \\
        Input Image &  \multicolumn{4}{c}{$\text{Preservation} \xleftrightarrow{\hspace{3cm}} \text{Instruction Adherence}$}

    \end{tabular}

    \caption{
    Our results for continuous preference control in image editing. Rows 1-5 were trained on the FFHQ editing data, rows 6-7 were trained on the EditScore \cite{luo2025editscore} data.
    }
    \label{fig:i2i_results_supp}
\end{figure*}

\begin{figure*}[t]
    \setlength{\tabcolsep}{0.002\textwidth}
    \scriptsize
    \centering
    \begin{tabular}{c c c c c c c c c}
        \multicolumn{9}{c}{\textit{``A white horse galloping along a sandy beach.''}} \\
        \includegraphics[width=0.105\linewidth]{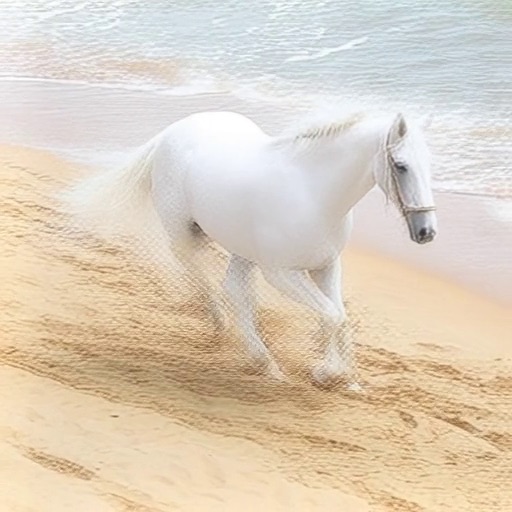} &
        \includegraphics[width=0.105\linewidth]{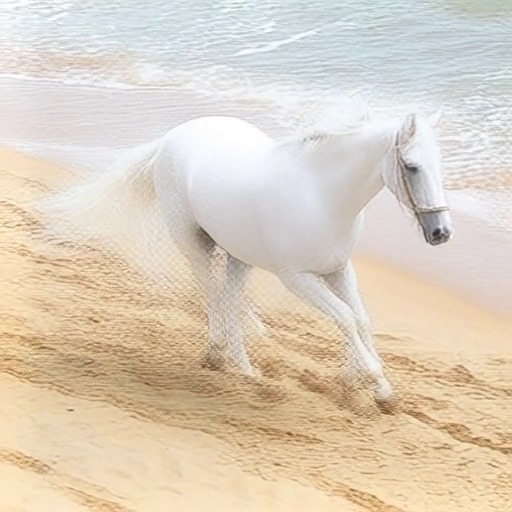} &
        \includegraphics[width=0.105\linewidth]{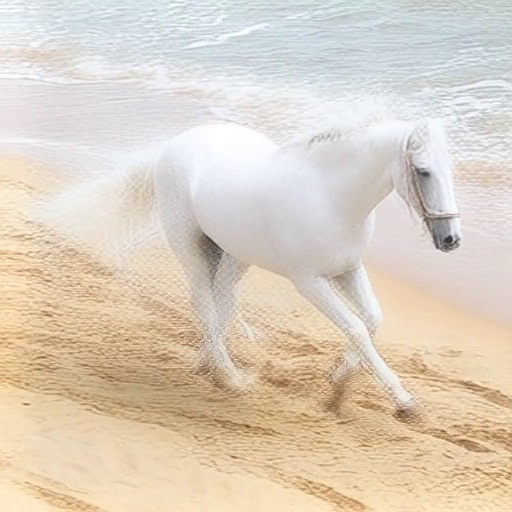} &
        \includegraphics[width=0.105\linewidth]{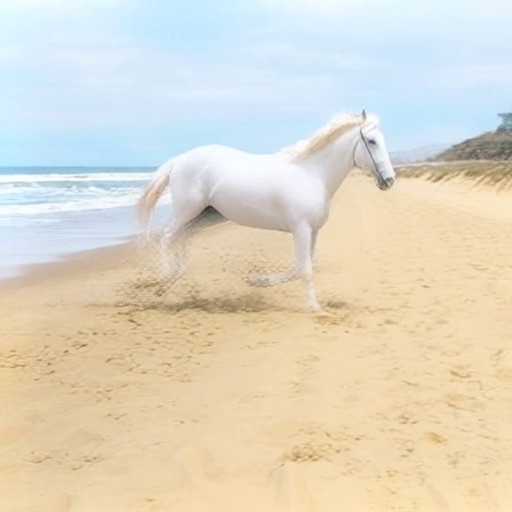} &
        \includegraphics[width=0.105\linewidth]{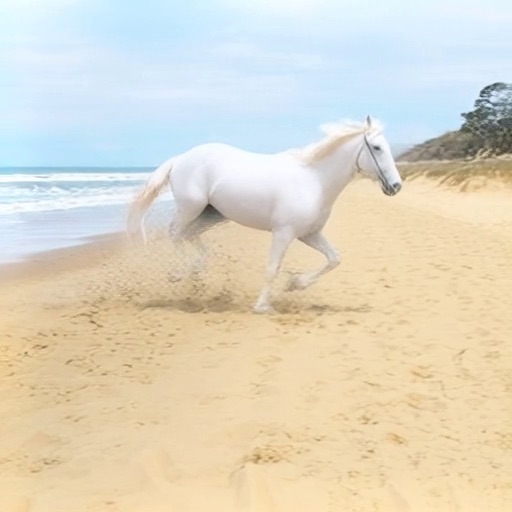} &
        \includegraphics[width=0.105\linewidth]{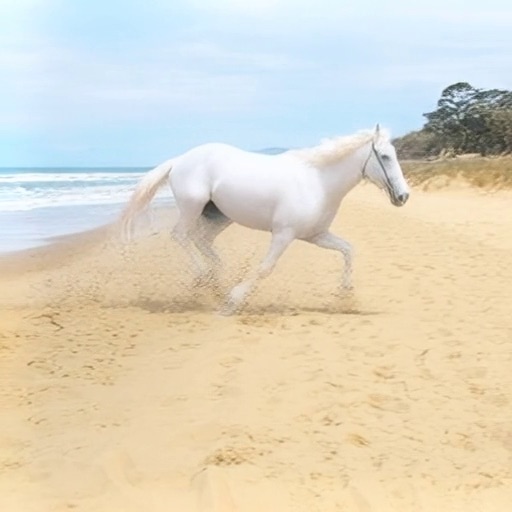} &
        \includegraphics[width=0.105\linewidth]{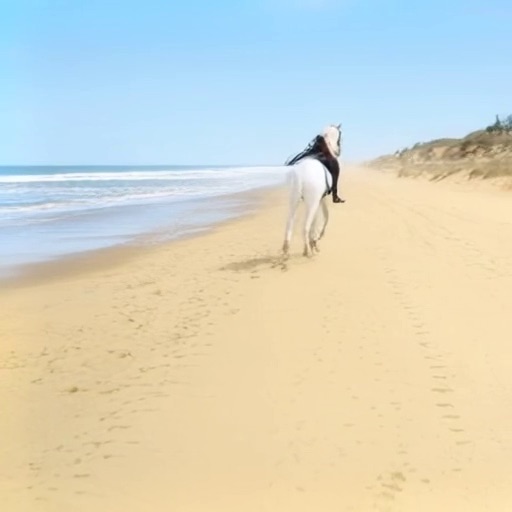} &
        \includegraphics[width=0.105\linewidth]{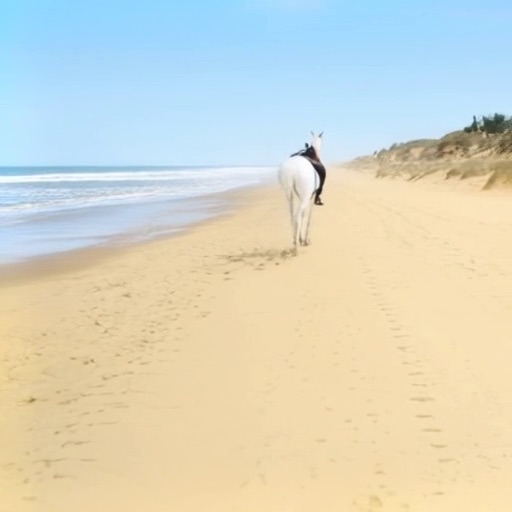} &
        \includegraphics[width=0.105\linewidth]{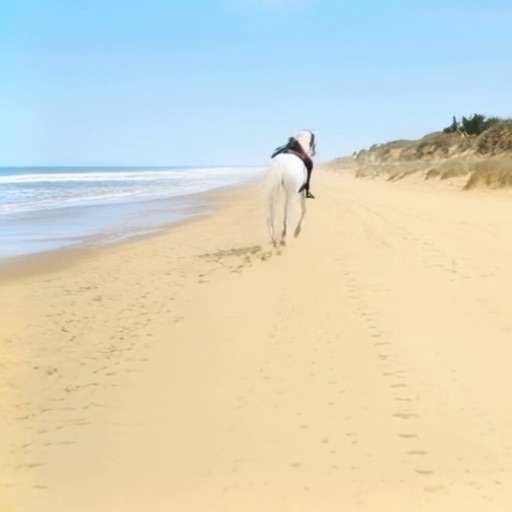} \\
        \multicolumn{9}{c}{\textit{``A fluffy kitten batting at a ball of yarn.''}} \\
        \includegraphics[width=0.105\linewidth]{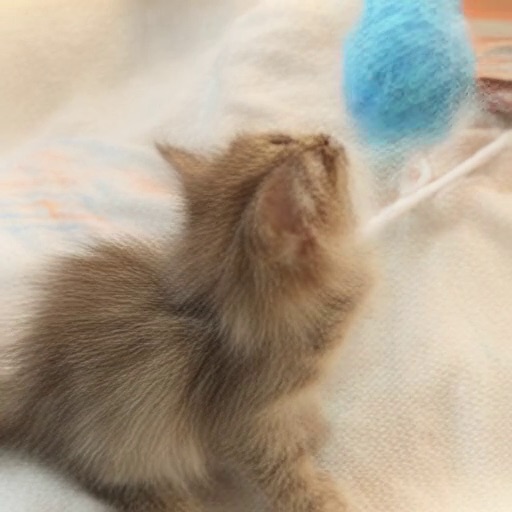} &
        \includegraphics[width=0.105\linewidth]{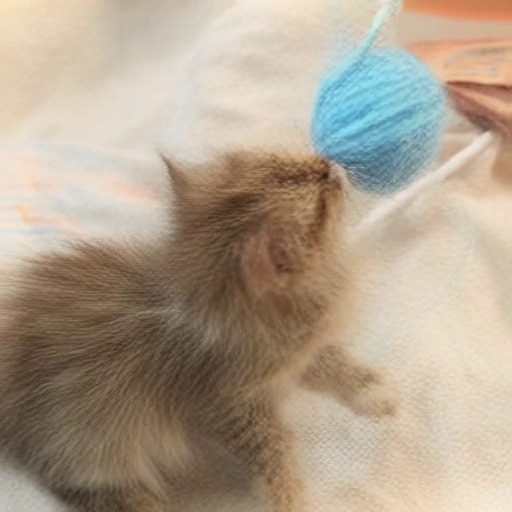} &
        \includegraphics[width=0.105\linewidth]{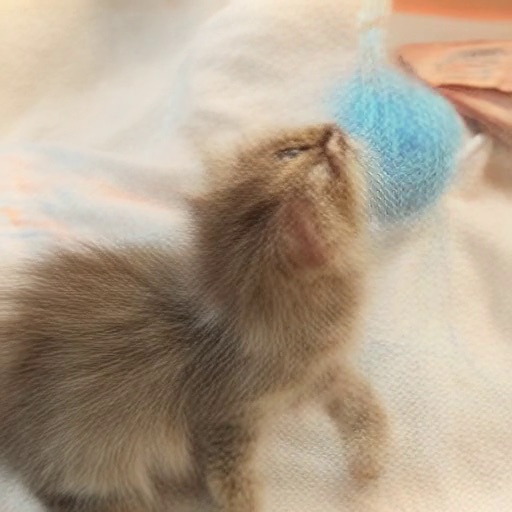} &
        \includegraphics[width=0.105\linewidth]{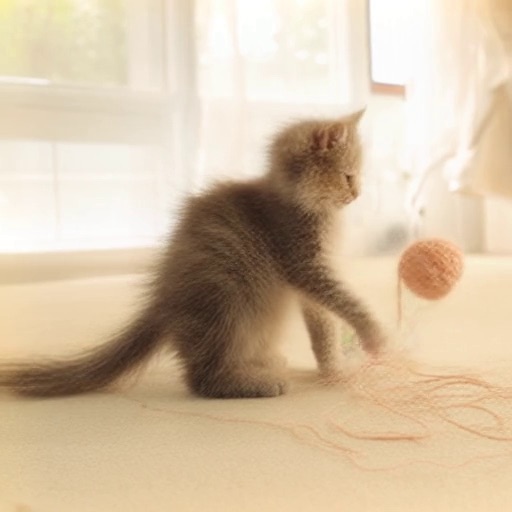} &
        \includegraphics[width=0.105\linewidth]{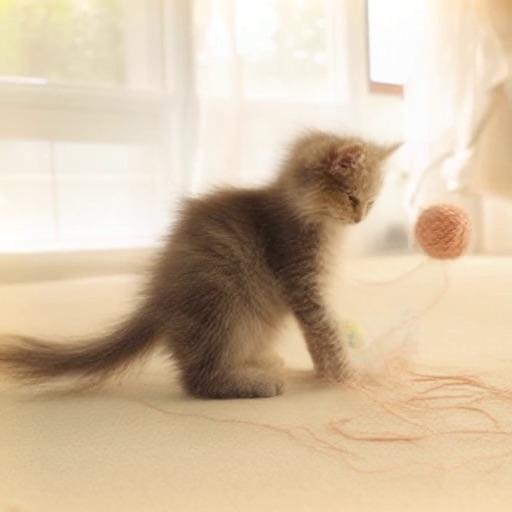} &
        \includegraphics[width=0.105\linewidth]{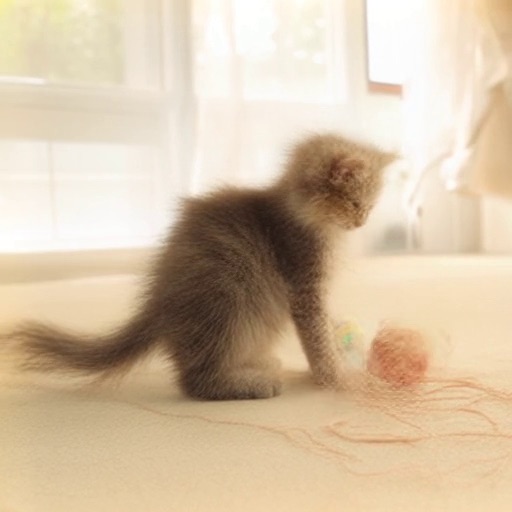} &
        \includegraphics[width=0.105\linewidth]{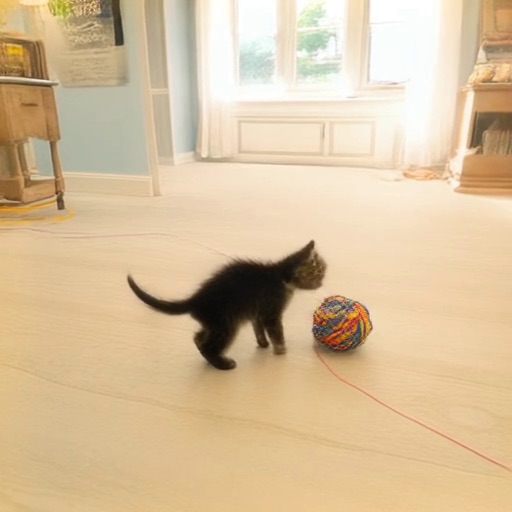} &
        \includegraphics[width=0.105\linewidth]{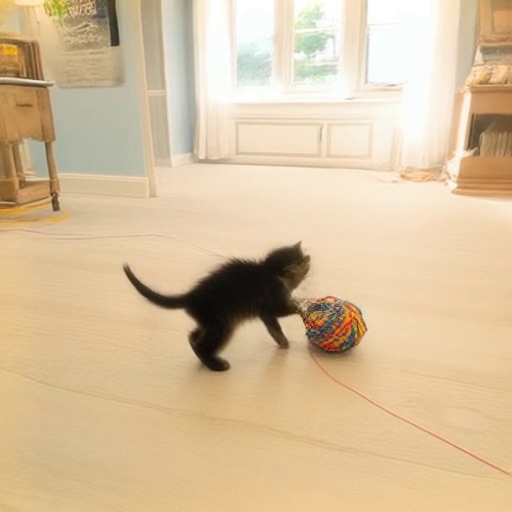} &
        \includegraphics[width=0.105\linewidth]{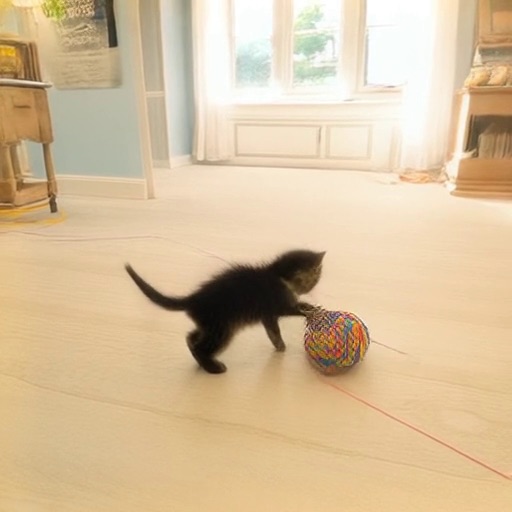} \\
        \multicolumn{9}{c}{$\text{Closeup} \xleftrightarrow{\hspace{10cm}} \text{Wideshot}$} \\
        \\
        \multicolumn{9}{c}{\textit{``A campfire crackling in the woods.''}} \\
        \includegraphics[width=0.105\linewidth]{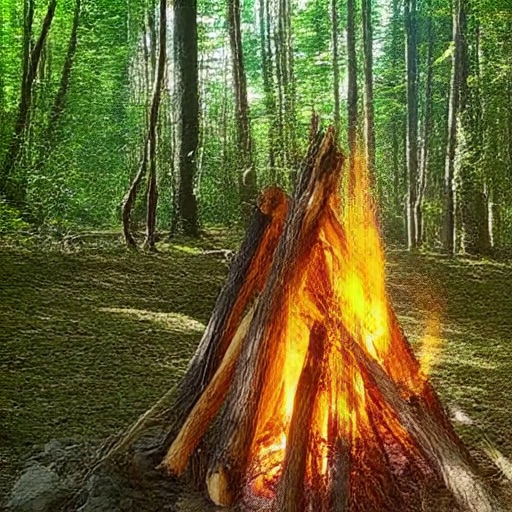} &
        \includegraphics[width=0.105\linewidth]{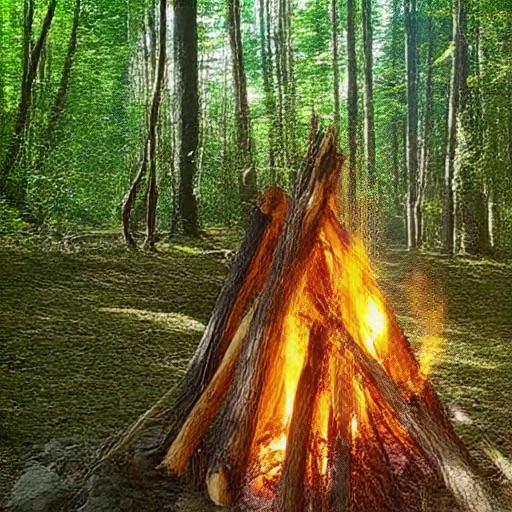} &
        \includegraphics[width=0.105\linewidth]{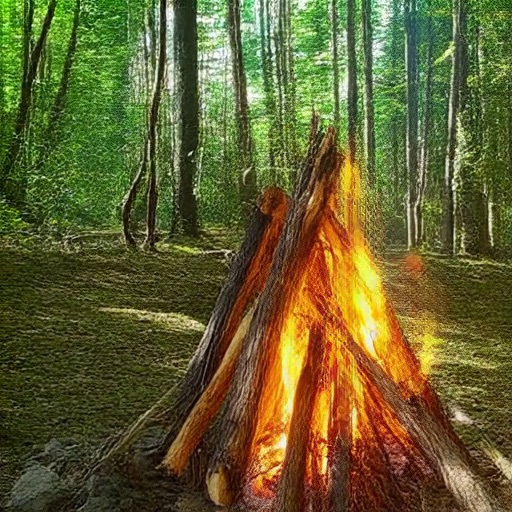} &
        \includegraphics[width=0.105\linewidth]{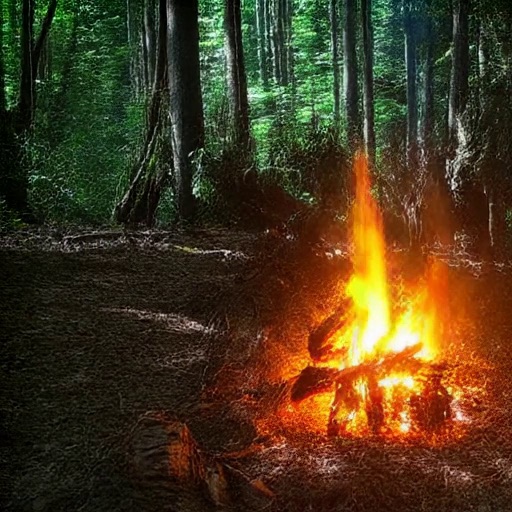} &
        \includegraphics[width=0.105\linewidth]{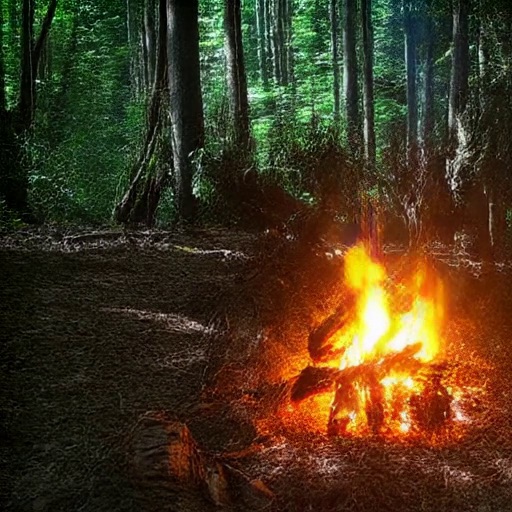} &
        \includegraphics[width=0.105\linewidth]{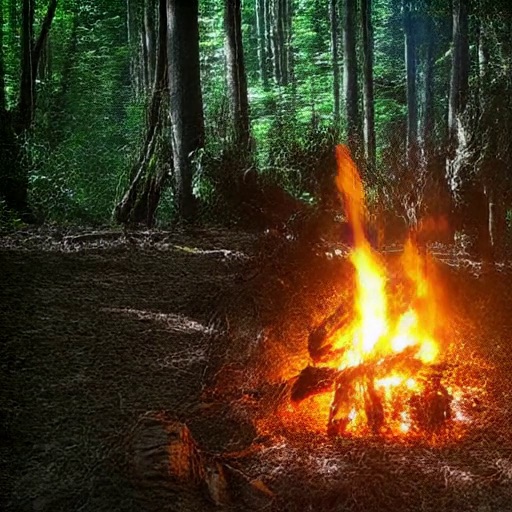} &
        \includegraphics[width=0.105\linewidth]{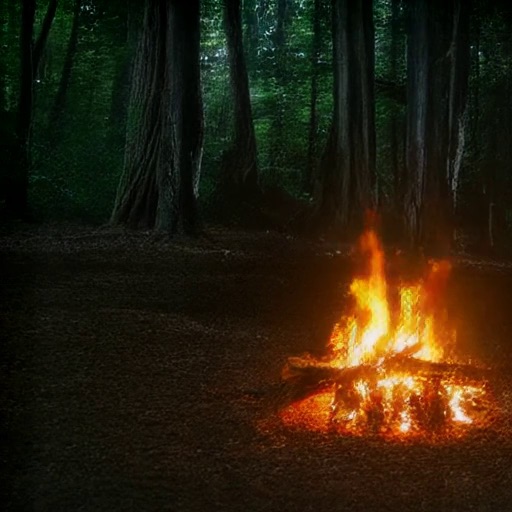} &
        \includegraphics[width=0.105\linewidth]{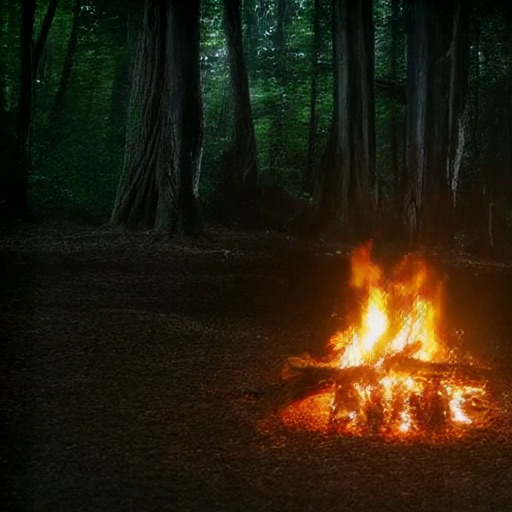} &
        \includegraphics[width=0.105\linewidth]{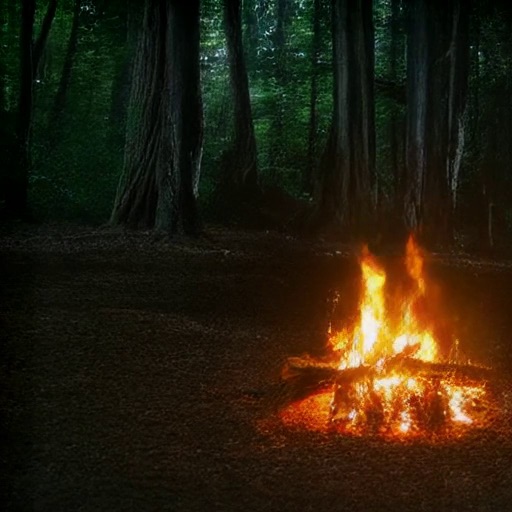} \\
        \multicolumn{9}{c}{\textit{``An owl blinking on a branch.''}} \\
        \includegraphics[width=0.105\linewidth]{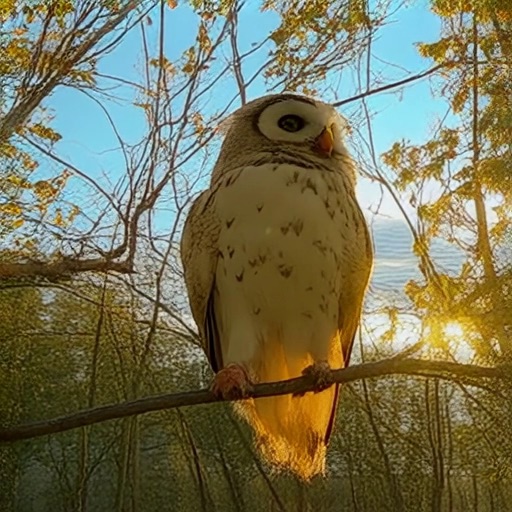} &
        \includegraphics[width=0.105\linewidth]{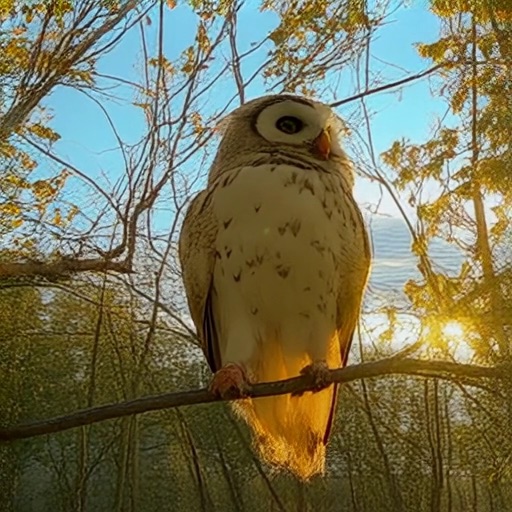} &
        \includegraphics[width=0.105\linewidth]{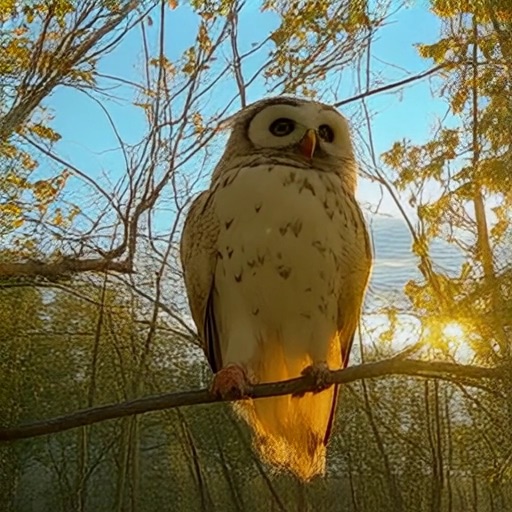} &
        \includegraphics[width=0.105\linewidth]{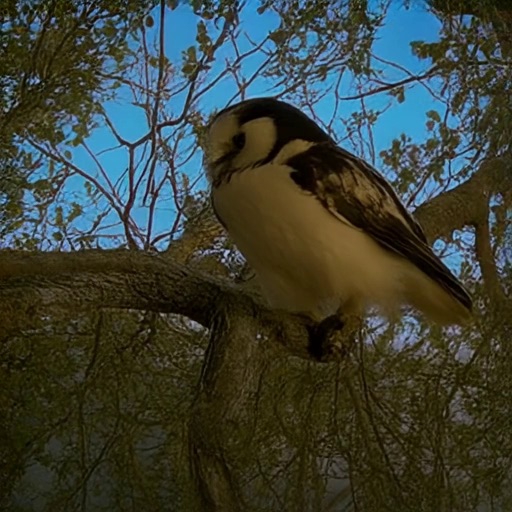} &
        \includegraphics[width=0.105\linewidth]{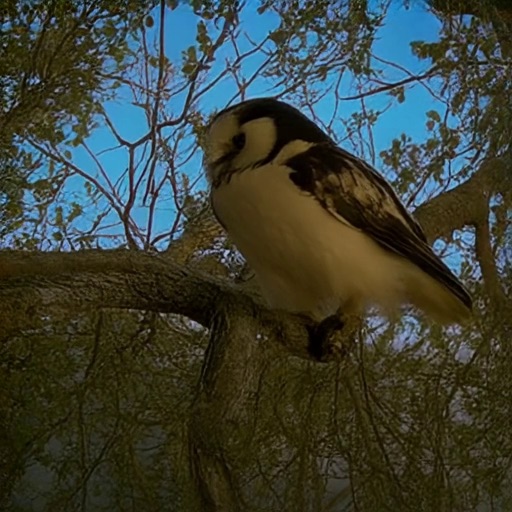} &
        \includegraphics[width=0.105\linewidth]{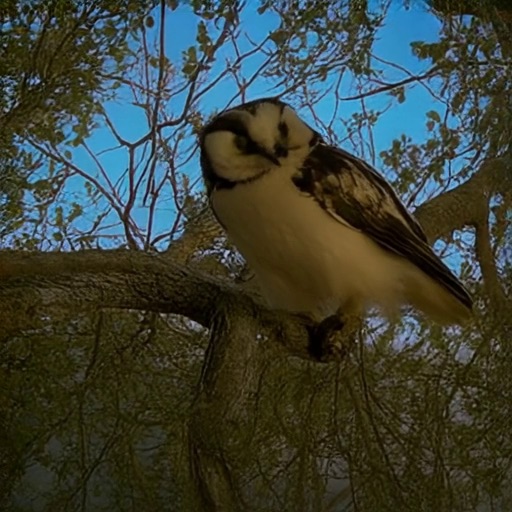} &
        \includegraphics[width=0.105\linewidth]{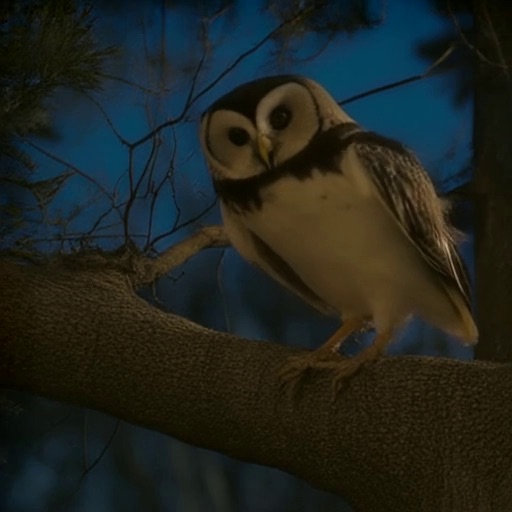} &
        \includegraphics[width=0.105\linewidth]{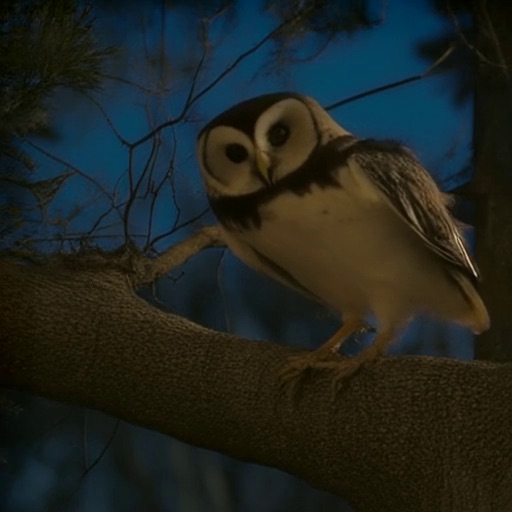} &
        \includegraphics[width=0.105\linewidth]{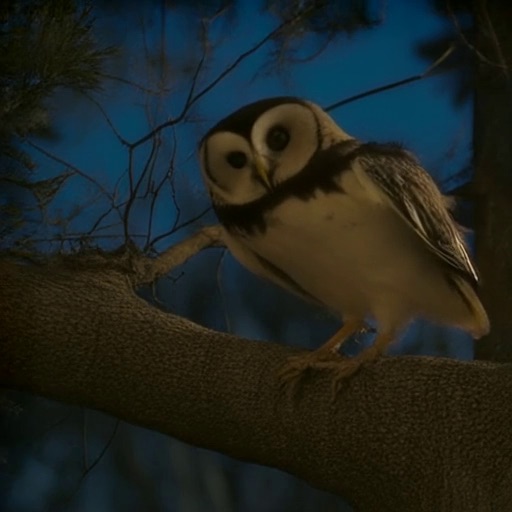} \\
        \multicolumn{9}{c}{$\text{Day} \xleftrightarrow{\hspace{10cm}} \text{Night}$} \\
        \\
        \multicolumn{9}{c}{\textit{``A golden retriever running through a grassy field.''}} \\
        \includegraphics[width=0.105\linewidth]{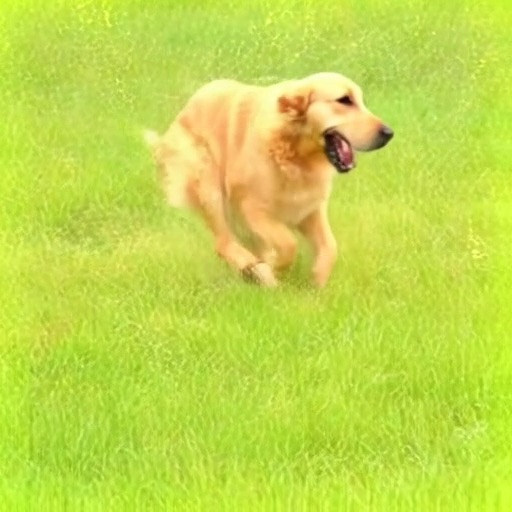} &
        \includegraphics[width=0.105\linewidth]{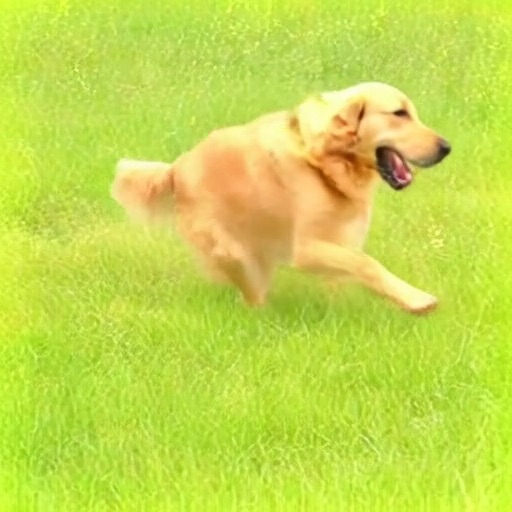} &
        \includegraphics[width=0.105\linewidth]{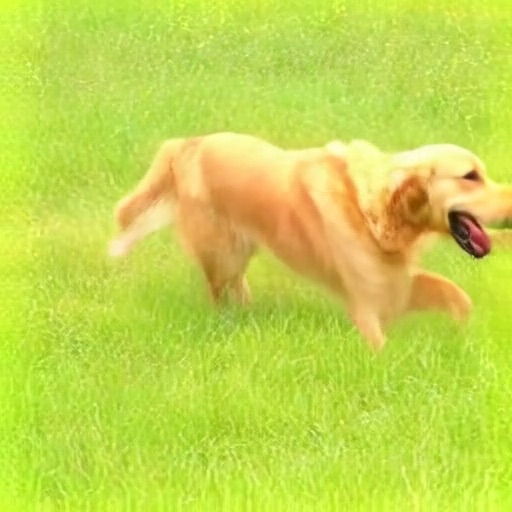} &
        \includegraphics[width=0.105\linewidth]{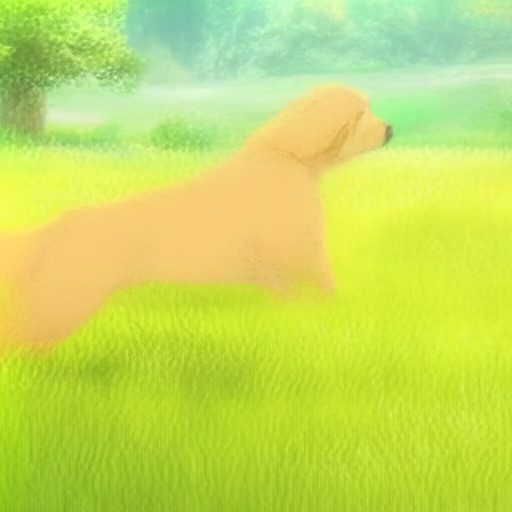} &
        \includegraphics[width=0.105\linewidth]{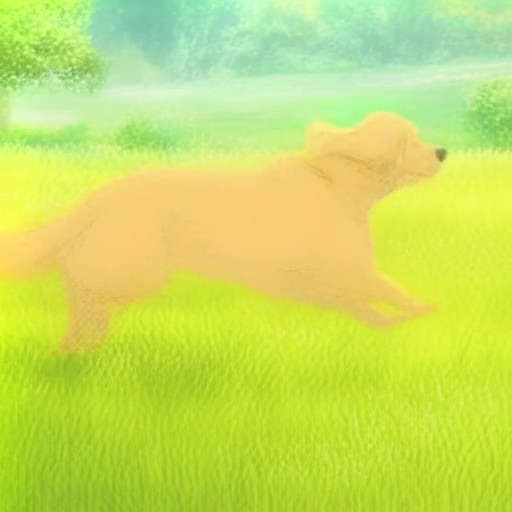} &
        \includegraphics[width=0.105\linewidth]{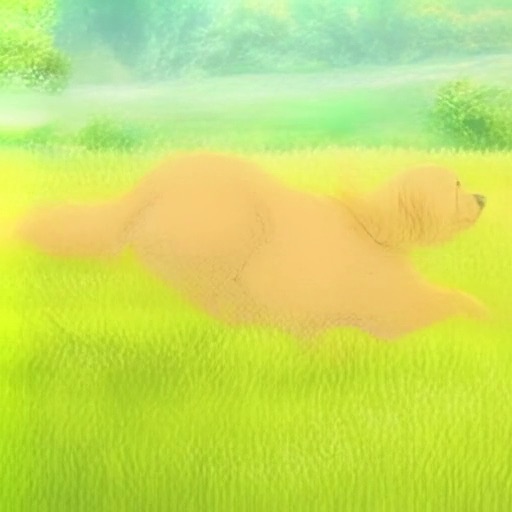} &
        \includegraphics[width=0.105\linewidth]{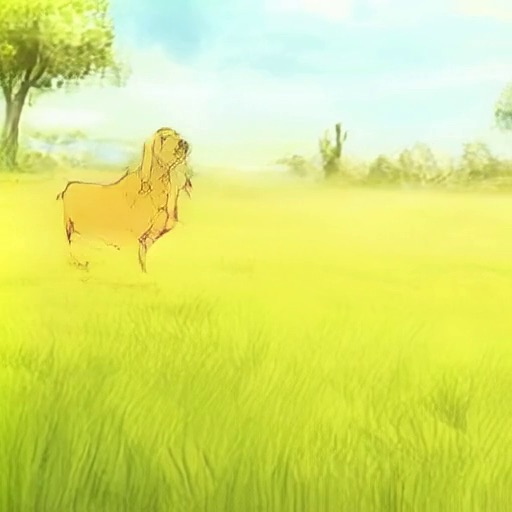} &
        \includegraphics[width=0.105\linewidth]{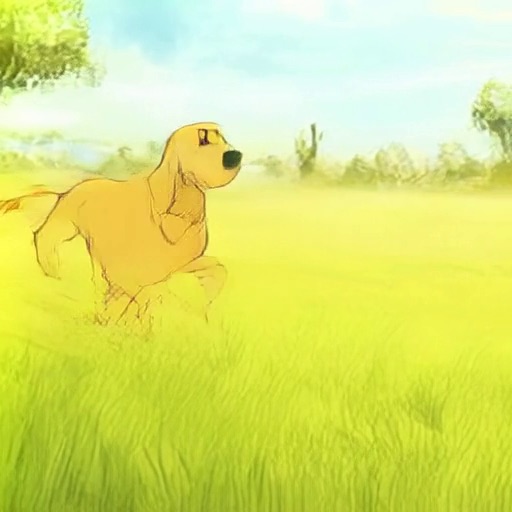} &
        \includegraphics[width=0.105\linewidth]{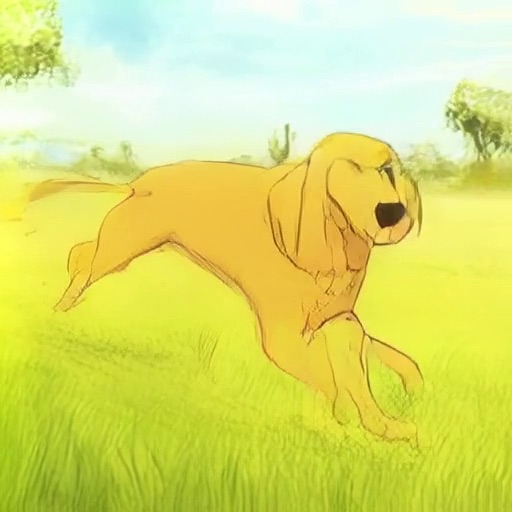} \\
        \multicolumn{9}{c}{\textit{``A sea turtle gliding through clear blue water.''}} \\
        \includegraphics[width=0.105\linewidth]{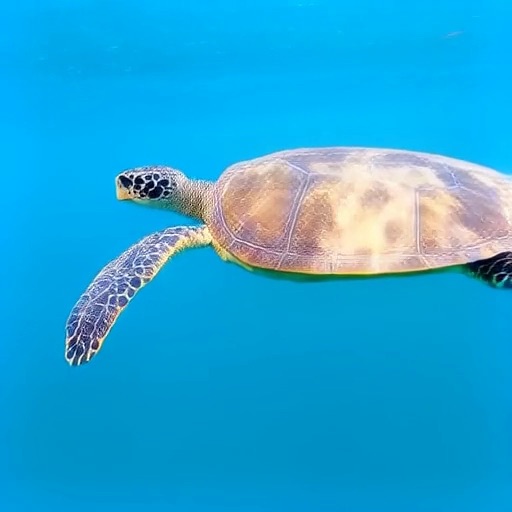} &
        \includegraphics[width=0.105\linewidth]{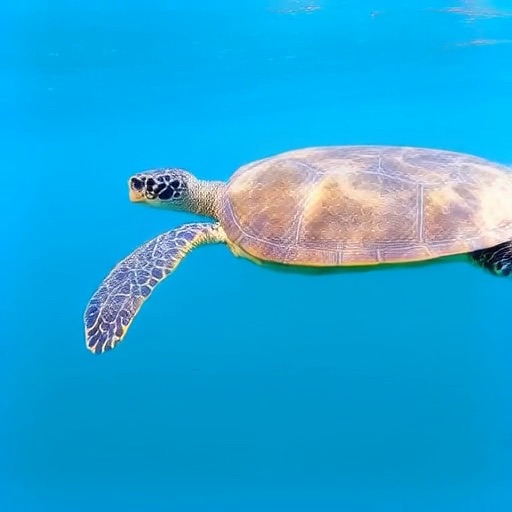} &
        \includegraphics[width=0.105\linewidth]{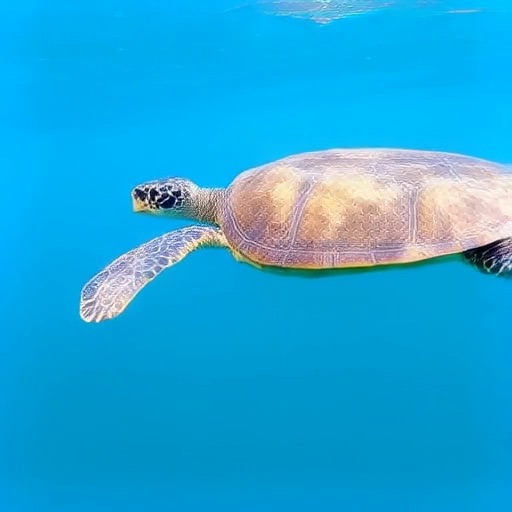} &
        \includegraphics[width=0.105\linewidth]{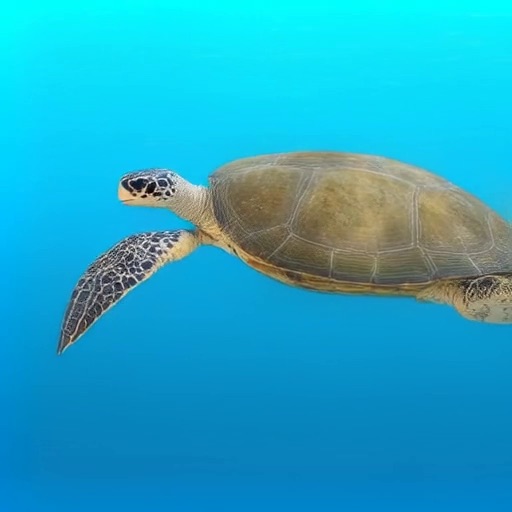} &
        \includegraphics[width=0.105\linewidth]{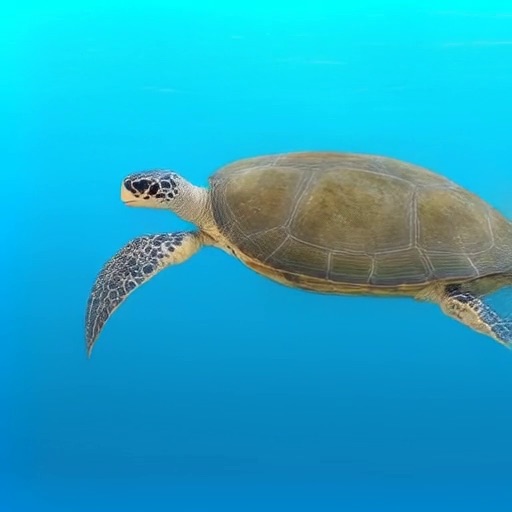} &
        \includegraphics[width=0.105\linewidth]{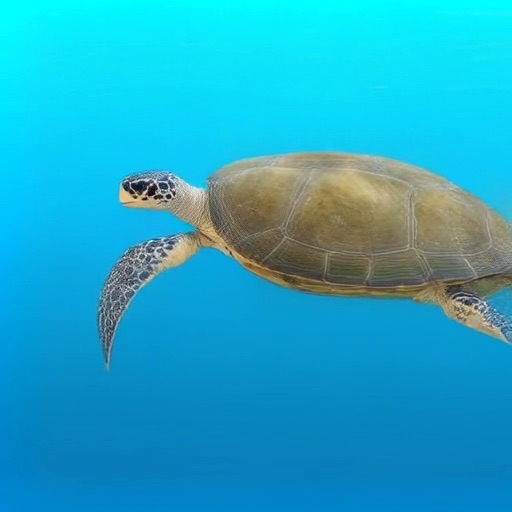} &
        \includegraphics[width=0.105\linewidth]{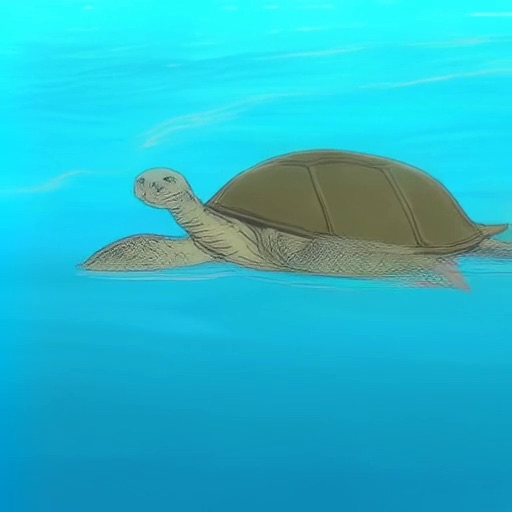} &
        \includegraphics[width=0.105\linewidth]{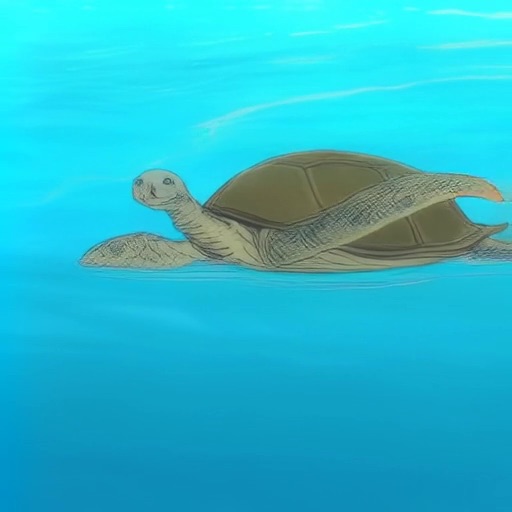} &
        \includegraphics[width=0.105\linewidth]{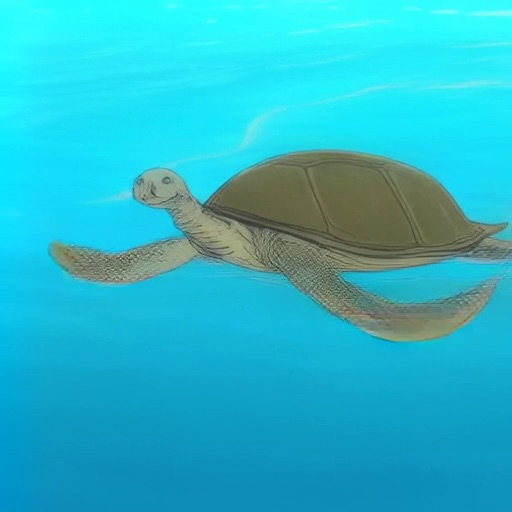} \\
        \multicolumn{9}{c}{$\text{Realistic} \xleftrightarrow{\hspace{10cm}} \text{Sketch}$} \\
        \\
    \end{tabular}
    \caption{Additional qualitative results on text-to-video generation.}
    \label{fig:ltx2_supp_extra}
\end{figure*}

\begin{figure*}
    \setlength{\tabcolsep}{0.002\textwidth}
    \renewcommand{\arraystretch}{0.0}
    \scriptsize
    \centering
    
    \begin{tabular}{c@{\hspace{4pt}}c c c c}
        & \textit{Photorealistic} & \textit{Sketch} & \textit{Flat-Vector} & \textit{Watercolor} \\[2pt]
        
        \raisebox{20pt}{\rotatebox{90}{\textit{Photorealistic}}} &
        \includegraphics[width=0.22\linewidth]{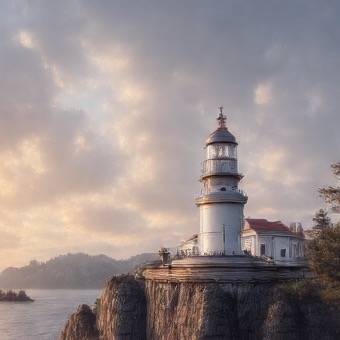} &
        \includegraphics[width=0.22\linewidth]{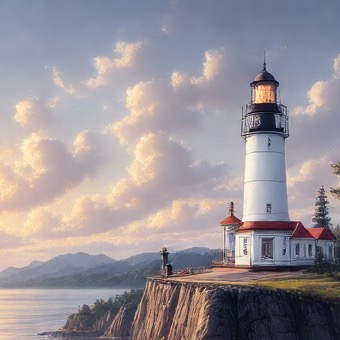} &
        \includegraphics[width=0.22\linewidth]{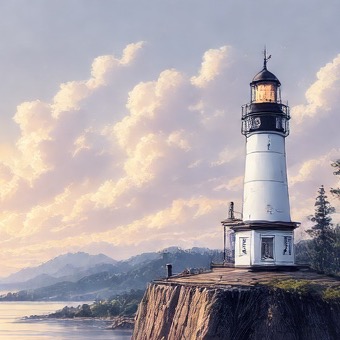} &
        \includegraphics[width=0.22\linewidth]{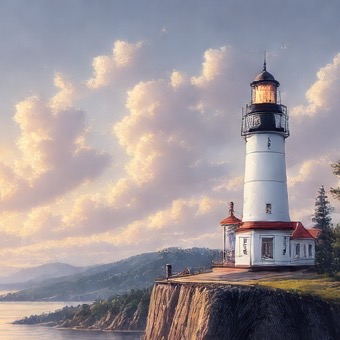} \\
        
        \raisebox{22pt}{\rotatebox{90}{\textit{Flat-Vector}}} &
        \includegraphics[width=0.22\linewidth]{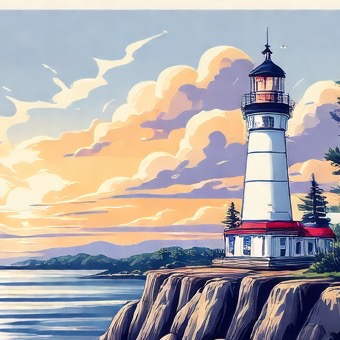} &
        \includegraphics[width=0.22\linewidth]{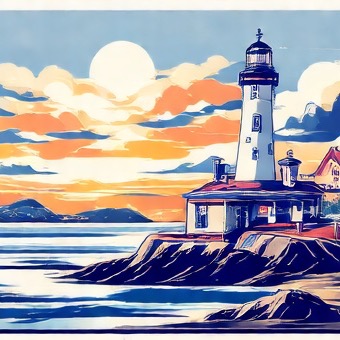} &
        \includegraphics[width=0.22\linewidth]{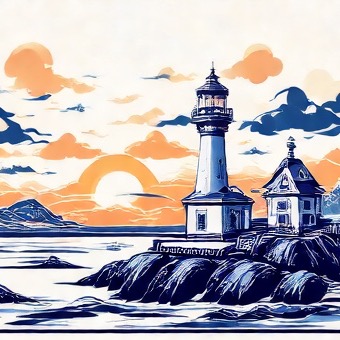} &
        \includegraphics[width=0.22\linewidth]{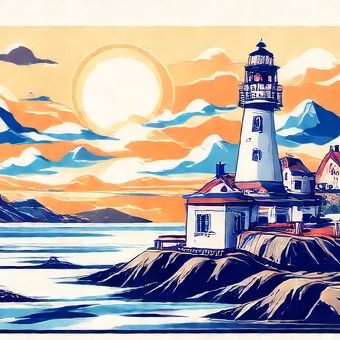} \\
        
        \raisebox{26pt}{\rotatebox{90}{\textit{Sketch}}} &
        \includegraphics[width=0.22\linewidth]{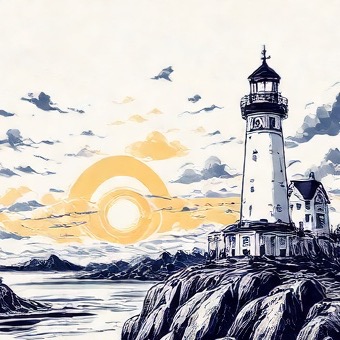} &
        \includegraphics[width=0.22\linewidth]{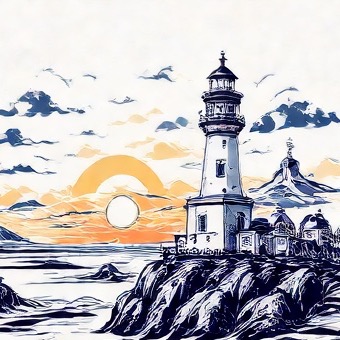} &
        \includegraphics[width=0.22\linewidth]{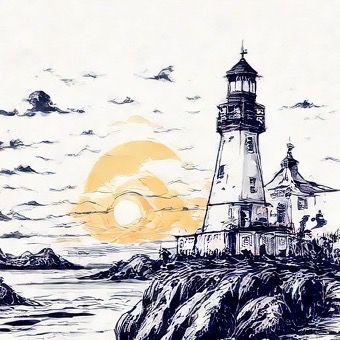} &
        \includegraphics[width=0.22\linewidth]{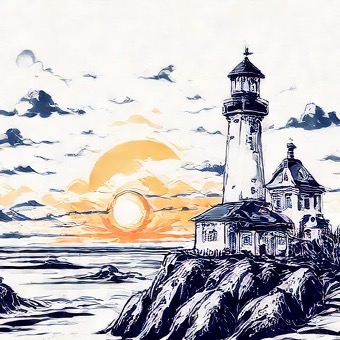} \\
        
        \raisebox{20pt}{\rotatebox{90}{{\textit{Watercolor}}}} &
        \includegraphics[width=0.22\linewidth]{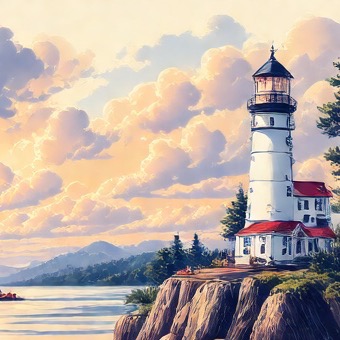} &
        \includegraphics[width=0.22\linewidth]{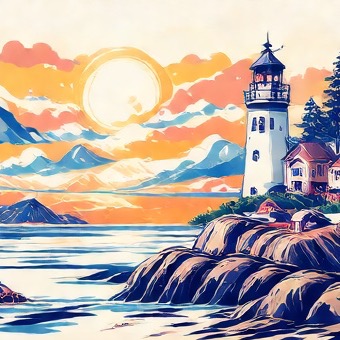} &
        \includegraphics[width=0.22\linewidth]{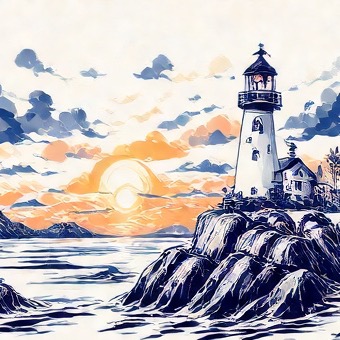} &
        \includegraphics[width=0.22\linewidth]{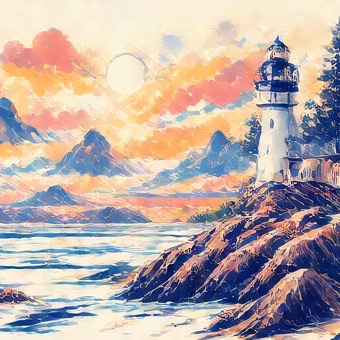} \\
    \end{tabular}

    \caption{
    Continuous control over four jointly-trained reward objectives. The diagonal shows each reward at full strength ($w_i{=}1$); off-diagonal cells show a blend of the row reward at weight $0.8$ and the column reward at weight $0.2$ (other rewards set to zero). Prompt: \textit{``a lighthouse on a rocky cliff at sunset''}.
    }
    \label{fig:four_reward_grid}
\end{figure*}

\clearpage

\vspace{-15pt}
\begin{tcolorbox}[colback=white,colframe=black!50, boxrule=0.5pt]
\label{box:qwenphoto}
\fontsize{8pt}{8pt}\selectfont \ttfamily
``You are judging sampled frames from a generated video against a caption.
Caption: ``A photorealistic video of [prompt].''

Task: Give ONE integer score from 0 to 5 based on BOTH content alignment and photorealistic style. Be strict. Do not hallucinate details.

Step 1) Style checks (pass/fail):
- Looks like real camera footage (not illustration, not cartoon, not 3D render).
- Realistic lighting and shadows consistent across frames.
- Realistic textures/materials (skin, fabric, metal, wood, etc.\ look natural if present).
- No painterly brush strokes, no heavy outlines, no flat shading.
- Temporal consistency: style is uniform across all frames.

Step 2) Content checks (pass/fail):
- Main subject(s) in [prompt] clearly present.
- Key attributes from [prompt] present (count, colors, distinctive parts).
- Key relationships/actions from [prompt] correct (if any).

Scoring rule:\\
5: All style checks pass AND all content checks pass; frames are clear and detailed.\\
4: Style passes AND content mostly correct with only minor issues.\\
3: Either (A) style passes but content has clear mistakes, or (B) content correct but one style check fails.\\
2: Multiple content mistakes and/or multiple style failures, but some intent visible.\\
1: Very weak match; most requirements unmet.\\
0: Totally wrong or unusable frames.

Response format: output ONLY the single digit 0 1 2 3 4 or 5.''
\end{tcolorbox}

\vspace{-15pt}

\begin{tcolorbox}[colback=white,colframe=black!50, boxrule=0.5pt]
\label{box:qwenpixar}
\fontsize{8pt}{8pt}\selectfont \ttfamily
``You are judging sampled frames from a generated video against a caption.
Caption: ``Pixar-style 3D animation of [prompt]. Stylized, expressive characters with smooth rendering.''

Task: Output ONE integer score 0 to 5 for BOTH content alignment and demanded Pixar animation style. Be strict. Do not guess unseen details.

Step 1) Style checks (pass/fail):
- 3D-rendered appearance with smooth, clean surfaces (not photorealistic, not flat 2D).
- Stylized proportions: exaggerated or cartoon-like features (large eyes, rounded shapes).
- Soft, diffuse lighting typical of Pixar/animated films (no harsh real-world shadows).
- Vibrant but natural-looking colors with clean CG material quality (not oversaturated or blown-out).
- No photorealistic textures, no painterly brush strokes, no anime/cel-shading.
- Temporal consistency: 3D animated style is uniform across all frames.

Step 2) Quality checks (pass/fail):
- No visual glitches, color banding, oversaturation, or distorted geometry.
- Faces and bodies are well-formed (no melted or deformed features).

Step 3) Content checks (pass/fail):
- Main subject(s) in [prompt] present and recognizable.
- Key attributes and relationships in [prompt] correct (if any).

Scoring rule:\\
5: All style, quality, and content checks pass; crisp and readable.\\
4: Style and quality pass AND content mostly correct (minor missing attribute/detail).\\
3: Either (A) style+quality pass but content has clear mistakes, or (B) content correct but 1 style/quality check fails.\\
2: Partial match; multiple failures but some intent visible.\\
1: Very weak match.\\
0: Totally wrong or unusable frames.

Response format: output ONLY the single digit 0 1 2 3 4 or 5.''
\end{tcolorbox}

\vspace{-15pt}

\begin{tcolorbox}[colback=white,colframe=black!50, boxrule=0.5pt]
\label{box:qwensketch}
\fontsize{8pt}{8pt}\selectfont \ttfamily
``You are presented with a generated image and its associated text caption. Your task is to evaluate how much the image resembles a hand-drawn sketch. Consider: visible line work, pencil/ink stroke quality, plain backgrounds, and artistic simplification.

Rate the sketchiness on a scale from 1 to 5:\\
1: No sketch qualities - fully rendered or photorealistic image \\
2: Mostly rendered with slight sketch-like elements \\
3: Mixed style - some sketch qualities but also rendered areas \\
4: Mostly sketch-like with strong line work and minimal rendering \\
5: Pure sketch - clear hand-drawn line work with no photorealistic rendering \\

Output your evaluation using the format below:
Sketch Score (1-5): X''

\end{tcolorbox}

\vspace{-10pt}

\begin{tcolorbox}[colback=white,colframe=black!50, boxrule=0.5pt]
\label{box:edit}
\fontsize{8pt}{8pt}\selectfont \ttfamily
``You are a professional digital artist. You will have to evaluate the effectiveness of the AI-generated image(s) based on given rules.

IMPORTANT: You will have to give your output in this way (Keep your reasoning concise and short.):
\{
"reasoning" : "...",
"score" : [...],
\}

RULES:

Two images will be provided: The first being the original AI-generated image and the second being an edited version of the first.
The objective is to evaluate how successfully the editing instruction has been executed in the second image.

Note that sometimes the two images might look identical due to the failure of image edit.

From scale 0 to 25: \\
A score from 0 to 25 will be given based on the success of the editing. (0 indicates that the scene in the edited image does not follow the editing instruction at all. 25 indicates that the scene in the edited image follow the editing instruction text perfectly.) \\
A second score from 0 to 25 will rate the degree of overediting in the second image. (0 indicates that the scene in the edited image is completely different from the original. 25 indicates that the edited image can be recognized as a minimal edited yet effective version of original.)
Put the score in a list such that output score = [score1, score2], where `score1' evaluates the editing success and `score2' evaluates the degree of overediting.

Editing instruction: [instruction]''
\end{tcolorbox}
\vspace{-15pt}

\end{document}